\documentclass{article}

\usepackage{microtype}
\usepackage{graphicx}
\usepackage{booktabs} 
\usepackage{subcaption}

\usepackage{hyperref}

\usepackage[accepted]{icml2024}

\usepackage{amsmath}
\usepackage{amssymb}
\usepackage{mathtools}
\usepackage{amsthm}
\usepackage{dsfont}

\usepackage[capitalize,noabbrev]{cleveref}

\theoremstyle{plain}
\newtheorem{theorem}{Theorem}[section]

\newtheorem{corollary}[theorem]{Corollary}
\theoremstyle{definition}

\theoremstyle{remark}
\newtheorem{remark}[theorem]{Remark}

\usepackage[textsize=tiny]{todonotes}

\usepackage{tikz}
\usetikzlibrary{calc,arrows.meta,positioning}

\definecolor{palette0}{HTML}{780000}
\definecolor{palette5}{HTML}{669BBC}

\def\ci{\perp\!\!\!\perp}

\icmltitlerunning{Conformal Validity Guarantees Exist for Any Data Distribution (and How to Find Them)}

\begin{document}

\twocolumn[


\icmltitle{Conformal Validity Guarantees Exist for Any Data Distribution \\ (and How to Find Them)}

\icmlsetsymbol{equal}{*}

\begin{icmlauthorlist}
\icmlauthor{Drew Prinster$^*$}{jhu}
\icmlauthor{Samuel Stanton$^*$}{gne}
\icmlauthor{Anqi Liu}{jhu}
\icmlauthor{Suchi Saria}{jhu}
\end{icmlauthorlist}

\icmlaffiliation{jhu}{Department of Computer Science, Johns Hopkins University, Baltimore, MD, U.S.A.}
\icmlaffiliation{gne}{Prescient Design, Genentech, New York City, NY, U.S.A.}

\icmlcorrespondingauthor{Drew Prinster}{drew@cs.jhu.edu}
\icmlcorrespondingauthor{Samuel Stanton}{stanton.samuel@gene.com}

\icmlkeywords{Machine Learning, ICML, conformal prediction, AI agents, exchangeability, covariate shift, active learning, black-box optimization}

\vskip 0.3in
]



\printAffiliationsAndNotice{\icmlEqualContribution} 

\begin{abstract}
    As artificial intelligence (AI) / machine learning (ML) gain widespread adoption, practitioners are increasingly seeking means to quantify and control the risk these systems incur. This challenge is especially salient when such systems have autonomy to collect their own data, such as in black-box optimization and active learning, where their actions induce sequential feedback-loop shifts in the data distribution. Conformal prediction is a promising approach to uncertainty and risk quantification, but prior variants' validity guarantees have assumed some form of ``quasi-exchangeability'' on the data distribution, thereby excluding many types of sequential shifts. In this paper we prove that conformal prediction  can theoretically be extended to \textit{any} joint data distribution, not just exchangeable or quasi-exchangeable ones. Although the most general case is exceedingly impractical to compute, for concrete practical applications we outline a procedure for deriving specific conformal algorithms for any data distribution, and we use this procedure to derive tractable algorithms for a series of AI/ML-agent-induced covariate shifts. We evaluate the proposed algorithms empirically on synthetic black-box optimization and active learning tasks.
\end{abstract}

\section{Introduction}
\label{sec:intro}

Quantifying the uncertainty of predictions from artificial intelligence (AI) and machine learning (ML) models is often imperative to managing the risks associated with their downstream use. Principled uncertainty quantification (UQ) is thus especially crucial in real-world scenarios where AI/ML systems are used to inform high-stakes decisions, or where they may even act autonomously. Unfortunately, it is precisely when uncertainty estimation is most vital when it is often most difficult to quantify reliably. For a prime example, consider the increasingly common setting where AI systems are able to take actions of their own, thereby transforming them from passive observers into active agents: common instances include active learning, black-box optimization, and reinforcement learning. In all these cases, merely enabling an AI model to explore by actively selecting the next datapoint induces feedback-loop shifts in the data distribution, which can accumulate over time and cause standard UQ methods to deteriorate or break down entirely. 

Conformal prediction (CP) \citep{vovk2005algorithmic} is an increasingly popular UQ framework because it provides coverage guarantees without knowledge of the exact distributional form of the data. Even so, standard CP methods do make one very strong assumption about the data distribution: namely, that the data are \textit{exchangeable} (IID is a special case). Informally, for time-series data, exchangeability means the distribution is ``time invariant’’; meanwhile in a batch setting, it means the distribution remains constant between training and test batches. Exchangeability thus excludes many real-world settings such as dynamic time series or data generated from (or queried by) active AI agents, which can exhibit stark time-dependent or batch-dependent distribution shifts. 

This work builds on that of \citet{tibshirani2019conformal}, who introduced a weighted generalization of CP where the traditional exchangeability requirement is relaxed to a condition called \textit{weighted exchangeability}. For instance, data that are independent but not identically distributed are weighted exchangeable. However, the conventional wisdom is still that conformal prediction methods require some notion of “quasi-exchangeability” to achieve a valid coverage guarantee.\footnote{For instance, \citet{fannjiang2022conformal} introduced a similar assumption called “pseudo-exchangeability” to study data shifts with limited dependencies of the test point on the observed data.}  We challenge this common perception via the following theoretical and practical contributions:
\vspace{-0.25cm}
\begin{itemize}
    \item Our main theoretical contribution is to present a general view of (weighted) conformal prediction, drawing heavily upon the analysis of \citet{tibshirani2019conformal}. In particular, we prove that conformal prediction can in theory be extended to \textit{any} joint distribution of data with a valid probability density function $f$, including non-exchangeable ones with sequential dependencies, although the fully general form is not practically feasible to compute (factorial runtime and requiring knowledge of $f$). From this general perspective, ``quasi-exchangeability’’ definitions (including exchangeability and weighted exchangeability) are no longer necessary conditions for conformal prediction theory; instead, from this view, these assumptions' only role is to specify conditions for practical tractability.
    \item For practical applications, we present a general procedure for deriving specific weighted CP algorithms for any given data distribution, and we demonstrate how this procedure can be used to derive weighted CP methods with valid coverage guarantees for sequential feedback-loop covariate shifts that characterize common AI/ML agent scenarios. We moreover show that these methods can be realized in practice via empirical evaluation on \textit{in silico} protein design and active learning tasks, which verify that we maintain coverage when state-of-the-art baselines fail.
\end{itemize}

\section{Background}
\label{sec:background}

Assume we are given an initial dataset $Z_{1:n} := \{Z_i\}_{i=1}^n$, where $Z_i := (X_i, Y_i) \in \mathcal{X} \times \mathcal{Y}$ are real-valued ($\mathcal{X} \times \mathcal{Y} \subseteq \mathbb{R}^p \times \mathbb{R}$) feature-label pairs with some distribution function $P_Z^{(0)}$.
At each future time $t$, we observe a test point's features $X_{n+t}$ and are interested in rigorous UQ for an ML estimate of the unknown true label $Y_{n+t}$. For simplicity we interpret $t$ as a timestep, but in general each datapoint only needs an identifying index; also, we often let $t=1$ to focus on a single test point indexed $n+1$. When distinction is needed, capital letters (e.g., $Z_i$) denote random variables, and lower case (e.g., $z_i$) denote observed values.




\subsection{Standard Conformal Prediction}

\label{subsec:background_standard_CP}

Conformal prediction (CP) is a framework for \textit{predictive inference}: the problem of computing a predictive confidence set $\widehat{\mathcal{C}}_n(x)$  (in regression, this is often an interval) likely to contain the true label at least at a user-specified target rate $1-\alpha\in (0, 1)$. This desiderata is called \textit{valid coverage}:\footnote{We focus on marginal coverage (i.e., on average over many repeated experiments) rather than conditional coverage; see \citet{foygel2021limits} for more details on this distinction.}
\begin{align}
    \mathbb{P}\big\{Y_{n+1} \in \widehat{\mathcal{C}}_n(X_{n+1})\big\} \geq 1 - \alpha. \label{eq:coverage_condition}
\end{align}
For an accessible introduction to CP see \citet{angelopoulos2022conformal} or for the definitive reference see \citet{vovk2005algorithmic}. Computing conformal predictive sets requires access to a real-valued score function $\mathcal{S}$ that takes as its inputs a point $(x, y)$ and a ``bag''\footnote{That is, a multiset. CP methods assume the score-fitting algorithm is symmetric with respect to permutations of its training inputs $\bar{Z}$; we maintain this assumption unless stated otherwise.} of other examples $\bar{Z}$, where $\mathcal{S}\big((x,y), \bar{Z}\big)$ can be thought of as quantifying how ``nonconforming'' or ``strange'' the point $(x,y)$ is relative to $\bar{Z}$. For example, the absolute value residual $|\widehat{\mu}_{\bar Z}(x) - y|$ is a common score function, where $\widehat{\mu}_{\bar Z}$ is an ML predictor trained on the examples $\bar Z$. For shorthand, we will sometimes write the fitted score function as $\widehat{\mathcal{S}}(x,y)=\mathcal{S}\big((x,y), \bar{Z}\big)$ and outputted score values as $V_i = \widehat{\mathcal{S}}(Z_i)$. 
With $Q_{1-\alpha}$ denoting an empirical quantile evaluated at level $1-\alpha$, a standard CP set $\widehat{\mathcal{C}}_n(x)$ is then constructed by taking a (conservative) quantile on the empirical distribution of score values for each $y\in \mathcal{Y}$:
\begin{align}
    \widehat{\mathcal{C}}_n(x) = \Big\{ y \in \mathcal{Y} : \widehat{\mathcal{S}}(x, y) \leq \mathrm{Q}_{1-\alpha}\big(V_{1:n} \cup \{\infty\}\big)\Big\}.\label{eq:exchangeable_split_cp}
\end{align}
Standard CP methods assume that all the data $Z_1, \dots, Z_{n + 1}$ are \textit{exchangeable}---that is, that their joint distribution is invariant to permutations, a special case being independent and identically distributed (IID). Under exchangeability, $\widehat{\mathcal{C}}_n(x)$ in Eq. \eqref{eq:exchangeable_split_cp} achieves the valid coverage guarantee in Eq. \eqref{eq:coverage_condition} with finite data samples.
This standard CP guarantee is \textit{distribution-free} in the sense of making no assumptions about the specific parametric
form of the data distribution---this is not to be confused with robustness to distribution \textit{shift}, which violates exchangeability for example when the distribution of $Z_{n+1}$ differs from that of $Z_{1:n}$. Exchangeability is thus still a strong assumption in its own way.
For instance, it excludes many environment-dependent or time-dependent distribution shifts which often occur in practice.

\subsection{Weighted Conformal Prediction for Covariate Shift}

\citet{tibshirani2019conformal} presented an alternate proof of the conformal coverage guarantee in a way that isolated the role of exchangeability and allowed them to generalize conformal prediction to a broader class of distributions they termed \textit{weighted exchangeable}.
In particular, they thoroughly developed a specific weighted CP algorithm for the \textit{standard covariate shift} setting \citep{shimodaira2000improving}, where the marginal distribution of the inputs $X$ may differ independently between training and test distributions, but the conditional distribution $Y\mid X$ is assumed to be invariant:
\begin{align}
    (X_i, Y_i) &\overset{\text{i.i.d.}}{\sim} P_Z^{(0)} = P_X^{(0)} \times P_{Y \mid X}, \; i = 1, \dots n, \nonumber \\
    (X_{n + 1}, Y_{n + 1}) &\hspace{0.5mm}\sim P_Z^{(1)} = P_X^{(1)} \times P_{Y \mid X}, \ \text{independently}. \nonumber
\end{align}
Note that in standard covariate shift the test distribution $P_Z^{(1)}$ is \textit{independent} from the training data; it is fixed \textit{a priori} and cannot change depending on different draws of $Z_{1:n}$. \citet{tibshirani2019conformal}'s weighted CP algorithm for standard covariate shift generalizes Eq. \eqref{eq:exchangeable_split_cp} with likelihood (or density) ratio function weights $w(x) = p_X^{(1)}(x)/p_X^{(0)}(x)$.

\citet{fannjiang2022conformal} followed the overall approach of \citet{tibshirani2019conformal} to derive a specific weighted CP algorithm with a valid coverage guarantee (Eq. \eqref{eq:coverage_condition}) in a setting they called (one-step) \textit{feedback covariate shift} (FCS), where $P_{X; Z_{1:n}}^{(1)}$ is explicitly \textit{dependent} on the realized draw $Z_{1:n}$. 
While this work was an important extension of CP to dependent data, they did not consider a more general ``multistep'' FCS case that could describe how data shifts may accumulate over time, for instance in active learning. Their focus on the one-step FCS case could be because they posit as the basis of their theoretical analysis an assumption they call \textit{pseudo-exchangeability}, which explicitly characterizes one-step FCS but does not formally describe its multistep analog (see Appendix \ref{app:pseudo_diff_def}), which we introduce next.

\subsection{Multistep Feedback Covariate Shift}
\label{sec:MFCS}

Whereas standard and feedback covariate shift are ``one-step'' shifts between training and test data batches, in this work we focus our practical evaluations on a more general setting we call \textit{multistep feedback covariate shift} (MFCS). MFCS is not only a clear example of a sequentially-dependent data distribution that goes far beyond exchangeability, but it moreover describes important instances of feedback-loop shifts inherent to the actions of AI/ML agents.
For each time $t \in \{1, ..., T\}$, MFCS allows $X_{n + t}$ to change depending on the past data $Z_{1:(n+t-1)}=\{Z_i\}_{i=1}^{n+t-1}$, while assuming that $Y | X$ remains invariant. 
More precisely, MFCS assumes
\begin{align}
    (X_i, Y_i) &\overset{\text{i.i.d.}}{\sim} P_X^{(0)} \times P_{Y \mid X}, \; i = 1, \dots n, \label{eq:multistep_fcs}\\
    (X_{n + t}, Y_{n + t}) &\hspace{0.5mm}\sim P_{X;Z_{1:(n+t-1)}}^{(t)} \times P_{Y \mid X}, \; t = 1, \dots, T, \nonumber
\end{align}
where $n$ is the size of an IID initialization, and $t$ is the number of subsequent MFCS steps. 
Setting $T=1$ reduces Eq. \eqref{eq:multistep_fcs} to one-step FCS \citep{fannjiang2022conformal} and moreover to standard covariate shift if independence is further assumed.
MFCS is commonly induced by ML agents able to actively select each datapoint $X_{n + t}$, informed by prior training observations, so long as the predictive goal or ``concept'' $Y | X$ remains the same. For instance, MFCS describes active learning, where the goal is to efficiently construct a training corpus to improve model performance, and black-box optimization, where the ML model is an instrument to optimize $X$ with respect to some unknown cost (or utility) function only indirectly accessible by each observation of $Y$.



\section{Related Work}
\label{sec:related_work}

\textbf{Conformal Prediction Under Distribution Shift} Prior work can largely be categorized into variants of \textit{weighted} CP \citep{tibshirani2019conformal, podkopaev2021distribution, xu2021conformal, fannjiang2022conformal, prinster2022jaws, prinster2023jaws, barber2023conformal, farinhas2023non, nair2023randomization, yang2024doubly} and \textit{adaptive} CP \citep{gibbs2021adaptive, gibbs2022conformal, zaffran2022adaptive, angelopoulos2023conformal, feldman2023achieving, bhatnagar2023improved}.
The former exploits knowledge or estimates of the data shift to proactively adjust inference and thereby achieve guarantees that hold at inference time; meanwhile, the latter aims to maintain a target risk threshold by retroactively responding to threshold violations and thus obtain guarantees most meaningful ``in the long run'' rather than at inference time. 

This work builds on the weighted CP literature,
and our practical applications focus on cases of MFCS where the shift is agent-induced and thus known.
Aside from \citet{tibshirani2019conformal} and \citet{fannjiang2022conformal}, \citet{prinster2022jaws} and \citet{prinster2023jaws} also studied CP under (one-step) standard or feedback covariate shift by extending cross-validation-style CP methods (from \citet{barber2021predictive}) to those settings for flexibly balancing compute-versus-sample efficiency. \citet{nair2023randomization} studied a ``$Y$-stationarity'' setting similar to MFCS with a Monte Carlo-based CP algorithm for discrete data. Otherwise, our work relates to \citet{barber2023conformal}, which bounds the worst-case ``coverage gap'' (loss of coverage below Eq. \eqref{eq:coverage_condition}) if the CP weights are unknown and instead fixed \textit{a priori} independently of the data. In contrast, we study CP for non-exchangeable distributions using data-dependent weights to achieve coverage validity guarantees as in Eq. \eqref{eq:coverage_condition} (without a coverage gap).


\textbf{Conformal Prediction for Decision Making} Ideas from CP are increasingly being applied to decision-making to inform utility estimates \citep{vovk2018conformal,  stanton2023bayesian, salinas2023optimizing} or satisfy constraints \citep{lekeufack2023conformal, zhang2023bayesian, dixit2023adaptive, laufer2023risk, jin2023selection}.
In particular, \citet{stanton2023bayesian} drew motivation from \citet{fannjiang2022conformal} to combine CP with Bayesian optimization but noted that a gap in theory prevented a formal coverage guarantee after the first round, a gap we address in this work.




\section{Theory and Method Contributions}
\label{sec:methods_and_theory}

\subsection{The Role of (Weighted) Exchangeability and Related Assumptions in Conformal Prediction}
\label{subsec:key_insight_role_of_definitions}

\textbf{Informal Statement of the Key Insight} One can think of CP as an inverted permutation test (with a conservative adjustment). Informally, the key insight is that---from this perspective of an inverted permutation test---the only role of exchangeability, weighted exchangeability \citep{tibshirani2019conformal}, or related distributional assumptions such as pseudo-exchangeability \citep{fannjiang2022conformal} in CP is to enable tractability by telling us how much to ``weigh'' or ``count'' a given ordering of datapoints; in particular, by facilitating computation of the relative likelihood of each sequence of observations, relative to some (permutation-invariant) core function, given the unordered ``bag'' of observed values (details in Appendix \ref{sec:roles_exchangeability_extensions}). So, extensions of CP should be possible so long as one can quantify the relative probability of any ordering of observed datapoints (or score values).

\textbf{Formal Sketch of the Key Insight} 
We first sketch analysis from \citet{tibshirani2019conformal} (which corresponds to viewing CP as an inverted permutation test) followed by the key insight for our current paper's main result.
Their analysis begins by conditioning on the set\footnote{For simplicity, we assume all the values are distinct to work with sets rather than multisets; in general, the argument holds with more complex notation using uniform randomness to break ties.} of data values (or scores)---that is, on the event $\{Z_1, ..., Z_{n+1}\} = \{z_{1}, ..., z_{n+1}\}$ denoted by $E_z$. To be clear, $E_z$ means the \textit{unordered} ``bag'' of values is known but \textit{not} whether $Z_i=z_{i}$, and so on; we know that the random variable $Z_i$ takes on an observed value in $\{z_1, ..., z_{n+1}\}$, but we do not yet know \textit{which} value (similarly for the scores $V_i$). Given $E_z$, the goal is to examine the probability that the test point's score $V_{n+1}$ was actually $v_i$, for each $i\in \{1, ..., n+1\}$, which they write as
\begin{align}
    \mathbb{P}\{V_{n+1}=v_i\ | \ &E_z\} = \mathbb{P}\{Z_{n+1}=z_i\ | \ E_z\} \nonumber \\ 
    = & \frac{\sum_{\sigma:\sigma(n+1)=i}f(z_{\sigma(1)}, ..., z_{\sigma(n+1)})}{\sum_{\sigma}f(z_{\sigma(1)}, ..., z_{\sigma(n+1)})},
    \label{eq:summary_perm_test}
\end{align}
where $f$ denotes the joint probability density function (PDF). Standard CP proceeds from this statement by substituting $f(z_{\sigma(1)}, ..., z_{\sigma(n+1)})=f(z_{1}, ..., z_{n+1})$ via exchangeability, reducing the problem to counting permutations; weighted CP proceeds by factorizing $f$ using weighted exchangeability, with the product of weight functions representing the ``weight'' given to each permutation $\sigma$; and, CP with pseudo-exchangeable data proceeds similarly to handle a specific instance of potential dependencies (details in Appendix \ref{sec:roles_exchangeability_extensions}).

Generally, the key insight in our paper is that simplifying Eq. \eqref{eq:summary_perm_test} is only a practical rather than a theoretically necessary step. For intuition on this, note that Eq. \eqref{eq:summary_perm_test} makes no assumptions on $f$; it follows from the definition of conditional probability 
and the law of total probability (LOTP):
\begin{align*}
    \mathbb{P}\{Z_{n+1}=z_i\ | \ E_z\} &= \frac{p\{Z_{n+1}=z_i,  E_z\} }{p\{E_z\} } \\
    &= \frac{\sum_{\sigma:\sigma(n+1)=i}f(z_{\sigma(1)}, ..., z_{\sigma(n+1)})}{\sum_{\sigma}f(z_{\sigma(1)}, ..., z_{\sigma(n+1)})}.
\end{align*}
That is, the last step uses LOTP in both the numerator and denominator---for instance, the denominator's sum is over every way the event $E_z$ could occur, which amounts to all the possible permutations $\sigma$ of the values. The crux of extending CP to any, possibly non-exchangeable $f$ can thus reduce to computing $f(z_{\sigma(1)}, ..., z_{\sigma(n+1)})$ directly for all orderings $\sigma$, or factoring $f$ into ``dynamic'' terms that are tractable to compute (e.g., ``weight functions'') and remaining permutation-invariant terms that cancel in the ratio. 
\vspace{-0.25cm}

\subsection{A General View of Conformal Prediction}
\label{subsec:general_result}

\textbf{Main Formal Result} With the insight from Section \ref{subsec:key_insight_role_of_definitions} in mind, we are now ready to see how---in theory---extensions of CP are possible for any, potentially non-exchangeable distribution of the calibration and test data $Z_1, ..., Z_{n+1}$ with a valid PDF\footnote{To be precise, we require a \textit{non-singular} distribution so that there exists a valid PDF $f$, or more generally, a valid Radon-Nikodym derivative with respect to an arbitrary base measure.
} $f$ (we let $t=1$ for simplicity, but in general our argument holds for $Z_1, ..., Z_{n+t}$).
This section's theorem formally states this result as the coverage guarantee for a generalized weighted CP set for arbitrary $f$. 

Let us define nonconformity scores more explicitly as 
\begin{align}
    \begin{cases}
    V_i^{(x,y)}&=\mathcal{S}(Z_i, Z_{1:n}\cup \{(x,y)\}), \ i\in \{1, ..., n\},  \\
    V_{n+1}^{(x,y)}&=\mathcal{S}\big((x,y), Z_{1:n}\cup\{(x,y)\}\big),
    \end{cases}
    \label{eq:def_scores_ordinary_full}
\end{align}
and for condensed notation, let us denote $\mathbb{P}_{n+1}\{z_i| E_z\} = \mathbb{P}\{Z_{n+1}=z_i| E_z\}$, recalling Eq. \eqref{eq:summary_perm_test}. To enable us to write a \textit{weighted} empirical distribution of values with probability weights $\mathbb{P}_{n+1}\{z_i| E_z\}$ applied to each value $v_i$, we will also now denote use $\delta_v$ to denote a point mass at the value $v$ (e.g., in the special case of uniform weights $\tfrac{1}{m}$ over values $v_1, ..., v_{m}$, then $Q_{1-\alpha}(v_{1:m})=Q_{1-\alpha}(\sum_{i=1}^m\tfrac{1}{m}\delta_{v_i})$). We now have the notation needed for our main theorem.

\begin{theorem}
\textit{Assume that $Z_i=(X_i, Y_i)\in\mathbb{R}^d\times \mathbb{R}, i=1, ..., n+1$ have the joint PDF $f$. For any measurable score function $\mathcal{S}$, and any $\alpha\in(0,1)$, define the generalized conformal prediction set (based on $n$ calibration samples) at a point $x\in \mathbb{R}^d$ by}
\begin{align}\label{eq:def_general_conformal_set}
&\widehat{\mathcal{C}}_n(x) = \bigg\{y\in \mathbb{R} : V_{n+1}^{(x,y)} \leq \\
&Q_{1-\alpha}\bigg(\sum_{i=1}^n\mathbb{P}_{n+1}\{Z_i| E_z\}\delta_{V_i^{(x,y)}}  + \mathbb{P}_{n+1}\{Z_{n+1}| E_z\}\delta_{\infty}\bigg)\bigg\}  \nonumber
\end{align}
\textit{where $V_i^{(x,y)}, i\in \{1, ..., n+1\}$ are as in \eqref{eq:def_scores_ordinary_full} and $\mathbb{P}_{n+1}\{z_i| E_z\}$ is notation for \eqref{eq:summary_perm_test}. Then, $\widehat{\mathcal{C}}_n$ satisfies}
\begin{align*}
    \mathbb{P}\big\{Y_{n+1}\in \widehat{\mathcal{C}}_n(X_{n+1})\big\}\geq 1-\alpha.
\end{align*}
    \label{thm:general_CP_def_coverage}
\end{theorem}
\vspace{-0.75cm}
\textbf{Proof Sketch} We defer the full proof for the result to Appendix \ref{sec:general_cp_coverage_proof}, but a sketch of the proof has two main steps:
\begin{enumerate}
    \item \textit{CP as an inverted (weighted) permutation test:} With setup as in Section \ref{subsec:key_insight_role_of_definitions}, use Eq. \eqref{eq:summary_perm_test} to derive a general quantile lemma, which informally states the following: For $\beta\in(0,1)$, the test point's score $V_{n+1}$ is contained within the level $\beta$ quantile of the weighted empirical distribution of score values (with weights defined as in Eq. \eqref{eq:summary_perm_test}) with probability at least $\beta$. 
    \item \textit{Conservative adjustment to connect to general CP set:} Connect the general quantile lemma from the first proof step to general CP set (Eq. \eqref{eq:def_general_conformal_set}) by conservatively adjusting for the fact that the true test point's score value is unknown. That is, replacing the test point's score value with $\infty$ in the empirical distribution maintains validity, and setting $\beta = 1-\alpha$ gives the theorem.
\end{enumerate}

\begin{remark}\label{remark:LOO_full_split}
    The score functions in Eq. \eqref{eq:def_scores_ordinary_full} correspond to the ``ordinary'' full CP method, but the same proof argument and result follows for the ``deleted'' full CP method \citep{vovk2005algorithmic}, which uses leave-one-out scores to prevent overfitting. Moreover, the same result follows for split conformal prediction as a special case. (See Appendix \ref{app:remark_details}.)
\end{remark}

\begin{remark}\label{remark:nonsymmetric_algo}
    As written, Theorem \ref{thm:general_CP_def_coverage} maintains the convention of requiring $\mathcal{S}$ to be symmetric with respect to any training data also used in calibration. This condition is relevant for full CP methods, but for split CP, this symmetry holds trivially due to training and calibration sets being separate. (See Appendix \ref{app:assymetric_algo_extension} for further details and discussion.)
\end{remark}





\textbf{Limitations and Benefits of the General View of CP} 
There are two main obstacles to computing Eq. \eqref{eq:def_general_conformal_set} in practice: a \textit{computational complexity} challenge and an \textit{epistemic} challenge regarding knowledge about $f$.
That is, in the general case for an arbitrary or unfactorized $f$, Eq. \eqref{eq:summary_perm_test} requires $\mathcal{O}((n+1)!)$ evaluations of $f$, which quickly becomes intractable. 
Secondly, we often do not know $f$; indeed, the primary motivation of distribution-free UQ is that very deficiency. Exchangeability, weighted exchangeability \citep{tibshirani2019conformal}, and related conditions such as pseudo-exchangeability \citep{fannjiang2022conformal} introduce distributional assumptions that alleviate these challenges.

One immediate benefit of the general presentation of conformal prediction in Theorem \ref{thm:general_CP_def_coverage} is a shift in perspective: because the theorem holds for any joint distribution $f$, for any \textit{particular} $f$ the question is not whether a CP set with valid coverage \textit{exists}, it is instead whether we can \textit{compute} the prediction set \textit{in practice}. Consequently---from this general perspective---``quasi-exchangeability'' definitions are no longer necessary conditions for CP theory; rather, the \textit{only role} of these definitions is to enable practical tractability by simplifying and eliminating terms from the computation (of Eq. \eqref{eq:summary_perm_test})---for example, at one extreme, exchangeability ``assumes away'' $f$ from the computation entirely.  (Appendix \ref{sec:roles_exchangeability_extensions} provides further formal details on this point, including how previous CP validity guarantees can be viewed as corollaries to Theorem \ref{thm:general_CP_def_coverage}.) Specific weighted CP algorithms with corresponding coverage validity guarantees can thus be derived for any (possibly non-exchangeable) distribution. In section \ref{sec:deriving_cp_any_dist_MFCS}, we outline how to perform such derivations in general and provide the MFCS case as a worked example. 

\textbf{Further Discussion of \citet{tibshirani2019conformal}} 
This general view of CP is implicit in a close reading of \citet{tibshirani2019conformal},\footnote{We note that the main insight in our paper is also invoked in the following lecture notes \citep{tibshiranilecture}: \url{https://www.stat.berkeley.edu/~ryantibs/statlearn-s23/lectures/conformal_ds.pdf}, which a sharp reader could use to infer Theorem \ref{thm:general_CP_def_coverage}. However, based on how \citet{tibshirani2019conformal} has been cited in the CP literature (i.e., primarily as extending CP to standard covariate shift or to weighted exchangeability), we believe this important insight has not been fully appreciated by the community and we hope that our contribution can help correct this oversight.} 
but in that paper the authors did not explicitly state or prove the general result we present in Theorem \ref{thm:general_CP_def_coverage}.
Instead, \citet{tibshirani2019conformal} premised their main result on their definition of weighted exchangeability, although this condition is not strictly necessary for CP in the broad sense. 
While we maintain that weighted exchangeability is still a useful definition (see Appendix \ref{app:weighted_exchangeability}), subsequent literature has seemingly been led to believe that new definitions of ``quasi-exchangeability'' are necessary for advancing and delineating the boundaries of CP theory---for example, \citet{fannjiang2022conformal}'s introduction of pseudo-exchangeability to study the single-step FCS setting. 
We hope our contribution will lead to wider recognition of the generality of \citet{tibshirani2019conformal}'s analysis and foster a broadened scope of possibility in the study of CP: in particular, we encourage a shift away from viewing ``quasi-exchangeability'' conditions as theoretical boundaries to be overcome, and instead towards primarily using such conditions to clearly specify the assumptions granted by a given setting of interest or adopted for practical tractability.













\subsection{How to Find Weighted CP Validity Guarantees and Algorithms for MFCS (or Any Data Distribution)}

\label{sec:deriving_cp_any_dist_MFCS}

\textbf{A General Procedure for Deriving Weighted CP Algorithms with Valid Coverage Guarantees} 
To improve the accessibility and practical utility of our paper's main theoretical contribution,
we now outline a general procedure that can be used to derive weighted CP algorithms for any specific joint PDF (or mass function) $f$ over the calibration and test data. Theorem \ref{thm:general_CP_def_coverage} implies the existence of this procedure without premising it on any assumptions on $f$, while emphasizing that the only role of distributional assumptions like exchangeability is for tractability in the last step.
\begin{enumerate}
    \item \textit{List assumptions (if any):} 
    For a given application setting, specify assumptions on $f$, if any, such as conditional independence or invariance assumptions. 
    (For example, the MFCS assumptions are formally given by Eq. \eqref{eq:multistep_fcs}; informally, MFCS assumes $Y | X$ is invariant, while $X$ is dynamic depending on past observations.) 
    Alternatively, one could begin by describing an application's data-generating process by a probablistic graphical model (e.g., a Markov random field or a causal directed acyclic graph) and use the conditional independence assumptions implied by that graphical model (for an example see Appendix \ref{app:subsec:alternate_derivations}; for details on such models \citet{koller2009probabilistic, pearl2009causality}).\footnote{Probablistic graphical models can intuitively represent a data-generating process as a graph with nodes corresponding to random variables (or random vectors). Missing edges in such a graph imply conditional independences in the joint PDF $f$; the converse depends on further assumptions \citep{pearl2009causality}.} 
    \item \textit{Factorize $f$:} Factorize the joint PDF $f$ using standard probability rules (e.g., conditional probability, chain rule, etc.) and assumptions from Step 1 to separate dynamic factors from those that are invariant (to permutations $\sigma$ of the data indices). 
    (For example, for MFCS $f$ can be written as a product of dynamic factors $X_j \mid Z_1,…,Z_{j-1}$ and invariant factors $Y\mid X$.)
    \item \textit{Compute or estimate weights:} Plug the factorized form of $f$ from Step 2 into Eq. \eqref{eq:summary_perm_test} to obtain the normalized calibration and test point weights, and simplify by canceling out permutation-invariant and constant factors. Compute or estimate the simplified weights, and plug them into \eqref{eq:def_general_conformal_set} to obtain the CP set. If computation is performed exactly, then the resulting CP set has a valid coverage guarantee premised on the assumptions from Step 1 (as a corollary to Theorem \ref{thm:general_CP_def_coverage}).
\end{enumerate}

\textbf{Deriving CP for Agent-Induced MFCS} For a concrete demonstration of how this general procedure can be used to derive weighted CP methods for a given practical setting, we now return to the MFCS setting introduced in Section \ref{sec:MFCS}, which characterizes common instances of ML-agent-induced feedback-loop data shifts. We provide only a sketch of key ideas here; for full details  see Appendix \ref{sec:MFCS_sequential_pseudo-exchangeability}. Step 1 of this procedure for MFCS begins by recalling Eq. \eqref{eq:multistep_fcs}, the conditional independence and invariance assumptions that we used to define MFCS. Using standard probability rules and leveraging these MFCS assumptions, for Step 2 we are able to factorize the joint PDF $f$ to separate the time-dependent factors corresponding to $X_j \mid Z_1,…,Z_{j-1}$ from the time-invariant factors corresponding to $Y\mid X$:
\begin{align}
    f(&z_1, ..., z_{n+t}) = \nonumber \\
    & \prod_{j=1}^{n+t}\Big[\underbrace{p(x_j\ | \ z_1, ..., z_{j-1})}_{\text{Time-dependent factors}}\Big] \cdot \underbrace{\prod_{j=1}^{n+t}\Big[p(y_j| x_{j} )\Big]}_{\text{Time-invariant factor}}.
    \label{eq:mfcs_f_factorization}
\end{align}
We focus on \textit{time} dependence and invariance here only because we assume that the data indices indicate timesteps; in general, Step 2 aims to separate terms that depend on permutations of the data indices from those that are invariant to such permutations (if any). Lastly, for Step 3, we use our factorization of $f$ to simplify the CP calibration and test point weights. To do so, we slightly modify our prior notation by letting $E_z^{(t)}$ denote the event $\{Z_1, ..., Z_{n+t}\} = \{z_1, ..., z_{n+t}\}$; then, plugging Eq. \eqref{eq:mfcs_f_factorization} into the analog of 
Eq. \eqref{eq:summary_perm_test} for test point $n+t$ yields 
\begin{align}\label{eq:mfcs_weights_exact}
    \mathbb{P}\{&Z_{n+t}=z_i\mid E_z^{(t)}\} =  \frac{\sum_{\sigma:\sigma(n+t)=i}f(z_{\sigma(1)}, ..., z_{\sigma(n+t)})}{\sum_{\sigma}f(z_{\sigma(1)}, ..., z_{\sigma(n+t)})} \nonumber \\
    = & \frac{\sum_{\sigma:\sigma(n+t)=i}\prod_{j=1}^{n+t}p(x_{\sigma(j)} \mid z_{\sigma(1)}, ..., z_{\sigma(j-1)})}{\sum_{\sigma}\prod_{j=1}^{n+t}p(x_{\sigma(j)} \mid z_{\sigma(1)}, ..., z_{\sigma(j-1)})},
\end{align}
after canceling out $\prod_{j=1}^{n+t}p(y_{\sigma(j)}| x_{\sigma(j)})$ from the numerator and denominator, as it is invariant to permutations $\sigma$. 

As a corollary to Theorem \ref{thm:general_CP_def_coverage}, a CP method with weights given by Eq. \eqref{eq:mfcs_weights_exact} has a valid coverage guarantee premised on the MFCS assumptions in Eq. \eqref{eq:multistep_fcs}. We can thus turn to practical estimation. Firstly, in the agent-induced MFCS settings that we focus on in our experimental evaluations (Section \ref{sec:experimental_results}), the epistemic challenge is overcome because the dynamic components of $f$ are known entirely.
That is, the invariant factor corresponding to $Y|X$ cancels in the ratio, and in our practical evaluations the remaining factors $p(x | Z_1, ..., Z_{j-1})$ represent ML-agent-controlled query probability functions at each timestep $j$. Next, we examine computational complexity. Due to the MFCS IID initialization implying $p(x|Z_1, ..., Z_{j-1})=p(x)$ for $j \leq n$, and when $p(x | Z_1, ..., Z_{j-1})$ is computed from an ML model that treats $Z_1, ..., Z_{j-1}$ symmetrically (as in our experiments), 
the exact computation of Eq. \eqref{eq:mfcs_weights_exact} has complexity $\mathcal{O}(\prod_{j=1}^t(n+j))$. Though reduced from $\mathcal{O}((n+t)!)$, the complexity for arbitrary $f$, this still quickly becomes intractable for large $t$. To alleviate this remaining complexity bottleneck, we propose estimating Eq. \eqref{eq:mfcs_weights_exact} by using only the ``highest-order'' terms from the $d$ most recent timesteps. That is, 
for a user-specified estimation depth $d \in \{1, ..., t\}$, we define our ``$d$-step'' estimate of Eq. \eqref{eq:mfcs_weights_exact} as 
\begin{align}\label{eq:def_3rd_order_weights}
    &\widehat{p}_{n+t}^{\ (d)}\{z_{i}  \mid E_z^{(t)}\} = \widehat{p}^{\ (d)}\{Z_{n+t}=z_{i} \mid E_z^{(t)}\}  \\   
    &=\frac{\sum_{\sigma:\sigma(n+t)=i}\prod^{n+t}_{j=\color{blue}n+t+1-d}p(x_{\sigma(j)} \mid z_{\sigma(1)}, ..., z_{\sigma(j-1)})}{\sum_{\sigma}\prod^{n+t}_{j=\color{blue}n+t+1-d}p(x_{\sigma(j)} \mid z_{\sigma(1)}, ..., z_{\sigma(j-1)})}, \nonumber
\end{align}
which reduces the complexity to $\mathcal{O}(\prod_{j=t+1-d}^t(n+j))$, a tractable polynomial when $d$ is small. 
Appendix \ref{subsec:mfcs_algo_pseudo_code} provides a derivation for our specific MFCS CP algorithms, which use a recursive implementation for further efficiency.

\label{subsec:CP_MFCS_presentation}


\begin{figure*}[!htb]
\centering
    \begin{subfigure}{0.85\textwidth}\includegraphics[width=\textwidth]{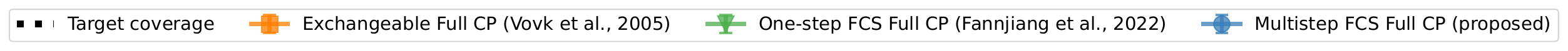}
    \end{subfigure}
    \\
    \hfill
    \begin{subfigure}{0.3\textwidth}
        \includegraphics[width=\textwidth]{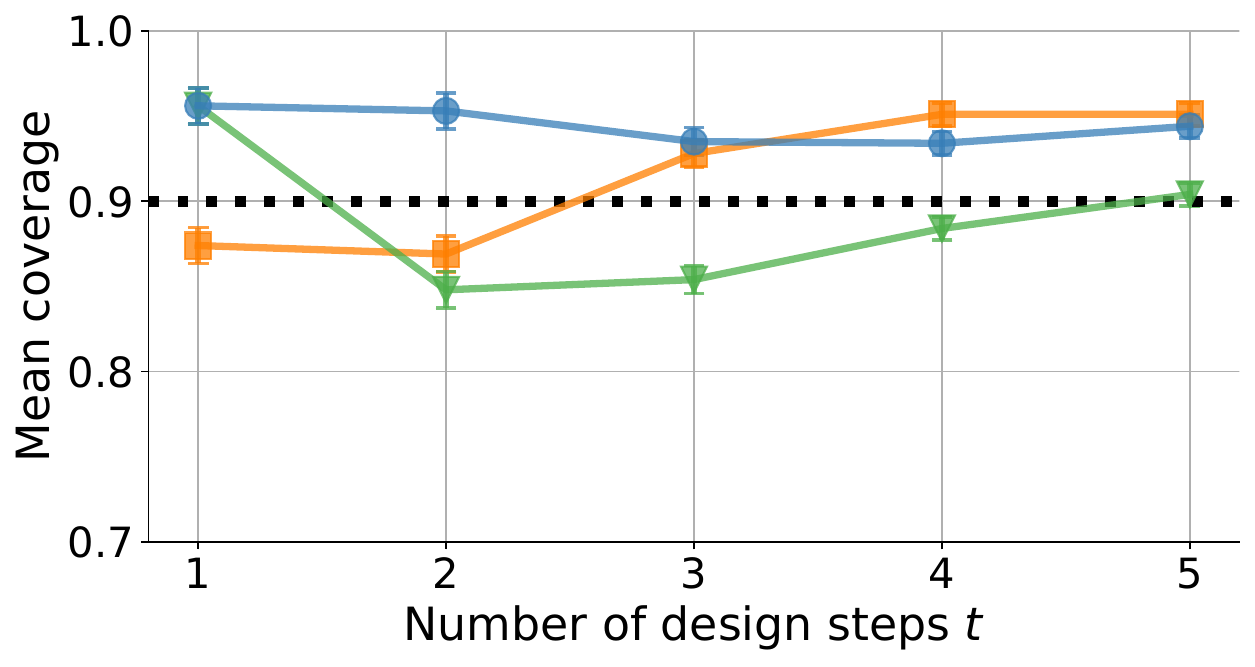}
    \end{subfigure}
    \hfill
    \begin{subfigure}{0.3\textwidth}
        \includegraphics[width=\textwidth]{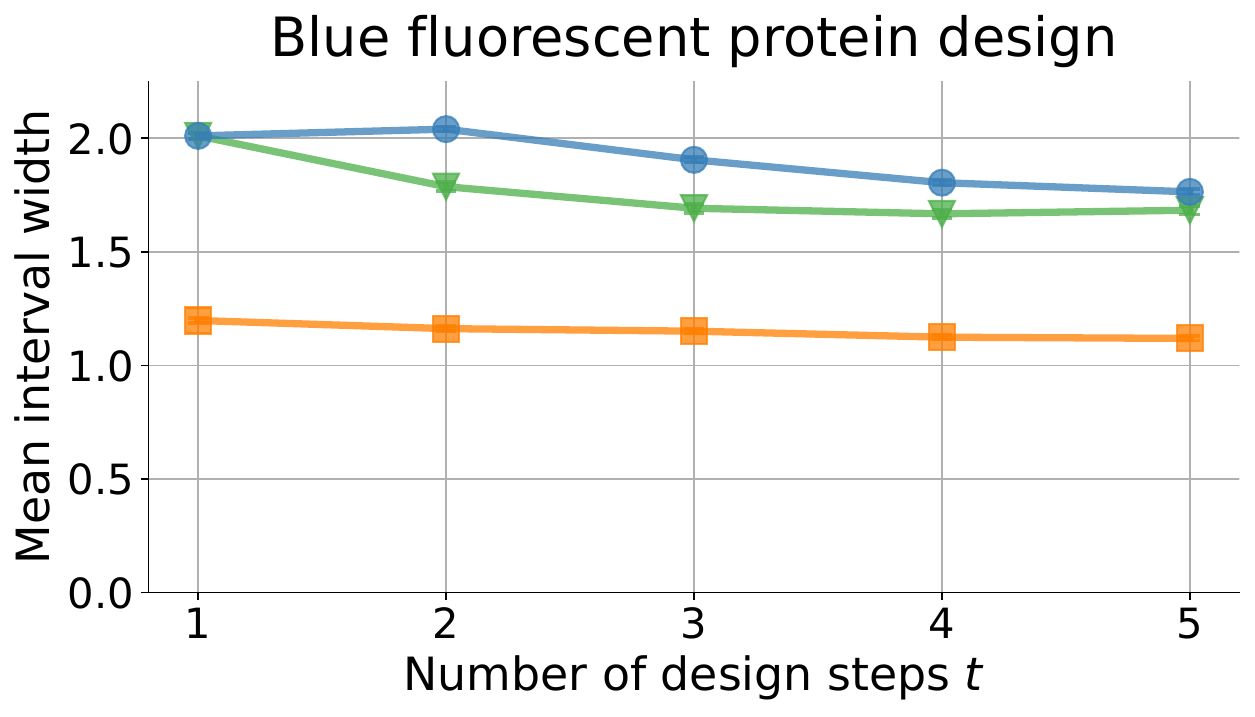}
    \end{subfigure}
    \hfill
    \begin{subfigure}{0.3\textwidth}
        \includegraphics[width=\textwidth]{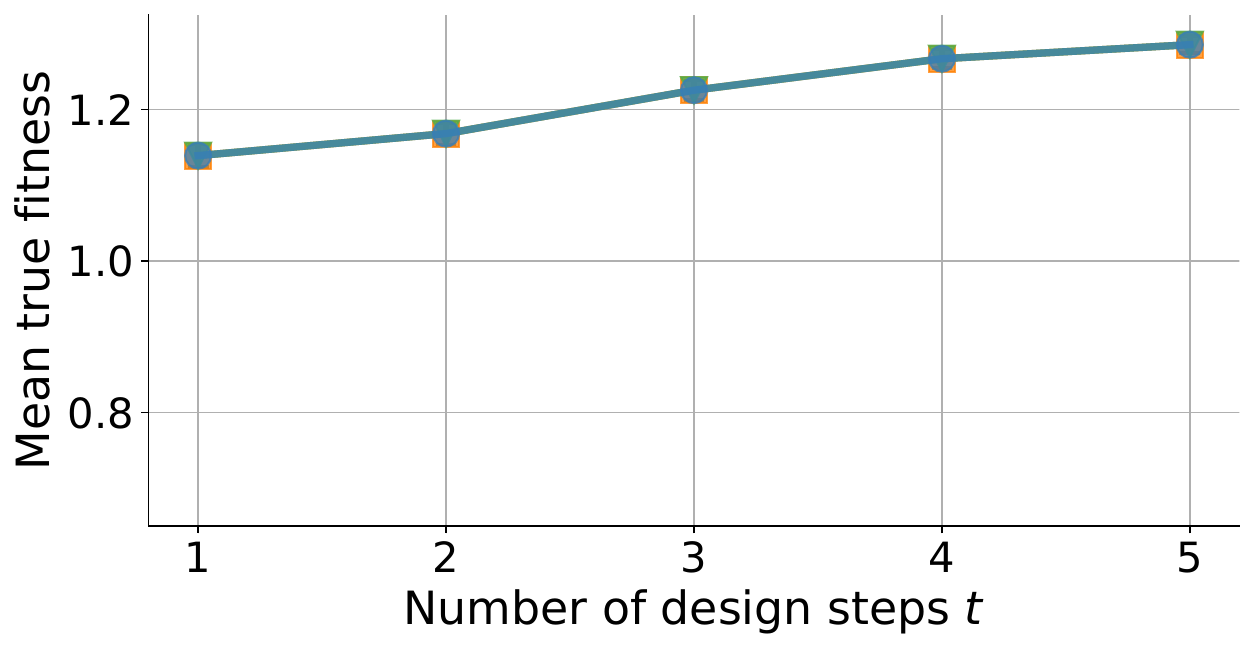}
    \end{subfigure}
    \hfill
    \\
    \hfill
    \begin{subfigure}{0.3\textwidth}
        \includegraphics[width=\textwidth]{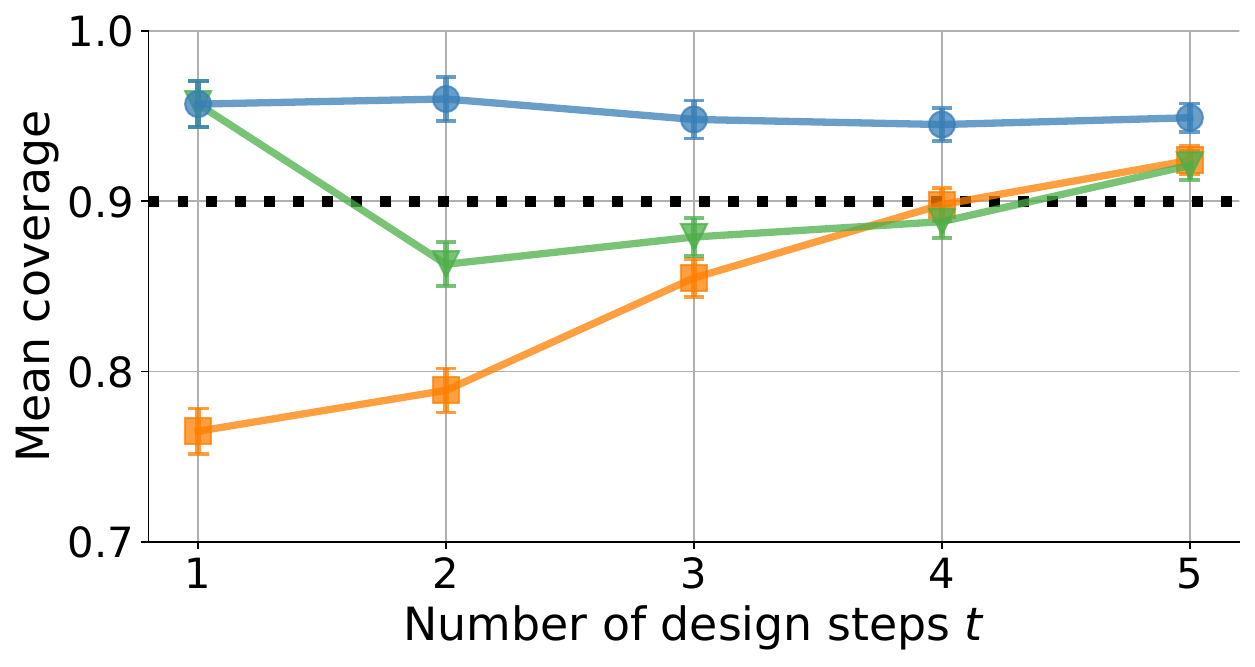}
    \end{subfigure}
    \hfill
    \begin{subfigure}{0.3\textwidth}
        \includegraphics[width=\textwidth]{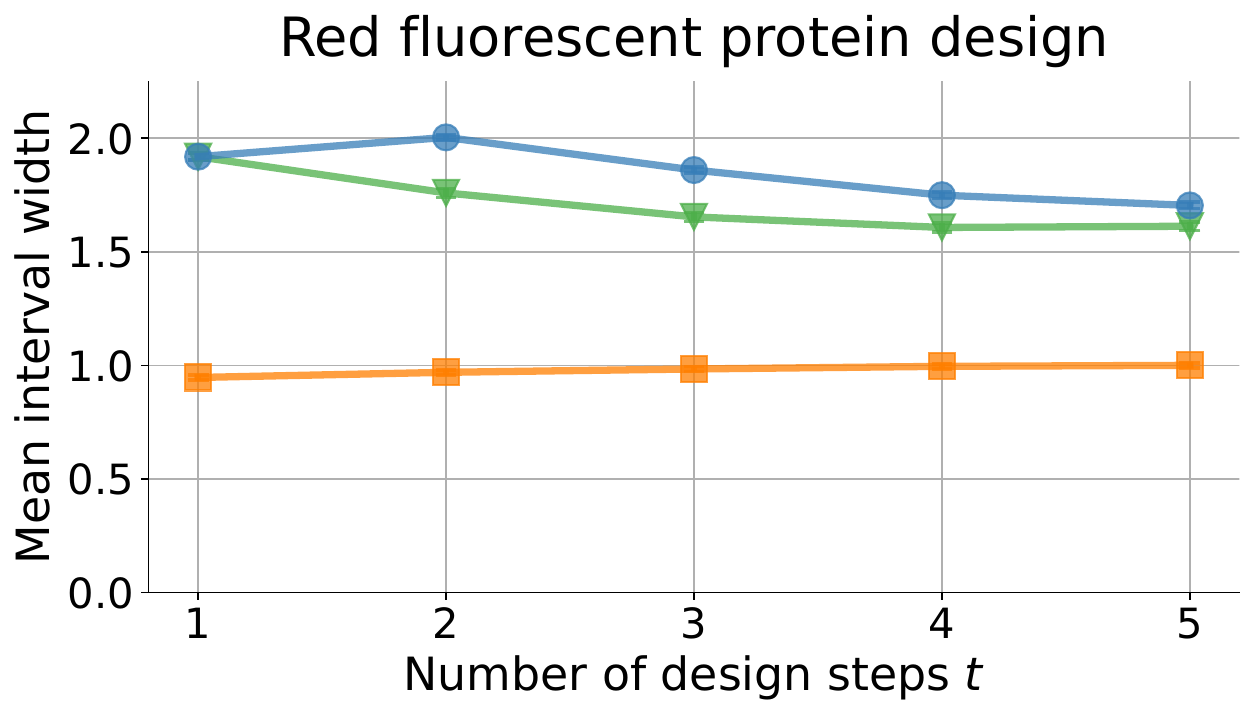}
    \end{subfigure}
    \hfill
    \begin{subfigure}{0.3\textwidth}
        \includegraphics[width=\textwidth]{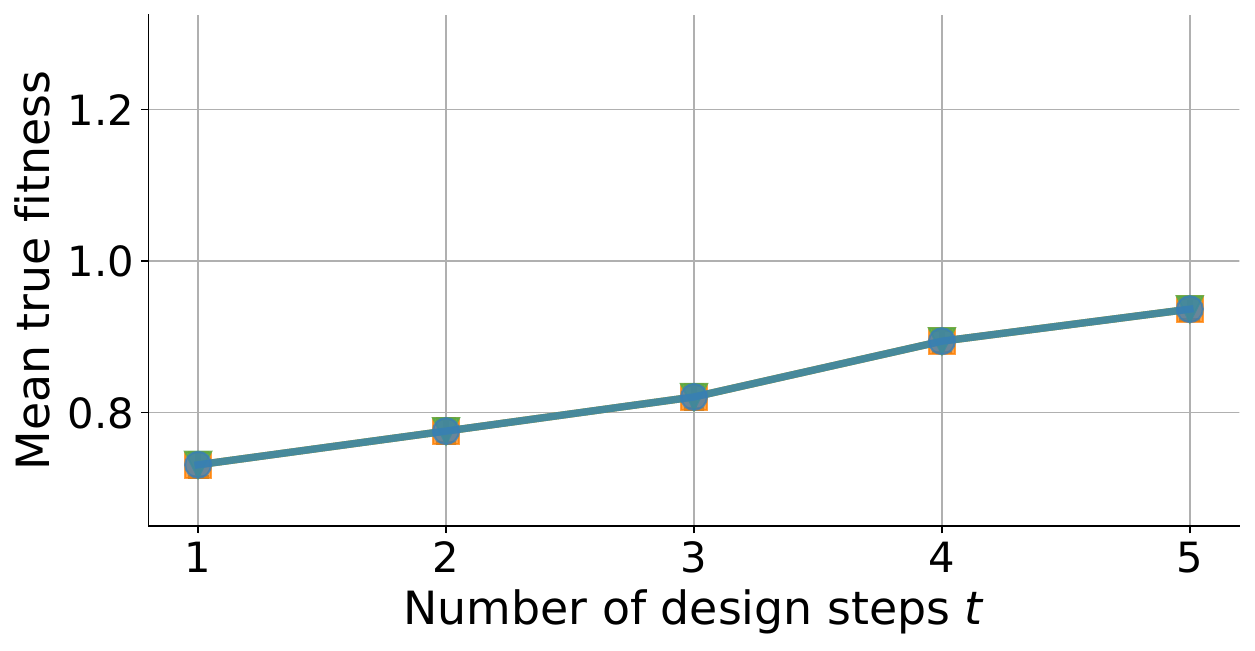}
    \end{subfigure}
    \hfill
    \\
\caption{
    Results for protein design ($n=32$, $\lambda = 8$) with linear ridge regression, comparing the proposed MFCS Full CP method ($d=2$) to standard Full CP and One-Step FCS Full CP. 
    Values are computed over 1,000 random seeds; error bars for mean coverage and mean predicted fitness are standard errors, while error bars for median interval width are interquartile ranges.
    MFCS Full CP maintains coverage where the baselines do not. Its intervals are wider than those of the one-step FCS baseline where the latter loses coverage ($t=2, 3, 4$), but similar where the baseline maintains coverage ($t=1, 5$). (Further experimental details in Appendix  \ref{subsec:black_box_opt_exp_details} Table \ref{tab:fig1}.) 
}
\label{fig:FullCP_MultistepDesignExpts}
\end{figure*}

\textbf{Full versus Split CP Algorithms} The derivation procedure and MFCS methods presented in this section apply to variants of both Full and Split CP. Full CP
efficiently uses the same dataset for both training and calibration (often resulting in more accurate and sharper prediction intervals). Full CP's data efficiency comes at a steep computational cost, with its complexity dominated by the number of predictors that must be \textit{trained} for each candidate label: For an arbitrary predictor, computing the (leave-one-out) Full CP MFCS weights in Eq. \eqref{eq:mfcs_weights_exact} requires training $|\mathcal{Y}|\cdot\prod_{j=1}^t(n+j)$ distinct predictors,  
where $|\mathcal{Y}|$ is the cardinality of the label space.\footnote{In practice for regression, a grid approximates the label space.}
Certain predictor classes (e.g., linear ridge regression) have closed-form solutions as a function of $y$ \citep{vovk2005algorithmic}, which reduces the training cost to 
$\prod_{j=1}^t(n+j)$.

Variants of Split CP \citep{papadopoulos2008inductive} avoid the training cost of Full CP by splitting the available data into a ``proper'' training set used only to fit a single ML model (for each updated training dataset), and a separate holdout ``calibration'' set on which the fitted model computes CP scores 
to construct the CP set. 
Split CP's computational complexity at inference time thus corresponds to the \textit{evaluation} cost of the pre-fitted model on the calibration and test data.








\begin{figure*}[!htb]
\centering
    \begin{subfigure}{0.85\textwidth}\includegraphics[width=\textwidth]{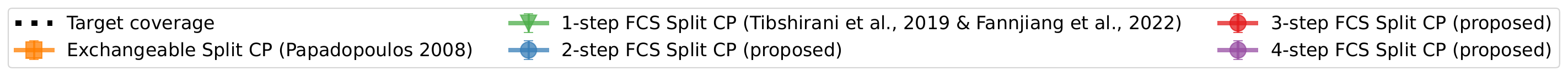}
    \end{subfigure}
    \\
    \hfill
    \begin{subfigure}{0.3\textwidth}
        \includegraphics[width=\textwidth]{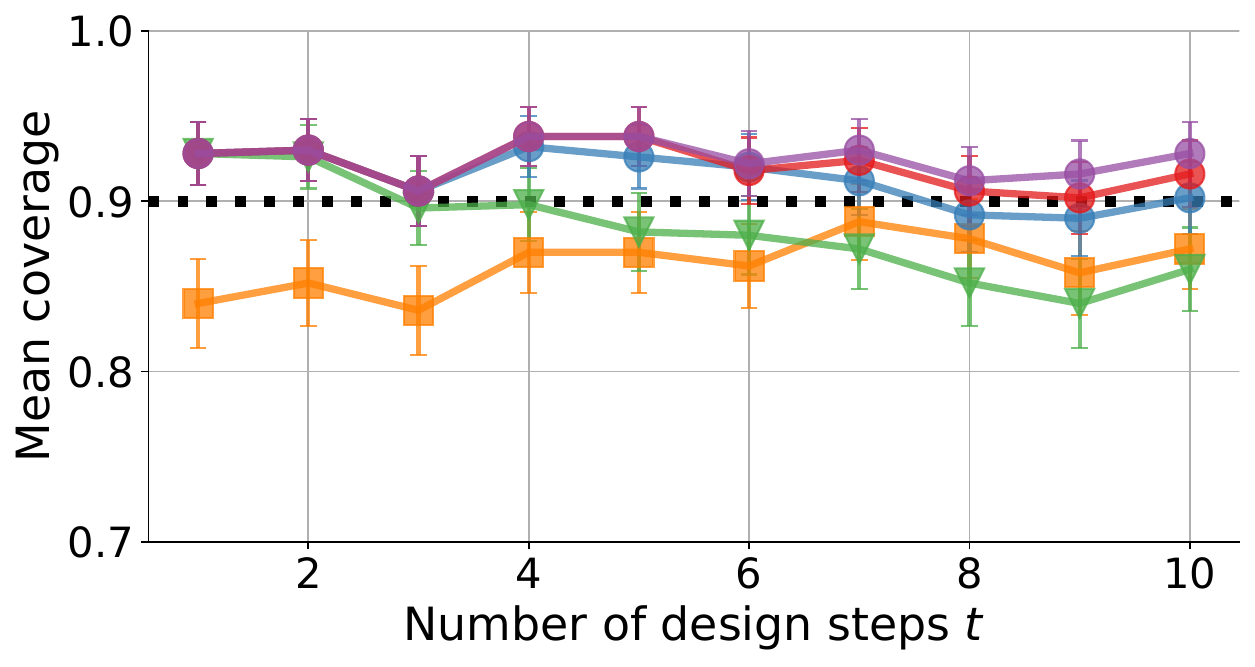}
    \end{subfigure}
    \hfill
    \begin{subfigure}{0.3\textwidth}
        \includegraphics[width=\textwidth]{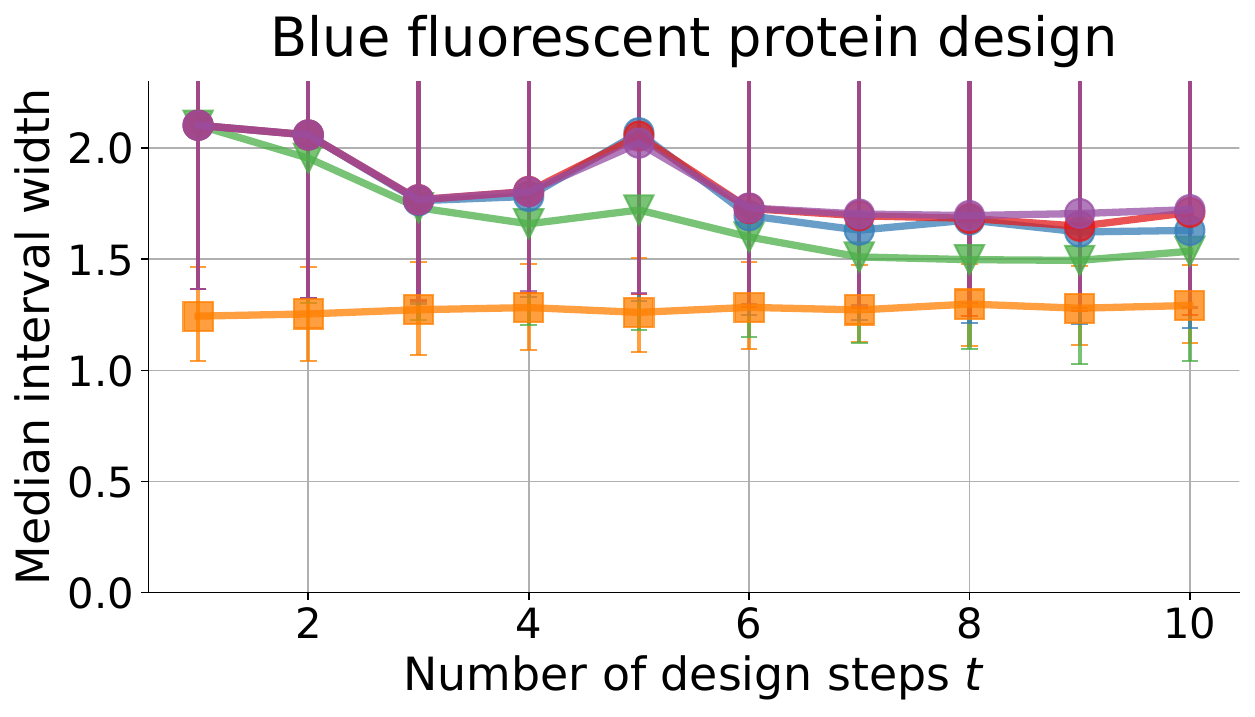}
    \end{subfigure}
    \hfill
    \begin{subfigure}{0.3\textwidth}
        \includegraphics[width=\textwidth]{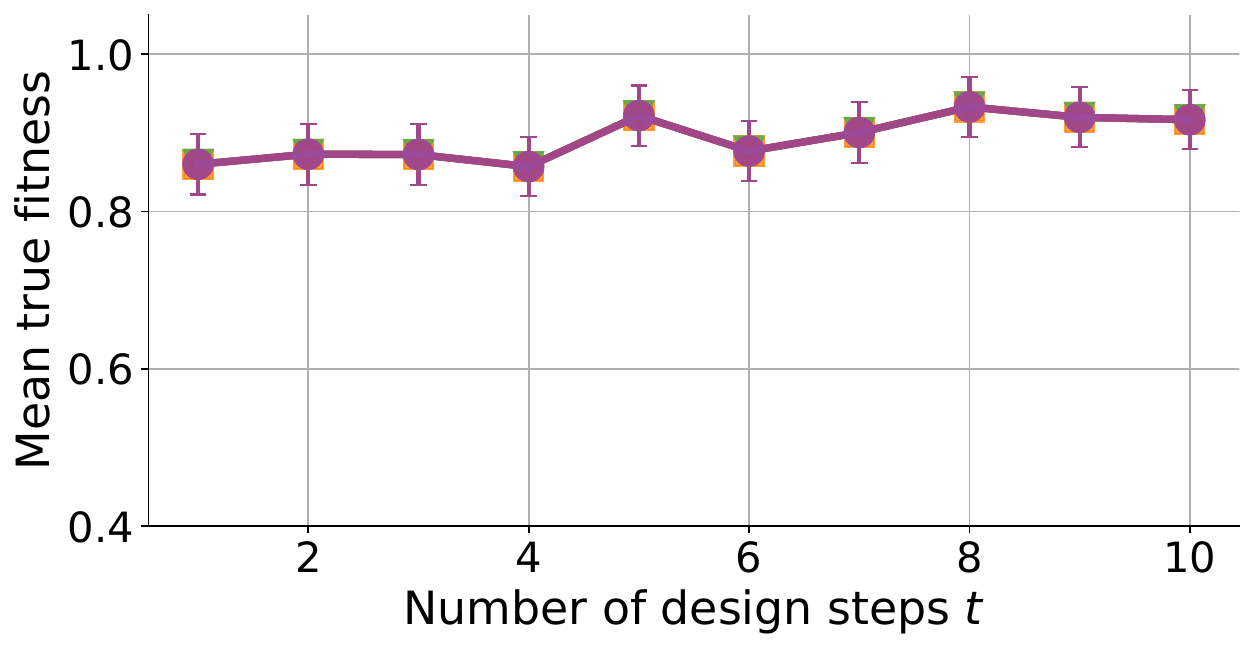}
    \end{subfigure}
    \hfill
    \\
    \hfill
    \begin{subfigure}{0.3\textwidth}
        \includegraphics[width=\textwidth]{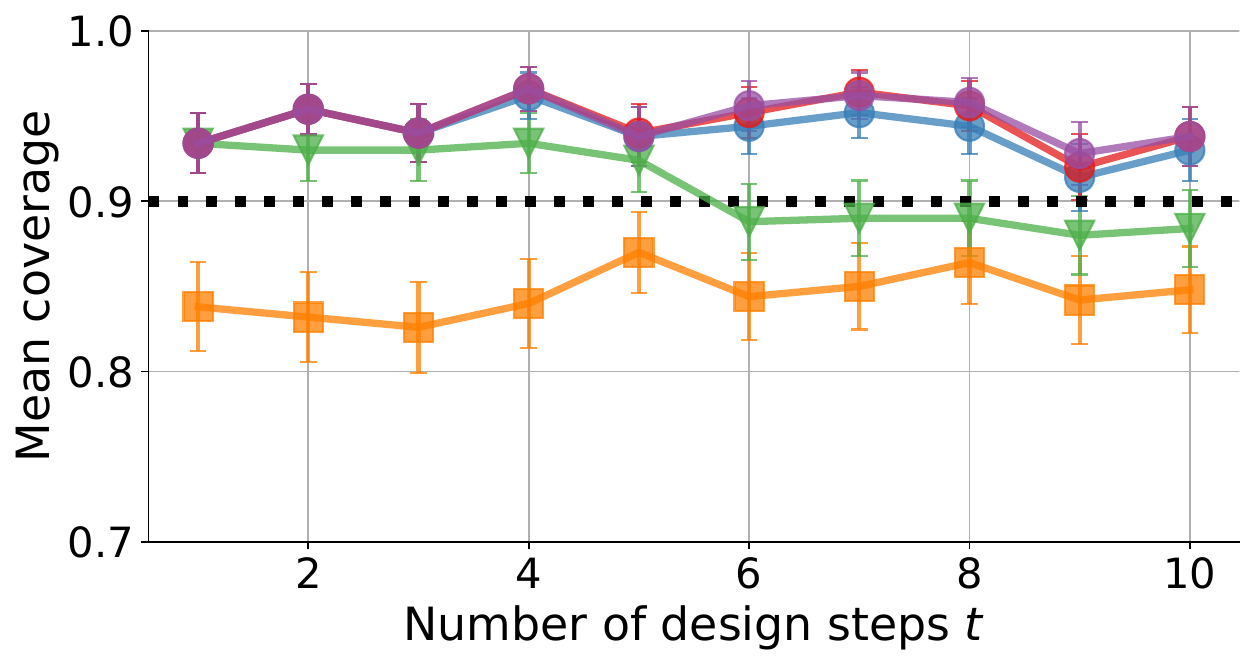}
    \end{subfigure}
    \hfill
    \begin{subfigure}{0.3\textwidth}
        \includegraphics[width=\textwidth]{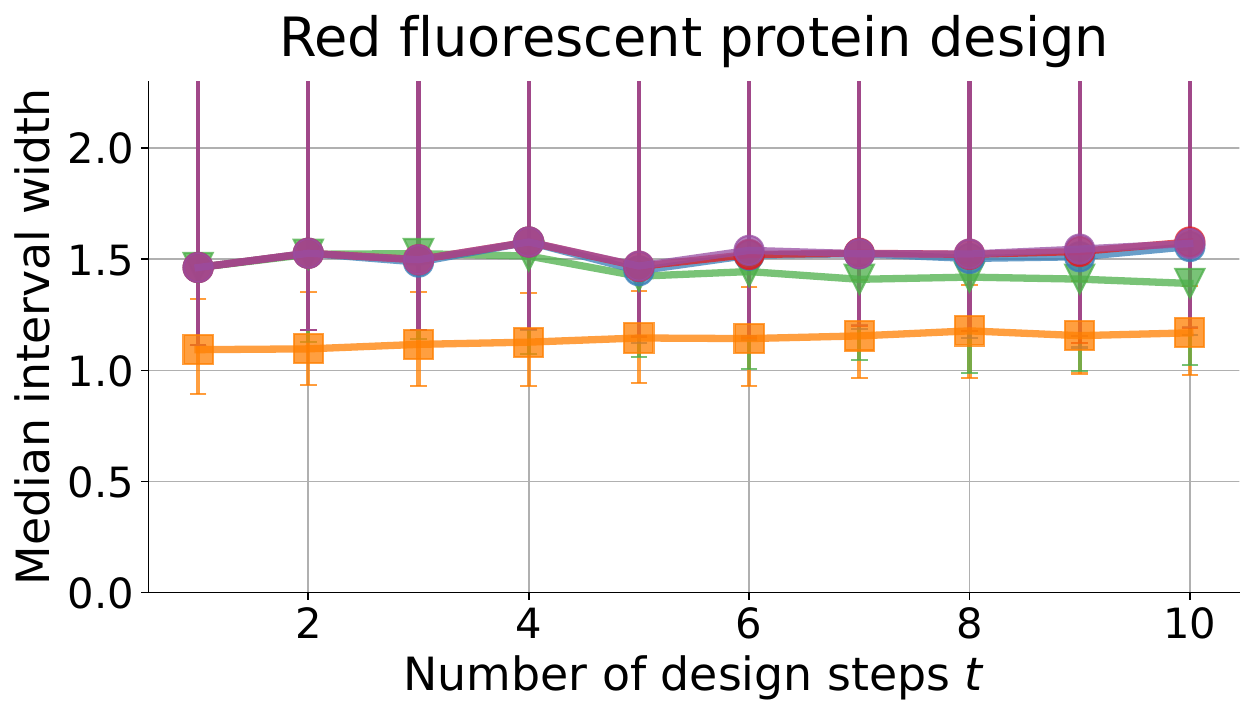}
    \end{subfigure}
    \hfill
    \begin{subfigure}{0.3\textwidth}
        \includegraphics[width=\textwidth]{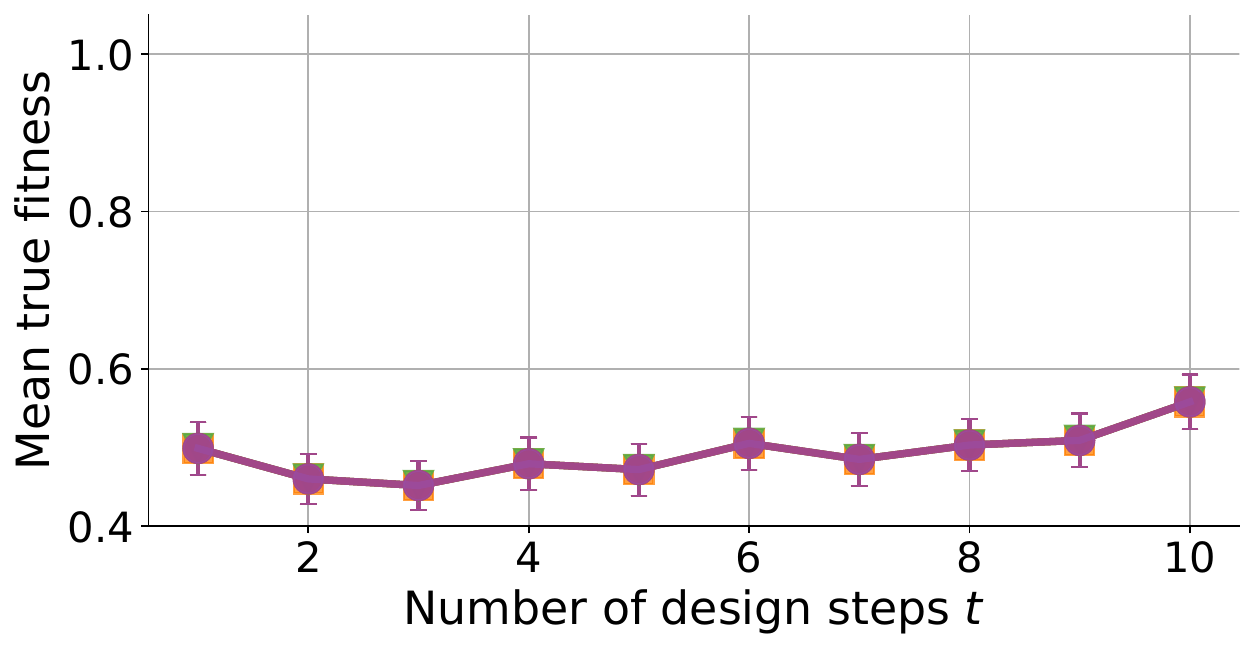}
    \end{subfigure}
    \hfill
    \\
\caption{
    Results for fluorescent protein design with a multi-layer perceptron (MLP) regressor (initial training and calibration data sizes: $n_{\text{train}}=n_{\text{cal}}=32$; $\lambda = 5$), comparing the proposed MFCS Split CP methods to Standard Split CP and the One-Step FCS Split CP baselines. 
    Values are computed over 500 random seeds. Error bars for mean coverage and mean predicted fitness are standard errors; for median interval width they are the interquartile ranges.
    Error bars extending beyond the top of the figure indicate infinite upper quartiles.
    MFCS Split CP maintains coverage where the baselines do not, but its intervals are occasionally very wide, suggesting the proposal distribution is too aggressive for dependably informative uncertainty estimation. (More experimental details in Appendix \ref{subsec:black_box_opt_exp_details} Table \ref{tab:fig2}.)
}
\label{fig:SplitCP_MultistepDesignExpts_NN_mu}
\end{figure*}

\section{Experimental Results}
\label{sec:experimental_results}

We evaluate the performance of our proposed CP methods for ML-agent-induced MFCS on synthetic black-box optimization (specifically protein design) and active learning tasks. All experiments are a regression setting with all CP methods using the absolute residual score $\widehat{\mathcal{S}}(x,y) = |y - \widehat{\mu}_t(x)|$, where $\widehat{\mu}_{t}(x)$ is the prediction of the model fit to $Z_{\text{train}}^{(t)}$, the training data available just prior to time $t$ (for Split CP these are separate from calibration data, for Full CP $Z_{\text{train}}^{(t)}=Z_{1:(n+t-1)}$). Our proposed methods are defined by estimating Eq. \eqref{eq:def_general_conformal_set} with weights as in Eq. \eqref{eq:def_3rd_order_weights}.
The primary desired criterion for reliability is whether its prediction intervals maintain empirical coverage at or above the target coverage level $1-\alpha$. Secondarily---if target coverage is achieved---then smaller predictive sets are more informative (i.e., predicting the full label space $\widehat{\mathcal{C}}_n(x)=\mathcal{Y}$ could trigger deferral to an expert but does not itself convey new insights from the predictor).
In all experiments we compare to Exchangeable CP \citep{vovk2005algorithmic} and One-Step FCS CP \citep{fannjiang2022conformal}, and in the active learning experiments we also compare with Adaptive Conformal Inference (ACI) \citep{gibbs2021adaptive}.
Our proposal distributions follow the form $p(x \mid Z_{\text{train}}^{(t)}) \propto \exp(\lambda \cdot u_{t}(x))$, where $u_t$ is some utility function and $\lambda \geq 0$ is a hyperparameter that controls the magnitude of agent-induced covariate shift.
In all experiments we corrupt labels with Gaussian noise to simulate noisy observations.
Each query point is added to training for full CP experiments, and for split CP, to training or calibration by flipping a fair coin. (See Appendix \ref{sec:exp_details} for details, Appendices \ref{app:sec:adaptive_exploration_results} and \ref{app:sec:additional_expts} for additional experiments.)

\subsection{Multistep Protein Design Experiments}


\textbf{Optimization-Induced MFCS Experimental Details} 
Our protein design procedure is similar to a common approach in the biomolecular engineering literature \citep{biswas2021low, zhu2024optimal} and related works in CP \citep{fannjiang2022conformal, prinster2023jaws}, but extended to a more realistic online setting where protein sequence proposals and model updates are interleaved and dependent \citep{stanton2022accelerating, gruver2023protein, angermueller2023high}.
All methods start with an IID initialization by sampling $n$ points uniformly at random from a combinatorially complete library \citep{poelwijk2019learning}, splitting the data equally into training and calibration sets if using split CP. 
(See Appendix \ref{subsec:black_box_opt_exp_details} for details.)
We evaluate the mean coverage and median intervals width as a function of $t$ (the number of optimization steps past initialization), fitting the regressor to training data observed prior to time $t$, and sampling one query point with replacement (with added measurement noise) from the library, each with probability $p(x\mid Z_{\text{train}}^{(t)}) \propto \exp(\lambda \cdot \widehat{\mu}_t(x))$.

 



\textbf{Full CP Results}
Figure \ref{fig:FullCP_MultistepDesignExpts} reports the results for our proposed MFCS Full CP method (blue circles) with hyperparameters reported in the caption. 
On both the blue and red fluorescent protein design datasets, and on all five evaluated design steps, our proposed MFCS Full CP method maintains coverage above the target level even when both baselines lose coverage below this level.
The MFCS method's coverage is maintained by increasing the conservativeness (size) of its prediction intervals relative to the One-Step baseline after the first design step.
The superimposed fitness curves on the right verify that differences in coverage can be attributed to the different CP methods, not the data collected.

\begin{figure*}[!htb]
    \centering
    \begin{subfigure}{0.85\textwidth}\includegraphics[width=\textwidth]{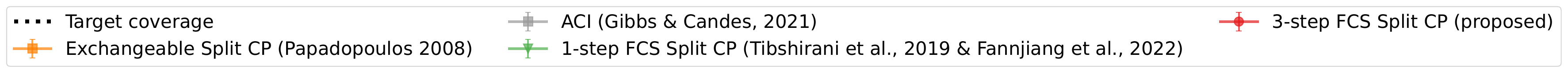}
    \end{subfigure}
    \\
    \hfill
    \begin{subfigure}{0.3\textwidth}
        \includegraphics[width=\textwidth]{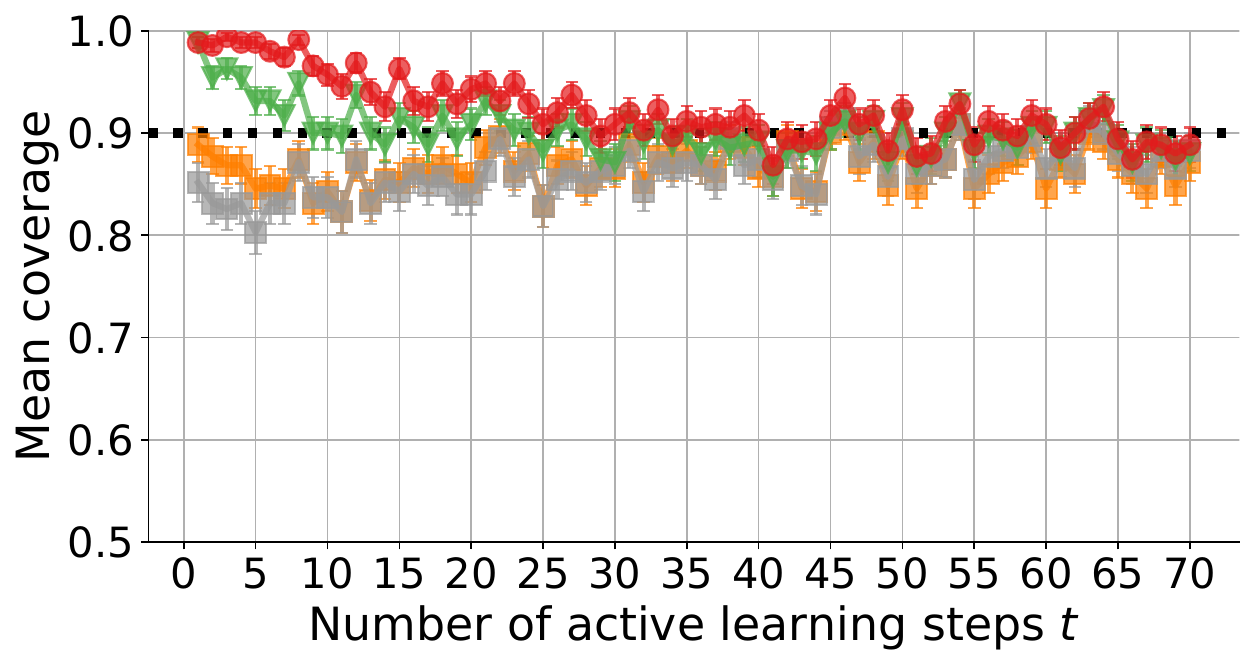}
    \end{subfigure}
    \hfill
    \begin{subfigure}{0.3\textwidth}
        \includegraphics[width=\textwidth]{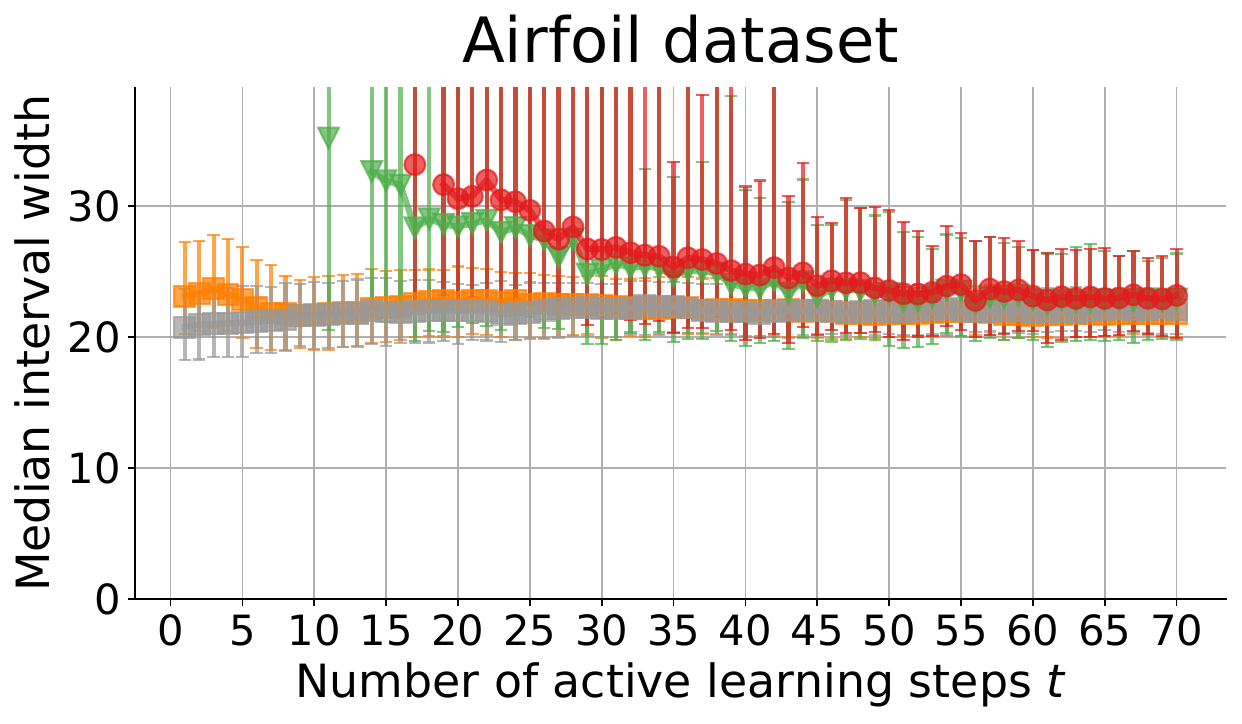}
    \end{subfigure}
    \hfill
    \begin{subfigure}{0.3\textwidth}
        \includegraphics[width=\textwidth]{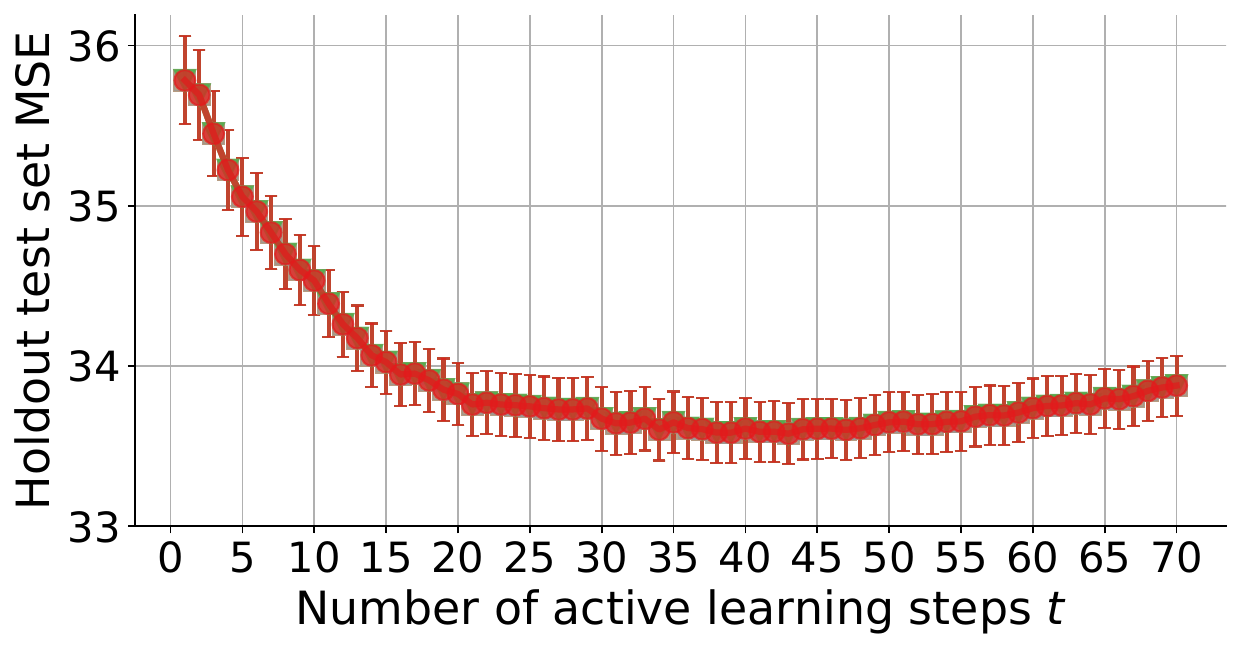}
    \end{subfigure}
    \hfill
    \\
    \hfill
    \begin{subfigure}{0.3\textwidth}
        \includegraphics[width=\textwidth]{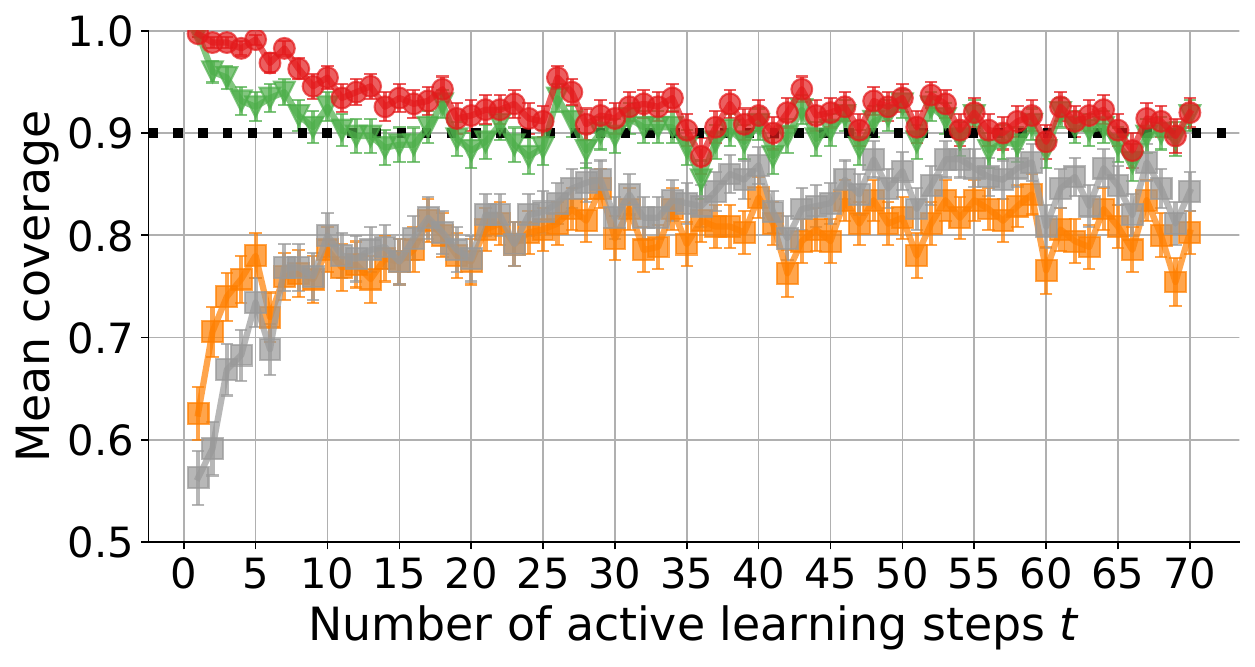}
    \end{subfigure}
    \hfill
    \begin{subfigure}{0.3\textwidth}
        \includegraphics[width=\textwidth]{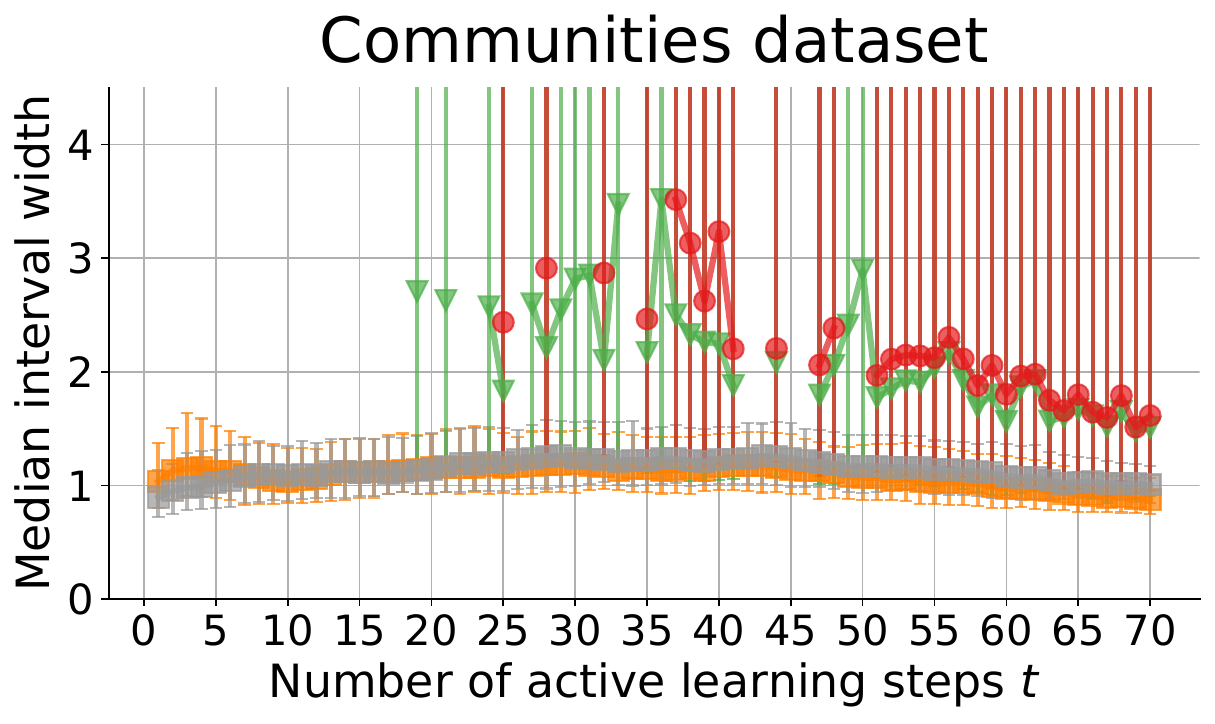}
    \end{subfigure}
    \hfill
    \begin{subfigure}{0.3\textwidth}
        \includegraphics[width=\textwidth]{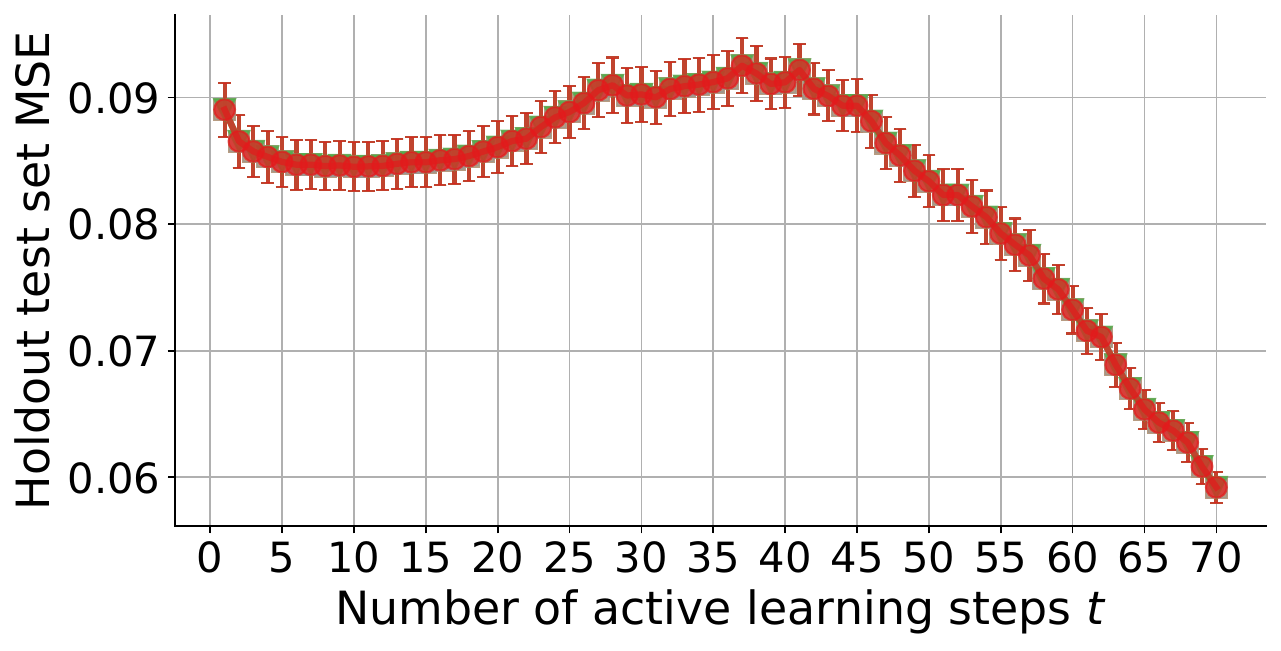}
    \end{subfigure}
    \hfill
    \\
    \hfill
    \begin{subfigure}{0.3\textwidth}
        \includegraphics[width=\textwidth]{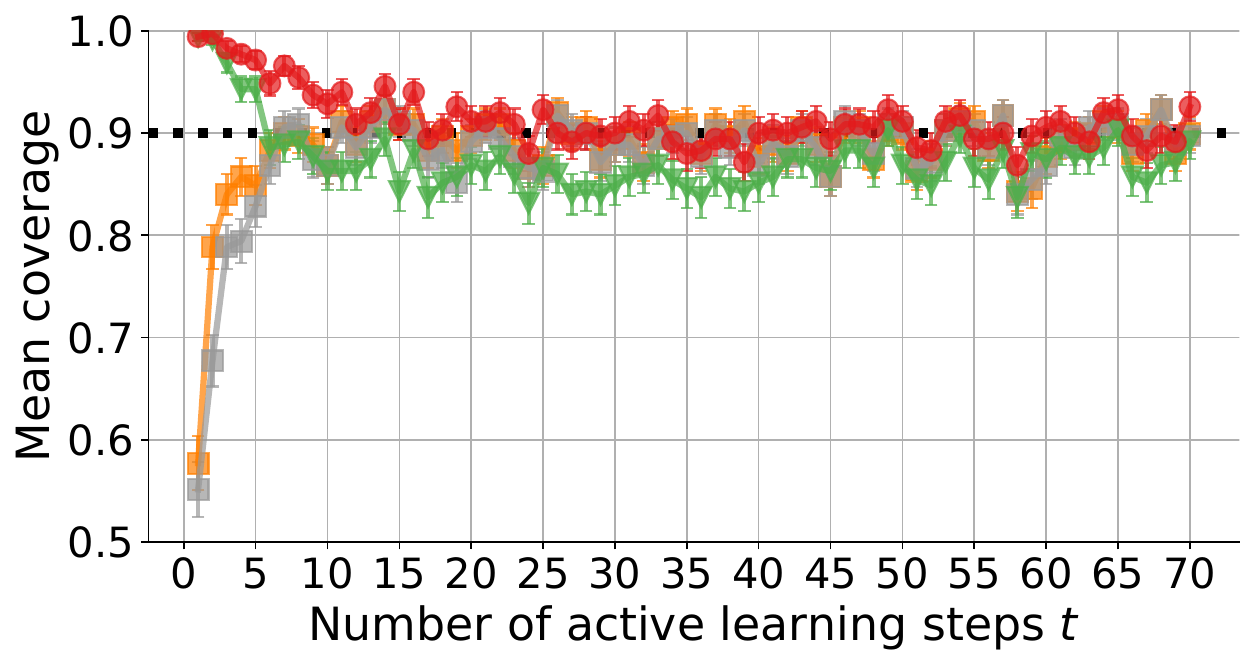}
    \end{subfigure}
    \hfill
    \begin{subfigure}{0.3\textwidth}
        \includegraphics[width=\textwidth]{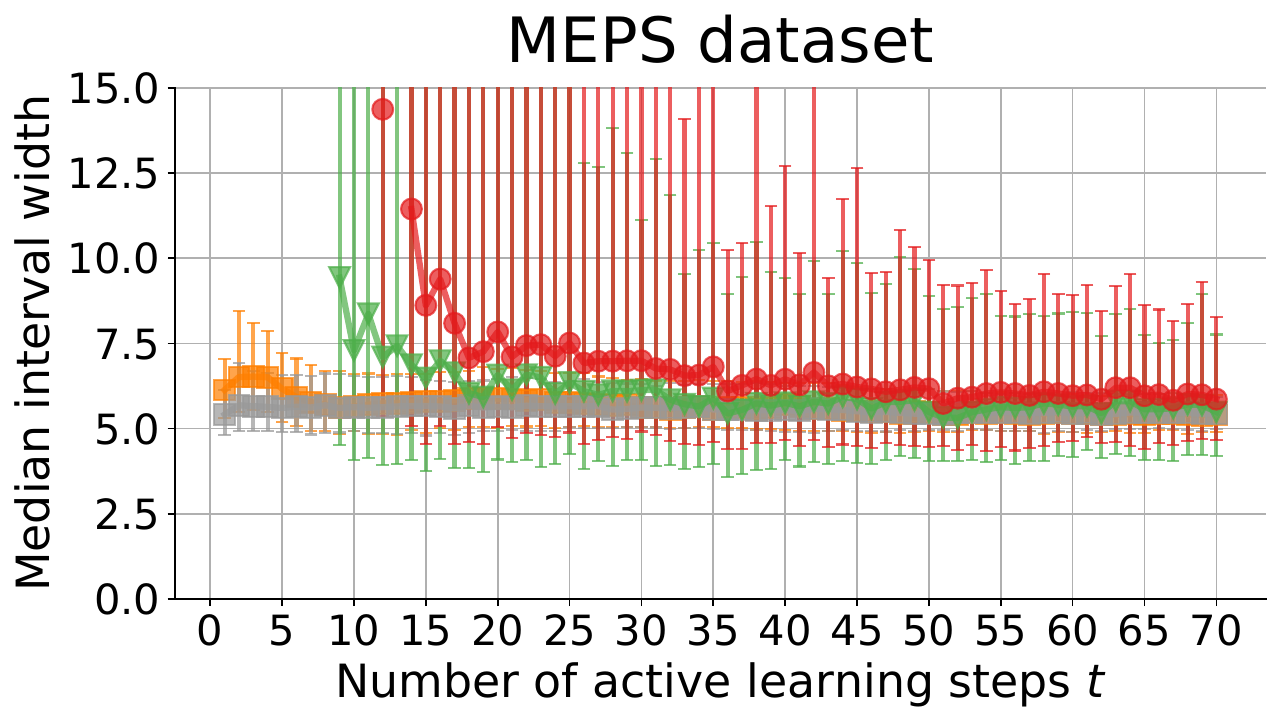}
    \end{subfigure}
    \hfill
    \begin{subfigure}{0.3\textwidth}
        \includegraphics[width=\textwidth]{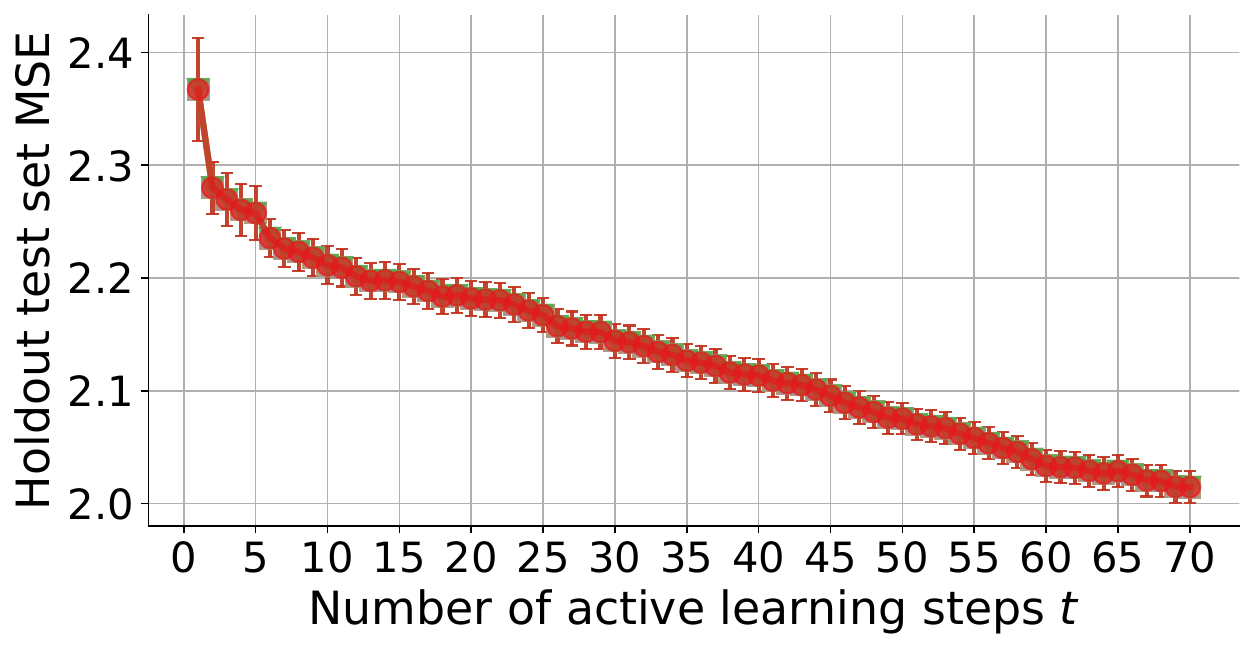}
    \end{subfigure}
    \hfill
    \\
    \hfill
    \begin{subfigure}{0.3\textwidth}
        \includegraphics[width=\textwidth]{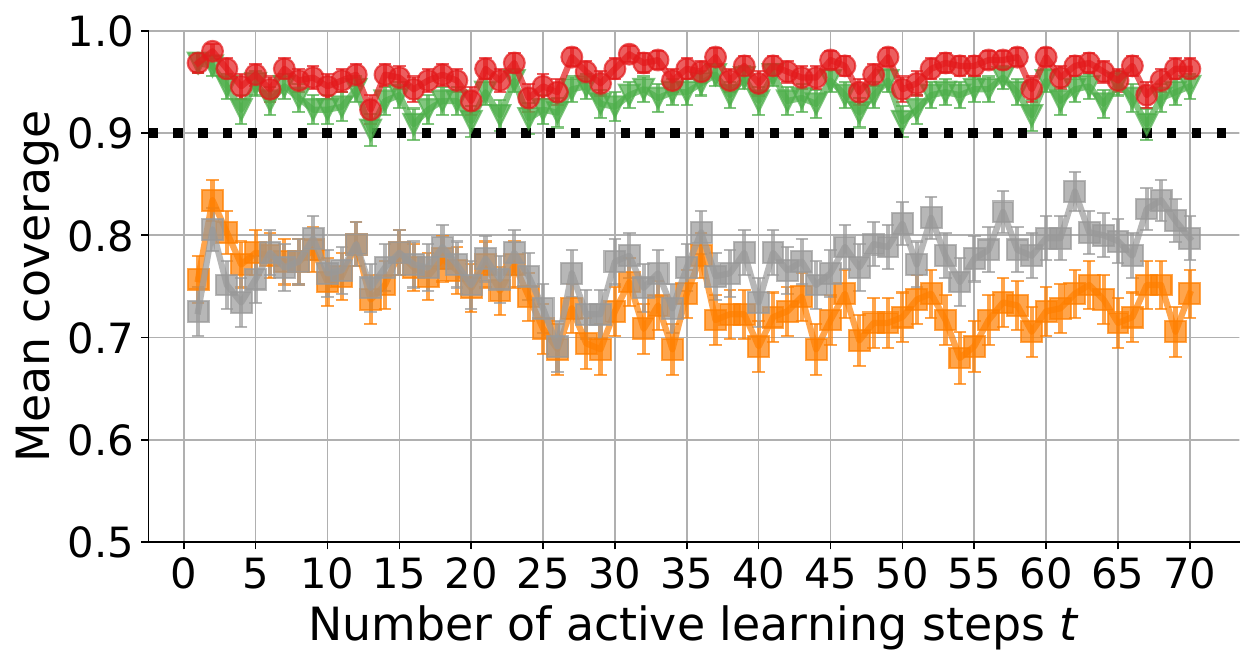}
    \end{subfigure}
    \hfill
    \begin{subfigure}{0.3\textwidth}
        \includegraphics[width=\textwidth]{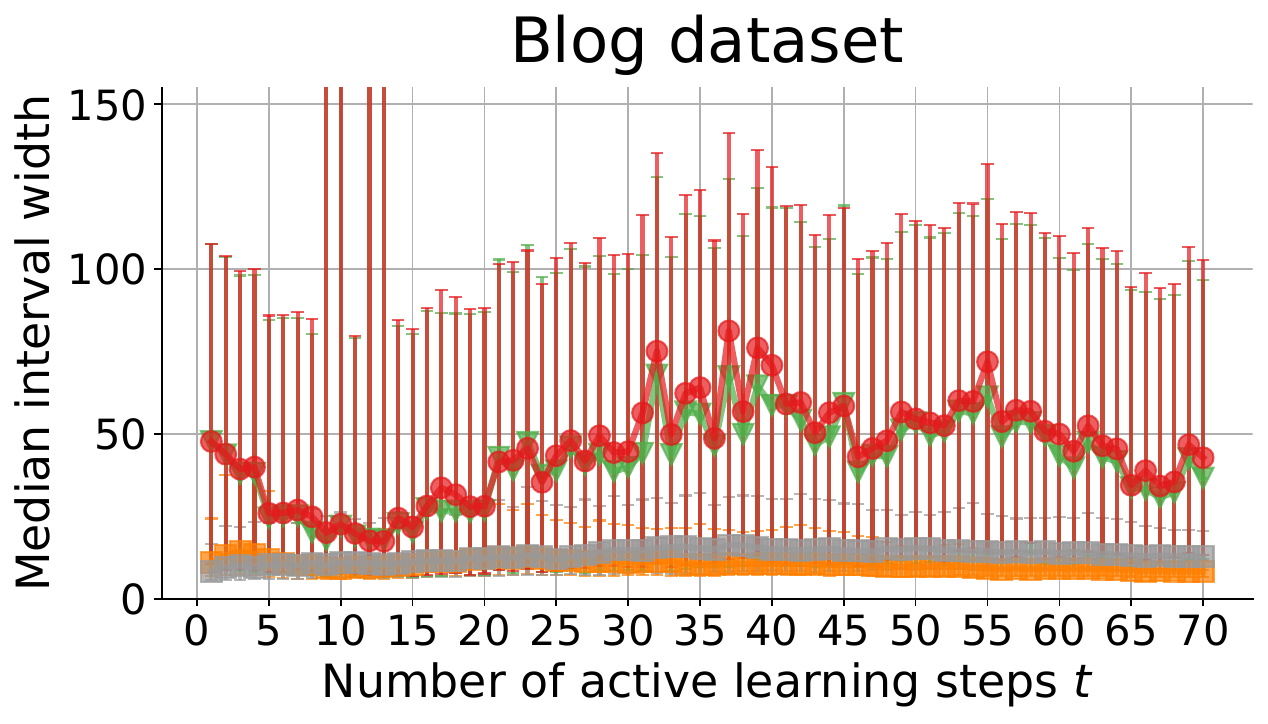}
    \end{subfigure}
    \hfill
    \begin{subfigure}{0.3\textwidth}
        \includegraphics[width=\textwidth]{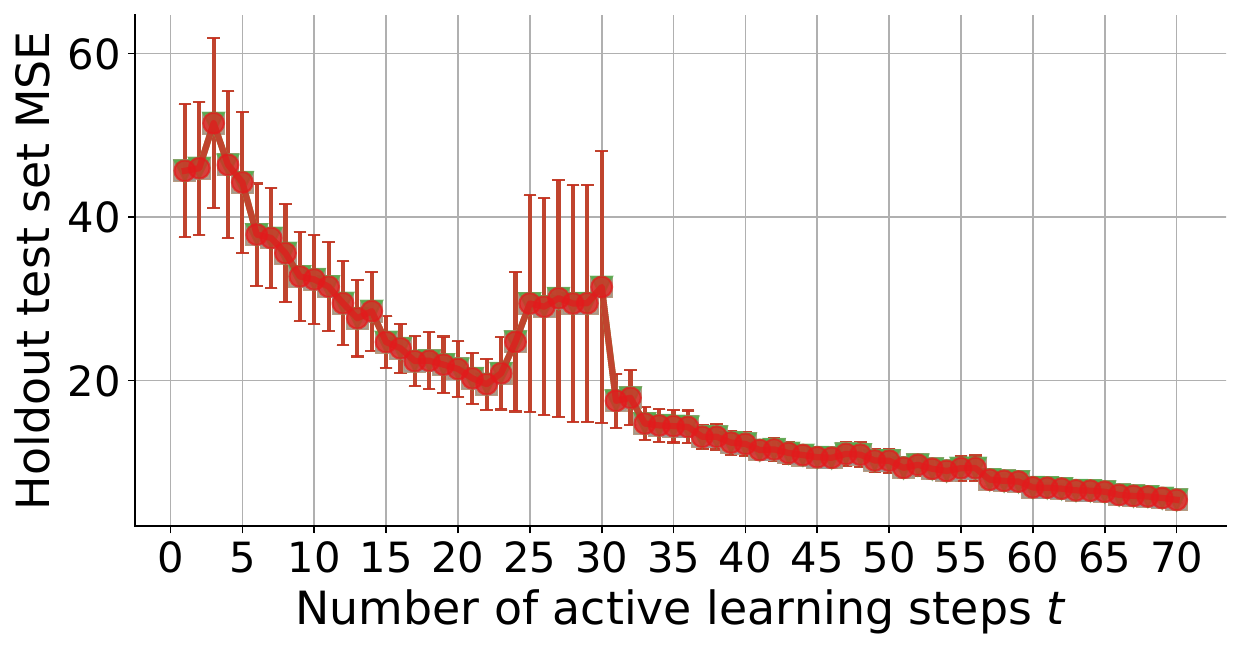}
    \end{subfigure}
    \hfill
\caption{
    Active learning experiments of proposed MFCS Split CP method for $d=3$ (red) compared to baselines of unweighted Split CP (orange), One-Step Split CP (green), and ACI (gray). 
    Y-axes represent mean coverage, median interval width, and mean squared error on a holdout test set to track the accuracy of the base predictor. X-axes correspond to the number of active learning query steps, with each query based on posterior variance of a GP regressor.
    All values are computed over 350 distinct randomm seeds (full experimental details in Appendix \ref{subsec:active_learning_exp_details}). The proposed MFCS CP method maintains target coverage over a long time horizon even where baselines do not.
    }
\label{fig:SplitCP_ActiveLearningExpts}
\end{figure*}

\textbf{Split CP Results with Neural Network Regressor}
The results in Figure \ref{fig:SplitCP_MultistepDesignExpts_NN_mu} use an MLP regressor 
on the protein design datasets to evaluate our MFCS Split CP method. We moreover compare the effect of several estimation depths $d$ for estimating the MFCS weights as in Eq. \eqref{eq:def_3rd_order_weights}. 
For weight estimation depths $d \in \{3, 4\}$, our proposed MFCS CP methods
maintain target coverage across all 10 evaluated design steps, for both the blue and red fluorescent protein design datasets, even when the baselines break down. 
Increasing the estimation depth $d$ tends to improve coverage at later design steps, for instance relative to the One-Step FCS method, apparently with diminishing returns (the red 3-step and purple 4-step methods have nearly overlapping performance). Coverage is maintained by the MFCS Split CP methods by slightly increasing the size of the intervals relative to the one-step baseline at later design steps. However, both the MFCS Split CP method and the One-Step FCS Split CP method occasionally have noninformative (infinite width) intervals, which could be unacceptable in some application settings. 
In Appendix \ref{app:sec:adaptive_exploration_results}, we show that adaptively bounding the ML agent's query probability function can achieve dependably informative (finite) interval widths without losing coverage, by restricting the ML agent's feedback shift.

\subsection{Active Learning-Induced MFCS}

Now we turn to an active learning setting where MFCS is induced by a max-entropy data acquisition strategy, where the ML agent selects new queries for annotation by maximizing its (Bayesian) model uncertainty.
In this experiment we use a Gaussian process (GP) regressor as the predictor and take $u_t(x) = \mathbb{V}(\mu(x) \mid Z_{\text{train}}^{(t)})$ (i.e. the GP posterior variance).
See Appendix \ref{subsec:active_learning_exp_details} for model and dataset details.

\textbf{Active Learning Experimental Results} Figure \ref{fig:SplitCP_ActiveLearningExpts} demonstrates the performance of our proposed MFCS Split Conformal methods for several depths of weight approximation.
Whereas the baseline methods lose coverage, the proposed MFCS Split CP methods maintain target coverage for all datasets and across the 70 evaluated active learning steps. 
As the active learning process reaches later iterations where more labeled data is available for both training and calibration, the proposed MFCS Split CP methods are still able to maintain target coverage with sharper intervals. 


\section{Discussion}
\label{sec:discussion}

In this paper we have demonstrated how weighted conformal prediction can theoretically generalize to any joint data distribution.
Exchangeability conditions are only necessary to rigorously specify when \textit{a specific practical algorithm} (e.g., standard CP) is equivalent to the more general formulation, which is always valid but may not be computable.
Promising future directions include further addressing practical bottlenecks, extending our general marginal coverage guarantees to other loss functions \citep{angelopoulos2022conformal}, and identifying minimal assumptions for strong conditional coverage in practice \citep{barber2021predictive, gibbs2023conformal}.

We have also provided practical CP algorithms for multiple rounds of agent-induced, data-dependent covariate shifts, which maintains coverage on data collected by active learning and black-box optimization agents.
Finally, we have shown that these agents can attain both empirically valid coverage and informative intervals by restricting the degree of shift introduced (Appendix \ref{app:sec:adaptive_exploration_results}).
These findings bring us to the control-consistency dilemma at the heart of statistical decision-making.
To control is to change, but statistics requires consistency.
Like the explore-exploit dilemma, the right tradeoff depends on the situation at hand.
As machine learning agents assume increasing levels of responsibility in gravely consequential settings, it is more important than ever that we face the control-consistency dilemma head-on.



\section*{Code Availability} Repository with code for all experiments: \url{https://github.com/drewprinster/conformal-mfcs}

\section*{Impact Statement}
\label{sec:impact}
This paper presents work whose goal is to advance the field of Machine Learning. There are many potential societal consequences of our work which we do not think must be specifically highlighted here. However, we note that our contributions are specifically aimed at improving the reliability of methods for uncertainty quantification of machine learning predictions, which is often critical for managing risks and improving the trustworthiness of such systems.




\section*{Acknowledgements}

D.P. was partially funded by a fellowship from the Johns Hopkins University (JHU) Mathematical Institute for Data Science (MINDS); D.P. and S. Saria were partially funded by National Science Foundation (NSF) grant \#1840088; D.P., A.L., and S. Saria were partially funded by the Gordon and Betty Moore Foundation grant \#12128. A.L. was also partially funded by a JHU Discovery Award and a JHU Institute for Assured Autonomy (IAA) seed grant. The authors would like to thank Anastasios Angelopoulos for early discussions that later helped inform this work and Clara Fannjiang for helpful feedback on a manuscript draft.




\bibliography{bibliography}
\bibliographystyle{icml2024}

\newpage
\appendix
\onecolumn
\section{Main Result Proof and Details}

\subsection{Proof of Theorem \ref{thm:general_CP_def_coverage}}

\label{sec:general_cp_coverage_proof}

In this section we present the proof for our main result given in Theorem \eqref{thm:general_CP_def_coverage}. Our proof builds on ideas in \citet{tibshirani2019conformal}'s alternate proof of their Lemma 1 (for conformal prediction assuming exchangeable data) and their proof of their Lemma 3 (for conformal prediction assuming weighted exchangeable data). However, those proofs make key substitutions using the definitions of exchangeability and weighted exchangeability respectively, which thereby yields guarantees that are premised on those definitions. In contrast, the proof we present here makes no such substitution, and thus holds for an arbitrary joint probability density function (PDF)\footnote{More generally, $f$ can be a valid Radon-Nikodym derivative with respect to an arbitrary base measure: in the discrete case $f$ is a probability mass function (PMF); it could also be a mixture of PDFs and PMFs.} $f$. We will call attention to the key step where our proof critically differs from those in \citet{tibshirani2019conformal} in substance, and even for steps that we have in common with \citet{tibshirani2019conformal} we aim to provide more explicit derivation explanations for increased accessibility (e.g., where explicit step justifications may be omitted in \citet{tibshirani2019conformal}). 


As we sketch in Section \ref{subsec:key_insight_role_of_definitions} of our main text, our proof corresponds to a view of CP as an inverted permutation test. Let $E_z$ denote the event of observing the unordered set\footnote{As in \citet{tibshirani2019conformal} we assume for simplicity that the score values $V_1, ..., V_{n+1}$ are distinct almost surely, though the result also holds in more general cases (but with more complicated notation involving randomness from uniform random variables to break ties).} of our data values. In particular, $E_z$ denotes the event that $\{Z_1, ..., Z_{n+1}\} = \{z_1, ..., z_{n+1}\}$, but importantly, the lack of ordering means that this does \textit{not} imply that $Z_i=z_i$ for all $i\in\{1, ..., n+1\}$; we do not yet know whether $z_1$ is the value obtained by $Z_1$, or by $Z_2$, or by $Z_{n+1}$, and so on. 
Similarly, let $E_v$ denote the event $\{V_1, ..., V_{n+1}\} = \{v_1, ..., v_{n+1}\}$ for the scores.
Also, recall the simplifying notation $v_i=\mathcal{\widehat{S}}(z_i)$ for the value computed by our fitted score function $\mathcal{\widehat{S}}$ with input $z_i$ (where $\mathcal{\widehat{S}}(z_i)=\mathcal{S}(z_i, z_{1:(n+1)})$, that is $\mathcal{\widehat{S}}$ is fit on the set $z_{1:(n+1)}=\{z_1, ..., z_{n+1}\}$, the second input to $\mathcal{S}$). 

With this notation, we consider the probability that, conditioned on $E_v$, the value corresponding to the random variable $V_{n+1}$ was actually $v_i$. That is, we consider
\begin{align*}
    \mathbb{P}\{V_{n+1}=v_i\ | \ E_v\} = \mathbb{P}\{Z_{n+1}=z_i\ | \ E_z\},\ i\in \{1, ..., n+1\},
\end{align*}
which follows since, conditioned on $E_z$, there are an equal number of scores $v_1, ..., v_{n+1}$ as data points $z_1, ..., z_{n+1}$, thus implying that the fitted (symmetric) score function $\mathcal{\widehat{S}}$ induces a bijection between the discrete set of distinct data observations and the discrete set of distinct scores.\footnote{To be thorough and precise, the equation $\mathbb{P}\{V_{n+1}=v_i\ | \ E_v\} = \mathbb{P}\{V_{n+1}=v_i\ | \ E_z\} = \mathbb{P}\{Z_{n+1}=z_i\ | \ E_z\}$ relies on the convention in conformal prediction that $\mathcal{S}$ is a \textit{symmetric} score function (see Remark \ref{remark:nonsymmetric_algo}). We elaborate on this point in Appendix \ref{app:assymetric_algo_extension}.} Then, using the definition of conditional probability and the law of total probability (LOTP), it follows that
\begin{align}
    \mathbb{P}\{Z_{n+1}=z_i\ | \ E_z\} &= \frac{p\{Z_{n+1}=z_i,  E_z\} }{p\{E_z\} } \nonumber \\
    &= \frac{\sum_{\sigma:\sigma(n+1)=i}f(z_{\sigma(1)}, ..., z_{\sigma(n+1)})}{\sum_{\sigma}f(z_{\sigma(1)}, ..., z_{\sigma(n+1)})} \label{eq:app_perm_test_statement}
\end{align}
where the last step follows by applying the LOTP in both the numerator and the denominator. For example, in the denominator the summation is over all possible permutations $\sigma$ of the value, i.e., all the different orderings in which the event $E_z$ could have occurred. Even more explicit notation could be to write out $f(z_{\sigma(1)}, ..., z_{\sigma(n+1)})=f_{Z_1, ..., Z_{n+1}}(z_{\sigma(1)}, ..., z_{\sigma(n+1)})$ if $f$ is a PDF, or $f(z_{\sigma(1)}, ..., z_{\sigma(n+1)})=f(Z_1 = z_{\sigma(1)}, ..., Z_{n+1}=z_{\sigma(n+1)})$ if $f$ is a probability mass function (PMF), to emphasize that the permutation is on the \textit{observed values} and not on the random variables in the joint distribution. 

Here is where our proof differs critically from the analogous proofs in \citet{tibshirani2019conformal}. At this point, their proof corresponding to exchangeable conformal prediction uses exchangeability to substitute in $f(z_{\sigma(1)}, ..., z_{\sigma(n+1)}) = f(z_{1}, ..., z_{n+1})$, since under exchangeability $f$ is invariant to permutations $\sigma$ (see Appendix \ref{app:exchangeability_role} for details); meanwhile, their proof corresponding to weighted conformal prediction substitutes in $f(z_{\sigma(1)}, ..., z_{\sigma(n+1)}) = \prod_{j=1}^{n+1}w_j(z_{\sigma(j)})g(z_{1}, ..., z_{n+1})$, since under their definition of weighted exchangeability $f$ factorizes into weight functions $w_j$ and a core function $g$ that is invariant to $\sigma$ (see Appendix \ref{app:weighted_exchangeability} for details on weighted exchangeability). Here, unlike in \citet{tibshirani2019conformal}, we instead maintain the generality of Eq. \eqref{eq:app_perm_test_statement} by avoiding any substitutions, and this will allow us to prove the more general conformal prediction result, which holds for any $f$. 

For simpler notation, let $\mathbb{P}\{Z_{n+1}=z_i\ | \ E_z\} = \mathbb{P}_{n+1}\{z_i\ | \ E_z\}$. Then, statement \eqref{eq:app_perm_test_statement} implies that, conditioned on $E_z$, the random variable $V_{n+1}$ is distributed according to the weighted empirical distribution of the values $\{v_1, ..., v_{n+1}\}$, with weight $\mathbb{P}\{Z_{n+1}=z_i\ | \ E_z\}$ at each $\delta_{v_i}$:
\begin{align*}
    V_{n+1}|E_v \sim \sum_{i=1}^{n+1}\mathbb{P}_{n+1}\{z_i\ | \ E_z\}\cdot\delta_{v_i}.
\end{align*}
Letting $Q_{\beta}$ denote a level-$\beta$ quantile function, this implies
\begin{align*}
    \mathbb{P}\bigg\{V_{n+1} \leq Q_{\beta}\bigg(\sum_{i=1}^{n+1}\mathbb{P}_{n+1}\{z_i\ | \ E_z\}\cdot\delta_{v_i}\bigg)\ \bigg | \ E_z\bigg\} \geq \beta.
\end{align*}
Due to the conditioning on $E_z$, this is equivalent to 
\begin{align*}
    \mathbb{P}\bigg\{V_{n+1} \leq Q_{\beta}\bigg(\sum_{i=1}^{n+1}\mathbb{P}_{n+1}\{Z_i\ | \ E_z\}\cdot\delta_{V_i}\bigg)\ \bigg | \ E_z\bigg\} \geq \beta,
\end{align*}
and marginalizing over draws of the event $E_z$ yields 
\begin{align}
    \mathbb{P}\bigg\{V_{n+1} \leq Q_{\beta}\bigg(\sum_{i=1}^{n+1}\mathbb{P}_{n+1}\{Z_i\ | \ E_z\}\cdot\delta_{V_i}\bigg)\bigg\} \geq \beta
    \label{eq:general_quantile_lemma_result}
\end{align}
We note that statement \eqref{eq:general_quantile_lemma_result} can be viewed as a general quantile lemma, which generalizes both Lemma 1 and Lemma 3 in \citet{tibshirani2019conformal}. We now connect statement \eqref{eq:general_quantile_lemma_result} to the general conformal prediction set defined in Theorem \ref{thm:general_CP_def_coverage} to obtain the result. To do so, we observe that 
\begin{align}
    Q_{\beta}\bigg(\sum_{i=1}^{n+1}\mathbb{P}_{n+1}\{Z_i\ | \ E_z\}\cdot\delta_{V_i}\bigg) \leq Q_{\beta}\bigg(\sum_{i=1}^{n}\Big[\mathbb{P}_{n+1}\{Z_i\ | \ E_z\}\cdot\delta_{V_i}\Big] + \mathbb{P}_{n+1}\{Z_{n+1}\ | \ E_z\}\cdot\delta_{\infty}\bigg),
    \label{eq:conservative_quantile_comparison}
\end{align}
where the right-hand side is obtained from the left-hand side by replacing $\delta_{V_{n+1}}$ with $\delta_{\infty}$ to conservatively increase the value of the right side relative to the left. Now, with the notation $V_i=V_i^{(X_{n+1}, Y_{n+1})}$ for $i\in \{1, ..., n+1\}$, note that the right hand side of \eqref{eq:conservative_quantile_comparison} is the same quantile as is used in the construction of the general conformal prediction interval in \ref{thm:general_CP_def_coverage}. That is, $Y_{n+1}\in \widehat{\mathcal{C}}_n(X_{n+1})$ if and only if 
\begin{align}
V_{n+1}\leq Q_{\beta}\bigg(\sum_{i=1}^{n}\Big[\mathbb{P}_{n+1}\{Z_i\ | \ E_z\}\cdot\delta_{V_i}\Big] + \mathbb{P}_{n+1}\{Z_{n+1}\ | \ E_z\}\cdot\delta_{\infty}\bigg). 
\label{eq:iff_conformal_statement_pf}
\end{align}
Due to statement \eqref{eq:conservative_quantile_comparison}, the probability of the event \eqref{eq:iff_conformal_statement_pf} is thus lower bounded by statement \eqref{eq:general_quantile_lemma_result}. Setting $\beta=1-\alpha$ in the quantile function then gives the result.

\subsection{Additional Details for Remark \ref{remark:LOO_full_split}}
\label{app:remark_details}

In Theorem \ref{thm:general_CP_def_coverage} and the proof in Appendix \ref{sec:general_cp_coverage_proof} we use score functions that correspond to the ``ordinary'' full conformal method \citet{vovk2005algorithmic}. However, the same result and proof technique would also apply to generalizing the ``deleted'' or ``leave-one-out'' full conformal method (which is computationally more expensive than ordinary full CP but can obtain more informative prediction sets by avoiding overfitting in its construction) \citep{vovk2005algorithmic} or the split conformal method \citep{papadopoulos2008inductive}. The score functions for the deleted full CP method would be $V_i^{(x,y)}=\mathcal{S}(Z_i, Z_{-i}\cup \{(x,y)\})$ for $i\in \{1, ..., n\}$, where $Z_{-i}=Z_{1:n}\backslash Z_i$, and $V_{n+1}^{(x,y)}=\mathcal{S}\big((x,y), Z_{1:n}\big)$. For split CP, the score functions could be defined using independent training and calibration sets. Letting $Z^0_{1:m} = \{Z_i^0\}_{i=1}^m$ denote the training data and $Z_{1:n} = \{Z_i\}_{i=1}^n$ denote the calibration data, the split CP score functions would be $V_i^{(x,y)}=\mathcal{S}(Z_i, Z_{1:m}^0)$ for $i\in \{1, ..., n\}$ and $V_{n+1}^{(x,y)}=\mathcal{S}\big((x,y), Z_{1:m}^0\big)$.

\subsection{Discussion on Symmetric Versus Nonsymmetric Score Functions}
\label{app:assymetric_algo_extension}

In the main paper, we maintained the convention in conformal prediction of requiring that $\mathcal{S}$ be a symmetric score function (e.g., see Remark \ref{remark:nonsymmetric_algo}). Here, we briefly consider how our main result would need to be modified for nonsymmetric score functions, while leaving more rigorous analysis for future work. First for an example, if $\widehat{\mu}_{\bar{Z}} \leftarrow \mathcal{A}(\bar{Z})$ is an ML predictor fit on training examples $\bar{Z}$ by an algorithm $\mathcal{A}$, then the absolute value residual score $\mathcal{S}\big((x,y), \bar{Z}\big) = |y - \widehat{\mu}_{\bar{Z}}(x)|$ is symmetric if the algorithm $\mathcal{A}$ treats its data symmetrically (i.e., if it is invariant to shuffling the ordering of its inputs); meanwhile, this score function is nonsymmetric if the fitted ML preditor output by the algorithm $\widehat{\mu}_{\bar{Z}} \leftarrow \mathcal{A}(\bar{Z})$ depends on the ordering of the elements in $\bar{Z}$. As we note in Remark \ref{remark:nonsymmetric_algo}, this distinction is relevant for full conformal methods, but it holds trivially for split conformal methods due to a separate dataset being used for training and calibration (the symmetry needed is for how the score function treats calibration and test data). 





For a more precise distinction, let $\mathcal{S}$ be a real-valued function that takes as its input a point $(x, y)$ and a \textit{sequence} of other examples $(z_1, ..., z_n)$, so that $\mathcal{S}\big((x, y), (z_1, ..., z_n)\big)$ can be interpreted as how ``nonconforming'' or ``strange'' $(x, y)$ is relative to the examples. Then, we can define symmetric and nonsymmetric score functions as follows:\footnote{Note that these definitions implicitly assume that any randomness is fixed, e.g., so that no equality or inequality is trivial due to differing random seeds.}
\begin{itemize}
    \item $\mathcal{S}$ is a \textbf{symmetric score function} if $\mathcal{S}\big((x, y), (z_1, ..., z_n)\big) = \mathcal{S}\big((x, y), (z_{\sigma(1)}, ..., z_{\sigma(n)})\big)$ for all $n\geq1$, all permutations $\sigma$ of the indices $\{1, ..., n\}$, all $(x, y) \in \mathcal{X}\times \mathcal{Y}$, and all $(z_1, ..., z_n)=((x_1, y_1), ..., (x_n, y_n))$. 
    \item Otherwise, $\mathcal{S}$ is a \textbf{nonsymmetric score function}; for example, if $\mathcal{S}$ is sensitive to the ordering of its examples such that $\mathcal{S}\big((x, y), (z_1, ..., z_n)\big) \neq \mathcal{S}\big((x, y), (z_{\sigma(1)}, ..., z_{\sigma(n)})\big)$ for some permutation $\sigma$ of the indices $\{1, ..., n\}$. 
\end{itemize}
In other words, symmetric score functions treat their training examples $(z_1, ..., z_n)$ symmetrically, so they can equivalently be defined (as in Section \ref{sec:background}) by requiring the examples to be a bag or multiset (rather than a sequence), wherein there is no ordering to the elements. In contrast, nonsymmetric score functions may be sensitive to the ordering of their training inputs.

The proof of Theorem \ref{thm:general_CP_def_coverage} provided in Appendix \ref{sec:general_cp_coverage_proof} relies on $\mathcal{S}$ being a symmetric score function to equate score probabilies to data probabilities; that is, the equation $\mathbb{P}\{V_{n+1}=v_i\ | \ E_v\} = \mathbb{P}\{Z_{n+1}=z_i\ | \ E_z\}$ relies on score-function symmetry. To see this, first recall that we justified this equality by stating that (conditioned on $E_z$), the \textit{fitted} and \textit{symmetric} score function $\mathcal{\widehat{S}}$ induces a bijection between the data points $z_1, ..., z_{n+1}$ and the scores $v_1, ..., v_{n+1}$. As we defined $\mathcal{\widehat{S}}(x, y) = \mathcal{S}\big((x, y), z_{1:(n+1)}\big)$, where $z_{1:(n+1)} = \{z_1, ..., z_{n+1}\}$ (an unordered set), this is equivalent to stating that the \textit{unfitted} (and symmetric) score function $\mathcal{S}$ induces a bijection between the $n+1$ ``data objects'' $(z_1, z_{1:(n+1)}), ..., (z_{n+1}, z_{1:(n+1)})$ and the $n+1$ scores $v_1, ..., v_{n+1}$. 

On the other hand, if we allow $\mathcal{S}$ to be a nonsymmetric score function, then in general it can be the case that $\mathbb{P}\{V_{n+1}=v_i\ | \ E_v\} \neq \mathbb{P}\{Z_{n+1}=z_i\ | \ E_z\}$. This can be seen for instance by comparing the events $E_v$ and $E_z$: observing an unordered set of data values (i.e., the event $E_z$) could result in many different possible sets of scores from a nonsymmetric score function. For nonsymmetric $\mathcal{S}$, we thus are not able to make this substitution to equate score probabilities $\mathbb{P}\{V_{n+1}=v_i\ | \ E_v\}$ to data probabilities $\mathbb{P}\{Z_{n+1}=z_i\ | \ E_z\}$. 

However, a similar argument to the proof in Appendix \ref{sec:general_cp_coverage_proof} could proceed without making this substitution. The main difference is that, rather than the corresponding general result requiring an arbitrary joint PDF $f$ of the \textit{data} in Eq. \eqref{eq:app_perm_test_statement}, instead a the result would require an arbitrary joint PDF of the \textit{score values}. By making this observation, we aim to point out that extensions to allow for nonsymmetric score functions could be a interesting direction, but one we leave for future work. Such future work could potentially leverage ideas from the ``swap step'' in \citet{barber2023conformal}, which the authors introduced to accomodate nonsymmetric algorithms.

\newpage

\section{Deriving Weighted Conformal Validity Guarantees and Algorithms for Data under MFCS}

\label{sec:MFCS_sequential_pseudo-exchangeability}

This section describes in detail how one can use the general procedure presented in Section \ref{sec:deriving_cp_any_dist_MFCS} to derive conformal prediction validity guarantees and practical algorithms for multistep feedback covariate shift (MFCS). 

Regarding notation: We will use $f$ to denote the joint probability density function (PDF) for the sequence of random variables $Z_{1}, ..., Z_{n+t}$ and $p$ to denote more specific (conditional) PDFs, for example $p_{Y|X}=dP_{Y|X}$. To lighten notation we will sometimes omit implicit random variables; for example, we will write $p_{X_j | Z_1, ..., Z_{j-1}}(x_{j} \mid Z_1 = z_1, ..., Z_{j-1}=z_{j-1})$ 
equivalently as $p_{X_j | Z_1, ..., Z_{j-1}}(x_{j} \mid z_1, ..., z_{j-1})$ 
or even as $p(x_{j} \mid z_1, ..., z_{j-1})$, if $X_j | Z_1, ..., Z_{j-1}$ are clear from context and if the shorthand may help declutter the notation.\footnote{We also use $p(x \mid Z_1, ..., Z_{j-1})$ to denote the generic the conditional density function that can change depending on different realized values of $Z_1, ..., Z_{j-1}$ (e.g., it could be that $p(x \mid Z_1=z_1, ..., Z_{j-1}=z_{j-1})\neq p(x \mid Z_1=z_{1}', ..., Z_{j-1}=z_{j-1}')$ for some $(z_1, ..., z_{j-1})\neq (z_1', ..., z_{j-1}')$), whereas $p(x \mid z_1, ..., z_{j-1})$ denotes the specific conditional density, conditional on the joint event $Z_1=z_1, ..., Z_{j-1}=z_{j-1}$.}

\subsection{Deriving MFCS Conformal Validity Guarantees (Sketch Given in Section \ref{sec:deriving_cp_any_dist_MFCS})}

\label{subsec:deriving_MFCS_CP_guarantees}

Here we use the general procedure outlined in Section \ref{sec:deriving_cp_any_dist_MFCS} to derive conformal validity guarantees for MFCS.

\textit{Step 1: List assumptions} 

We begin by restating the formal assumptions we used to characterize MFCS in Eq. \eqref{eq:multistep_fcs}:
\begin{align*}
    Z_i = (X_i, Y_i) &\overset{\text{i.i.d.}}{\sim} P_X^{(0)} \times P_{Y \mid X}, \; i = 1, \dots n, \\
    Z_{n+t} = (X_{n + t}, Y_{n + t}) &\hspace{0.5mm}\sim P_{X;Z_{1:(n+t-1)}}^{(t)} \times P_{Y \mid X}, \; t = 1, \dots, T. \nonumber
\end{align*}
Recall that informally, these assumptions are stating that (after an IID initialization of $n$ datapoints) the distribution of the covariates $X$ at time $t$ can change arbitrarily depending on past observations $Z_{1:(n+t-1)}=\{Z_1, ..., Z_{n+t-1}\}$, while the conditional label distribution $Y\mid X$ is assumed to remain invariant. Note that $Y\mid X$ being invariant implies $Z_1, ..., Z_{n+t-1} \ci Y_{n+t} \mid X_{n+t}$, where the latter can be read as ``$Y_{n+t}$ is independent of past observations $Z_1, ..., Z_{n+t-1}$, given $X_{n+t}$,'' and moreover that $Y_{i}|X_{i} \stackrel{\mathclap{d}}{=} Y|X$ for all $i \in \{1, ..., n+T\}$.


\textit{Step 2: Factorize $f$} 

Next, we factorize the joint probability density function (PDF) $f$ into ``dynamic'' and ``invariant'' factors---specifically, into factors that are dependent on versus are invariant to \textit{permutations} of the data indices. Because in MFCS it is useful to interpret the data indices as timesteps (where the $n$ IID initialized points as distinct observations all occuring at timestep $t=0$), we can think of this distinction as one of time-dependence versus time-invariance.


First, for simplicity, let us focus on factorizing out terms for the most recent timestep $n+t$:
\begin{align*}
    f&(z_1, ..., z_{n+t}) \\
    & = f(z_1, ..., z_{n+t-1}, x_{n+t}, y_{n+t}) && z_{n+t}=(x_{n+t}, y_{n+t}) \\
    & \ \begin{matrix*}[l] = \ p_{Z_1, ..., Z_{n+t-1}}(z_1, ..., z_{n+t-1}) \cdot  p_{X_{n+t}|Z_1, ..., Z_{n+t-1}}(x_{n+t} \mid z_1, ..., z_{n+t-1})  \\ \qquad \cdot \ p_{Y_{n+t}|Z_1, ..., Z_{n+t-1}, X_{n+t}}(y_{n+t} \mid z_1, ..., z_{n+t-1}, x_{n+t})  \end{matrix*} && \begin{matrix}\text{Chain rule of probability or} \\ \text{conditional probability density def.}\end{matrix} \\
    & \ \begin{matrix*}[l] = \ p_{Z_1, ..., Z_{n+t-1}}(z_1, ..., z_{n+t-1}) \cdot  p_{X_{n+t}|Z_1, ..., Z_{n+t-1}}(x_{n+t} \mid z_1, ..., z_{n+t-1})  \\ \qquad \cdot \ p_{Y | X}(y_{n+t} \mid  x_{n+t})  \end{matrix*} && \begin{matrix} \text{MFCS invariance assumption:} \\ Y \ | \ X \text{ invariant } \implies \\ Z_1, ..., Z_{n+t-1} \ci Y_{n+t} \mid X_{n+t} \\  \text{and } Y_{n+t}|X_{n+t} \stackrel{\mathclap{d}}{=} Y|X \end{matrix}
\end{align*}
These steps are not specific to $n+t$, however; we can repeat the same procedure for $p_{Z_1, ..., Z_{n+t-1}}(z_1, ...,  z_{n+t-1})$ and the index $n+t-1$, and so on. When these steps are performed for all indices $i\in \{1, ..., n+t\}$, we obtain
\begin{align*}
    f(z_1, ..., z_{n+t}) & = \prod_{j=1}^{n+t}\Big[p_{X_j|Z_1, ..., Z_{j-1}}(x_{j} \mid z_1, ..., z_{j-1}) \cdot p_{Y|X}(y_{j}| x_{j} )\Big] \\
    & = \prod_{j=1}^{n+t}\Big[\underbrace{p_{X_j|Z_1, ..., Z_{j-1}}(x_{j} \mid z_1, ..., z_{j-1})}_{\begin{matrix}\text{Time-dependent factors} \end{matrix}}\Big] \cdot \underbrace{\prod_{j=1}^{n+t}\Big[p_{Y|X}(y_{j}| x_{j} )\Big]}_{\begin{matrix}\text{Time-invariant factor} \end{matrix}},
\end{align*}
that is, we obtain the factorization into time-dependent and time-invariant factors provided in the main paper Eq. \eqref{eq:mfcs_f_factorization} (where in the main paper we omitted the random variables from the subscripts for ligher notation).


\textit{Step 3: Compute or estimate weights}

Lastly, we plug the result of our factorization from Step 2 into Eq. \eqref{eq:summary_perm_test} to compute or estimate the conformal weights (for the calibration and test points). Recall that with slightly modified notation from our proof in Appendix \ref{sec:general_cp_coverage_proof} to accommodate $n+t$ data indices, we let $E_z^{(t)}$ denote the event $\{Z_1, ..., Z_{n+t}\} = \{z_1, ..., z_{n+t}\}$.


\begin{align*} 
    \mathbb{P}\{Z_{n+t} = z_i \mid E_z^{(t)}\} & = \frac{\sum_{\sigma:\sigma(n+t)=i}f(z_{\sigma(1)}, ..., z_{\sigma(n+t)})}{\sum_{\sigma}f(z_{\sigma(1)}, ..., z_{\sigma(n+t)})} \nonumber \\
    & = \frac{\sum_{\sigma:\sigma(n+t)=i}\prod_{j=1}^{n+t}\Big[p_{X_j|Z_1, ..., Z_{j-1}}(x_{\sigma(j)} \mid z_{\sigma(1)}, ..., z_{\sigma(j-1)})\Big] \cdot \prod_{j=1}^{n+t}\Big[p_{Y|X}(y_{\sigma(j)} \mid x_{\sigma(j)} )\Big]}{\sum_{\sigma}\prod_{j=1}^{n+t}\Big[p_{X_j|Z_1, ..., Z_{j-1}}(x_{\sigma(j)} \mid z_{\sigma(1)}, ..., z_{\sigma(j-1)})\Big] \cdot \prod_{j=1}^{n+t}\Big[p_{Y|X}(y_{\sigma(j)} \mid x_{\sigma(j)} )\Big]} 
    \\
    & = \frac{\sum_{\sigma:\sigma(n+t)=i}\prod_{j=1}^{n+t}p_{X_j|Z_1, ..., Z_{j-1}}(x_{\sigma(j)} \mid z_{\sigma(1)}, ..., z_{\sigma(j-1)})}{\sum_{\sigma}\prod_{j=1}^{n+t}p_{X_j|Z_1, ..., Z_{j-1}}(x_{\sigma(j)}\ | \ z_{\sigma(1)}, ..., z_{\sigma(j-1)})}, 
\end{align*}


where this result is equivalent to Eq. \eqref{eq:mfcs_weights_exact} in the main paper; it follows because $\prod_{j=1}^{n+t}p_{Y|X}(y_{\sigma(j)} \mid x_{\sigma(j)})$ is invariant to permutations $\sigma$, and thus cancels out in the ratio. For more concise notation to refer to these weights derived using MFCS assumptions, let us define
\begin{align}\label{appeq:mfcs_weights_def_exact}
    \mathbb{P}_{n+t}^{(\text{MFCS})}\{z_i \mid E_z^{(t)}\} := \frac{\sum_{\sigma:\sigma(n+t)=i}\prod_{j=1}^{n+t}p_{X_j|Z_1, ..., Z_{j-1}}(x_{\sigma(j)} \mid z_{\sigma(1)}, ..., z_{\sigma(j-1)})}{\sum_{\sigma}\prod_{j=1}^{n+t}p_{X_j|Z_1, ..., Z_{j-1}}(x_{\sigma(j)}\ | \ z_{\sigma(1)}, ..., z_{\sigma(j-1)})}.
\end{align}

A weighted conformal prediction algorithm with these weights $\mathbb{P}_{n+t}^{(\text{MFCS})}\{z_i \mid E_z^{(t)}\}$ for $i\in \{1, ..., n+t\}$ has a valid coverage guarantee as a corollary\footnote{We call this result a ``corollary'' rather than a ``theorem'' primarily to emphasize that it follows from our main result, Theorem \ref{thm:general_CP_def_coverage}, when assuming data are generated under MFCS; however, we encourage future authors to use the term they find most fitting for this or other conformal validity guarantees derived from Theorem \ref{thm:general_CP_def_coverage}.} to Theorem \ref{thm:general_CP_def_coverage}, premised on the MFCS assumptions. This is because our derivation only relied on standard probability rules that hold for any joint PDF\footnote{Or more generally, for any valid Radon-Nikodym derivative with respect to an arbitrary base measure; this also includes discrete probability mass functions (PMFs) and mixtures of PDFs and PMFs.} $f$ along with the MFCS assumptions listed in Step 1. We now state this corollary for completeness.

\begin{corollary}
\textit{Assume that $Z_i=(X_i, Y_i)\in\mathbb{R}^d\times \mathbb{R}, i=1, ..., n+t$ have the joint PDF $f$ and are generated under multistep feedback covariate shift as in Eq. \eqref{eq:multistep_fcs}. For any measurable score function $\mathcal{S}$, and any $\alpha\in(0,1)$, define the generalized conformal prediction set (based on $n$ calibration samples) at a point $x\in \mathbb{R}^d$ by}
\begin{align}\label{appeq:def_mfcs_conformal_set}
&\widehat{\mathcal{C}}_n(x) = \bigg\{y\in \mathbb{R} : V_{n+t}^{(x,y)} \leq Q_{1-\alpha}\bigg(\sum_{i=1}^n\mathbb{P}_{n+t}^{(\text{MFCS)}}\{Z_i \mid E_z\}\cdot\delta_{V_i^{(x,y)}}  + \mathbb{P}_{n+t}^{(\text{MFCS)}}\{Z_{n+t}\mid E_z\}\cdot\delta_{\infty}\bigg)\bigg\}  
\end{align}
\textit{where $V_i^{(x,y)}, i\in \{1, ..., n+t\}$ are as in \eqref{eq:def_scores_ordinary_full} and $\mathbb{P}_{n+t}^{(\text{MFCS)}}\{Z_i \mid E_z\}$ is defined in \eqref{appeq:mfcs_weights_def_exact}. Then, $\widehat{\mathcal{C}}_n$ satisfies}
\begin{align*}
    \mathbb{P}\big\{Y_{n+1}\in \widehat{\mathcal{C}}_n(X_{n+1})\big\}\geq 1-\alpha.
\end{align*}
    \label{corollary:MFCS_CP_coverage}
\end{corollary}

\subsection{Deriving Practical Conformal Algorithms under Agent-Induced MFCS}
\label{subsec:mfcs_algo_pseudo_code}

With the conformal coverage validity guarantee for MFCS in Corollary \ref{corollary:MFCS_CP_coverage} at hand, we now turn to deriving practical algorithms for computing or estimating the conformal prediction set defined in Eq. \eqref{appeq:def_mfcs_conformal_set} under ML-agent-induced MFCS. As we discuss in Section \ref{sec:deriving_cp_any_dist_MFCS} of our main paper, the primary practical bottleneck in this case is computational complexity: While the epistemic challenge is overcome when MFCS is agent-induced (due to $p(x_j \mid Z_1, ..., Z_{j-1})$ representing an ML-agent-controlled query function at time $j$), the complexity for computing the weights in Eq. \eqref{appeq:mfcs_weights_def_exact} (equivalent to our main paper's Eq. \eqref{eq:mfcs_weights_exact}) is still $\mathcal{O}(\prod_{j=1}^t(n+j))$, which quickly becomes intractable for large $t$. To alleviate this bottleneck, we thus proposed an estimation that only uses the ``highest-order'' terms from the $d$ most recent timesteps,

\begin{align*}
     \widehat{p}_{n+t}^{\ (d)}\{z_{i}  \mid E_z^{(t)}\} = & \frac{\sum_{\sigma:\sigma(n+t)=i}\prod^{n+t}_{\color{blue}j=n+t+1-d}p_{X_j|Z_1, ..., Z_{j-1}}(x_{\sigma(j)} \mid z_{\sigma(1)}, ..., z_{\sigma(j-1)})}{\sum_{\sigma}\prod_{{\color{blue}j=n+t+1-d}}^{n+t}p_{X_j|Z_1, ..., Z_{j-1}}(x_{\sigma(j)} \mid z_{\sigma(1)}, ..., z_{\sigma(j-1)})},  
\end{align*}

where we call $d$ our ``estimation depth'' and $\widehat{p}_{n+t}^{\ (d)}\{z_{i}  \mid E_z^{(t)}\}$ is our ``$d$-step'' approximation of the MFCS weights.\footnote{We note that, due to assuming the first $n$ points are IID in MFCS it follows that $p_{X_j|Z_1, ..., Z_{j-1}}(x\mid Z_1, ..., Z_{j-1})=p_{X}(x)$ for $j \in \{1, ..., n\}$, which can allow for including the terms corresponding to $j \in \{1, ..., n\}$ in the product with only marginally more computation, regardless of $d$ (whereas as we have defined $\widehat{p}_{n+t}^{\ (d)}\{z_{i}  \mid E_z^{(t)}\}$, selecting $d \in \{1, ..., t\}$ excludes these terms from the product). However, in certain cases such as uniform random IID initialization where all points initial density on each point is a constant, these terms cancel out in the ratio and thus can be ignored without cost. Accordingly, for this reason and for ease of exposition, we focus on the estimated weights $\widehat{p}_{n+t}^{\ (d)}\{z_{i}  \mid E_z^{(t)}\}$ as we have introduced above.} This $d$-step approximation of the MFCS thus has reduced complexity $\mathcal{O}(\prod_{j=t+1-d}^t(n+j))$, which is a tractable polynomial when $d$ is small. This estimation depth $d$ can be specified by a user based on their computational budget, with larger values of $d$ better approximating the coverage validity guarantee at the expense of increasing computational demand.

When the ML model controlling the query function $p(x_j \mid Z_1, ..., Z_{j-1})$ treats its training data $Z_1, ..., Z_{j-1}$ symmetrically at each timestep (as in our experiments), we can factorize this even further and derive a recursive algorithm for computing the weights, wherein $d$ corresponds to the recursion depth. To provide intuition for this derivation and recursive implementation, let us start by letting $d=3$ and return to the simpler notation used in the main text, which drops the random variables (i.e., leaves them implicit to be more concise):
\begin{align*}
&\widehat{p}_{n+t}^{\ ({\color{blue}3})}\{z_{i}  \mid E_z^{(t)}\} 
     =  \frac{\sum_{\sigma:\sigma(n+t)=i}\prod^{n+t}_{{\color{blue}j=n+t-2}}p(x_{\sigma(j)}\ | \ z_{\sigma(1)}, ..., z_{\sigma(j-1)})}{\sum_{\sigma}\prod_{{\color{blue}j=n+t-2}}^{n+t}p(x_{\sigma(j)}\ | \ z_{\sigma(1)}, ..., z_{\sigma(j-1)})} \\
     & =  \frac{\sum_{\sigma:\sigma(n+t)=i}p(x_{\sigma(n+t)}\ | \ z_{\sigma(1)}, ..., z_{\sigma(n+t-1)})\cdot p(x_{\sigma(n+t-1)}\ | \ z_{\sigma(1)}, ..., z_{\sigma(n+t-2)})\cdot p(x_{\sigma(n+t-2)}\ | \ z_{\sigma(1)}, ..., z_{\sigma(n+t-3)})}{\sum_{\sigma}p(x_{\sigma(n+t)}\ | \ z_{\sigma(1)}, ..., z_{\sigma(n+t-1)})\cdot p(x_{\sigma(n+t-1)}\ | \ z_{\sigma(1)}, ..., z_{\sigma(n+t-2)})\cdot p(x_{\sigma(n+t-2)}\ | \ z_{\sigma(1)}, ..., z_{\sigma(n+t-3)})},
\end{align*}
where the second line simply writes out the product. Next, it will soon help us to write the denominator's summation $\sum_{\sigma}$ equivalently as $\sum_{i}\sum_{\sigma:\sigma(n+t)=i}$, for $i\in \{1, ..., n+t\}=[n+t]$
\begin{align*}
&\widehat{p}_{n+t}^{\ (3)}\{z_{i}  \mid E_z^{(t)}\} =\\
     & \frac{\sum_{\sigma:\sigma(n+t)=i}p(x_{\sigma(n+t)}\ | \ z_{\sigma(1)}, ..., z_{\sigma(n+t-1)})\cdot p(x_{\sigma(n+t-1)}\ | \ z_{\sigma(1)}, ..., z_{\sigma(n+t-2)})\cdot p(x_{\sigma(n+t-2)}\ | \ z_{\sigma(1)}, ..., z_{\sigma(n+t-3)})}{{\color{blue}\sum_{i}\sum_{\sigma:\sigma(n+t)=i}}p(x_{\sigma(n+t)}\ | \ z_{\sigma(1)}, ..., z_{\sigma(n+t-1)})\cdot p(x_{\sigma(n+t-1)}\ | \ z_{\sigma(1)}, ..., z_{\sigma(n+t-2)})\cdot p(x_{\sigma(n+t-2)}\ | \ z_{\sigma(1)}, ..., z_{\sigma(n+t-3)})},
\end{align*}
as we will now make use of knowing that $\sigma(n+t)=i$ within each of the summations $\sum_{\sigma:\sigma(n+t)=i}$ in the numerator and denominator. To do so, we need slightly more notation: 
for any subset of the data indices $K \subseteq \{1, ..., n+t\} = [n+t]$, we write $z_{-K}=\{z_i : i\in [n+t] \backslash K\}$ to denote the set of data whose indices are \textit{not} in $K$; for example, $z_{-\{i, j\}}=\{1, ..., n+t\} \backslash \{i, j\}$. And, since we are assuming ML models that treat their data symmetrically at each timestep, we allow our conditional probabilities to ignore the ordering of data observations being conditioned on (e.g., $p(x \mid z_{-K})$ does not depend on the ordering of the indices in $z_{-K}$). So, with this assumption and notation, the fact that $\sigma(n+t)=i$ within the summation $\sum_{\sigma:\sigma(n+t)=i}$ implies $p(x_{\sigma(n+t)}\ | \ z_{\sigma(1)}, ..., z_{\sigma(n+t-1)}) = p({x_{i}}\ | \ {z_{-\{i\}}})$. So, we can make this substitution and factor out $p({x_{i}}\ | \ {z_{-\{i\}}})$ from $\sum_{\sigma:\sigma(n+t)=i}$, since it no longer depends on $\sigma$:
\begin{align*}
\widehat{p}_{n+t}^{\ ({3})}&\{z_{i}  \mid E_z^{(t)}\} \\
     = & \frac{\sum_{\sigma:\sigma(n+t)=i}{\color{blue}p({x_{i}}\ | \ {z_{-\{i\}}})}\cdot p(x_{\sigma(n+t-1)}\ | \ z_{\sigma(1)}, ..., z_{\sigma(n+t-2)})\cdot p(x_{\sigma(n+t-2)}\ | \ z_{\sigma(1)}, ..., z_{\sigma(n+t-3)})}{\sum_i\sum_{\sigma:\sigma(n+t)=i}{\color{blue}p({x_{i}}\ | \ {z_{-\{i\}}})}\cdot p(x_{\sigma(n+t-1)}\ | \ z_{\sigma(1)}, ..., z_{\sigma(n+t-2)})\cdot p(x_{\sigma(n+t-2)}\ | \ z_{\sigma(1)}, ..., z_{\sigma(n+t-3)})} \\
     = & \frac{{\color{blue}p({x_{i}}\ | \ {z_{-\{i\}}})}\sum_{\sigma:\sigma(n+t)=i}p(x_{\sigma(n+t-1)}\ | \ z_{\sigma(1)}, ..., z_{\sigma(n+t-2)})\cdot p(x_{\sigma(n+t-2)}\ | \ z_{\sigma(1)}, ..., z_{\sigma(n+t-3)})}{\sum_i{\color{blue}p({x_{i}}\ | \ {z_{-\{i\}}})}\sum_{\sigma:\sigma(n+t)=i}p(x_{\sigma(n+t-1)}\ | \ z_{\sigma(1)}, ..., z_{\sigma(n+t-2)})\cdot p(x_{\sigma(n+t-2)}\ | \ z_{\sigma(1)}, ..., z_{\sigma(n+t-3)})}.
\end{align*}

Next, observe that the index $\sigma(n+t)$ no longer appears within the summation $\sum_{\sigma:\sigma(n+t)=i}$ in either the numerator or denominator (it has been factored out). So, similarly as we did before in the denominator, we can equivalently write the summation $\sum_{\sigma:\sigma(n+t)=i}$  equivalently as $\sum_{j\neq i}\sum_{\sigma:\sigma(n+t-1)=j}$, for

\begin{align*}
\widehat{p}_{n+t}^{\ ({3})}&\{z_{i}  \mid E_z^{(t)}\} \\
     = & \frac{p({x_{i}}\ | \ {z_{-\{i\}}}){\color{blue}\sum_{j\neq i}\sum_{\sigma:\sigma(n+t-1)=j}}p(x_{\sigma(n+t-1)}\ | \ z_{\sigma(1)}, ..., z_{\sigma(n+t-2)})\cdot p(x_{\sigma(n+t-2)}\ | \ z_{\sigma(1)}, ..., z_{\sigma(n+t-3)})}{\sum_ip({x_{i}}\ | \ {z_{-\{i\}}}){\color{blue}\sum_{j\neq i}\sum_{\sigma:\sigma(n+t-1)=j}}p(x_{\sigma(n+t-1)}\ | \ z_{\sigma(1)}, ..., z_{\sigma(n+t-2)})\cdot p(x_{\sigma(n+t-2)}\ | \ z_{\sigma(1)}, ..., z_{\sigma(n+t-3)})},
\end{align*}
and similarly as before, within the summation $\sigma(n+t-1)=j$, so $p(x_{\sigma(n+t-1)}\ | \ z_{\sigma(1)}, ..., z_{\sigma(n+t-2)}) = p(x_j\ | \ z_{-\{i, j\}})$. Making this substitution and factoring, and then repeating the pattern again for the third term ultimately gives
\begin{align*}
\widehat{p}_{n+t}^{\ ({3})}&\{z_{i}  \mid E_z^{(t)}\} \\
     = & \frac{p({x_{i}}\ | \ {z_{-\{i\}}})\sum_{j\neq i}\sum_{\sigma:\sigma(n+t-1)=j}{\color{blue}p(x_j\ | \ z_{-\{i, j\}})}\cdot p(x_{\sigma(n+t-2)}\ | \ z_{\sigma(1)}, ..., z_{\sigma(n+t-3)})}{\sum_ip({x_{i}}\ | \ {z_{-\{i\}}})\sum_{j\neq i}\sum_{\sigma:\sigma(n+t-1)=j}{\color{blue}p(x_j\ | \ z_{-\{i, j\}})}\cdot p(x_{\sigma(n+t-2)}\ | \ z_{\sigma(1)}, ..., z_{\sigma(n+t-3)})} \\
     = & \frac{p({x_{i}}\ | \ {z_{-\{i\}}})\sum_{j\neq i}{\color{blue}p(x_j\ | \ z_{-\{i, j\}})}\cdot\sum_{\sigma:\sigma(n+t-1)=j}p(x_{\sigma(n+t-2)}\ | \ z_{\sigma(1)}, ..., z_{\sigma(n+t-3)})}{\sum_ip({x_{i}}\ | \ {z_{-\{i\}}})\sum_{j\neq i}{\color{blue}p(x_j\ | \ z_{-\{i, j\}})}\cdot\sum_{\sigma:\sigma(n+t-1)=j}p(x_{\sigma(n+t-2)}\ | \ z_{\sigma(1)}, ..., z_{\sigma(n+t-3)})} \\
     = & \frac{p({x_{i}}\ | \ {z_{-\{i\}}})\sum_{j\neq i}{p(x_j\ | \ z_{-\{i, j\}})}\cdot{\color{blue}\sum_{k\neq i, j}\sum_{\sigma:\sigma(n+t-2)=k}}p(x_{\sigma(n+t-2)}\ | \ z_{\sigma(1)}, ..., z_{\sigma(n+t-3)})}{\sum_ip({x_{i}}\ | \ {z_{-\{i\}}})\sum_{j\neq i}{p(x_j\ | \ z_{-\{i, j\}})}\cdot{\color{blue}\sum_{k\neq i, j}\sum_{\sigma:\sigma(n+t-2)=k}}p(x_{\sigma(n+t-2)}\ | \ z_{\sigma(1)}, ..., z_{\sigma(n+t-3)})} \\
     = & \frac{p({x_{i}}\ | \ {z_{-\{i\}}})\sum_{j\neq i}{p(x_j\ | \ z_{-\{i, j\}})}\cdot\sum_{k\neq i, j}p(x_{k}\ | \ z_{-\{i,j, k\}})}{\sum_ip({x_{i}}\ | \ {z_{-\{i\}}})\sum_{j\neq i}{p(x_j\ | \ z_{-\{i, j\}})}\cdot\sum_{k\neq i, j}p(x_{k}\ | \ z_{-\{i,j, k\}})}.
\end{align*}

More generally, for an arbitrary estimation depth $d \in [n+t]$ we have the following, with annotations added to indicate how increasing our the depth of recursion $d$ will influence the computation,

\begin{align}\label{app:eq:recursive_mfcs_weights}
\widehat{p}_{n+t}^{\ ({d})}\{z_{i_1}  \mid E_z^{(t)}\} = \frac{\overbrace{p({x_{i_1}}\ | \ {z_{-\{i_1\}}})}^{\text{recursion depth 1}}\overbrace{\textstyle\sum_{i_2\neq i_1}{p(x_{i_2}\ | \ z_{-\{i_1, i_2\}})}}^{\text{recursion depth 2}}\cdots\overbrace{\textstyle\sum_{i_d\neq i_1, ..., i_{d-1}}p(x_{i_d}\ | \ z_{-\{i_1,i_2, ..., i_d\}})}^{\text{recursion depth }d}}{\sum_{i_1}p({x_{i_1}}\ | \ {z_{-\{i_1\}}})\sum_{i_2\neq i_1}{p(x_{i_2}\ | \ z_{-\{i_1, i_2\}})}\cdots\sum_{i_d\neq i_1, ..., i_{d-1}}p(x_{i_d}\ | \ z_{-\{i_1,i_2, ..., i_d\}})},
\end{align}
and where the denominator can be obtained by summing over all the numerator values for $i_1 \in [n+t]$.

Code for our specific implementations of MFCS Full CP and Split CP methods are available at the following GitHub repository: \url{https://github.com/drewprinster/conformal-mfcs}.

\newpage

\subsection{Alternative Approaches to Deriving MFCS Conformal Guarantees and Algorithms}

\label{app:subsec:alternate_derivations}

Here we describe certain alternative approaches to the derivation steps in Appendix \ref{subsec:deriving_MFCS_CP_guarantees}, and we briefly discuss how these alternate steps could have implications for practical algorithm design for MFCS settings and more broadly.

\textit{Alternative Step 1: Assumptions via a Probabilistic Graphical Model}

\begin{figure}[h]
\begin{center}
\begin{tikzpicture}[>=stealth, node distance=1.75cm]
    \tikzstyle{format} = [draw, very thick, circle, minimum size=9.0mm,
	inner sep=0pt]

	\begin{scope}
		\path[->, very thick]
                node[format] (Z1n) {$Z_{1:n}$}
			node[format, right of=Z1n] (Xn1) {$X_{n+1}$}
			node[format, right of=Xn1] (Yn1) {$Y_{n+1}$}
			node[format, right of=Yn1] (Xn2) {$X_{n+2}$}
			node[format, right of=Xn2] (Yn2) {$Y_{n+2}$}
                node[right of=Yn2] (dots)  {\textbf{\ldots}}
                node[format, right of=dots] (XnT)  {$X_{n+T}$}
                node[format, right of=XnT] (YnT)  {$Y_{n+T}$}

                (Z1n) edge[black] (Xn1)
                (Z1n) edge[black, bend left=30] (Xn2)
                (Z1n) edge[black, bend right=30] (XnT)
			(Xn1) edge[blue] (Yn1)
                (Yn1) edge[black] (Xn2)
			(Xn1) edge[black, bend right=30] (Xn2)
                (Xn1) edge[black,bend left=30] (XnT)
                (Yn1) edge[black,bend left=25] (XnT)
                (Xn2) edge[black,bend right=25] (XnT)
                (Yn2) edge[black,bend right=20] (XnT)
                (Xn2) edge[blue] (Yn2)
                (XnT) edge[blue] (YnT)
		;
	\end{scope}

\end{tikzpicture}
\end{center}
\caption{A causal directed acyclic graph (DAG) model that implies $Z_1, ..., Z_{n+t-1} \ci Y_{n+t} \mid X_{n+t}$ (as does MFCS). Moreover, the blue edges represent relationships that are further assumed to be equivalent, implying $Y_{n+t}|X_{n+t}\stackrel{\mathclap{d}}{=} Y|X$ for all $t\in \{1, ..., T\}$.}
\label{fig:DAG}
\end{figure}
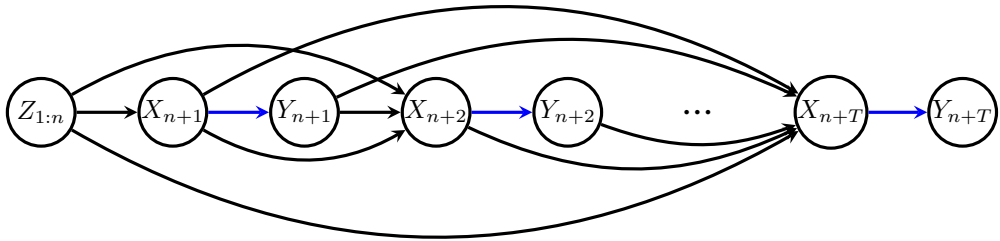

Rather than beginning with the MFCS conditional independence assumptions (Eq. \eqref{eq:multistep_fcs}) for Step 1, we could have instead begun by describing our data-generating process using a probabilistic graphical model \citep{koller2009probabilistic, pearl2009causality} and used the conditional independence conditions implied by the missing edges in that model.\footnote{Missing edges in probabilistic graphical models imply conditional independences in the joint PDF $f$; the converse depends on further assumptions \citep{pearl2009causality}.} One advantage of graphical models is that they can be more visually intuitive than conditional independence or invariance assumptions. For example, consider the directed acyclic graph (DAG) model in Figure \ref{fig:DAG}, on which we take a causal interpretation to view each directed edge $A\rightarrow B$ as representing that $A$ has a causal effect on $B$.\footnote{We take posit a \textit{causal} DAG here so that the edges better map onto our intuition about cause and effect; however, in other settings different graphical models could instead be used so long as the appropriate Markov property is used to obtain the implied conditional independence relations.}
As in MFCS, in this causal DAG model we first assume $Z_{1:n}=\{Z_1, ..., Z_n\}$ are independent and identically distributed (IID), and in the DAG we represent these IID variables with a single node to be concise; for the remaining nodes, we allow each $X_{n+t}$ to depend causally on all prior observations (i.e., each $X_{n+t}$ has incoming edges from all prior nodes), while we only allow each $Y_{n+t}$ to be affected by $X_{n+t}$ (i.e., $Y_{n+t}$ only has an incoming edge from $X_{n+t}$). We use blue arrows to represent relationships that we further assume are equivalent (in distribution).

By the rules of $d$-separation for DAGs \citep{pearl1988probabilistic}, for any $t \in \{1, ..., T\}$, conditioning on $X_{n+t}$ blocks all paths from prior nodes to $Y_{n+t}$; so, this DAG implies the same conditional independence relation that we obtained from our original MFCS invariance assumptions: $Z_1, ..., Z_{n+t-1} \ci Y_{n+t} \mid X_{n+t}$. Moreover, the blue edges represent relationships that are further assumed to be equivalent in distribution, implying $Y_{n+t}|X_{n+t}\stackrel{\mathclap{d}}{=} Y|X$ for all $t\in \{1, ..., T\}$. 

\vspace{1cm}

\textit{Alternative Steps 2 and 3: Factoring $f$ with likelihood-ratio factors}

The following is an alternative approach to Steps 2 and 3 in Appendix \ref{subsec:deriving_MFCS_CP_guarantees}, which factorizes $f$ using likelihood-\textit{ratio} factors, which more closely resemble the specific CP algorithms for standard covariate shift in \citet{tibshirani2019conformal} and (one-step) feedback covariate shift in \citet{fannjiang2022conformal}. We will refer to the factorization given in Step 2 of Appendix \ref{subsec:deriving_MFCS_CP_guarantees} as the \textit{direct} likelihood factorization, and the alternative derivation we present here as the likelihood \textit{ratio}-based factorization; we will soon discuss tradeoffs of these different viewpoints regarding interpretability and implementation.

First, recall the result of Step 2 in Appendix \ref{subsec:deriving_MFCS_CP_guarantees}:
\begin{align*}
    f&(z_1, ..., z_{n+t})  = \prod_{j=1}^{n+t}\Big[p_{X_j|Z_1, ..., Z_{j-1}}(x_{j}\mid z_1, ..., z_{j-1})\Big] \cdot \prod_{j=1}^{n+t}\Big[p_{Y|X}(y_{j}| x_{j} )\Big].
\end{align*}
Whereas the direct factorization in Appendix \ref{subsec:deriving_MFCS_CP_guarantees} proceeds immediately from here to Step 3 by plugging this statement into Eq. \eqref{eq:summary_perm_test}, now we consider further steps to obtain explicit likelihood-ratio factors. To do so, we simplify using our assumption that the first $n$ points are IID, which implies $p_{X_j|Z_1, ..., Z_{j-1}}(x_{j}\ | \ z_1, ..., z_{j-1}) = p_{X}(x_{j})$ for $j \in \{1, ..., n\}$, and then use a ``multiply by one'' trick to multiply by $1=\tfrac{\prod_{j=n+1}^{n+t}p_{X}(x_{j})}{\prod_{j=n+1}^{n+t}p_{X}(x_{j})}$, where $p_{X} := dP_X^{(0)}$ denotes the density for the features in the initial IID distribution $P_Z^{(0)}=P_X^{(0)}\times P_{Y|X}$:
\begin{align*}
    f(z_1, ..., z_{n+t}) & = \prod_{j=1}^{n}\Big[p_{X}(x_{j})\Big]\cdot\prod_{j=n+1}^{n+t}\Big[p_{X_j|Z_1, ..., Z_{j-1}}(x_{j}\ | \ z_1, ..., z_{j-1})\Big] \cdot \prod_{j=1}^{n+t}\Big[p_{Y|X}(y_{j}| x_{j} )\Big] \\
    & = \prod_{j=1}^{n}\Big[p_{X}(x_{j})\Big]\cdot\prod_{j=n+1}^{n+t}\Big[p_{X_j|Z_1, ..., Z_{j-1}}(x_{j}\ | \ z_1, ..., z_{j-1})\Big] \cdot \prod_{j=1}^{n+t}\Big[p_{Y|X}(y_{j}| x_{j} )\Big]\cdot \frac{\prod_{j=n+1}^{n+t}p_{X}(x_{j})}{\prod_{j=n+1}^{n+t}p_{X}(x_{j})} \\
    & = \prod_{j=1}^{n+t}\Big[p_{X}(x_{j})\Big]\cdot\prod_{j=n+1}^{n+t}\Big[\frac{p_{X_j|Z_1, ..., Z_{j-1}}(x_{j}\ | \ z_1, ..., z_{j-1})}{p_{X}(x_{j})}\Big] \cdot \prod_{j=1}^{n+t}\Big[p_{Y|X}(y_{j}| x_{j} )\Big] \\
    & = \prod_{j=n+1}^{n+t}\Big[\frac{p_{X_j|Z_1, ..., Z_{j-1}}(x_{j}\ | \ z_1, ..., z_{j-1})}{p_{X}(x_{j})}\Big] \cdot \prod_{j=1}^{n+t}\Big[p_{Y|X}(y_{j}| x_{j} )\cdot p_{X}(x_{j})\Big] \\
    & = \prod_{j=n+1}^{n+t}\Big[\underbrace{\frac{p_{X_j|Z_1, ...,Z_{j-1}}(x_{j}\ | \ z_1, ..., z_{j-1})}{p_{X}(x_{j})}}_{\begin{matrix}\text{Time-dependent } \\ \text{likelihood-ratio factors} \end{matrix}}\Big] \cdot \underbrace{\prod_{j=1}^{n+t}\Big[p_{Z}(z_{j})\Big]}_{\begin{matrix}\text{Time-invariant} \\ \text{``core'' function }\end{matrix}}.
\end{align*}


There are two main tradeoffs between the direct likelihood factorization (Step 2 of Appendix \ref{subsec:deriving_MFCS_CP_guarantees}) and the alternative likelihood-ratio factorization presented here---the first tradeoff is with respect to interpretability, and the second tradeoff is with respect to implementation. First, regarding interpretability: a benefit of the factorization presented in Appendix \ref{subsec:deriving_MFCS_CP_guarantees} is that its permutation-invariant factor $\prod_{j=1}^{n+t}p_{Y|X}(y_{j} \mid x_j)$ clearly and intuitively corresponds to the assumption that $Y\mid X$ is invariant in MFCS, which is one reason we chose to use that factorization in our main paper's presentation; on the other hand, a benefit of the current likelihood-ratio factorization is that it can be interpreted more clearly as ``the likelihood obtained by \textit{adjusting relative to if all the data were IID},'' since the permutation-invariant factor $\prod_{j=1}^{n+t}p_{Z}(z_{j})$ represents the probability of the sequence of observations \textit{if} they were all IID, and the likelihood ratios represent the necessary adjustment factors needed to compute the true probability. The current likelihood ratio-based factorization also relates more closely to the analysis and algorithms in \citet{tibshirani2019conformal} and \citet{fannjiang2022conformal}.


Next, to discuss implementation tradeoffs, it may be helpful to first complete Step 3 by plugging our factorization into the equation (Eq. \eqref{eq:summary_perm_test}) for the CP weights:
\begin{align*}
    \frac{\sum_{\sigma:\sigma(n+t)=i}f(z_{\sigma(1)}, ..., z_{\sigma(n+1)})}{\sum_{\sigma}f(z_{\sigma(1)}, ..., z_{\sigma(n+1)})} & = \frac{\sum_{\sigma:\sigma(n+t)=i}\prod_{j=n+1}^{n+t}\Big[\tfrac{p_{X_j|Z_1,...,Z_{j-1}}(x_{\sigma(j)}\ | \ z_{\sigma(1)}, ..., z_{\sigma(j-1)})}{p_{X}(x_{\sigma(j)})}\Big]\prod_{j=1}^{n+1}\Big[p_{Z}(z_{\sigma(j)} )\Big]}{\sum_{\sigma}\prod_{j=n+1}^{n+t}\Big[\tfrac{p_{X_j|Z_1,...,Z_{j-1}}(x_{\sigma(j)}\ | \  z_{\sigma(1)}, ..., z_{\sigma(j-1)})}{p_{X}(x_{\sigma(j)})}\Big]\prod_{j=1}^{n+1}\Big[p_{Z}(z_{\sigma(j)})\Big]}
    \\
    & = \frac{\sum_{\sigma:\sigma(n+t)=i}\prod_{j=n+1}^{n+t}\Big[\tfrac{p_{X_j|Z_1,...,Z_{j-1}}(x_{\sigma(j)}\ | \ z_{\sigma(1)}, ...,  z_{\sigma(j-1)})}{p_{X}(x_{\sigma(j)})}\Big]}{\sum_{\sigma}\prod_{j=n+1}^{n+t}\Big[\tfrac{p_{X_j|Z_1,...,Z_{j-1}}(x_{\sigma(j)}\ | \ z_{\sigma(1)}, ..., z_{\sigma(j-1)})}{p_{X}(x_{\sigma(j)})}\Big]}.
\end{align*}
Completing this Step 3 emphasizes that it is the ``dynamic'' factors (that are not permutation invariant) which are of practical concern for implementation, since the permutation-invariant factor $\prod_{j=1}^{n+t}p_{Z}(z_{\sigma(j)})$ has cancelled out in the ratio. 

One implementation tradeoff between the direct (Appendix \ref{subsec:deriving_MFCS_CP_guarantees}) and the likelihood ratio-based factorization here is an epistemic one: Implementations using the direct factorization rely on computing or estimating the product of likelihoods $\prod_{j=1}^{n+t}p_{X_j|Z_1,...,Z_{j-1}}(x_{\sigma(j)}\ | \ z_{\sigma(1)}, ..., z_{\sigma(j-1)})$; we leverage this direct approach in our main paper largely because our practical experiments focus on ML-agent-induced MFCS where $p_{X_j|Z_1,...,Z_{j-1}}(x \mid Z_1, ..., Z_{j-1})$ is known (it is the ML agent's query function). On the other hand, for a hypothetical MFCS instance where $p_{X_j|Z_1,...,Z_{j-1}}(x \mid Z_1, ..., Z_{j-1})$ is \textit{not} known, an implementation based on the likelihood-ratio factorization may  be preferred, because likelihood-ratio (density-ratio) estimation can often be easier than direct likelihood (density) estimation. 

Another implementation consideration to note is the ``failure mode'' of specific permutations $\sigma$ that have zero probability density, for example due to $p_{X}(x_{\sigma(j)}) = 0$ for some $j \in \{n+1, ..., n+t\}$, which could occur  if $X_{n+t}=x_{n+t}$ is observed out of the support of $P_X^{(0)}$, the initial distribution that the first $n$ points are assumed to be drawn IID from in MFCS. Implementations based on the likelihood ratio-based factorization would be undefined in such cases due to requiring dividing by zero, and thus they may need to exclude these cases by assuming that the density in the numerator is absolutely continuous with respect to that of the denominator (e.g., \citet{tibshirani2019conformal} and \citet{fannjiang2022conformal} made this assumption for their specific CP algorithms for one-step standard and feedback covariate shift). 
On the other hand, implementations using the direct factorization (Appendix \ref{subsec:deriving_MFCS_CP_guarantees}) are not necessarily undefined in these cases---that is, if one commits to a definition of conditional probability densities where $p(A|B)=0$ if $p(B)=0$, rather allowing the conditional density to be undefined in such cases (see e.g., \citet{kolmogorov1956foundations}). For a specific example, let us assume $T=1$ and that $X_{n+1}=x_{n+1}$ is observed out of the support of $P_X^{(0)}$, so $p_{X}(x_{n+1})=0$. Then (assuming a conditional probability density definition where $p(A|B)=0$ if $p(B)=0$), the MFCS weights obtained via direct factorization in Eq. \eqref{appeq:mfcs_weights_def_exact} would reduce to $\mathbb{P}_{n+1}^{(\text{MFCS})}\{z_i \mid E_z^{(t)}\} = 0$ for all $i \neq n+1$, and $\mathbb{P}_{n+1}^{(\text{MFCS})}\{z_{n+1} \mid E_z^{(t)}\} = 1$. Accordingly, the resulting CP set would maintain coverage trivially by concentrating all the weight on $\delta_{\infty}$ (the point mass at $\infty$) in the (conservative) empirical score distribution,  thus forcing a non-informative prediction set $\widehat{C}_{n}(x)=\mathcal{Y}$.

\newpage

\section{Bounded Query Function Experiments for Adaptive Exploration}

\label{app:sec:adaptive_exploration_results}

\subsection{Bounding the Query Function to Enforce Informative (Finite) Prediction Intervals}

Because noninformative prediction sets ($\widehat{\mathcal{C}}_n(x)=\mathcal{Y}$) occur in our weighted CP MFCS experiments when a test point's normalized weight as in Eq. \eqref{eq:def_3rd_order_weights} (implemented as in Eq. \eqref{app:eq:recursive_mfcs_weights}) exceeds $\alpha$, we now consider redesigning the ML query probability functions to prevent this from happening. For simplicity, we impose this constraint on the one-step weights. In particular, we consider bounding the query probability function $p(x \mid Z_1, ..., Z_{j-1})$ so that instead of being proportional to $\exp(\lambda\cdot u_t(x))$, the probabilities are instead proportional to a bounded function:
\begin{align*}
    p(x \mid Z_1, ..., Z_{j-1}) \propto &\min\big(\exp(\lambda \cdot u_t(x)), B \big), \quad \text{where } B = \sup \bigg\{b > 0 : \frac{b}{\sum_{i=1}^n\min\big(\exp(\lambda \cdot u_t(X_i)), b \big) + b} < \alpha \bigg\}.
\end{align*}
That is, we select $B$ as the largest bound that we could use to avoid infinite-width prediction intervals (in the depth $d=1$ weights).
The selection of $B$ above avoids noninformative prediction sets because
\begin{align*}
& \min\big(\exp(\lambda \cdot u_t(x)), B \big) \leq B \\
& \implies 
\frac{\min\big(\exp(\lambda \cdot u_t(x)), B \big)}{\sum_{i=1}^n\min\big(\exp(\lambda \cdot  u_t(X_i)), B \big) + \min\big(\exp(\lambda \cdot  u_t(x)), B \big)} \leq \frac{B}{\sum_{i=1}^n\min\big(\exp(\lambda \cdot  u_t(X_i)), B \big) + B} < \alpha.
\end{align*}
In our implementations, we solve for $B$ via a binary search algorithm over all values of the unbounded function $\exp(\lambda\cdot u_t(x))$ for data in the input space $x \in \mathcal{X}$, which is tractable in our experiments where $\mathcal{X}$ is the finite pool or library of possible query points. That is, we specifically solve for $B = \max\big\{b = \exp(\lambda\cdot u_t(x)) \ : \ x\in\mathcal{X}, \ \tfrac{b}{\sum_{i=1}^n\min(\exp(\lambda \cdot u_t(X_i)), b ) + b} < \alpha \big\}$. Code for our implementations is available here: \url{https://github.com/drewprinster/conformal-mfcs}.

Across the four real datasets and the same experimental conditions we used for our active learning experiments in Figure \ref{fig:SplitCP_ActiveLearningExpts} of our main paper, the following results empirically validate that imposing the proposed adaptive bound on our MFCS CP method’s query function prevents any noninformative (infinite width) intervals, all without sacrificing coverage even over a long time horizon (70 active learning steps). 
For easy comparison, for each dataset we provide the corresponding results for the original unbounded query function from Figure \ref{fig:SplitCP_ActiveLearningExpts}. We moreover plot the ``relative magnitude'' of the bound $B$---i.e., we plot $B / B_{\max}$, where $B_{\max} = \max_{x\in \mathcal{X}}(\exp(\lambda \cdot \hat{\sigma}(x)))$---across the same trajectory of active learning steps.


\textbf{Airfoil dataset}

\begin{figure*}[!htb]
    \centering
    \begin{subfigure}{0.85\textwidth}\includegraphics[width=\textwidth]{figures/SplitCPActiveLearning_bounded_legend.pdf}
    \end{subfigure}
    \\
    \begin{subfigure}{0.35\textwidth}\includegraphics[width=\textwidth]{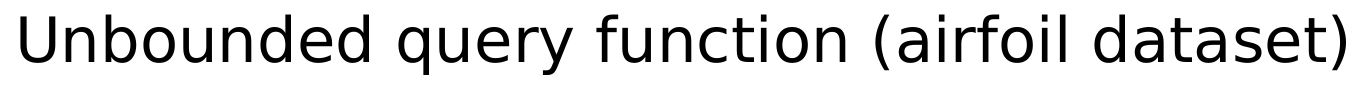}
    \end{subfigure}
    \\
    \hfill
    \begin{subfigure}{0.2\textwidth}
        \includegraphics[width=\textwidth]{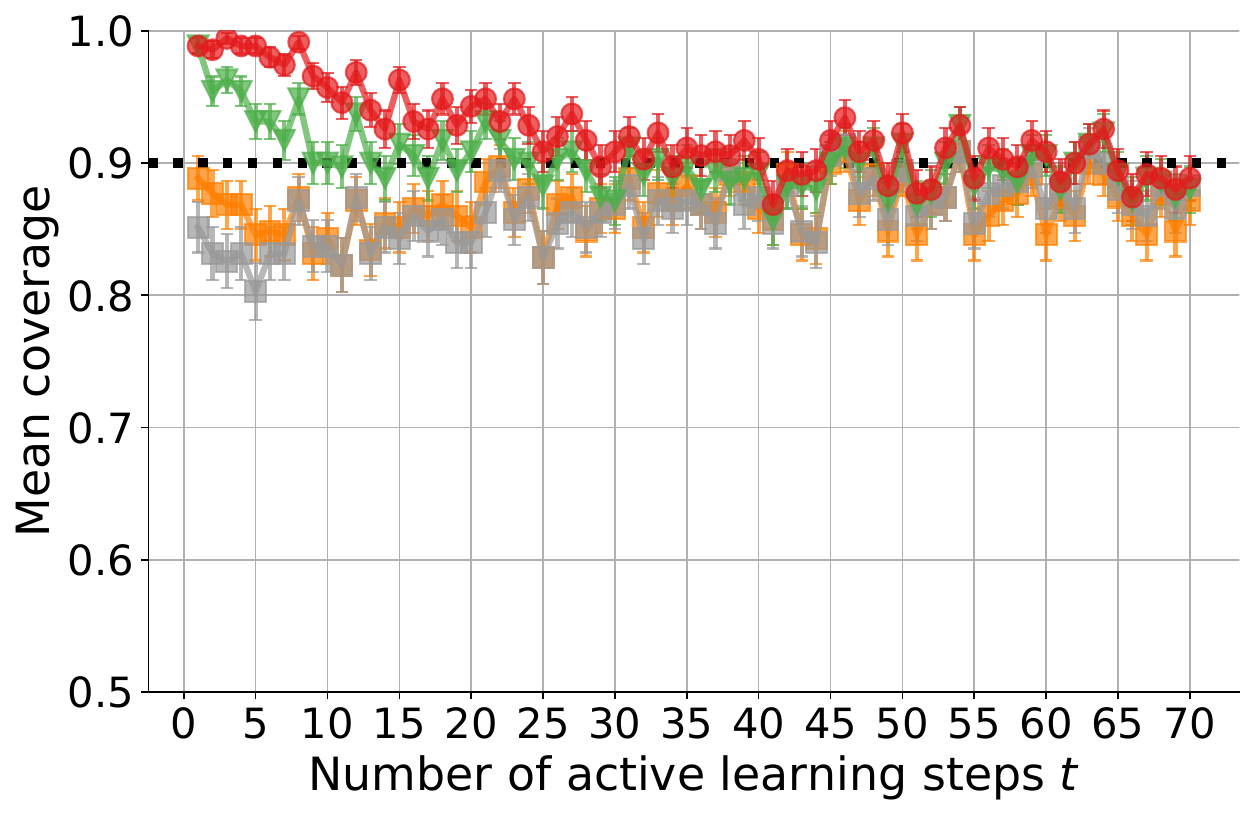}
    \end{subfigure}
    \hfill
    \begin{subfigure}{0.2\textwidth}
        \includegraphics[width=\textwidth]{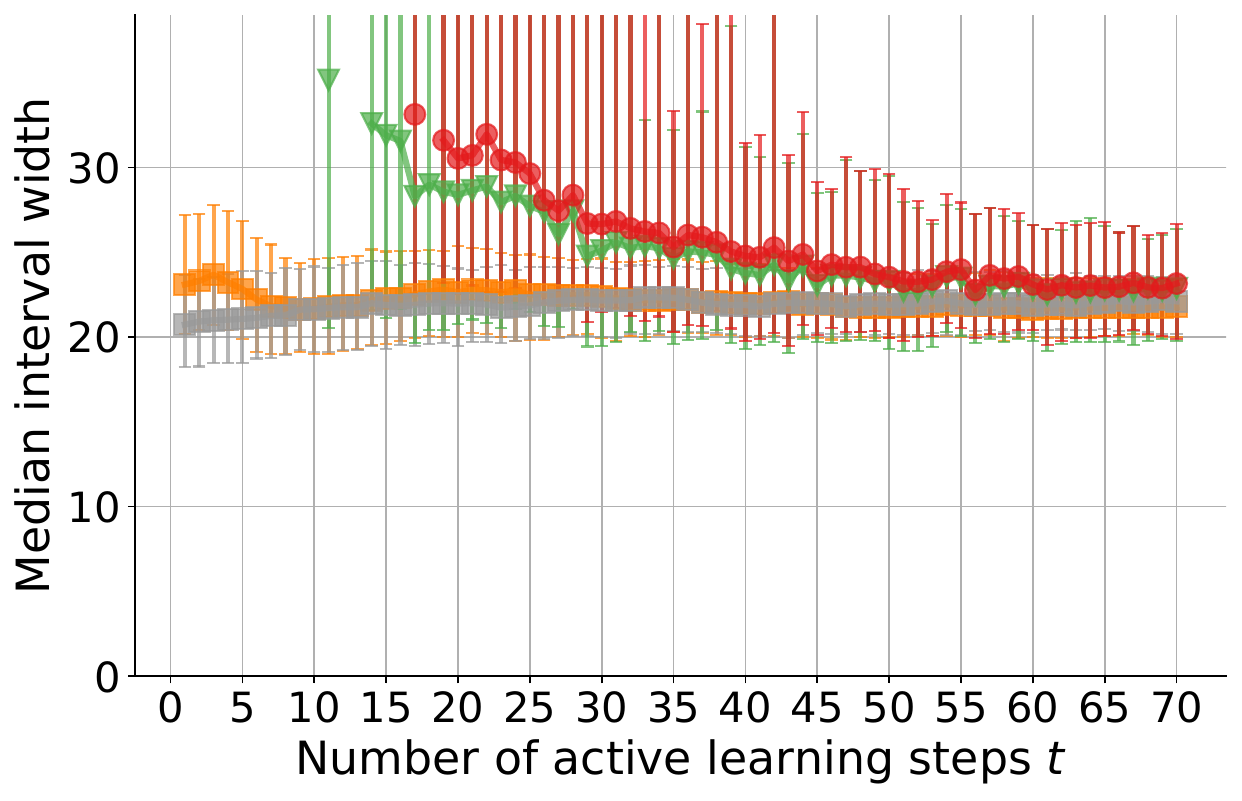}
    \end{subfigure}
    \hfill
    \begin{subfigure}{0.2\textwidth}
        \includegraphics[width=\textwidth]{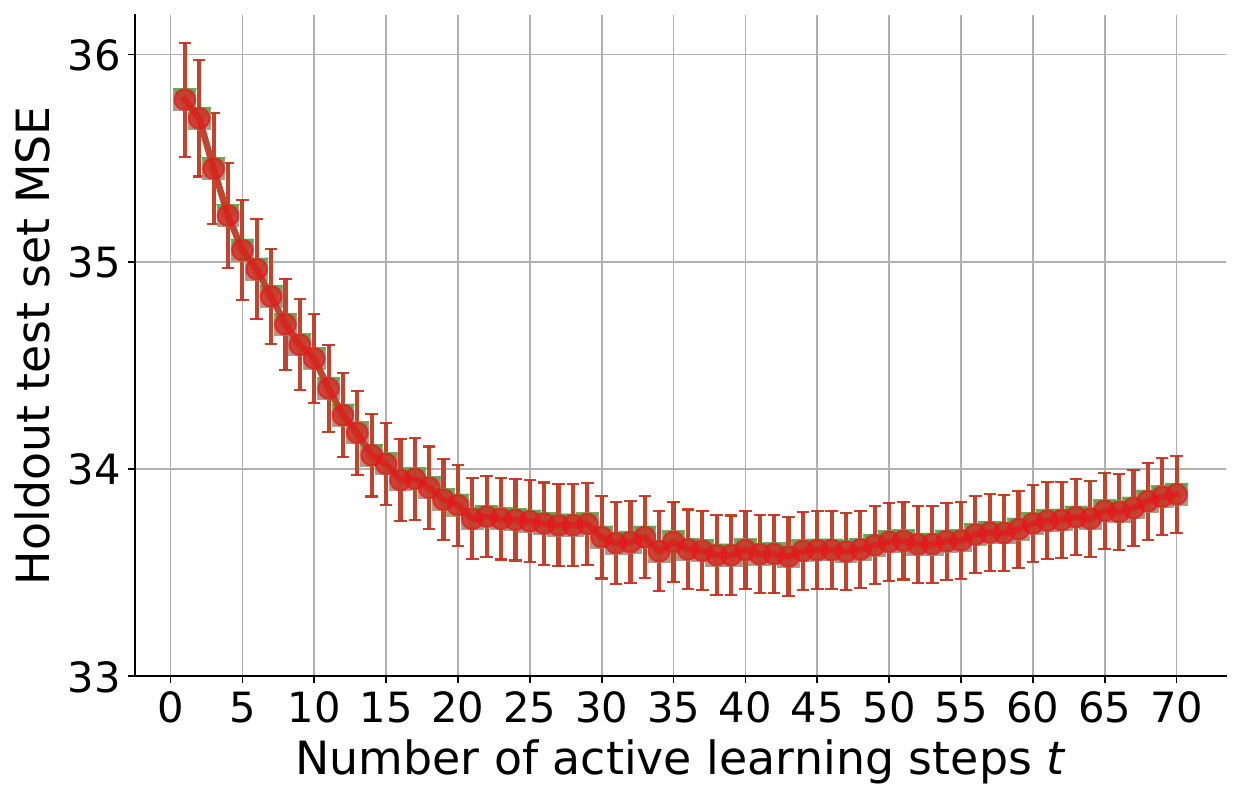}
    \end{subfigure}
    \hfill
    \begin{subfigure}{0.2\textwidth}
        \hfill
    \end{subfigure}
    \hfill
    \\
    \begin{subfigure}{0.35\textwidth}\includegraphics[width=\textwidth]{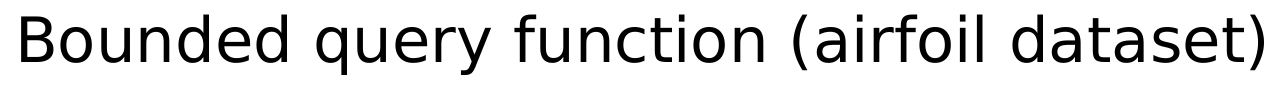}
    \end{subfigure}
    \\
    \hfill
    \begin{subfigure}{0.2\textwidth}
        \includegraphics[width=\textwidth]{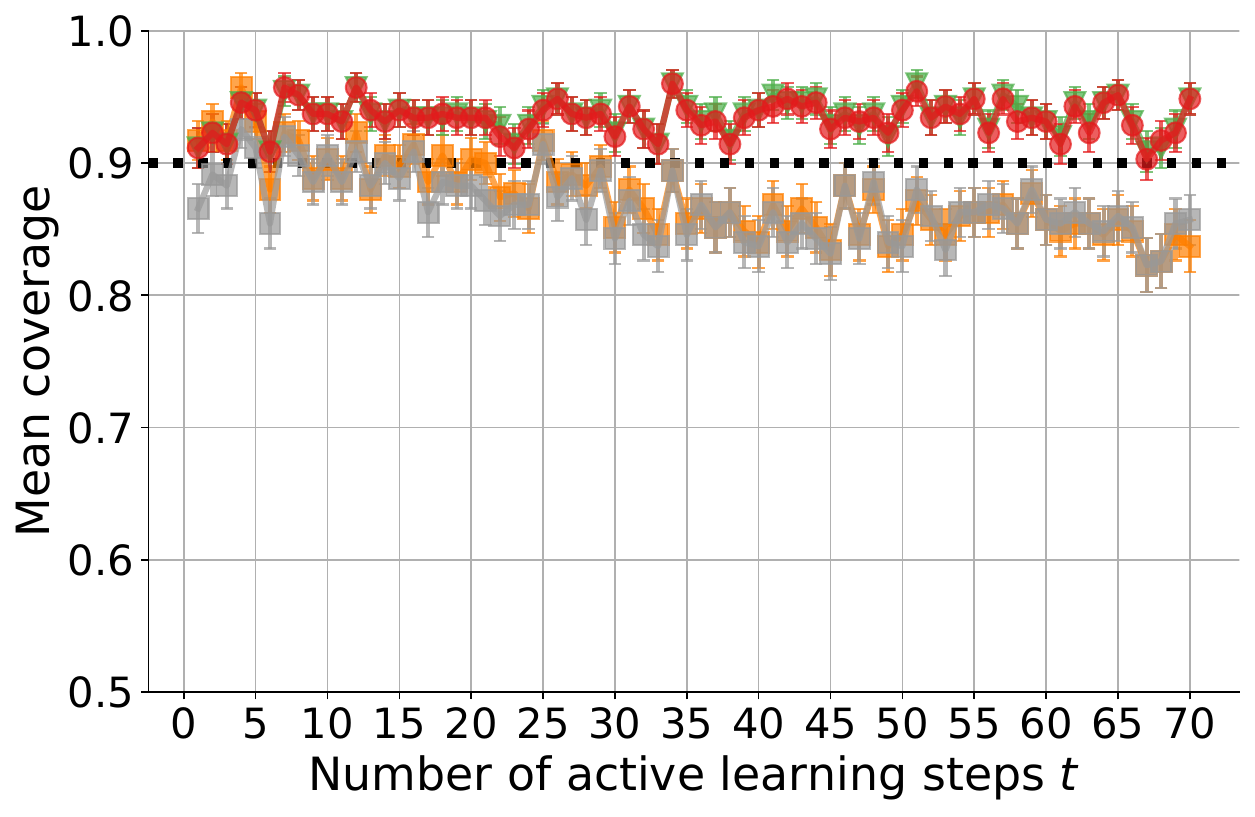}
    \end{subfigure}
    \hfill
    \begin{subfigure}{0.2\textwidth}
        \includegraphics[width=\textwidth]{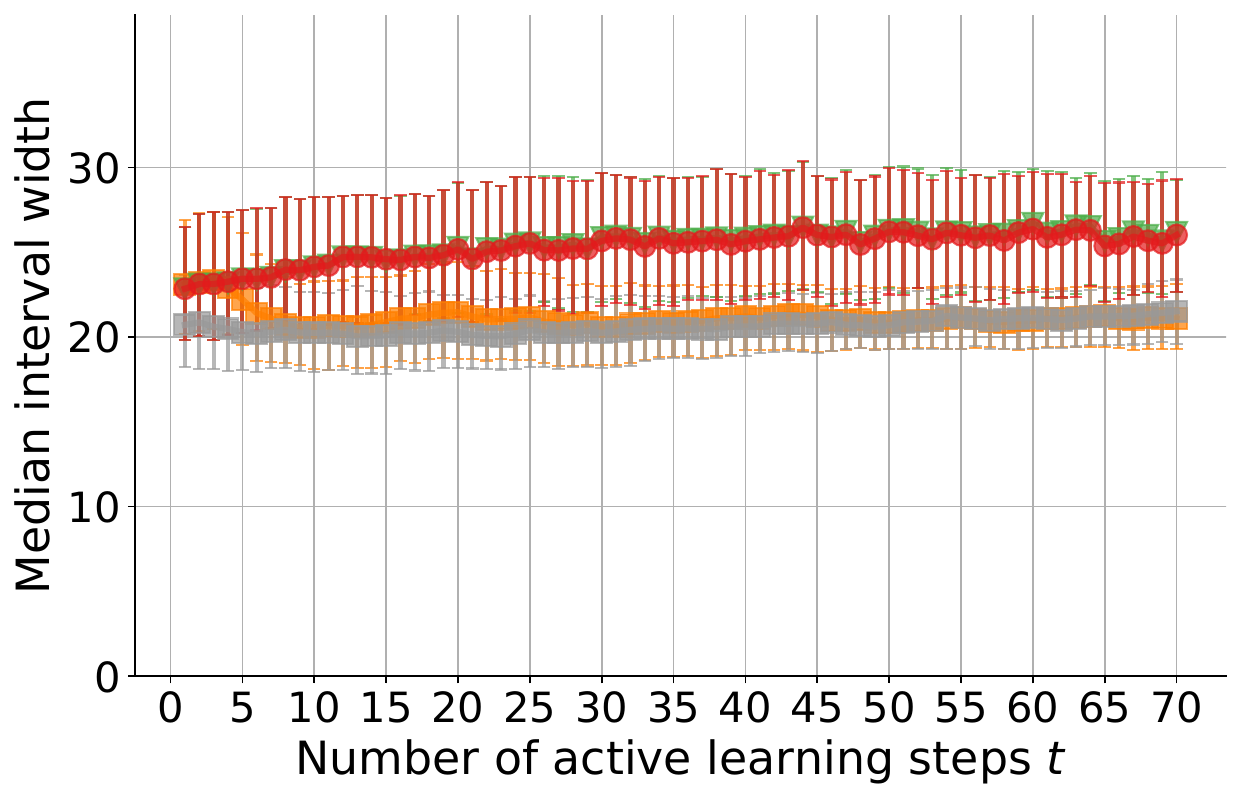}
    \end{subfigure}
    \hfill
    \begin{subfigure}{0.2\textwidth}
        \includegraphics[width=\textwidth]{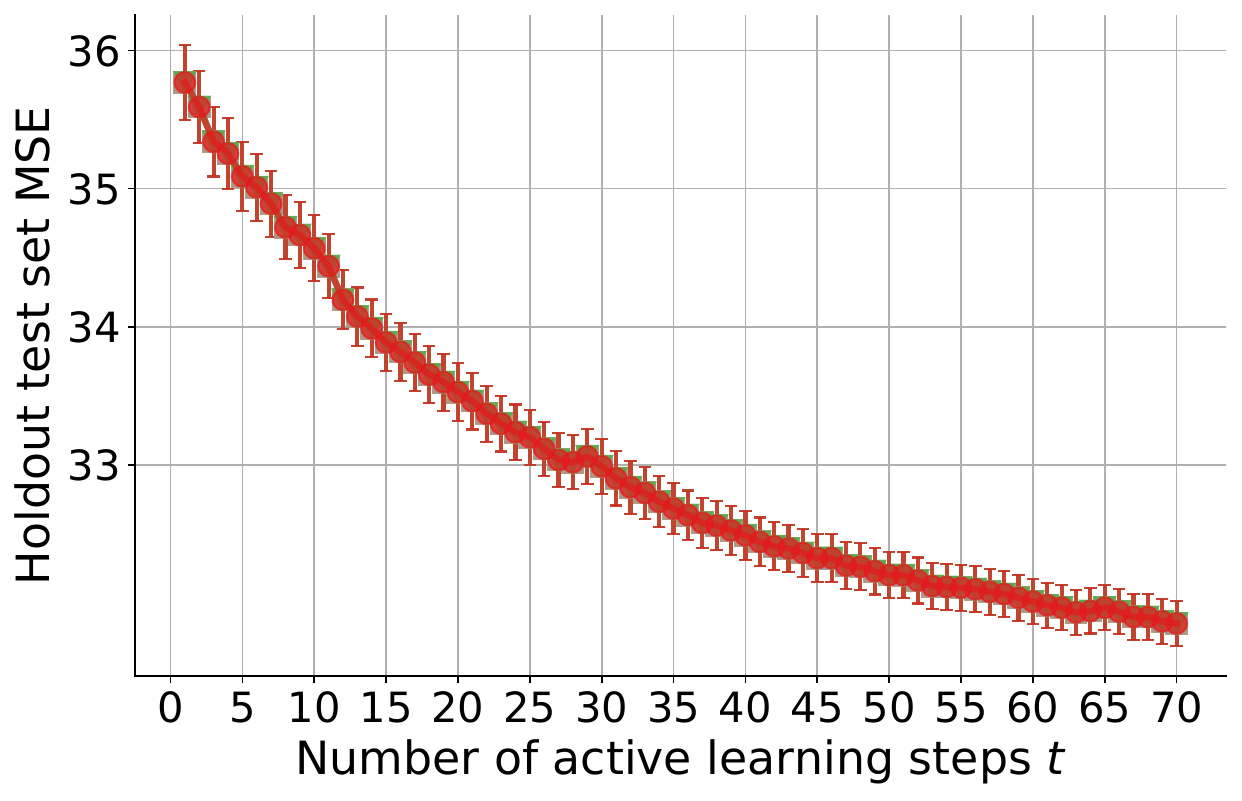}
    \end{subfigure}
    \hfill
    \begin{subfigure}{0.2\textwidth}
        \includegraphics[width=\textwidth]{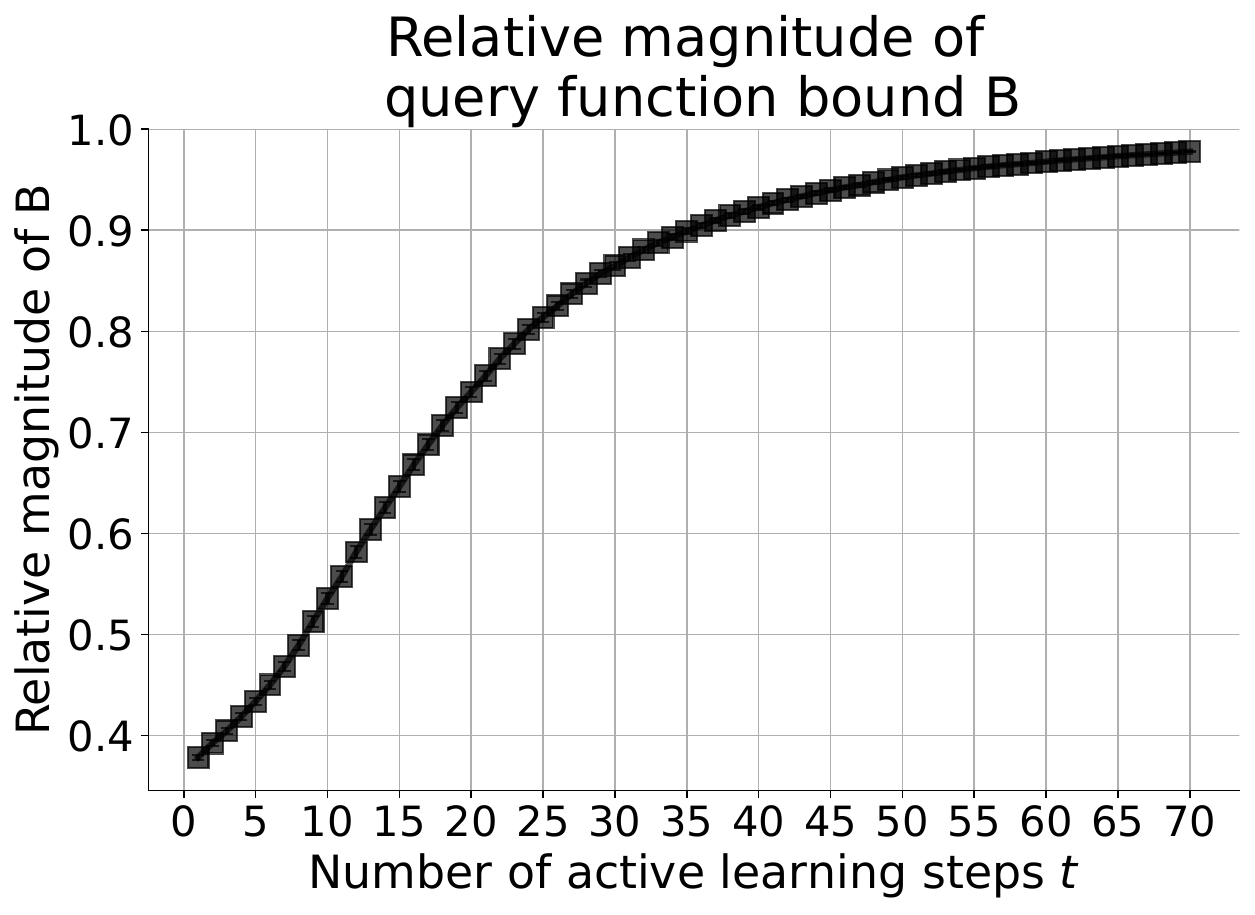}
    \end{subfigure}
    \hfill
    \\
\caption{Unbounded (top row) versus bounded (bottom row) active learning experiments of proposed multistep split CP method for $d=3$ (red circles) compared to baselines of unweighted split CP (orange squares), one-step split CP (green triangles), and ACI (gray squares) on the airfoil dataset.
    The Y-axes represent mean coverage, median interval width, and mean squared error on a holdout test set; the X-axes correspond to the number of active learning query steps, with each query based on posterior variance of a GP regressor.
    All values are computed over 350 distinct random seeds. Hyperparameters for the experiments are given in Appendix \ref{subsec:active_learning_exp_details} Table \ref{tab:fig3_hyperparams}.
    }
\label{appfig:BoundedActiveLearningExpts_airfoil}
\end{figure*}

\newpage

\textbf{Communities dataset}

\begin{figure*}[h]
    \centering
    \begin{subfigure}{0.85\textwidth}\includegraphics[width=\textwidth]{figures/SplitCPActiveLearning_bounded_legend.pdf}
    \end{subfigure}
    \\
    \begin{subfigure}{0.4\textwidth}\includegraphics[width=\textwidth]{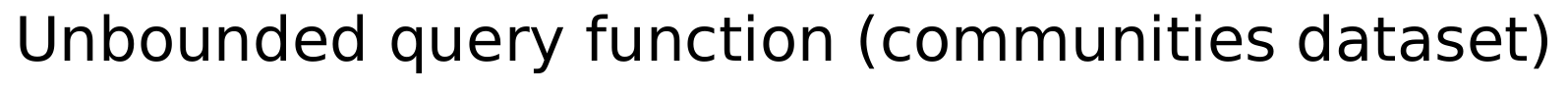}
    \end{subfigure}
    \\
    \hfill
    \begin{subfigure}{0.2\textwidth}
        \includegraphics[width=\textwidth]{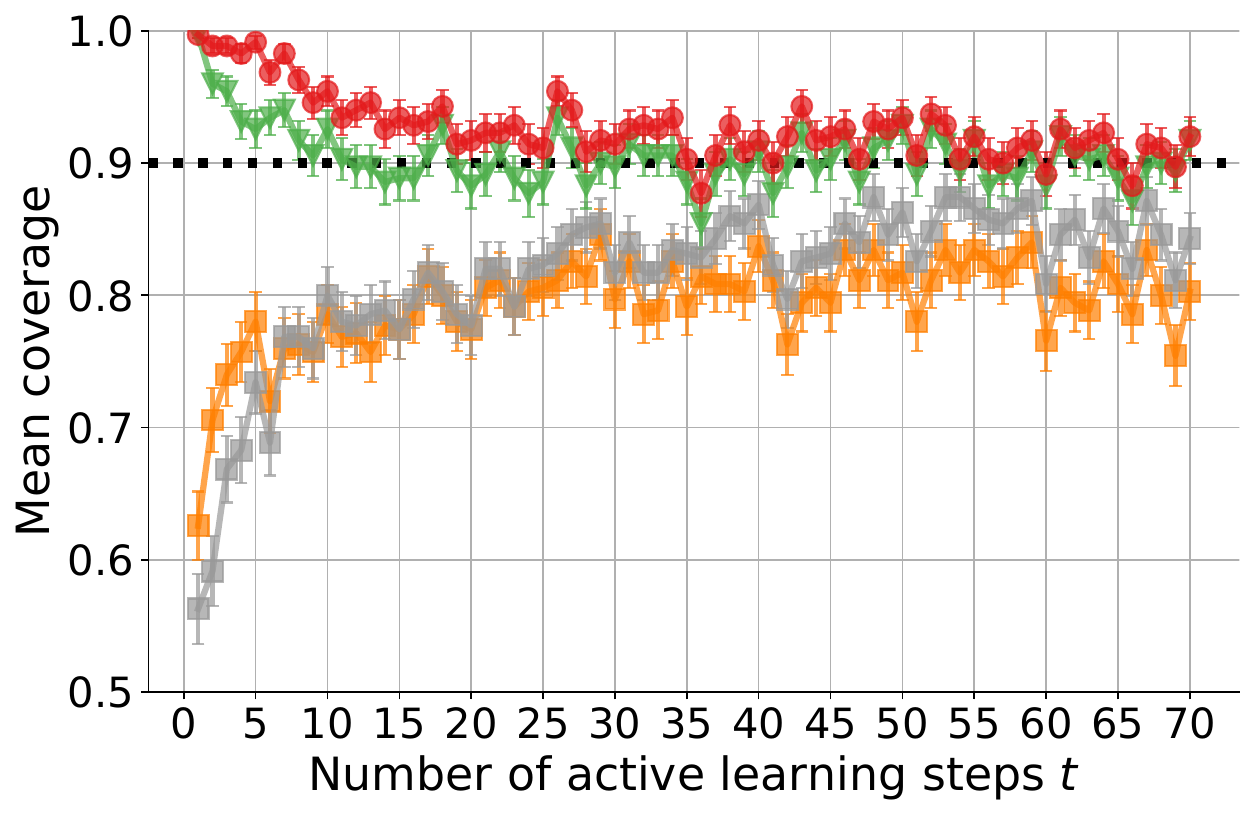}
    \end{subfigure}
    \hfill
    \begin{subfigure}{0.2\textwidth}
        \includegraphics[width=\textwidth]{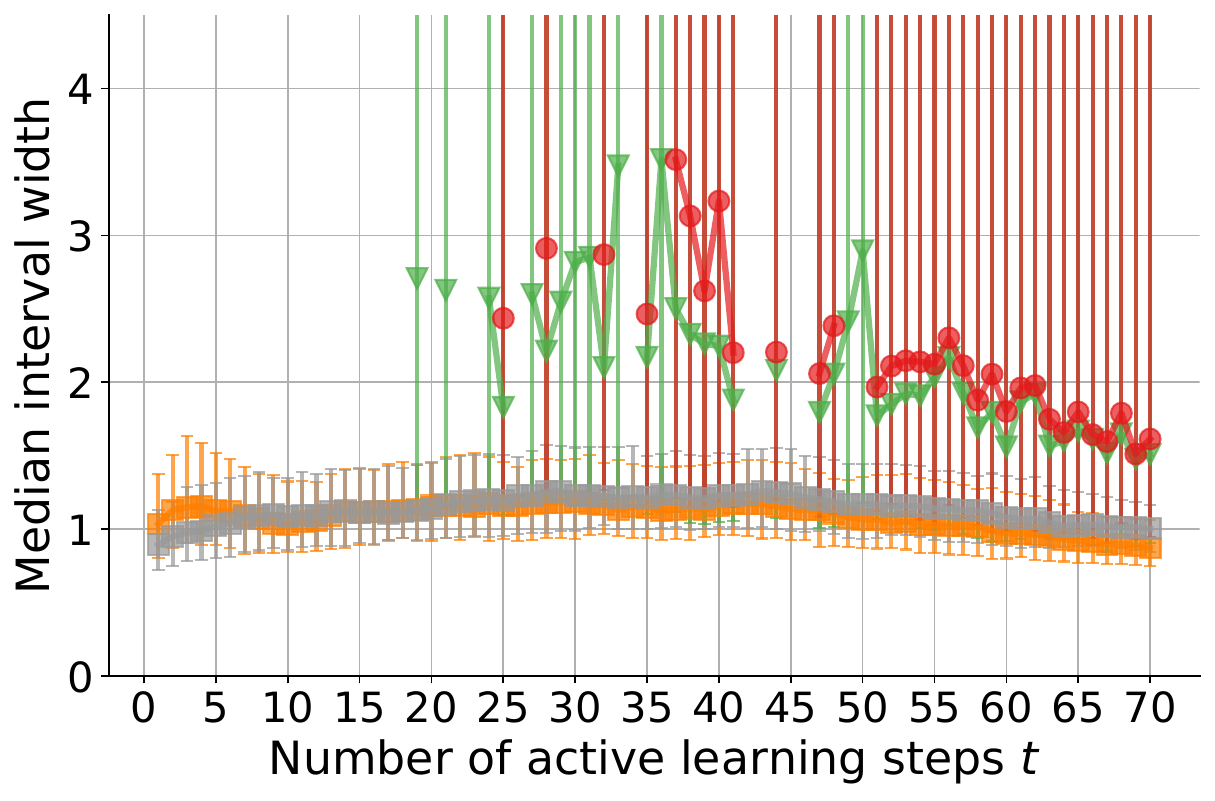}
    \end{subfigure}
    \hfill
    \begin{subfigure}{0.2\textwidth}
        \includegraphics[width=\textwidth]{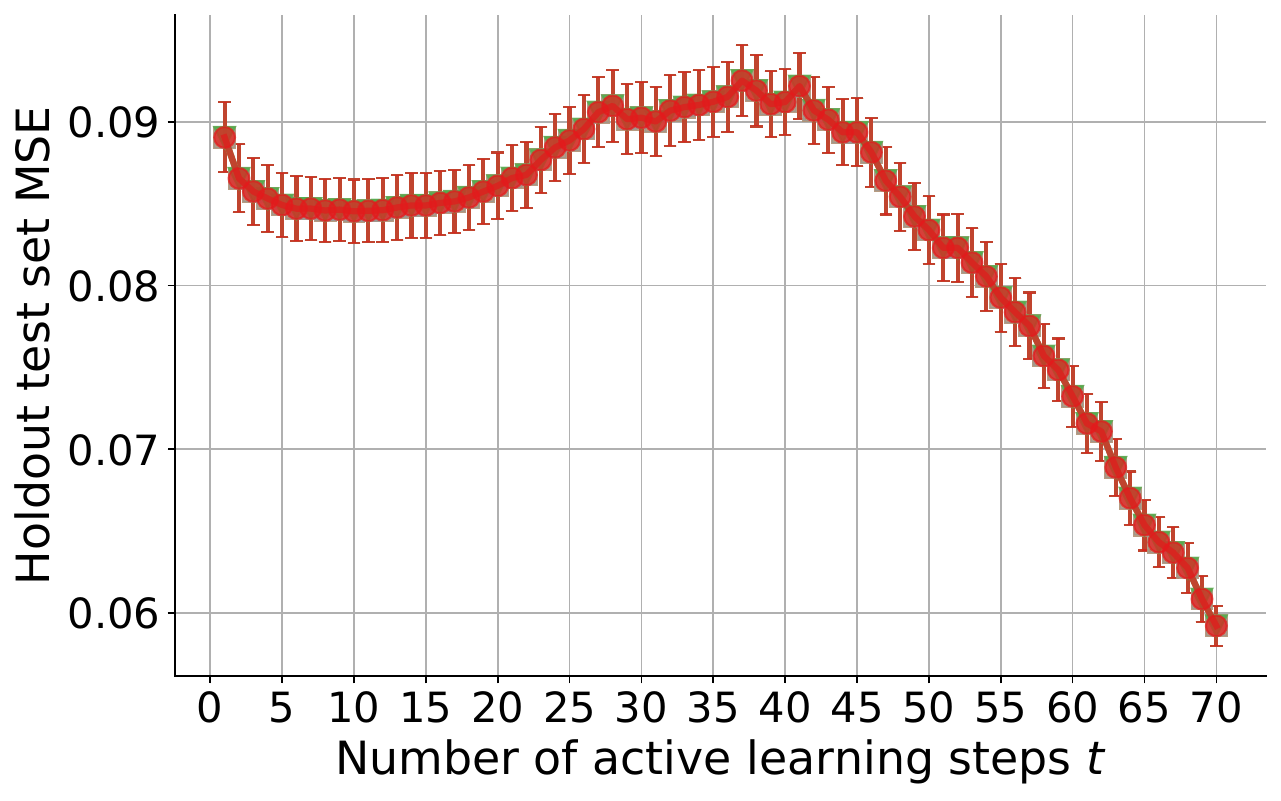}
    \end{subfigure}
    \hfill
    \begin{subfigure}{0.2\textwidth}
        \hfill
    \end{subfigure}
    \hfill
    \\
    \begin{subfigure}{0.4\textwidth}\includegraphics[width=\textwidth]{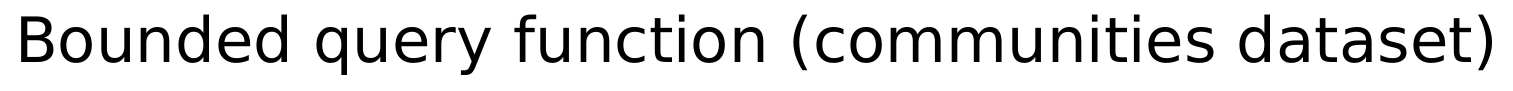}
    \end{subfigure}
    \\
    \hfill
    \begin{subfigure}{0.2\textwidth}
        \includegraphics[width=\textwidth]{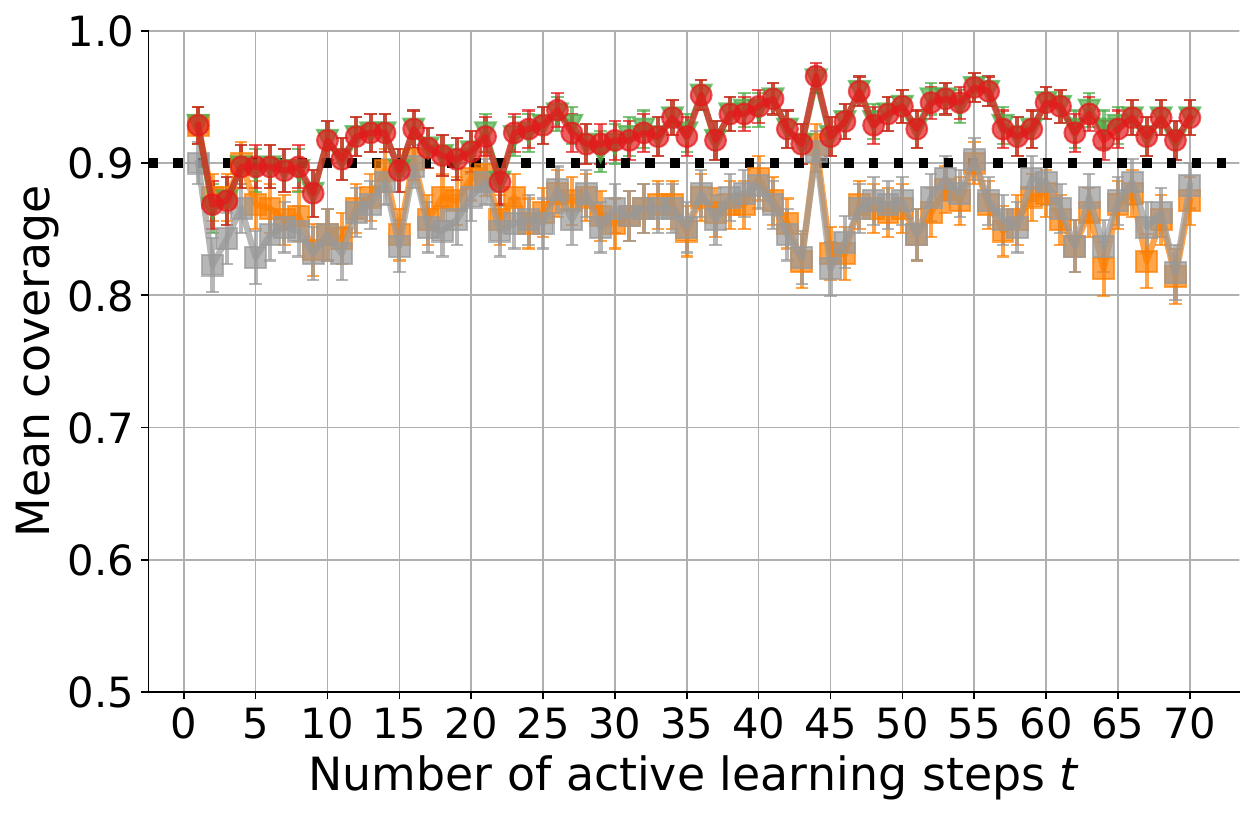}
    \end{subfigure}
    \hfill
    \begin{subfigure}{0.2\textwidth}
        \includegraphics[width=\textwidth]{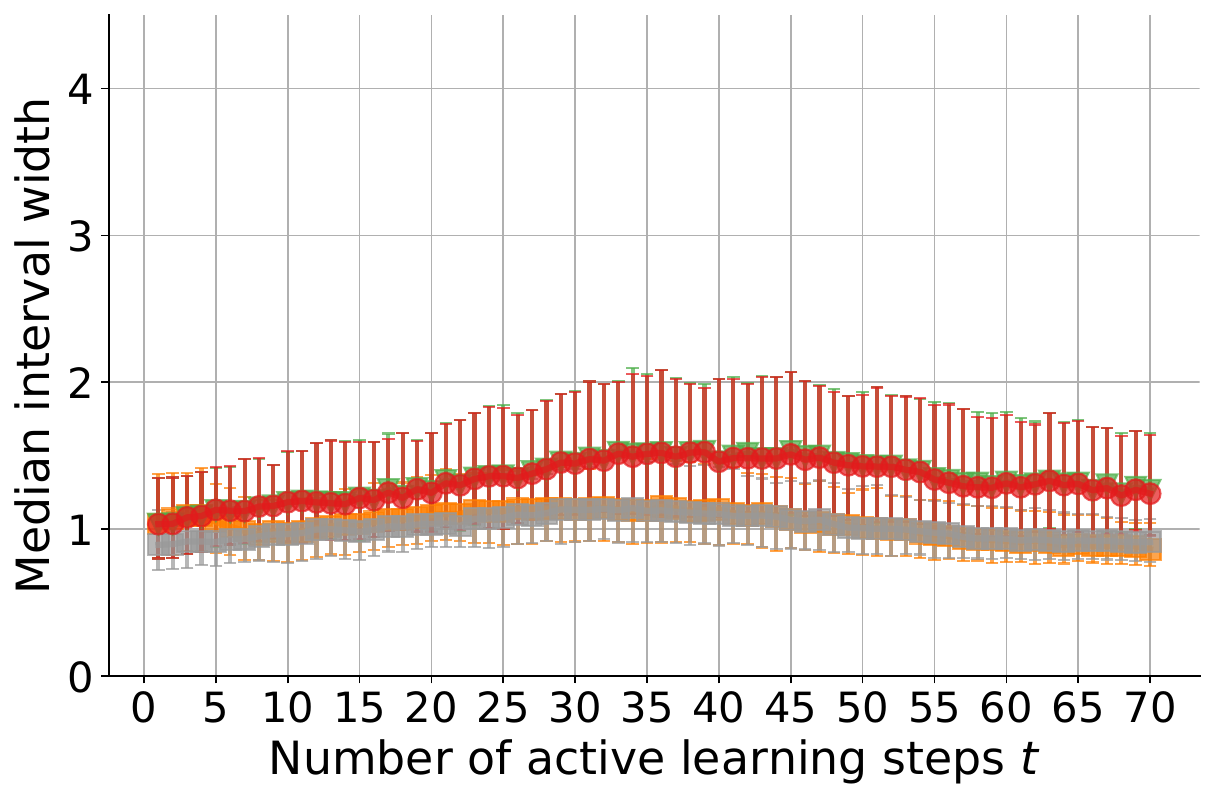}
    \end{subfigure}
    \hfill
    \begin{subfigure}{0.2\textwidth}
        \includegraphics[width=\textwidth]{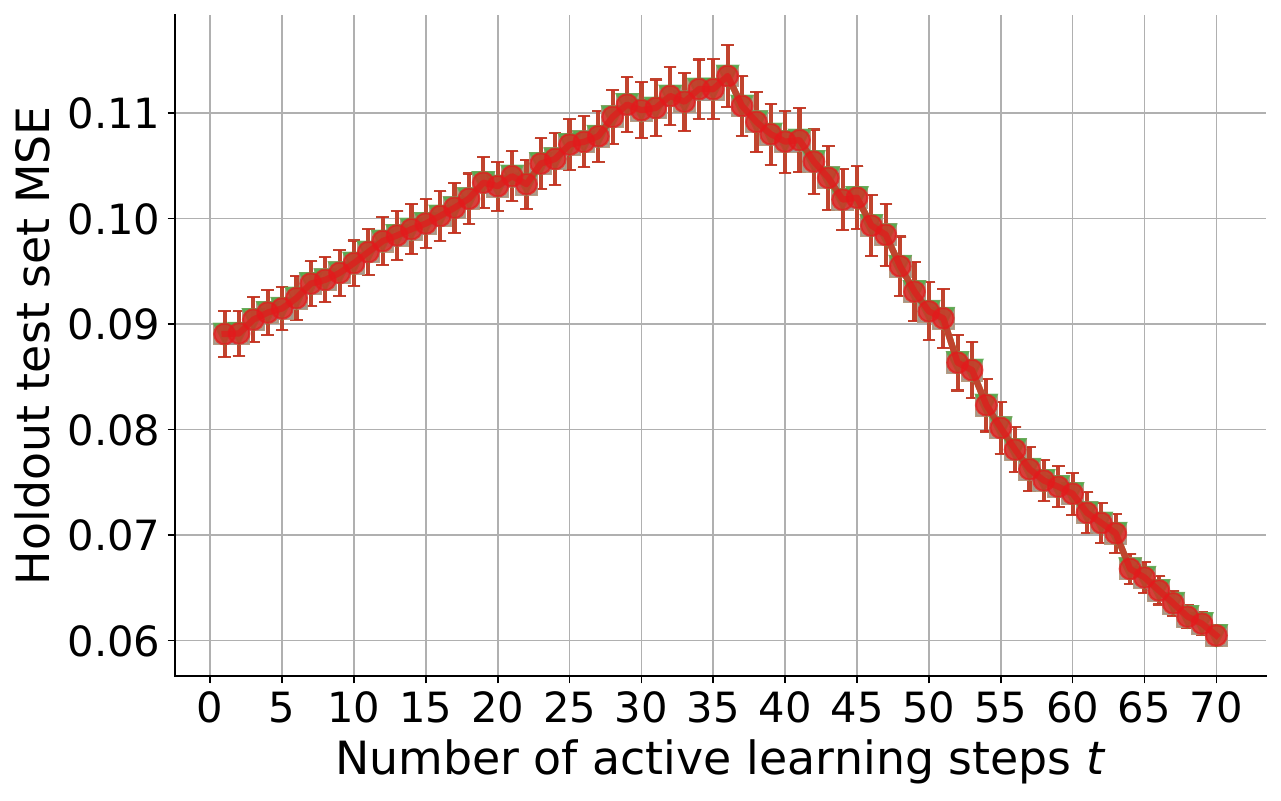}
    \end{subfigure}
    \hfill
    \begin{subfigure}{0.2\textwidth}
        \includegraphics[width=\textwidth]{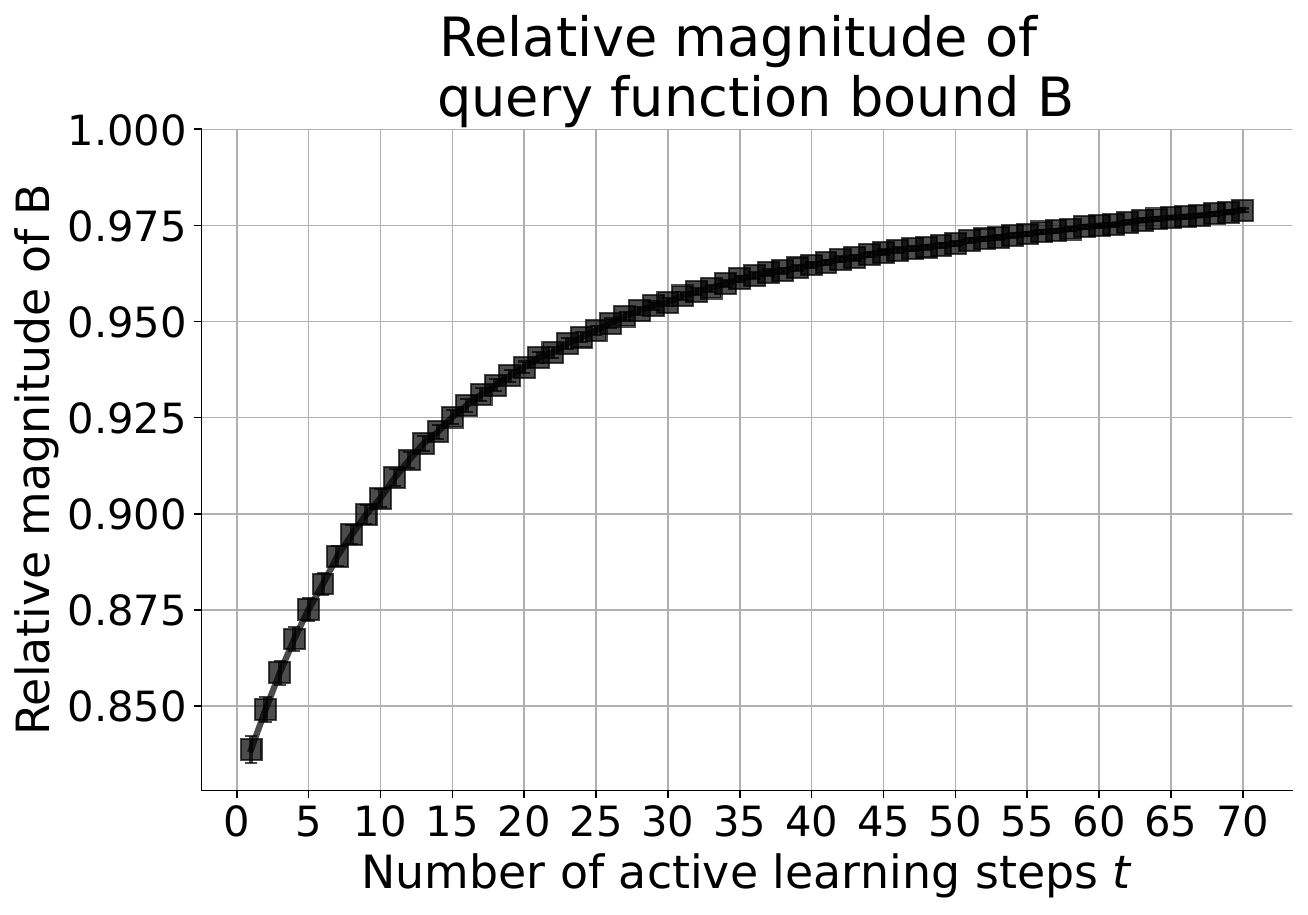}
    \end{subfigure}
    \hfill
    \\
\caption{Unbounded (top row) versus bounded (bottom row) active learning experiments of proposed multistep split CP method for $d=3$ (red circles) compared to baselines of unweighted split CP (orange squares), one-step split CP (green triangles), and ACI (gray squares) on the communities dataset.
    The Y-axes represent mean coverage, median interval width, and mean squared error on a holdout test set; the X-axes correspond to the number of active learning query steps, with each query based on posterior variance of a GP regressor.
    All values are computed over 350 distinct random seeds. Hyperparameters for the experiments are given in Appendix \ref{subsec:active_learning_exp_details} Table \ref{tab:fig3_hyperparams}.
    }
\label{appfig:BoundedActiveLearningExpts_communities}
\end{figure*}

\textbf{Medical Expenditure Panel Survey (MEPS) dataset}

\begin{figure*}[!hb]
    \centering
    \begin{subfigure}{0.85\textwidth}\includegraphics[width=\textwidth]{figures/SplitCPActiveLearning_bounded_legend.pdf}
    \end{subfigure}
    \\
    \begin{subfigure}{0.35\textwidth}\includegraphics[width=\textwidth]{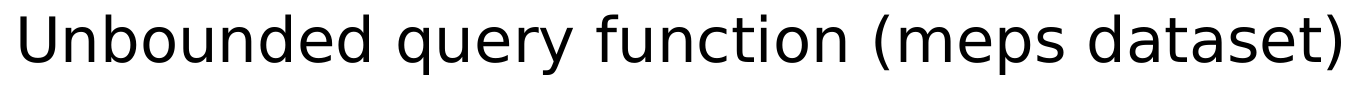}
    \end{subfigure}
    \\
    \hfill
    \begin{subfigure}{0.2\textwidth}
        \includegraphics[width=\textwidth]{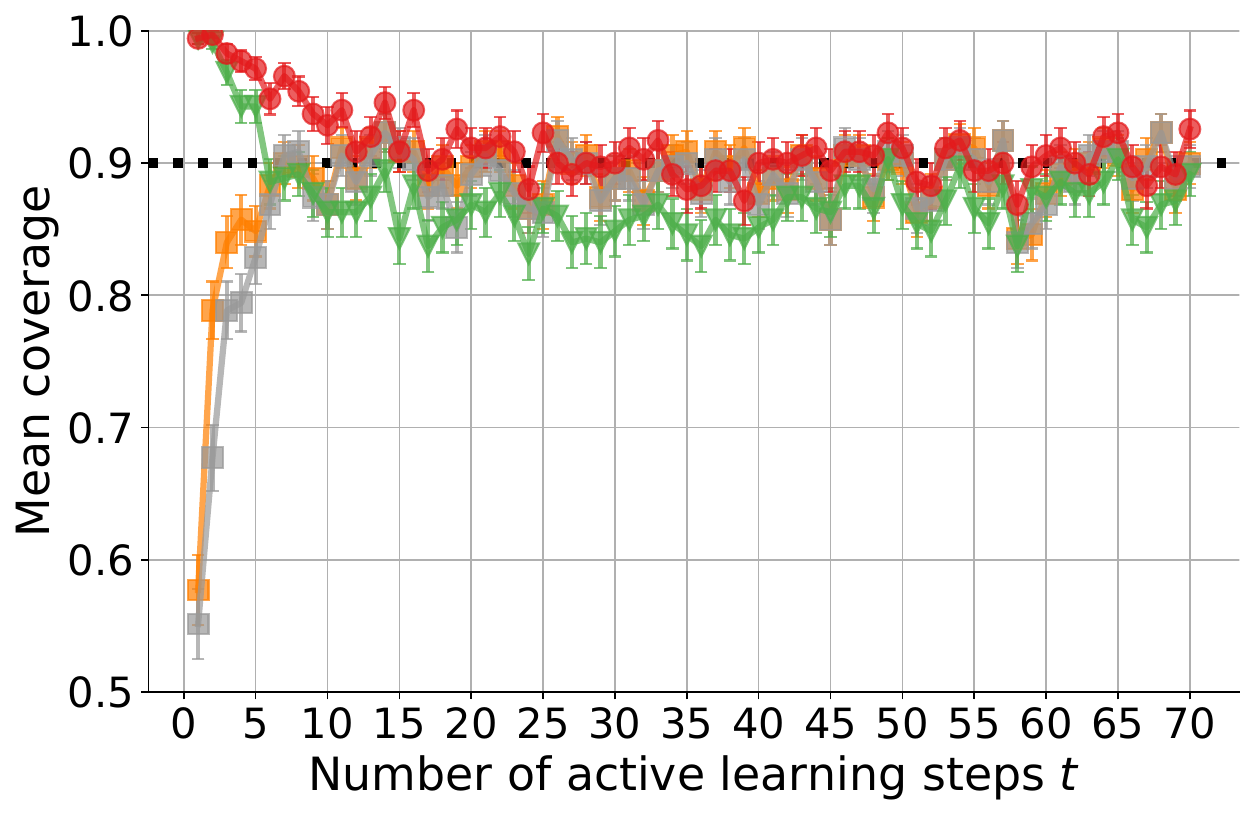}
    \end{subfigure}
    \hfill
    \begin{subfigure}{0.2\textwidth}
        \includegraphics[width=\textwidth]{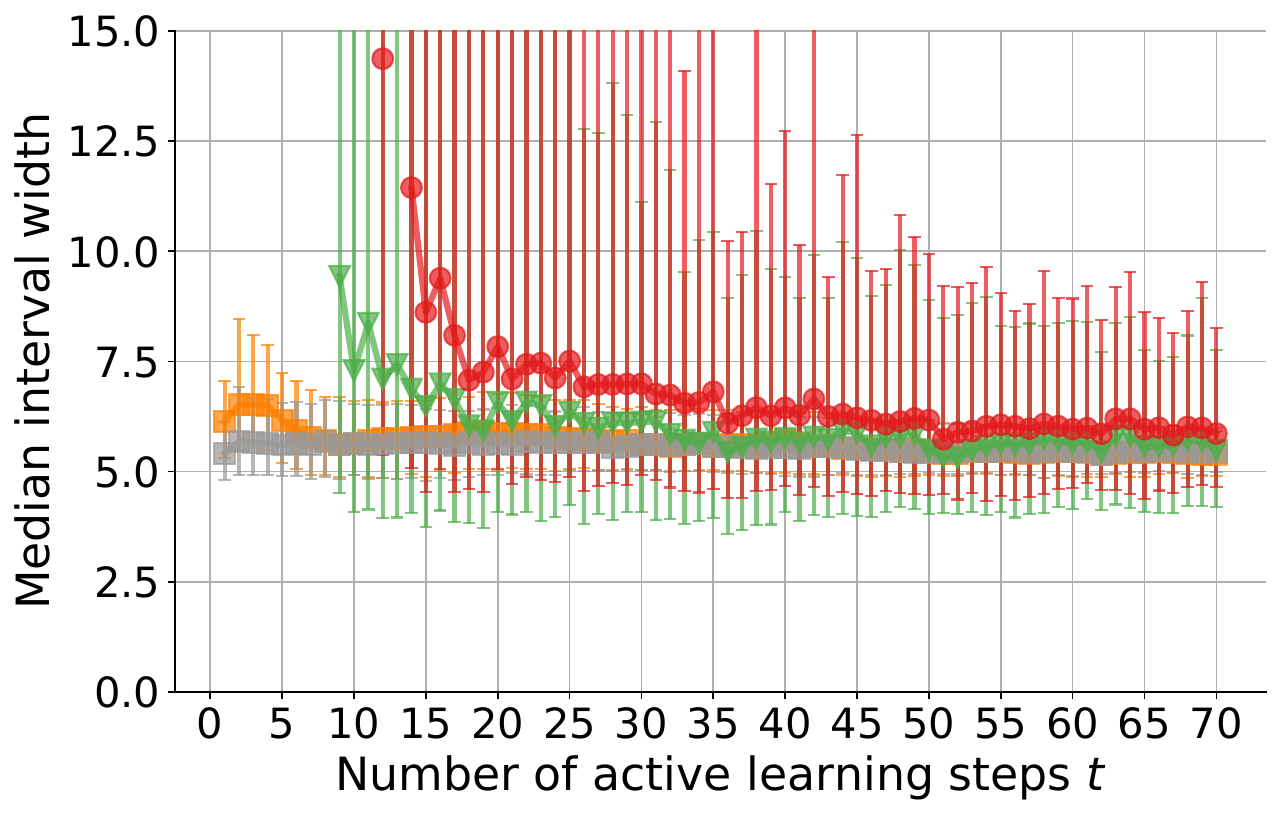}
    \end{subfigure}
    \hfill
    \begin{subfigure}{0.2\textwidth}
        \includegraphics[width=\textwidth]{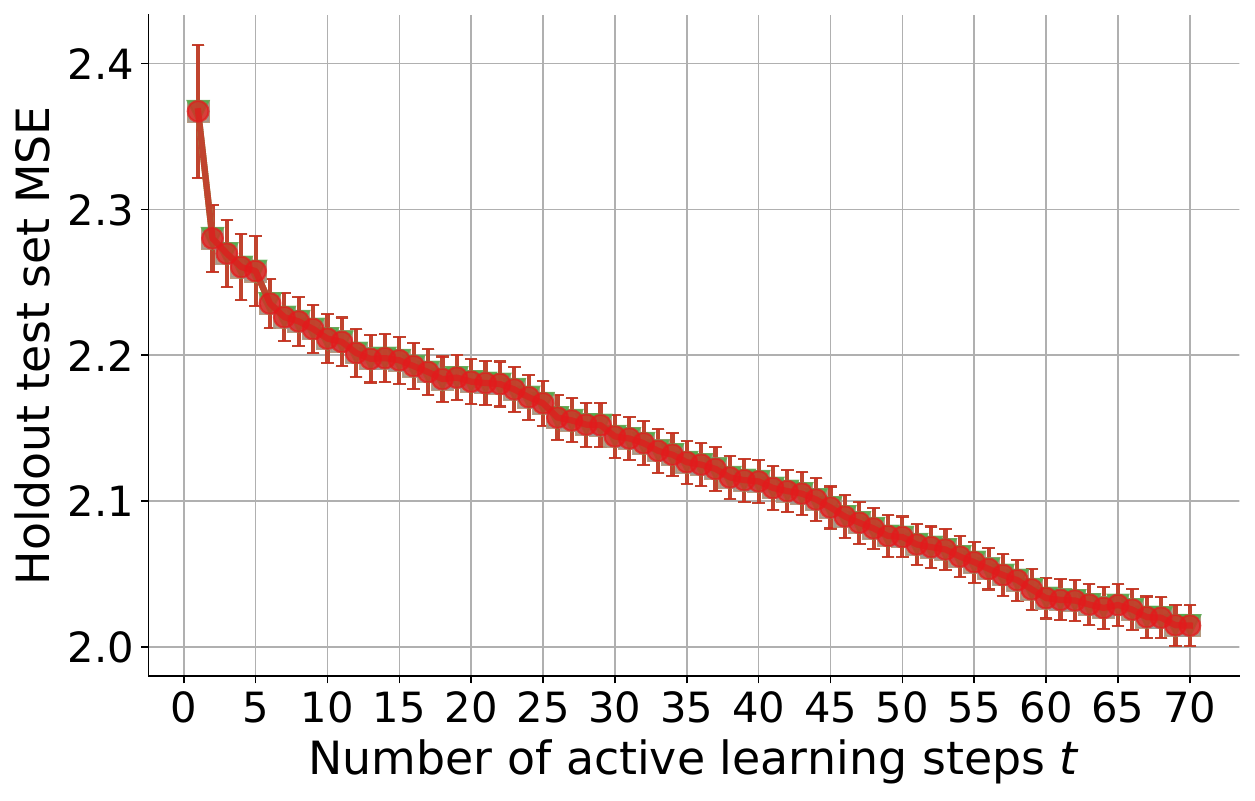}
    \end{subfigure}
    \hfill
    \begin{subfigure}{0.2\textwidth}
        \hfill
    \end{subfigure}
    \hfill
    \\
    \begin{subfigure}{0.35\textwidth}\includegraphics[width=\textwidth]{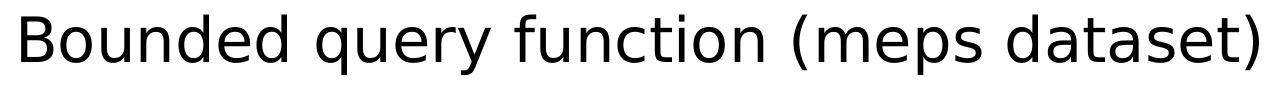}
    \end{subfigure}
    \\
    \hfill
    \begin{subfigure}{0.2\textwidth}
        \includegraphics[width=\textwidth]{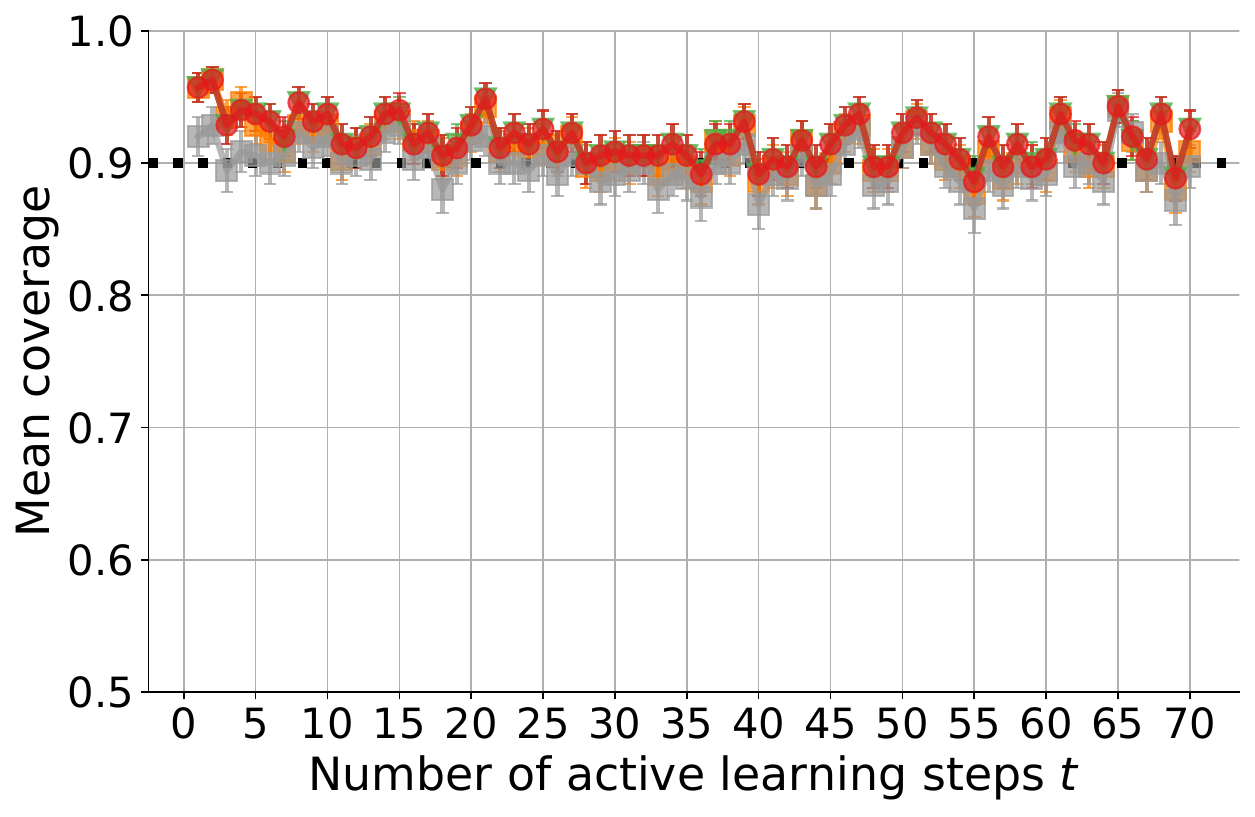}
    \end{subfigure}
    \hfill
    \begin{subfigure}{0.2\textwidth}
        \includegraphics[width=\textwidth]{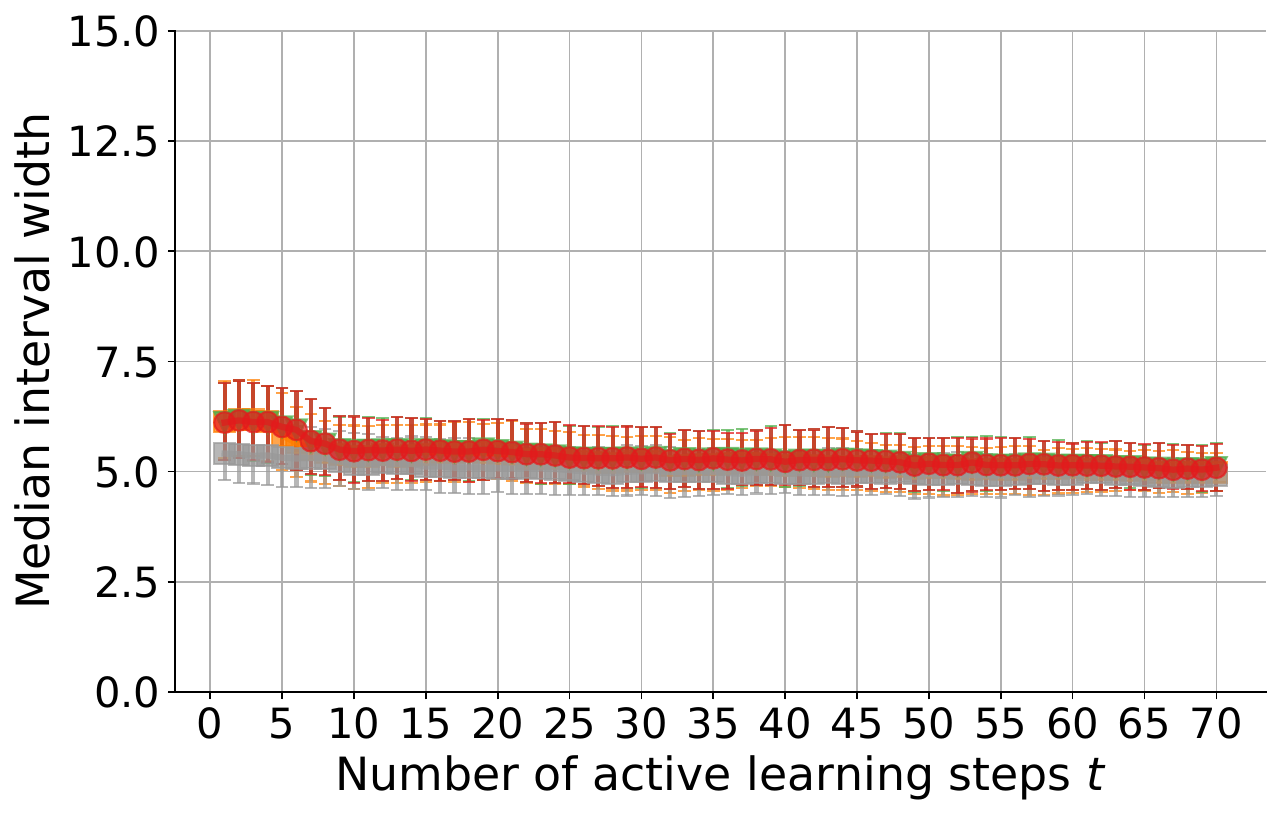}
    \end{subfigure}
    \hfill
    \begin{subfigure}{0.2\textwidth}
        \includegraphics[width=\textwidth]{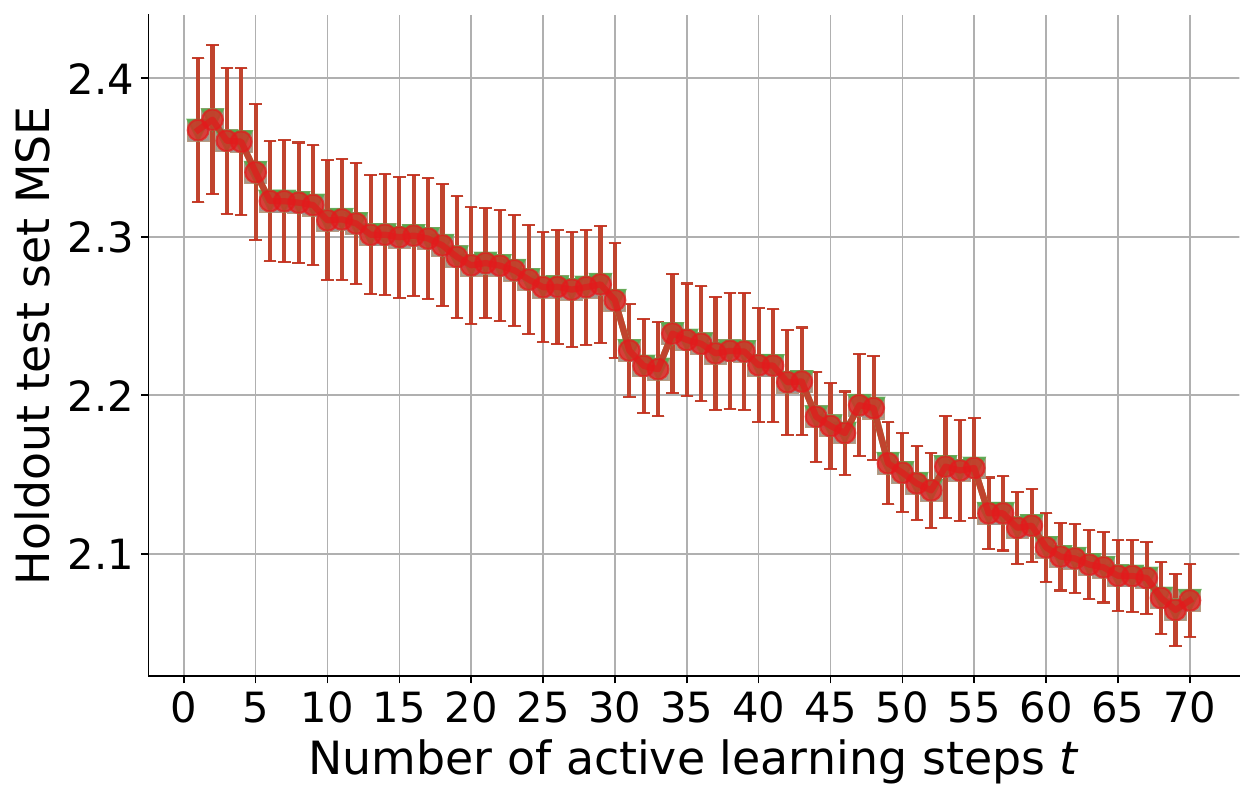}
    \end{subfigure}
    \hfill
    \begin{subfigure}{0.2\textwidth}
        \includegraphics[width=\textwidth]{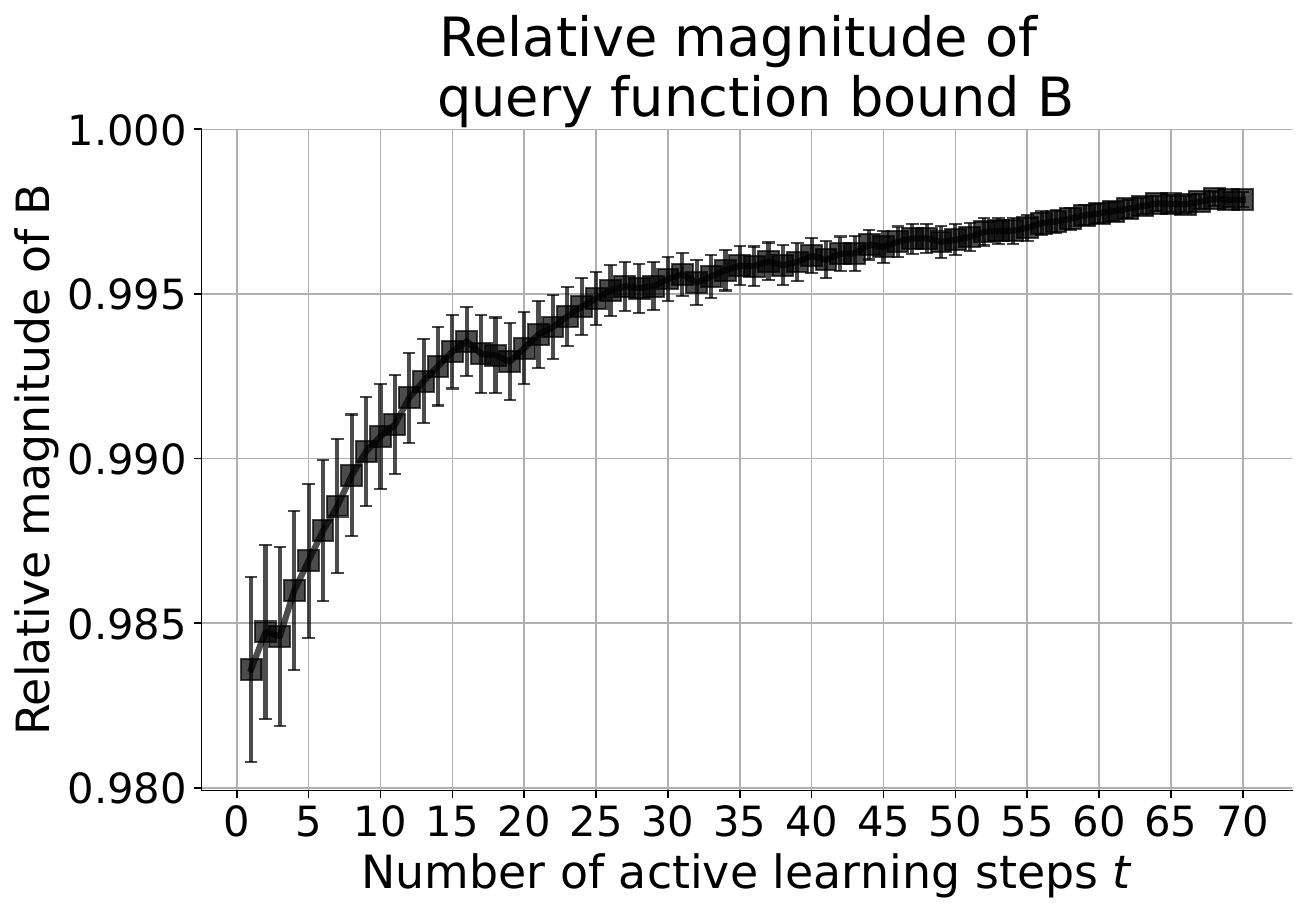}
    \end{subfigure}
    \hfill
    \\
\caption{Unbounded (top row) versus bounded (bottom row) active learning experiments with proposed multistep split CP method for estimation depth $d=3$ (red circles) compared to baselines of unweighted split CP (orange squares), one-step split CP (green triangles), and ACI (gray squares) on the Medical Expenditure Panel Survey (MEPS) dataset.
    The Y-axes represent mean coverage, median interval width, mean squared error on a holdout test set, and the relative bound magnitude $B/B_{\text{max}}$; the X-axes correspond to the number of active learning query steps, with each query based on posterior variance of a GP regressor. All values are computed over 350 distinct random seeds. Hyperparameters for the experiments are given in Appendix \ref{subsec:active_learning_exp_details} Table \ref{tab:fig3_hyperparams}.
    }
\label{appfig:BoundedActiveLearningExpts_meps}
\end{figure*}

\newpage

\textbf{Blog dataset}

\begin{figure*}[h]
    \centering
    \begin{subfigure}{0.85\textwidth}\includegraphics[width=\textwidth]{figures/SplitCPActiveLearning_bounded_legend.pdf}
    \end{subfigure}
    \\
    \begin{subfigure}{0.35\textwidth}\includegraphics[width=\textwidth]{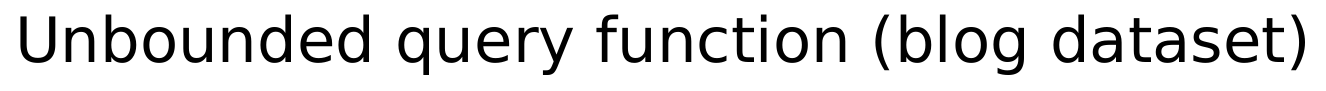}
    \end{subfigure}
    \\
    \hfill
    \begin{subfigure}{0.2\textwidth}
        \includegraphics[width=\textwidth]{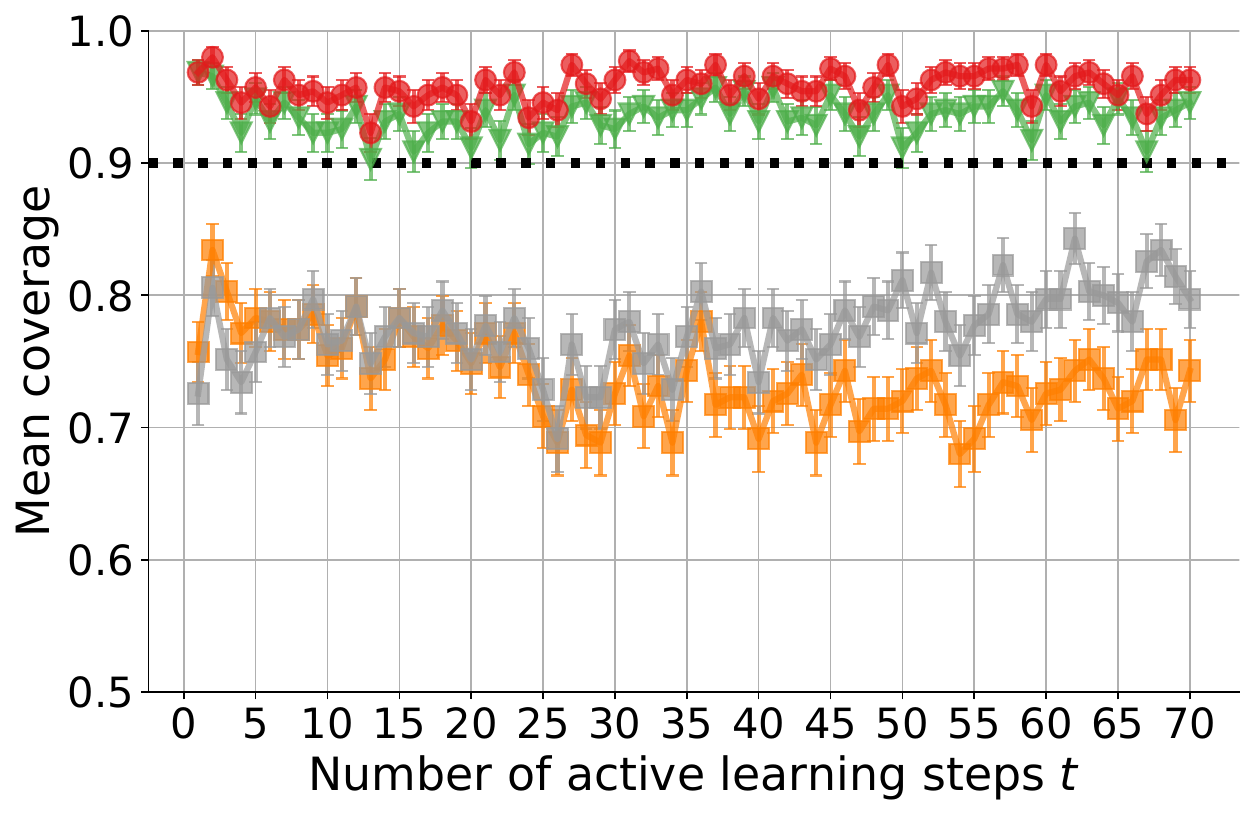}
    \end{subfigure}
    \hfill
    \begin{subfigure}{0.2\textwidth}
        \includegraphics[width=\textwidth]{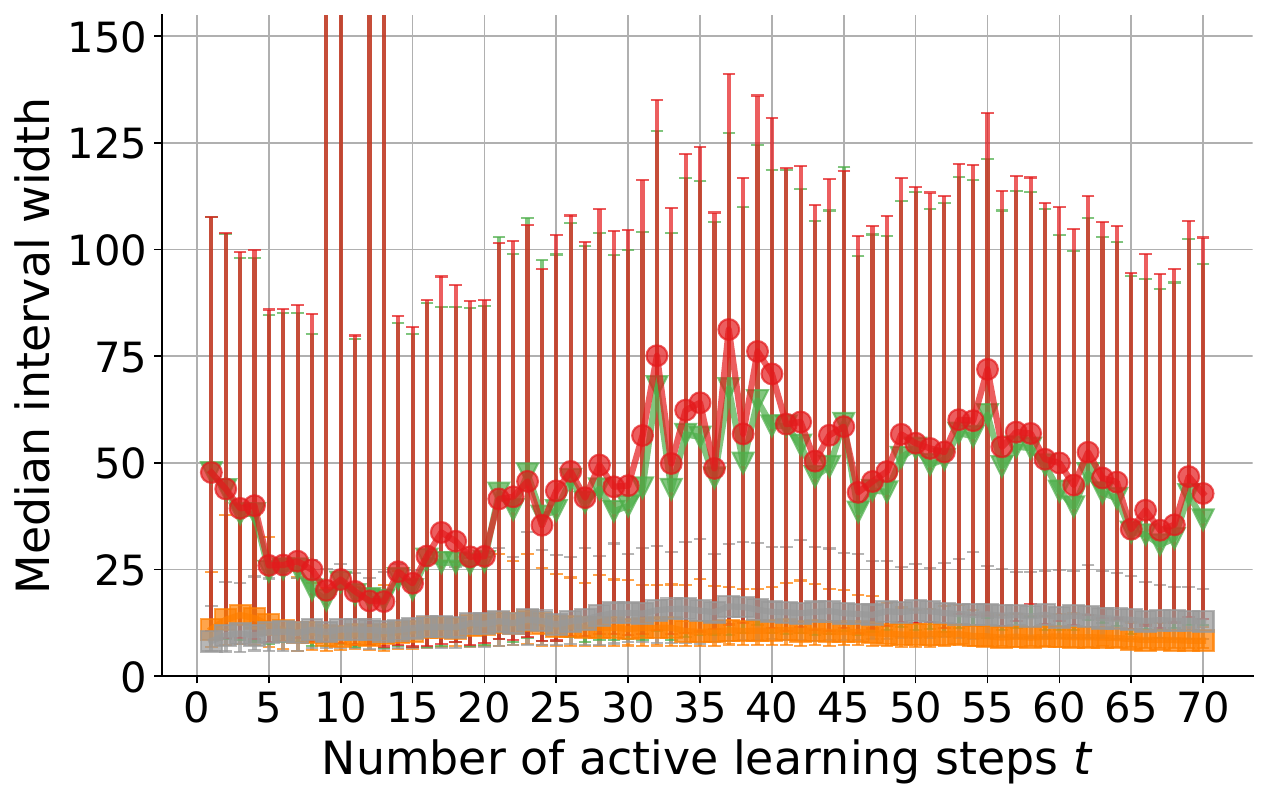}
    \end{subfigure}
    \hfill
    \begin{subfigure}{0.2\textwidth}
        \includegraphics[width=\textwidth]{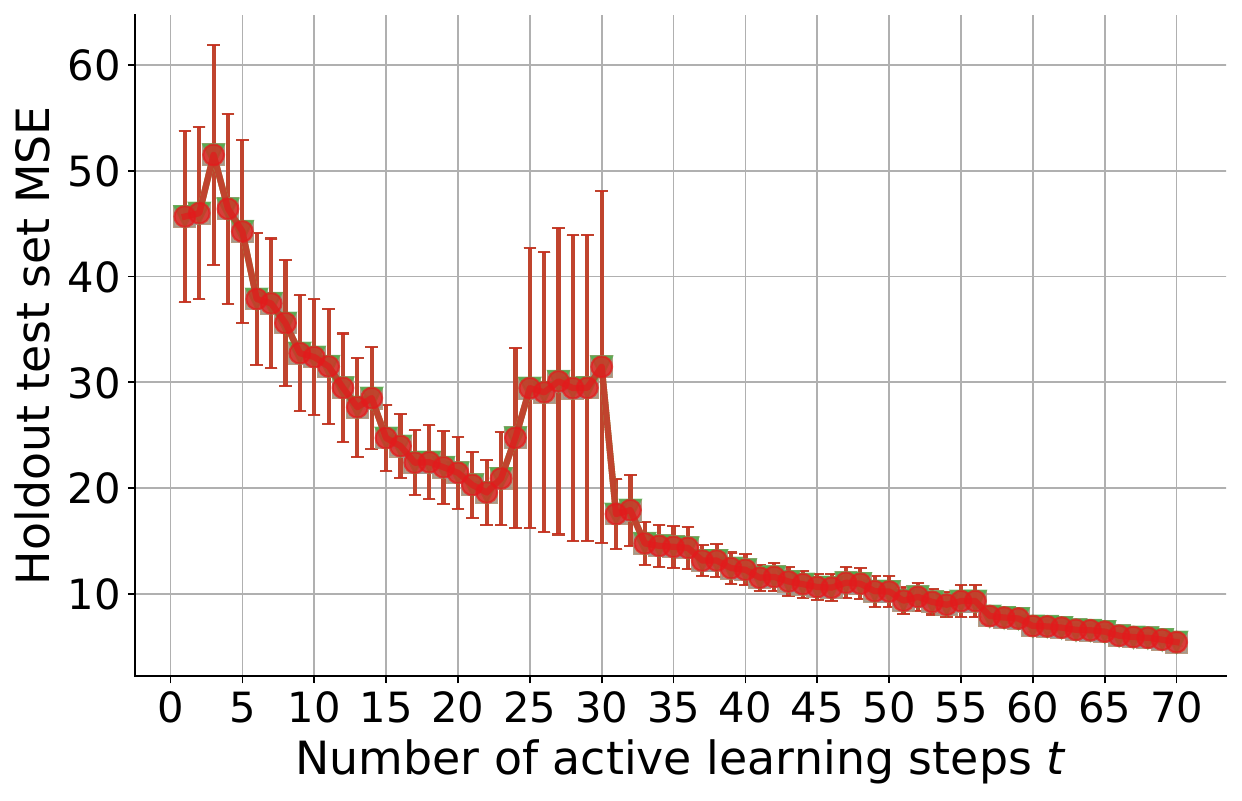}
    \end{subfigure}
    \hfill
    \begin{subfigure}{0.2\textwidth}
        \hfill
    \end{subfigure}
    \hfill
    \\
    \begin{subfigure}{0.35\textwidth}\includegraphics[width=\textwidth]{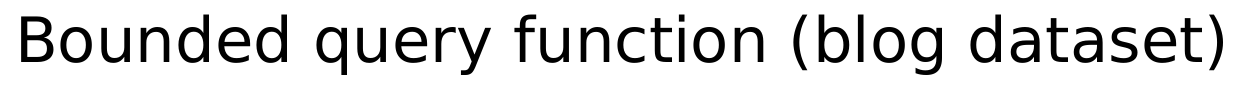}
    \end{subfigure}
    \\
    \hfill
    \begin{subfigure}{0.2\textwidth}
        \includegraphics[width=\textwidth]{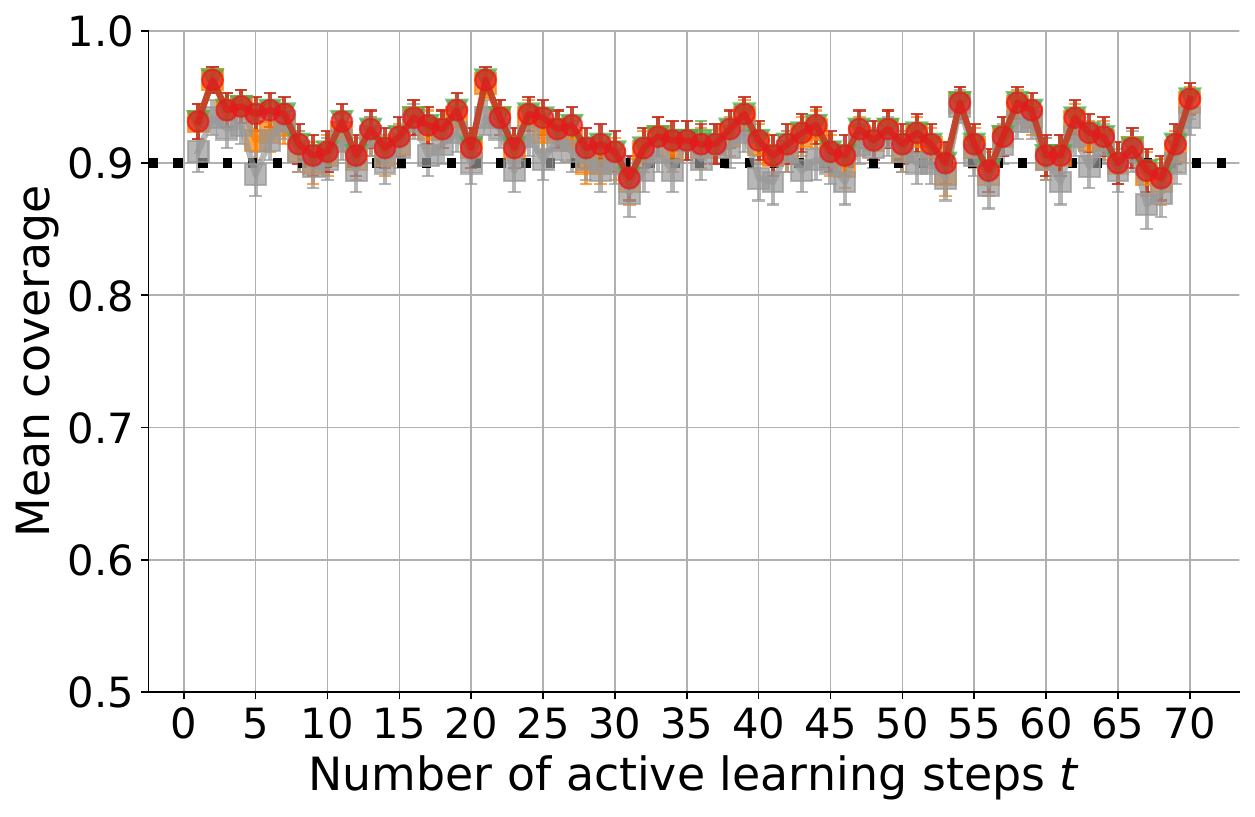}
    \end{subfigure}
    \hfill
    \begin{subfigure}{0.2\textwidth}
        \includegraphics[width=\textwidth]{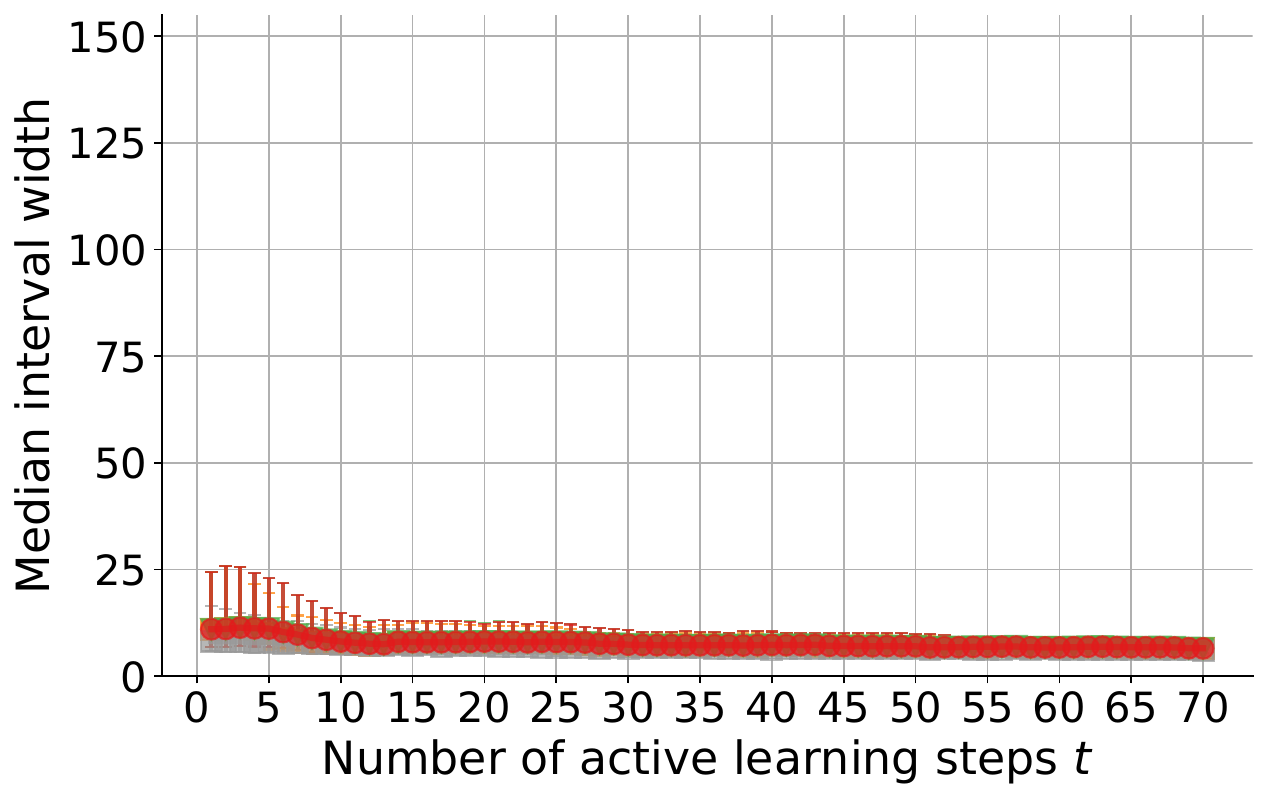}
    \end{subfigure}
    \hfill
    \begin{subfigure}{0.2\textwidth}
        \includegraphics[width=\textwidth]{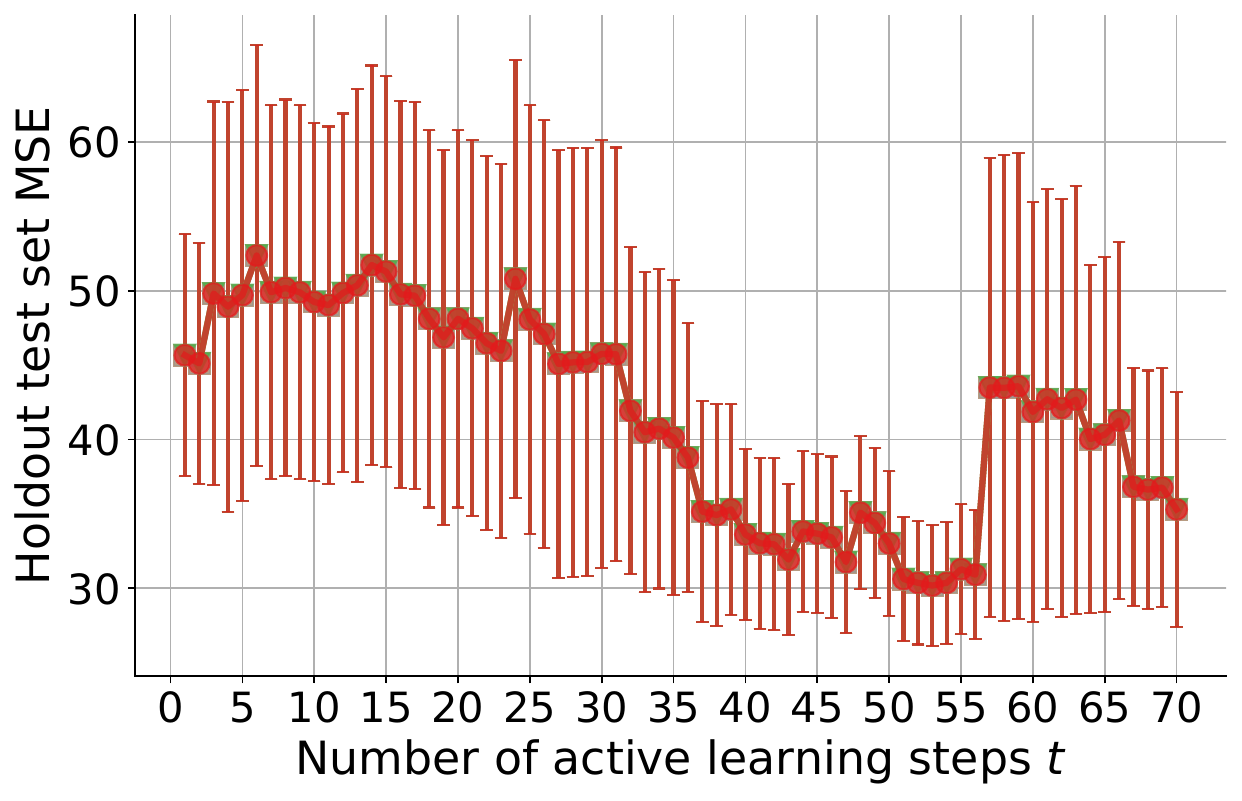}
    \end{subfigure}
    \hfill
    \begin{subfigure}{0.2\textwidth}
        \includegraphics[width=\textwidth]{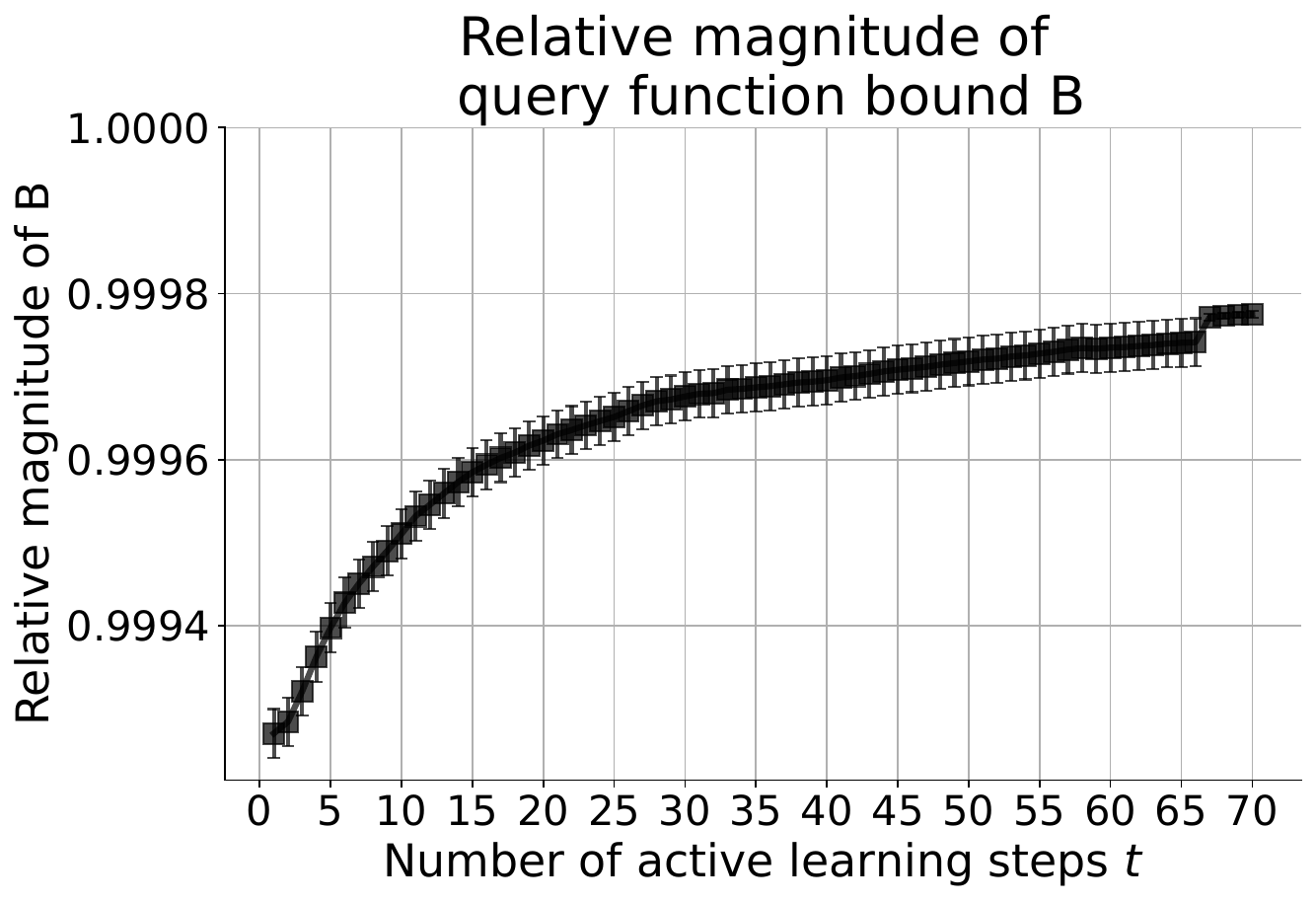}
    \end{subfigure}
    \hfill
    \\
\caption{Unbounded (top row) versus bounded (bottom row) active learning experiments of proposed multistep split CP method for $d=3$ (red circles) compared to baselines of unweighted split CP (orange squares), one-step split CP (green triangles), and ACI (gray squares) on the blog dataset.
    The Y-axes represent mean coverage, median interval width, and mean squared error on a holdout test set; the X-axes correspond to the number of active learning query steps, with each query based on posterior variance of a GP regressor.
    All values are computed over 350 distinct random seeds. Hyperparameters for the experiments are given in Appendix \ref{subsec:active_learning_exp_details} Table \ref{tab:fig3_hyperparams}.
    }
\label{appfig:BoundedActiveLearningExpts_blog}
\end{figure*}

The bounded query function’s improvement in prediction-interval informativeness (smaller prediction intervals relative to the unbounded query function) sometimes comes at the cost of an initially delayed decrease in MSE, though the long-run MSE for the bounded and unbounded query functions are comparable for most datasets. The bounded query function can thus be thought of as an ``informativeness constraint'' that forces the AI/ML agent to initially be more ``cautious'' or ``slow'' in its exploration until it has accumulated enough training and calibration data to explore more aggressively (as evidenced by the increasing bound magnitude over time).

\newpage

\section{Details on the Roles of Exchangeability, Weighted Exchangeability, and Pseudo-Exchangeability in Conformal Prediction}

\label{sec:roles_exchangeability_extensions}


\subsection{The Role of Exchangeability in Conformal Prediction}

\label{app:exchangeability_role}

\textbf{Summary:} Standard conformal prediction can be viewed as an inverted permutation test, where exchangeability allows us to ``weigh'' or count each permutation equally.

Appendix A.6 of \citet{tibshirani2019conformal} provides an alternate proof to a key Lemma underlying (standard) conformal prediction's validity, which corresponds to viewing conformal prediction as an inverted permutation test (though they do not explicitly state this). We reproduce most of that argument here, with some additional explanatory commentary to highlight the role played by the exchangeability assumption in standard conformal prediction.

The general strategy in \citet{tibshirani2019conformal} is ``to condition on the unlabeled multiset of values $\{v_1, ..., v_{n+1}\}$ obtained by our random variables $V_1, ..., V_{n+1}$, and then inspect the probabilities that the last random variable $V_{n+1}$ attains each one of these values.'' That is, \citet{tibshirani2019conformal} condition on the event $\{V_1, ..., V_{n+1}\}=\{v_1, ..., v_{n+1}\}$, which they denote by $E_v$---recall that this is \textit{not} saying that $V_i=v_i$ for all $i \in \{1, ..., n+1\}$. Rather, it is only saying that the set or ``bag''\footnote{For simplicity, \citet{tibshirani2019conformal} assume there are almost surely no ties among those scores (to work with sets rather than multisets/bags); the argument applies in the general case but with more cumbersome notation.} of values $\{v_1, ..., v_{n+1}\}$ are the values obtained by the random variables $V_1, ..., V_{n+1}$; we do not yet know whether $v_1$ is the value obtained by $V_1$, or by $V_2$, or by $V_{n+1}$, etc. 

With this notation, \citet{tibshirani2019conformal} consider the following probability:
\begin{align*}
    \mathbb{P}\{V_{n+1}=v_i\ | \ E_v\}, i = 1, ..., n+1, 
\end{align*}
that is, the probability that the value obtained by $V_{n+1}$ is $v_i$, conditioned on $E_v$ (conditioned on the bag of values). 

Next is where \citet{tibshirani2019conformal} implicitly use an inverted permutation test. With $f$ denoting the joint density function, they state the following, which is analogous to Eq. \eqref{eq:summary_perm_test} in our main paper:
\begin{align*}
    \mathbb{P}\{V_{n+1}=v_i\ | \ E_v\} = \frac{\sum_{\sigma:\sigma(n+1)=i}f(v_{\sigma(1)}, ..., v_{\sigma(n+1)})}{\sum_{\sigma}f(v_{\sigma(1)}, ..., v_{\sigma(n+1)})}.
\end{align*}
This statement can be thought of as an inverted permutation test as it equates the probability $\mathbb{P}\{V_{n+1}=v_i\ | \ E_v\}$ to the total probability mass of all permutations $\sigma$ that map $n+1$ to $i$, normalized across all possible permutations of the $n+1$ values. For intuition, more explicit notation with random variables can be helpful: That is, if $f$ is a PDF, then writing $f(v_{\sigma(1)}, ..., v_{\sigma(n+1)}) = f_{V_1, ..., V_{n+1}}(v_{\sigma(1)}, ..., v_{\sigma(n+1)})$; if $f$ is a PMF, then writing $f(v_{\sigma(1)}, ..., v_{\sigma(n+1)}) = f(V_1 = v_{\sigma(1)}, ..., V_{n+1}= v_{\sigma(n+1)})$.
In either case, writing the random variables explicitly can help to emphasize that each argument in $f$ still corresponds to the index of the \textit{random variable}; meanwhile, the \textit{observed values} are being permuted to different argument locations.

Next is where exchangeability comes in. That is, exchangeability means $f(v_{\sigma(1)}, ..., v_{\sigma(n+1)}) = f(v_{1}, ..., v_{n+1})$ for permutations $\sigma$ of the indices $\{1, ..., n+1\}$, so by plugging in this substitution and factoring out $f(v_{1}, ..., v_{n+1})$ from the summation, we can simplify to reduce our statement to counting permutations: 
\begin{align*}
    \mathbb{P}\{V_{n+1}=v_i\ | \ E_v\} & = \frac{\sum_{\sigma:\sigma(n+1)=i}f(v_{1}, ..., v_{n+1})}{\sum_{\sigma}f(v_{1}, ..., v_{n+1})} \\
    & = \frac{f(v_{1}, ..., v_{n+1})\sum_{\sigma:\sigma(n+1)=i}1}{f(v_{1}, ..., v_{n+1})\sum_{\sigma}1} \\
    & = \frac{\sum_{\sigma:\sigma(n+1)=i}1}{\sum_{\sigma}1} \\
    & = \frac{n!}{(n+1)!} = \frac{1}{n+1}.
\end{align*}
That is, the last line follows because (in the denominator's summation) there are $(n+1)!$ possible permutations of $n+1$ values and (in the numerator's summation) there are $n!$ possible permutations of $n$ values (since in the numerator's summation we only consider $\sigma$ where $\sigma(n+1)=1$, there are only $n$ values unknown). In other words, under the assumption of exchangeability, every permutation of the data is equally likely, so the probability of the event $\{V_{n+1}=v_i\ | \ E_v\}$ can be computed as the total number of permutations (of the $n+1$ values) where $\sigma(n+1)=i$, divided by the total number of all possible permutations (of the $n+1$ values).

The remainder of the argument in \citet{tibshirani2019conformal} then follows similarly as in the proof given in Appendix \ref{sec:general_cp_coverage_proof}, but premised on the assumption of exchangeability. That is, \citet{tibshirani2019conformal} explain that the above result shows that the distribution of $V_{n+1}|E_z$ is uniform on the set $\{v_1, ..., v_{n+1}\}$:
\begin{align*}
    V_{n+1}|E_v \sim \frac{1}{n+1}\sum_{i=1}^{n+1}\delta_{v_i},
\end{align*}
where $\delta_{v}$ denotes a point mass at the value $v$. The remaining steps of the proof then follow by the definition of a quantile, by the meaning of conditioning on $E_z$, and by marginalizing (similarly as in Appendix \ref{sec:general_cp_coverage_proof}). The resulting standard CP coverage guarantee can be viewed as a corollary to Theorem \ref{thm:general_CP_def_coverage} because its proof (as presented by \citet{tibshirani2019conformal}) follows the same steps, except also relying on the simplifying assumption of exchangeability to allow us to count every permutation equally.

Lastly, we note that while standard conformal prediction is often defined by taking the level $(1-\alpha)\cdot \tfrac{n+1}{n}$ quantile over the values $\{v_1, ..., v_n\}$, this is equivalent to taking the level $1-\alpha$ quantile of values $\{v_1, ..., v_n\} \cup \{\infty\}$. \citet{tibshirani2019conformal} introduced the latter with the point mass $\delta$ notation we have been using to enable writing a \textit{weighted} empirical distribution of score values. 


\subsection{The Role of Weighted Exchangeability in Conformal Prediction}
\label{app:weighted_exchangeability}

\textbf{Summary}: Weighted CP can be viewed as an inversion of \textit{weighted} permutation tests, where the weight functions tell us how much to ``weigh'' or count each permutation.

First, we restate the definition of weighted exchangeability from \citet{tibshirani2019conformal}:

\textbf{Definition 1:} \textit{Random variables $V_1, ..., V_{n}$ are said to be } weighted exchangeable, \textit{with weight functions $w_1, ..., w_{n}$, if the density\footnote{\citet{tibshirani2019conformal} give the following footnote here: ``As before, $f$ may be the Radon-Nikodym derivative with respect to an arbitrary base measure.''} $f$, of their joint distribution can be factorized as}
\begin{align}
    f(v_1, ..., v_{n}) = \prod_{i=1}^{n}w_i(v_i)\cdot g(v_1, ..., v_{n})
    \label{def:weighted-exchangeability}
\end{align}
\textit{where $g$ is any function that does not depend on the ordering of its inputs, i.e., $g(v_{\sigma(1)}, ..., v_{\sigma(n)}) = g(v_{1}, ..., v_{n})$ for any permutation $\sigma$ of $1, ..., n$.}

We now sketch the argument in appendix A.7 of \citet{tibshirani2019conformal}, which proves a key lemma underlying the validity of weighted CP, with the goal of understanding the role of the weight functions in an inversion of a weighted permutation test.

Similarly as before and as in equation \eqref{eq:summary_perm_test},\footnote{Again, for simplicity, \citet{tibshirani2019conformal} assume that there are almost surely no ties in the values $\{v_1, ..., v_{n+1}\}$ to work with sets rather than multisets/bags, but the results holds more generally (with more cumbersome notation).}
\begin{align*}
    \mathbb{P}\{V_{n+1}=v_i\ | \ E_z\} = & \mathbb{P}\{Z_{n+1}=z_i\ | \ E_z\} \\
    = & \frac{\sum_{\sigma:\sigma(n+1)=i}f(z_{\sigma(1)}, ..., z_{\sigma(n+1)})}{\sum_{\sigma}f(z_{\sigma(1)}, ..., z_{\sigma(n+1)})}.
\end{align*}
Assuming that the data are weighted exchangeability, this becomes
\begin{align*}
    \mathbb{P}\{V_{n+1}=v_i\ | \ E_z\} = & \frac{\sum_{\sigma:\sigma(n+1)=i}\prod_{j=1}^{n+1}w_j(z_{\sigma(j)})g(z_{\sigma(1)}, ..., z_{\sigma(n+1)})}{\sum_{\sigma}\prod_{j=1}^{n+1}w_j(z_{\sigma(j)})g(z_{\sigma(1)}, ..., z_{\sigma(n+1)})}.
\end{align*}
Recalling the definition of weighted exchangeability wherein $g$ is a permutation-invariant function, we are able to substitute $g(z_{\sigma(1)}, ..., z_{\sigma(n+1)}) = g(z_{1}, ..., z_{n+1})$, analogously to how we substituted $f(z_{\sigma(1)}, ..., z_{\sigma(n+1)}) = f(z_{1}, ..., z_{n+1})$ using exchangeability in Appendix \ref{app:exchangeability_role}:

\begin{align*}
    \mathbb{P}\{V_{n+1}=v_i\ | \ E_z\} = & \frac{\sum_{\sigma:\sigma(n+1)=i}\prod_{j=1}^{n+1}w_j( z_{\sigma(j)})g(z_{1}, ..., z_{n+1})}{\sum_{\sigma}\prod_{j=1}^{n+1}w_j(z_{\sigma(j)})g( z_{1}, ...,  z_{n+1})} \\
     = & \frac{\sum_{\sigma:\sigma(n+1)=i}\prod_{j=1}^{n+1}w_j(z_{\sigma(j)})}{\sum_{\sigma}\prod_{j=1}^{n+1}w_j(z_{\sigma(j)})}.
\end{align*}

From this simplified statement, it is relatively easier to see that the 
product $\prod_{j=1}^{n+1}w_j(z_{\sigma(j)})$ can be thought of as how much more likely it would be to observe $Z_1=z_{\sigma(1)}, ..., Z_{n+1}=z_{\sigma(n+1)}$ (given $E_z$), relative to the measure that our core function $g$ assigns to this event---i.e., relative to $g(z_{\sigma(1)}, ..., z_{\sigma(n+1)}) = g(z_{1}, ..., z_{n+1})$. Put more simply, $\prod_{j=1}^{n+1}w_j(z_{\sigma(j)})$ tells us how much we should ``weigh'' or ``count'' a given permutation $\sigma$.

\subsection{The Role of Pseudo-Exchangeability in Conformal Prediction}

\label{app:pseudo_diff_def}

\textbf{Summary}: Pseudo-exchangeability, as defined in \citet{fannjiang2022conformal}, has a similar role as weighted exchangeability but with ``factor functions'' in the place of weight functions, and where the factor functions are written to explicitly depend on other datapoints. While pseudo-exchangeability describes \textit{one-step} feedback covariate shift as a special case (as it was seemingly introduced mainly for this purpose), it does not formally describe MFCS, as an artifact of the definition. 

First, we recall the definition of pseudo-exchangeability from (the appendix of) \citet{fannjiang2022conformal}:

\textbf{Definition 1:} \textit{Random variables $V_1, ..., V_{n+1}$ are } pseudo-exchangeable \textit{with factor functions $g_1, ..., g_{n+1}$ and core function $h$ if the density, $f$, of their joint distribution can be factorized as}
\begin{align}
    f(v_1, ..., v_{n+1}) = \prod_{i=1}^{n+1}g_i(v_i;v_{-i})\cdot h(v_1, ..., v_{n+1})
    \label{def:pseudo-exchangeability}
\end{align}
\textit{where $v_{-i}=v_{1:(n+1)}\backslash v_i$, each $g_i(\cdot ; v_{-i})$ is a function that depends on the multiset $v_{-i}$ (that is, on the values in $v_{-i}$ but not on their ordering), and $h$ is a function that does not depend on the ordering of its $n+1$ inputs.}




Note that pseudo-exchangeability is similar to weighted exchangeability, except the ``factor functions'' $g_i(v_i; v_{-i})$ take the place of the weight functions $w_i(v_i)$ to explicitly describe how $v_i$ can depend (but only symmetrically, as per the definition) on the other observations $v_{-i}$.
The analogous argument in \citet{fannjiang2022conformal} is very similar to that of \citet{tibshirani2019conformal} except using pseudo-exchangeability.\footnote{Again, for simplicity, \citet{fannjiang2022conformal} assume that there are almost surely no ties to work with sets rather than multisets/bags, but the results holds more generally (with more cumbersome notation).} As before, we begin with the statement of equation \eqref{eq:summary_perm_test}:
\begin{align*}
    \mathbb{P}\{V_{n+1}=v_i\ | \ E_z\} = & \mathbb{P}\{Z_{n+1}=z_i\ | \ E_z\} \nonumber \\
    = & \frac{\sum_{\sigma:\sigma(n+1)=i}f(z_{\sigma(1)}, ..., z_{\sigma(n+1)})}{\sum_{\sigma}f(z_{\sigma(1)}, ..., z_{\sigma(n+1)})},
\end{align*}
Using pseudo-exchangeability, this becomes
\begin{align*}
    \mathbb{P}\{Z_{n+1}=z_i\ | \ E_z\} = & \frac{\sum_{\sigma:\sigma(n+1)=i}\prod_{j=1}^{n+1}g_j(z_{\sigma(j)}; z_{-\sigma(j)})h(z_{\sigma(1)}, ..., z_{\sigma(n+1)})}{\sum_{\sigma}\prod_{j=1}^{n+1}g_j(z_{\sigma(j)}; z_{-\sigma(j)})h(z_{\sigma(1)}, ..., z_{\sigma(n+1)})},
\end{align*}
where the core function $h$ does not depend on the ordering of its inputs, so we have
\begin{align}
    \mathbb{P}\{Z_{n+1}=z_i\ | \ E_z\} = & \frac{\sum_{\sigma:\sigma(n+1)=i}\prod_{j=1}^{n+1}g_j(z_{\sigma(j)}; z_{-\sigma(j)})h(z_{1}, ..., z_{n+1})}{\sum_{\sigma}\prod_{j=1}^{n+1}g_j(z_{\sigma(j)}; z_{-\sigma(j)})h(z_{1}, ..., z_{n+1})} \nonumber \\
    = & \frac{\sum_{\sigma:\sigma(n+1)=i}\prod_{j=1}^{n+1}g_j(z_{\sigma(j)}; z_{-\sigma(j)})}{\sum_{\sigma}\prod_{j=1}^{n+1}g_j(z_{\sigma(j)}; z_{-\sigma(j)})}.
    \label{pseudo:factor_funcs_result}
\end{align}
Here, in the same way that for weighted exchangeable data the product $\prod_{j=1}^{n+1}w_j(z_{\sigma(j)})$ represents the (relative) likelihood of observing the joint sequence of events $Z_1 = z_{\sigma(1)}, ..., Z_{n+1} = z_{\sigma(n+1)}$ (given $E_Z$), for pseudo-exchangeable data the analogous role is played by the product $\prod_{j=1}^{n+1}g_j(z_{\sigma(j)}; z_{-\sigma(j)})$: the role of this product is to tell us how much to ``weigh'' or ``count'' a given permutation $\sigma$. However, the specification in pseudo-exchangeability's definition that ``$g_i(\cdot ; v_{-i})$ is a function that depends on the \textit{multiset} $v_{-i}$ (that is, on the values in $v_{-i}$ but not on their ordering)'' (emphasis added) prevents pseudo-exchangeability from formally describing multistep feedback covariate shift. We state this more formally in the following remark.


\begin{remark}
    Pseudo-exchangeability does not formally characterize \textit{multistep} FCS. Consider a two-step instance of FCS for a counterexample, that is with data $Z_1, ..., Z_{n+2}$ generated under MFCS. The factor function for the first step $g_{n+1}(z_{n+1} \ ; \ Z_1 = z_1, ..., Z_{n} = z_{n})$ depends on $Z_1, ..., Z_n$ differently than it does on $Z_{n+2}$ (i.e., it does not depend on $Z_{n+2}$ at all). So, the factor function $g_{n+1}(\cdot \ ; \ Z_1 = z_1, ..., Z_{n} = z_{n})$ cannot be said to depend on the multiset of values $z_{-(n+1)}=\{z_1, ..., z_n, z_{n+2}\}$ independently of the ordering of the values (i.e., it requires knowing which of the values corresponds to $Z_{1:n}$ versus $Z_{n+2}$), as is required by the defintion of pseudo-exchangeability from \citet{fannjiang2022conformal}. 
\end{remark}

\newpage

\section{Additonal Details for Main Paper Experiments}
\label{sec:exp_details}


\subsection{Black-box Optimization Experimental Details}
\label{subsec:black_box_opt_exp_details}

\textbf{Protein Design Datasets} Our multistep biomolecular design experiments leverage fluorescent protein data from \citet{poelwijk2019learning}, previously used for evaluation of related single-step FCS CP methods in \citet{fannjiang2022conformal} and \citet{prinster2023jaws}. This dataset has the benefit of a combinatorially complete set of labels, which simulates measuring a design in an experiment. In particular, both a ``blue'' and ``red'' wavelength fluorescence strength were experimentally measured for each of $2^{13}=8,192$ possible combinations of binary variations to a wild-type (natural) fluorescent protein at 13 positions on the primary sequence. 

\textbf{Measurement Noise for Protein Design Experiments} We followed the same procedure for adding measurement noise as described in  \citet{fannjiang2022conformal}'s Supplementary Materials S3. That is, every time the $i$-th sequence was sampled from the combinatorially complete library (including for initial training and calibration data as well as for each query point), zero-mean Guassian noise was added (to the label value) with standard deviation set to the absolute residual between that $i$-th sequence's true label and the prediction on that $i$-th sequence from a separately fit linear model. The separately-fit linear model had as its input covariates the up to seventh-order terms identified as statistically significant by \citet{poelwijk2019learning}, as described by \citet{fannjiang2022conformal}.

\begin{table}[h]
\caption{Hyperparameters for Full CP protein design experiments with ridge regression predictor (Figure \ref{fig:FullCP_MultistepDesignExpts}).}
\label{tab:fig1}
\vskip 0.15in
\begin{center}
\begin{tabular}{lc}
\toprule
Hyperparameter name & Value(s) \\
\midrule
$\alpha$ (Target miscoverage rate)  & 0.1 \\
$n$ (Initial number of training/calibration samples) & 32 \\
$\lambda$ (Shift magnitude or ``inverse temperature'') & 8.0 \\
$d$ (MFCS estimation depth) & $\{1, 2\}$ \\
$T$ (Number of MFCS steps past IID initialization) & 5 \\
Ridge regression regularization strength  & 0.01 \\
Random seeds & $\{0, ..., 999\}$ \\
\bottomrule
\end{tabular}
\end{center}
\vskip -0.1in
\end{table}

\begin{table}[h]
\caption{Hyperparameters for Split CP protein design neural network experiments (Figure \ref{fig:SplitCP_MultistepDesignExpts_NN_mu}) with the \texttt{MLPRegressor} from \texttt{scikit-learn} (L-BFGS solver and logistic activation function, default parameters otherwise).}
\label{tab:fig2}
\vskip 0.15in
\begin{center}
\begin{tabular}{lc}
\toprule
Hyperparameter name & Value(s) \\
\midrule
$\alpha$ (Target miscoverage rate)  & 0.1 \\
$n_{\text{train}}$ (Initial number of training samples) & 32 \\
$n_{\text{cal}}$ (Initial number of calibration samples) & 32 \\
$\lambda$ (Shift magnitude or ``inverse temperature'') & 5.0 \\
$d$ (MFCS estimation depth) & $\{1, 2, 3, 4\}$ \\
$T$ (Number of MFCS steps past IID initialization) & 10 \\
Probability each queried point is added to training (versus calibration) data & 0.5 \\
Random seeds & $\{0, ..., 499\}$ \\
\bottomrule
\end{tabular}
\end{center}
\vskip -0.1in
\end{table}

\subsection{Active Learning Experimental Details}
\label{subsec:active_learning_exp_details}

\textbf{Datasets for Active Learning Experiments} 
We evaluated on four datasets from the UCI Machine Learning Repository \citep{frank2010uci} that are commonly used for evaluation in the CP literature \citep{tibshirani2019conformal, barber2021predictive, prinster2022jaws, prinster2023jaws} and represent a range of sample sizes and dimensionalities: the NASA airfoil self-noise dataset (1503 samples, $p=5$) \citep{misc_airfoil_self-noise_291}, the communities and crime dataset (1994 samples, $p=99$) \citep{misc_communities_and_crime_183}, the Medical Expenditure Panel Survey 2016 data set\footnote{\url{https://meps.ahrq.gov/mepsweb/data_stats/download_data_files_detail.jsp?cboPufNumber=HC-192}} (33005 samples, $p=107$) (preprocessed as in \citet{barber2021predictive}; details for an older version of the data are in \citet{ezzati2008sample}), and the blog feedback dataset (60021 samples, $p=281$) \citep{misc_blogfeedback_304}. 


\textbf{Sampling Holdout Test Set and Initial Training Set} For all experiments, a holdout test set of 250 samples was first sampled uniformly at random from the full dataset to track the accuracy of the base predictor as the active learning progressed. To simulate a practical active learning setting where there often tends to be sample-selection bias in how the initial training data is obtained relative to the desired test distribution, we sampled the initial training and calibration datasets from the remaining non-test-set data with probabilities proportional to $\exp(\gamma\cdot X_{\text{PCA}1}^{\text{normed}})$, where $X_{\text{PCA}1}^{\text{normed}}$ is the min-max-normed first principal component representation of the data (that is, the dataset projected onto the first principal component extracted from the same dataset).

\textbf{Measurement Noise for Active Learning Experiments} Every time the $i$-th datapoint was sampled (including for the holdout test set, initial training/calibration data, and each queried point), zero-mean Gaussian noise was added to the label value. The standard deviation for the Gaussian noise for the $i$-th datapoint was set as 0.05 times the absolute residual between that $i$-th point's true label value and the prediction for that $i$-th point by a separately fit kernel ridge regression model that was fit to all of the data samples. In particular, the separately-fit regressor was the \texttt{KernelRidge} method from \texttt{scikit-learn}, with regularization strength 1.0 and default parameters otherwise.

\begin{table}[h]
\caption{Hyperparameters for Split CP active learning experiments (Figure \ref{fig:SplitCP_ActiveLearningExpts}) with the \texttt{GaussianProcessRegressor} from \texttt{scikit-learn} as the active learning agent (kernel=DotProduct(sigma\_0=0.05)+WhiteKernel(noise\_level=0.05), default parameters otherwise).}
\label{tab:fig3_hyperparams}
\vskip 0.15in
\begin{center}
\begin{tabular}{lc}
\toprule
Hyperparameter name & Value(s) \\
\midrule
$\alpha$ (Target miscoverage rate)  & 0.1 \\
$n_{\text{train}}$ (Initial number of training samples) & 64 \\
$n_{\text{cal}}$ (Initial number of calibration samples) & 16 \\
$\lambda$ (Shift magnitude) & 10.0 \\
$d$ (MFCS estimation depth) & $\{1, 3\}$ \\
$T$ (Number of MFCS steps past IID initialization) & 70 \\
$\gamma$ (Bias magnitude in initial training data IID sampling) & 3.0 \\
ACI baseline ``step size'' (default parameter in \citet{gibbs2021adaptive}) & 0.005 \\
Probability each queried point is added to training (versus calibration) data & 0.5 \\
Random seeds & $\{0, ..., 349\}$ \\
\bottomrule
\end{tabular}
\end{center}
\vskip -0.1in
\end{table}


\newpage

\section{Additional Experiments}
\label{app:sec:additional_expts}

\subsection{Shift Magnitude Ablation Study}
\label{app:subsec:shift_mag_ablation}

These ablation studies examine the effect of changing the shift magnitude $\lambda$, or the ``aggressiveness'' of the ML agent's query strategy, focusing on the Split CP blue fluorescent protein design case. Other than the different values of $\lambda$, the experimental setting and other hyperparameters are the same as in the blue protein results reported in the main paper Figure \ref{fig:SplitCP_MultistepDesignExpts_NN_mu}, where $\lambda=5.0$ (full hyperparameters reported in Appendix \ref{subsec:black_box_opt_exp_details}, Table \ref{tab:fig2}). Increasing $\lambda$ tends to exacerbate the discrepancy in coverage between the $\{1,2,3,4\}$-step MFCS CP methods (with higher-order methods maintaining stronger coverage); increase interval widths of the MFCS CP methods; and increase the fitness values of the designed protein sequence.

\begin{figure*}[!htb]
\centering
    \begin{subfigure}{0.85\textwidth}\includegraphics[width=\textwidth]{figures/SplitCPDesignExpts_lam5.0_legend.pdf}
    \end{subfigure}
    \\
    \hfill
    \begin{subfigure}{0.3\textwidth}
        \includegraphics[width=\textwidth]{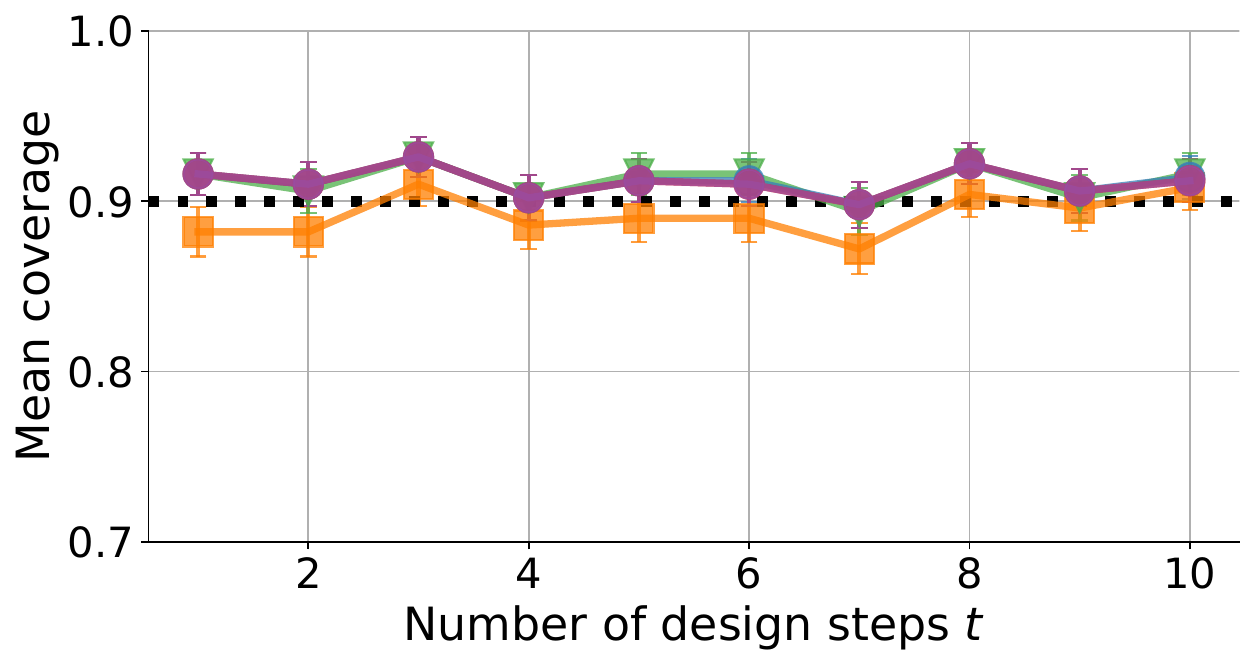}
    \end{subfigure}
    \hfill
    \begin{subfigure}{0.3\textwidth}
        \includegraphics[width=\textwidth]{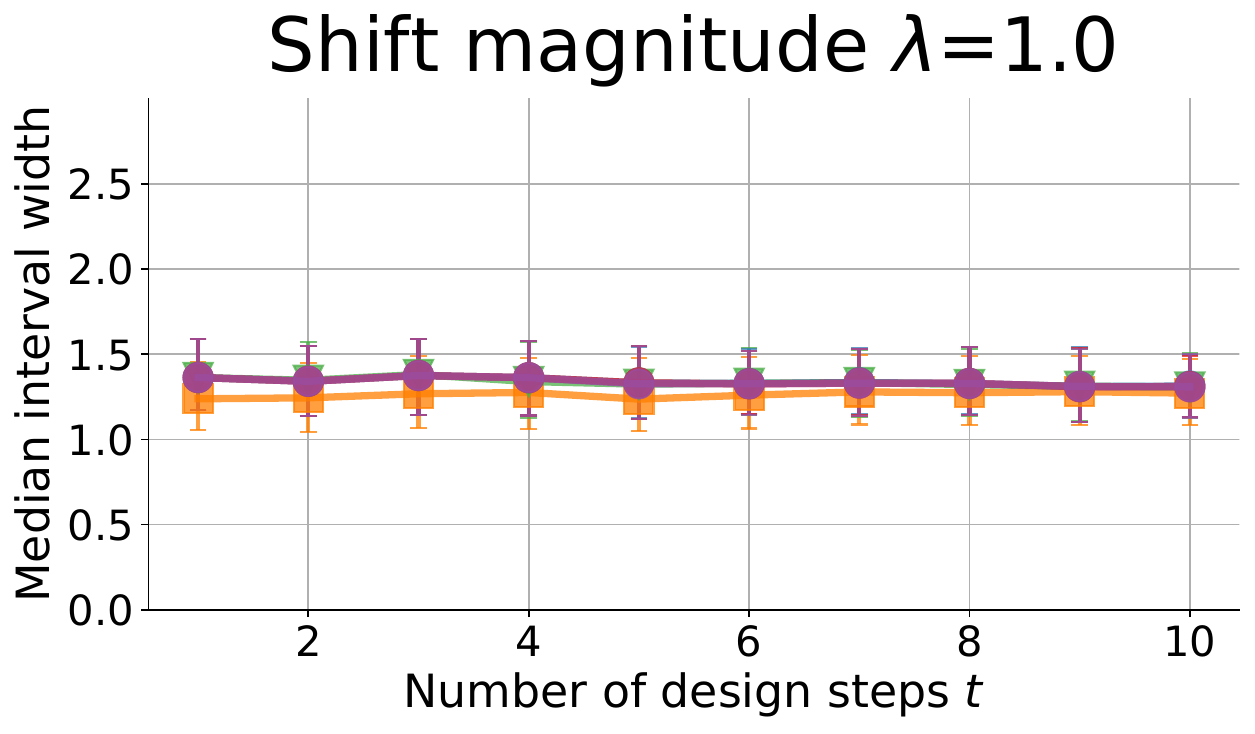}
    \end{subfigure}
    \hfill
    \begin{subfigure}{0.3\textwidth}
        \includegraphics[width=\textwidth]{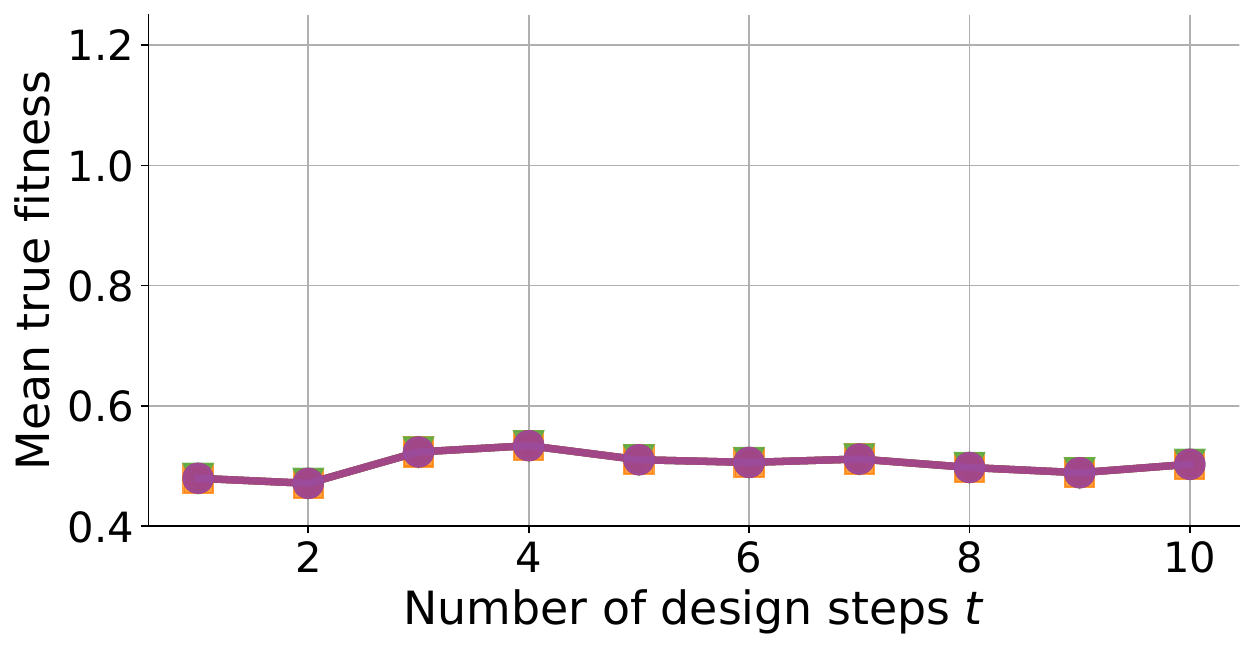}
    \end{subfigure}
    \hfill
    \\
    \hfill
    \begin{subfigure}{0.3\textwidth}
        \includegraphics[width=\textwidth]{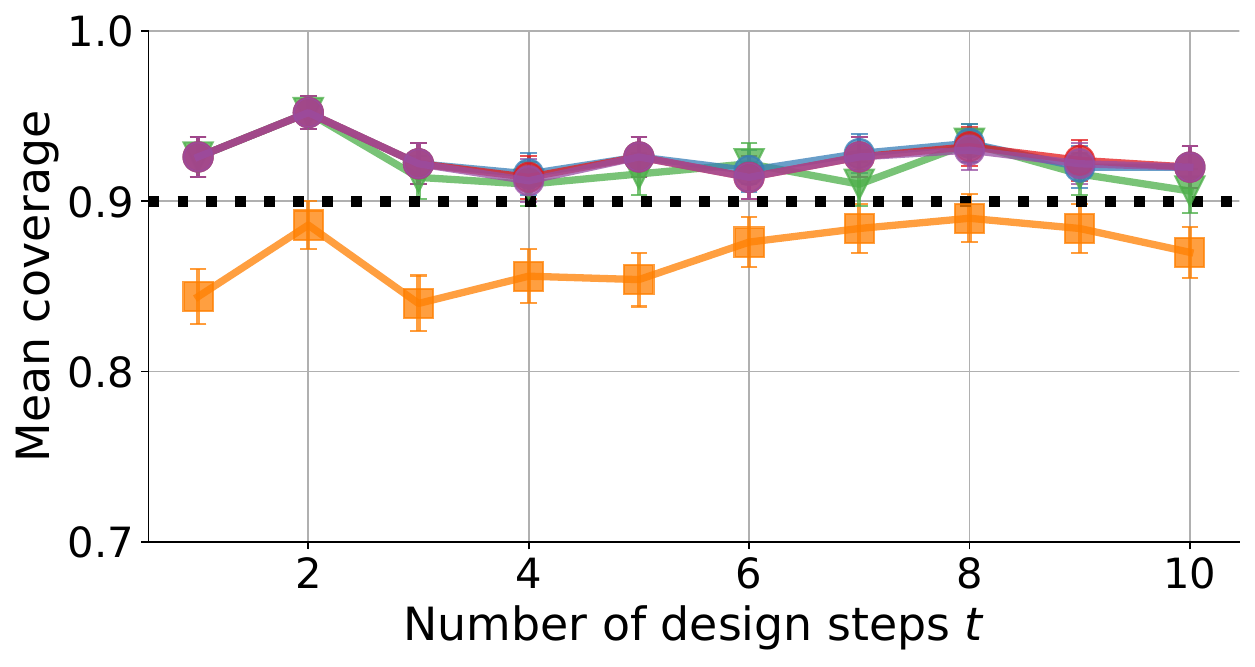}
    \end{subfigure}
    \hfill
    \begin{subfigure}{0.3\textwidth}
        \includegraphics[width=\textwidth]{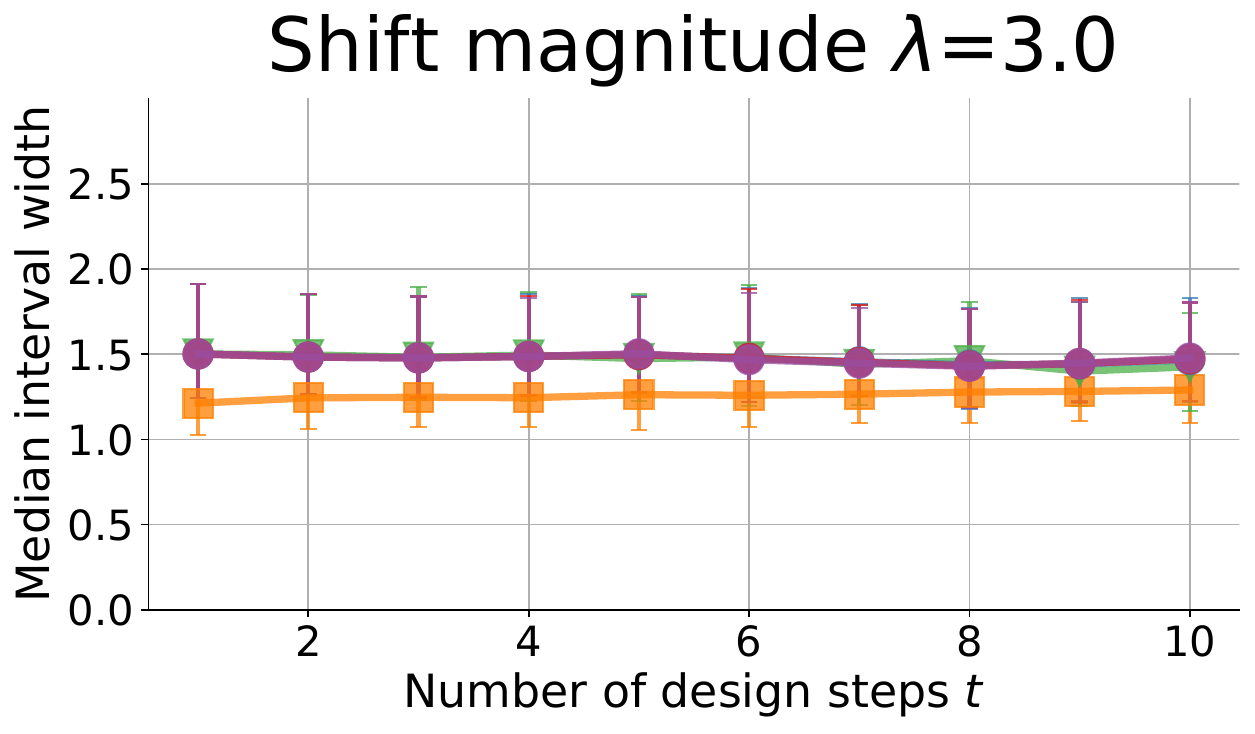}
    \end{subfigure}
    \hfill
    \begin{subfigure}{0.3\textwidth}
        \includegraphics[width=\textwidth]{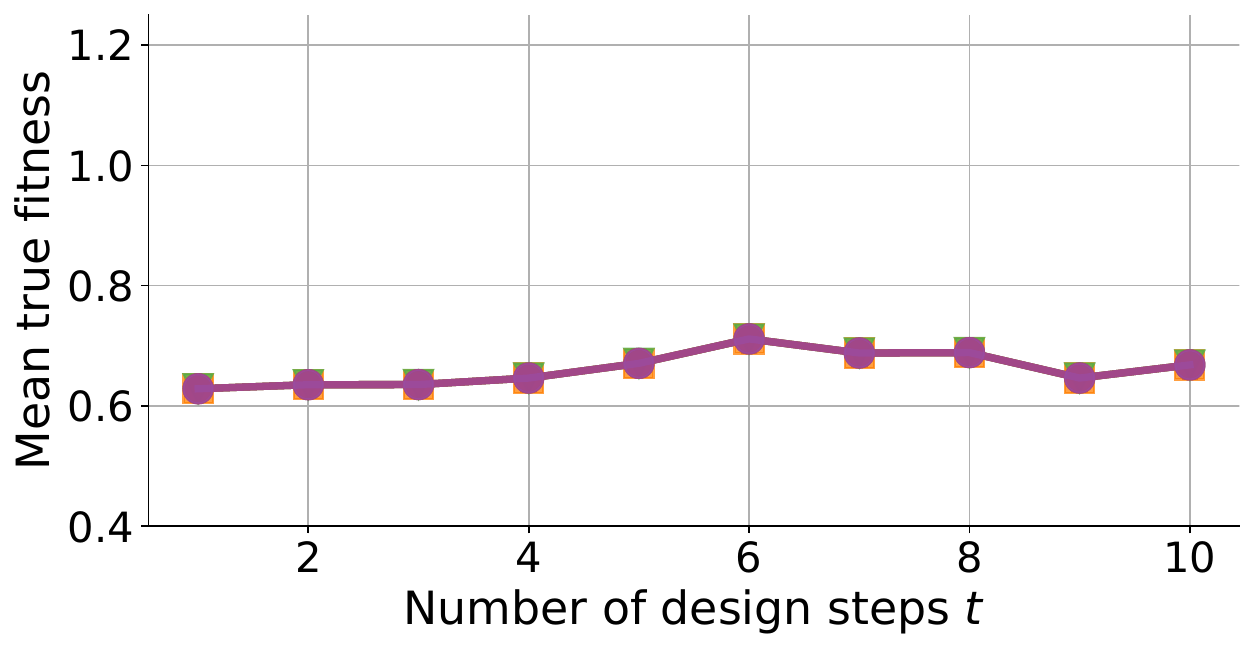}
    \end{subfigure}
    \hfill
    \\
    \hfill
    \begin{subfigure}{0.3\textwidth}
        \includegraphics[width=\textwidth]{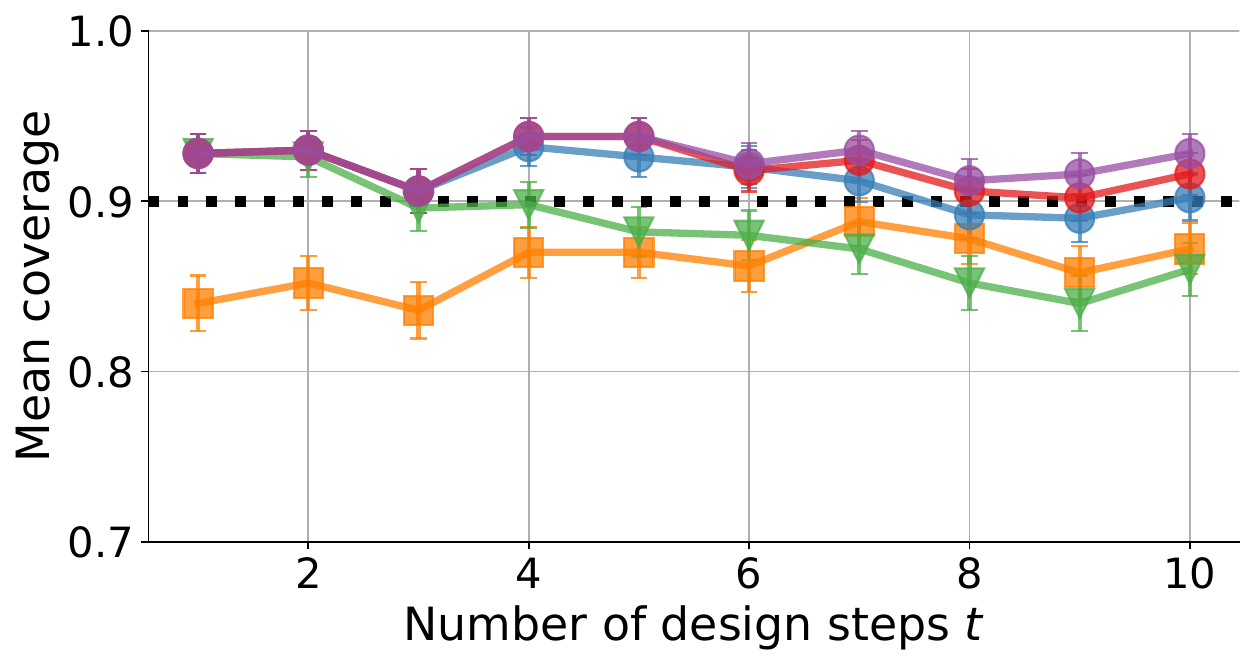}
    \end{subfigure}
    \hfill
    \begin{subfigure}{0.3\textwidth}
        \includegraphics[width=\textwidth]{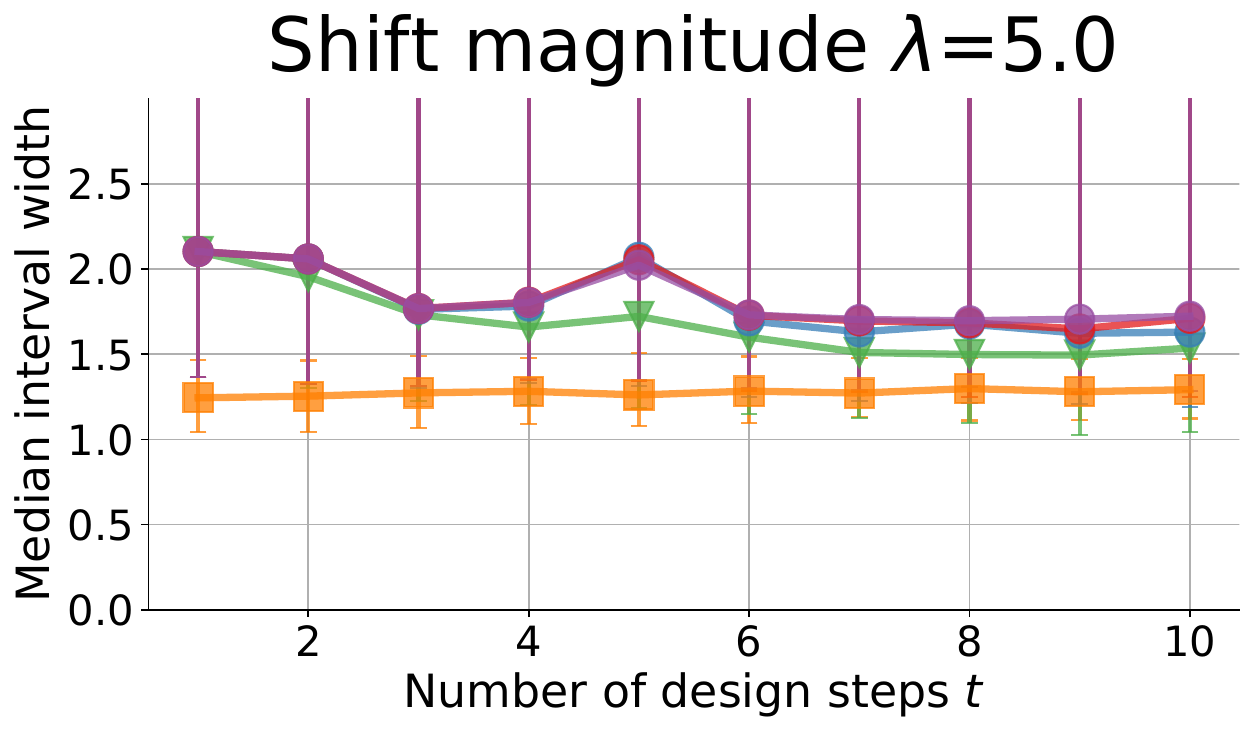}
    \end{subfigure}
    \hfill
    \begin{subfigure}{0.3\textwidth}
        \includegraphics[width=\textwidth]{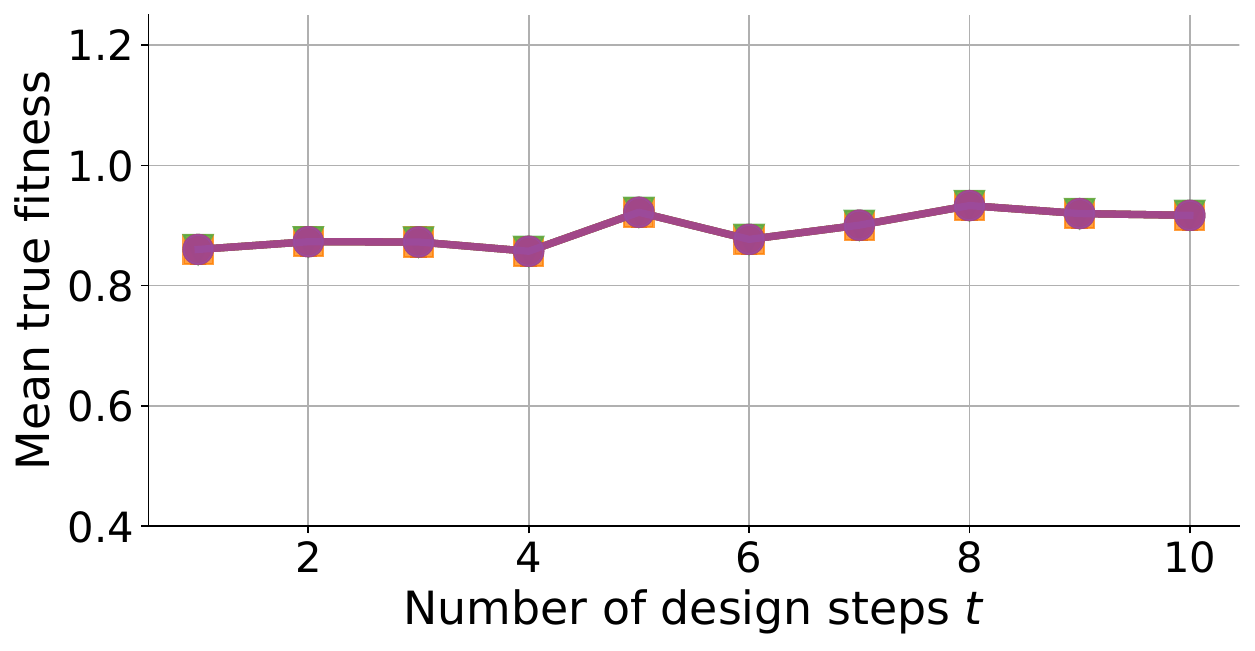}
    \end{subfigure}
    \hfill
    \\
    \hfill
    \begin{subfigure}{0.3\textwidth}
        \includegraphics[width=\textwidth]{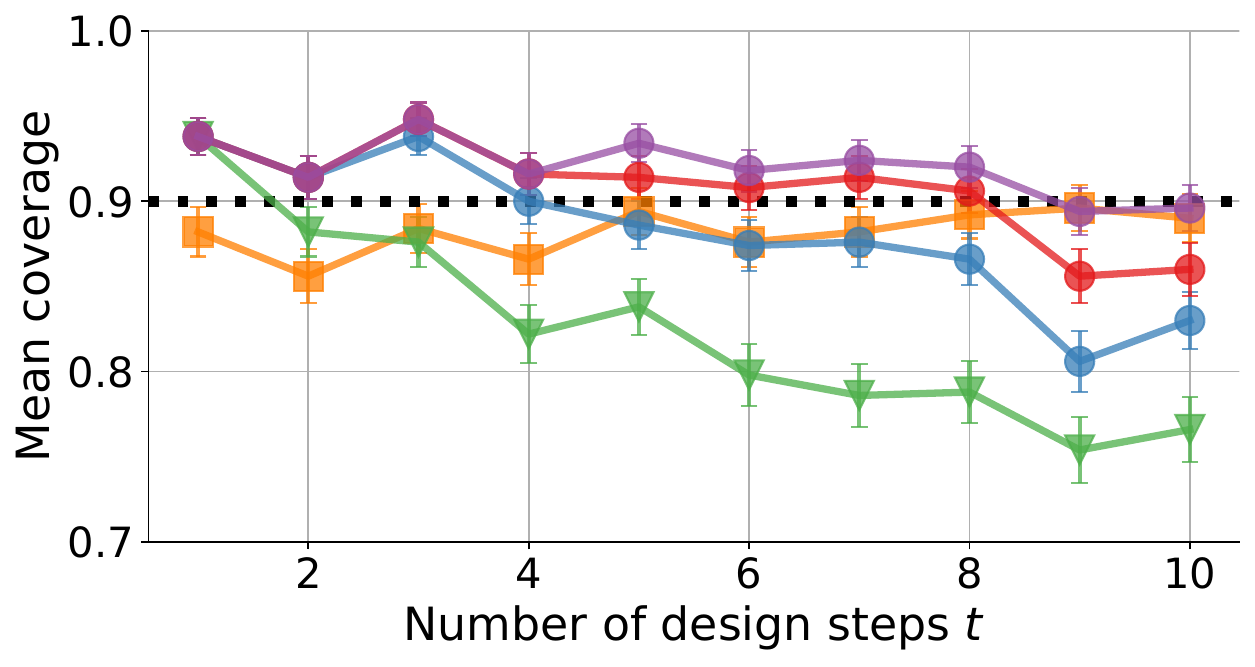}
    \end{subfigure}
    \hfill
    \begin{subfigure}{0.3\textwidth}
        \includegraphics[width=\textwidth]{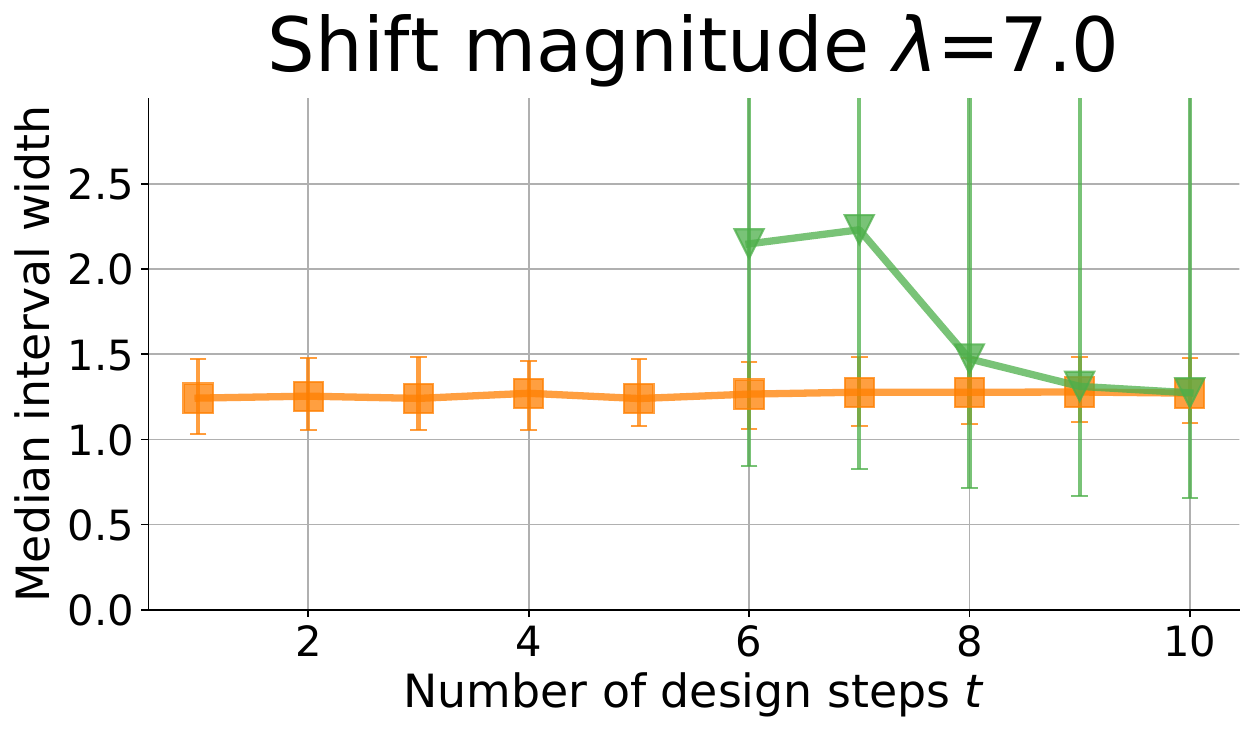}
    \end{subfigure}
    \hfill
    \begin{subfigure}{0.3\textwidth}
        \includegraphics[width=\textwidth]{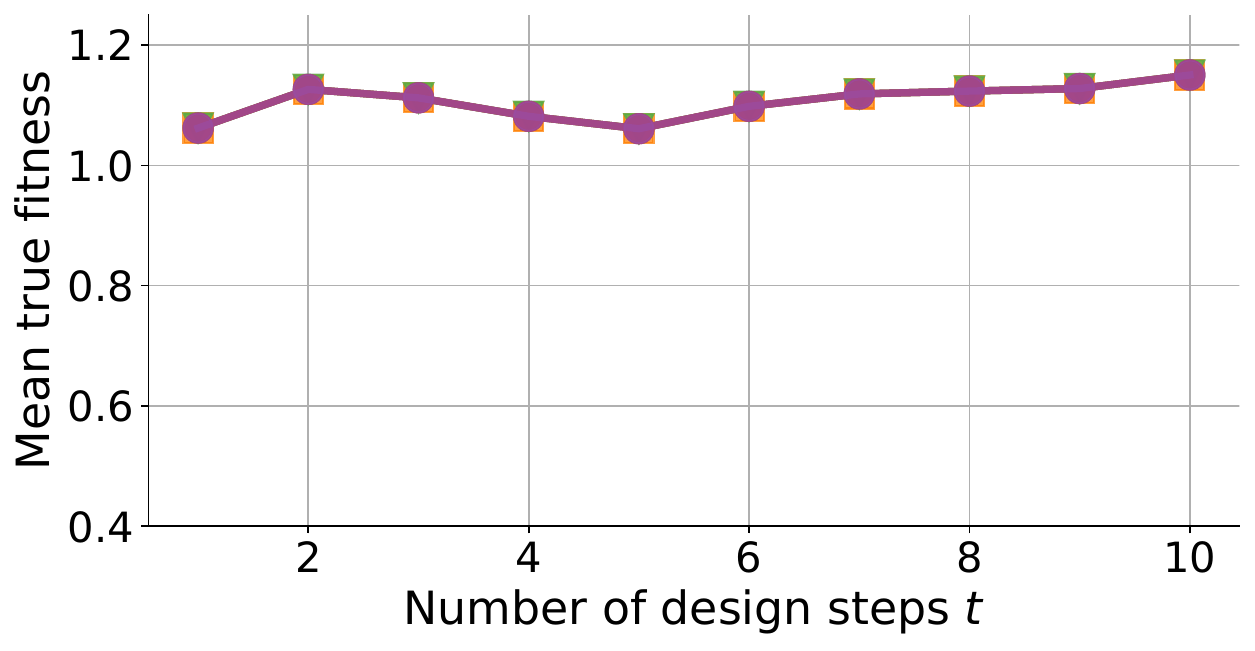}
    \end{subfigure}
    \hfill
    \\
\caption{ Ablation studies for shift magnitude $\lambda$, focused on Split CP experiments on the blue protein design dataset, where each row corresponds to a different shift magnitude $\lambda \in \{1.0, 3.0, 5.0, 7.0\}$.
    All experiments are conducted with the \texttt{MLPRegressor} from \texttt{scikit-learn} (with LBFGS solver, logistic activation, and default parameters otherwise) as the ML predictor, initially trained on 32 datapoints and with 32 points in the calibration set. All values are computed over 500 random seeds. Other than $\lambda$, full hyperparameters are as in Appendix \ref{subsec:black_box_opt_exp_details}, Table \ref{tab:fig2}.
}
\label{fig:ShiftMagnitudeAblation}
\end{figure*}

\newpage

\subsection{Ablation Study for Initial Amount of Training and Calibration Data}
\label{app:subsec:n_initial}

These ablation studies examine the effect of changing the initial number of training points ($n_{\text{train}}$) and the initial number of calibration points ($n_{\text{cal}}$), focusing on the Spit CP red fluorescent protein design case. Other than the different initial sample sizes (where $n_{\text{train}}= n_{\text{cal}}$), the experimental setting and other hyperparameters are the same as in the blue protein results reported in the main paper Figure \ref{fig:SplitCP_MultistepDesignExpts_NN_mu}, where $n_{\text{train}}= n_{\text{cal}}=32$ (full hyperparameters reported in Appendix \ref{subsec:black_box_opt_exp_details}, Table \ref{tab:fig2}). All values are computed over 500 random seeds.

Increasing the initial training and calibration sample size tends to hurt the coverage of baselines and increase the discrepancy in coverage between the $\{1, 2, 3\}$-step CP methods in these experiments. This may be because more training data improves the performance of the ML model, which can enable it to propose more extreme design sequences. This explanation is plausible because the increase in mean protein fitness values with larger initial training sets is an indication of those design procedures being more aggressive. Interval widths of the MFCS Split CP methods tend to increase with larger initial sample sizes, potentially for the same reason, to maintain coverage when the shift is more extreme.


\begin{figure*}[!htb]
\centering
    \begin{subfigure}{0.85\textwidth}\includegraphics[width=\textwidth]{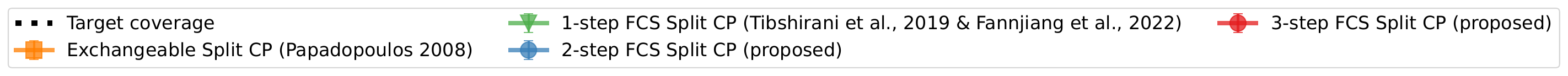}
    \end{subfigure}
    \hfill
    \\
    \begin{center}
    \begin{subfigure}{0.5\textwidth}\includegraphics[width=\textwidth]{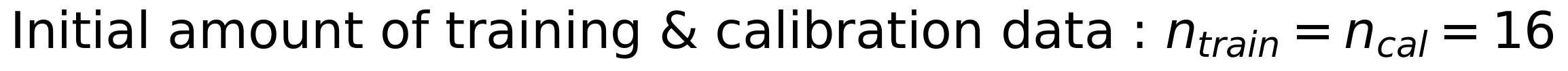}
    \end{subfigure}
    \end{center}
    \hfill
    \\
    \begin{subfigure}{0.3\textwidth}
        \includegraphics[width=\textwidth]{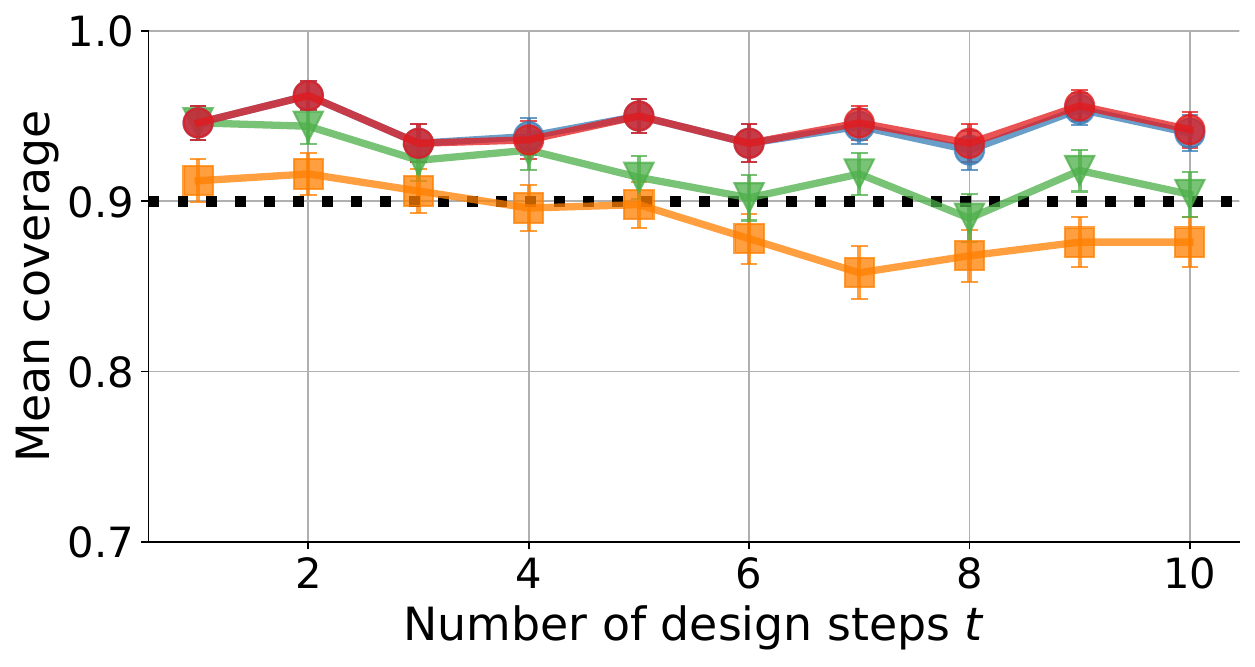}
    \end{subfigure}
    \hfill
    \begin{subfigure}{0.3\textwidth}
        \includegraphics[width=\textwidth]{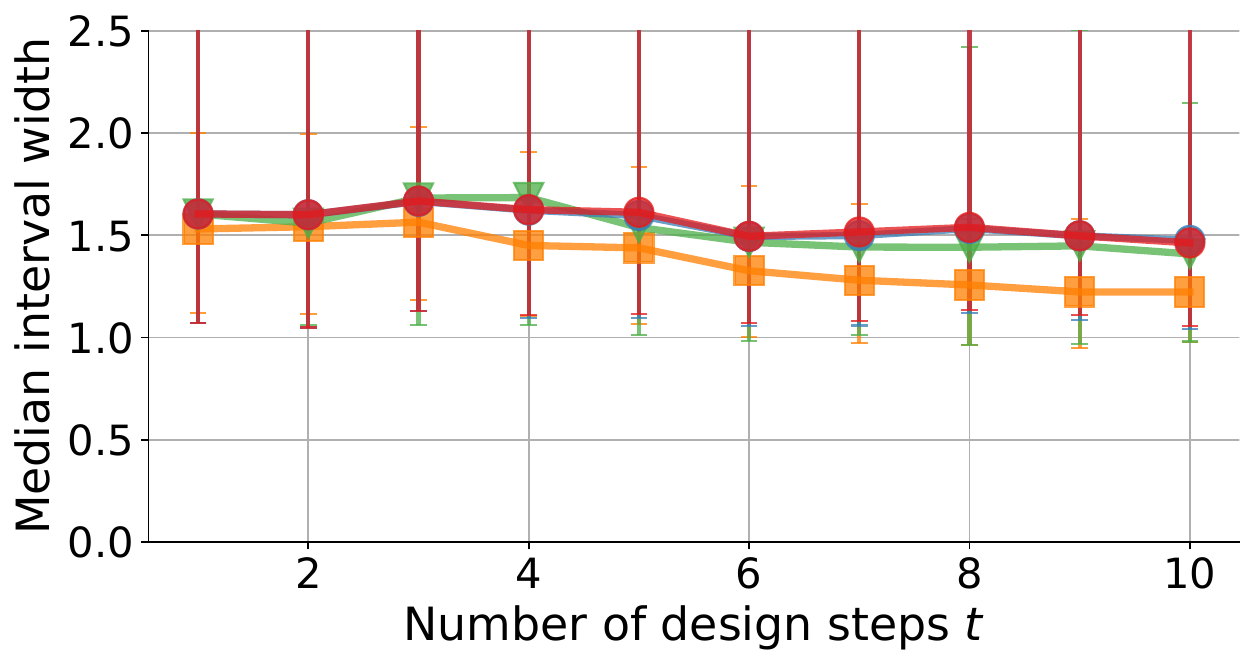}
    \end{subfigure}
    \hfill
    \begin{subfigure}{0.3\textwidth}
        \includegraphics[width=\textwidth]{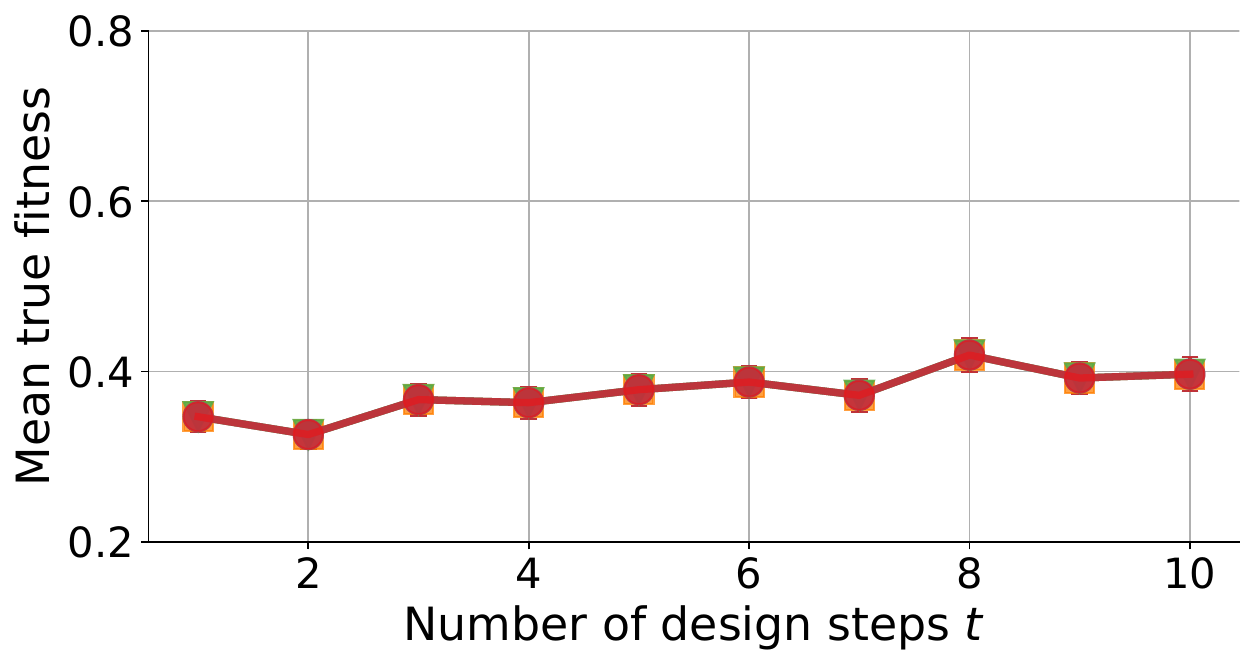}
    \end{subfigure}
    \hfill
    \\
    \begin{center}
    \begin{subfigure}{0.5\textwidth}\includegraphics[width=\textwidth]{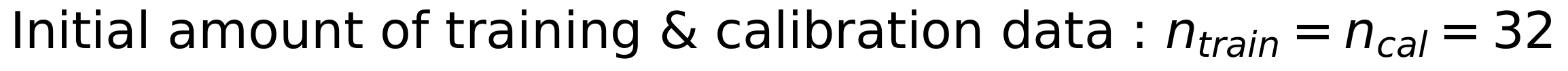}
    \end{subfigure}
    \end{center}
    \hfill
    \\
    \hfill
    \begin{subfigure}{0.3\textwidth}
        \includegraphics[width=\textwidth]{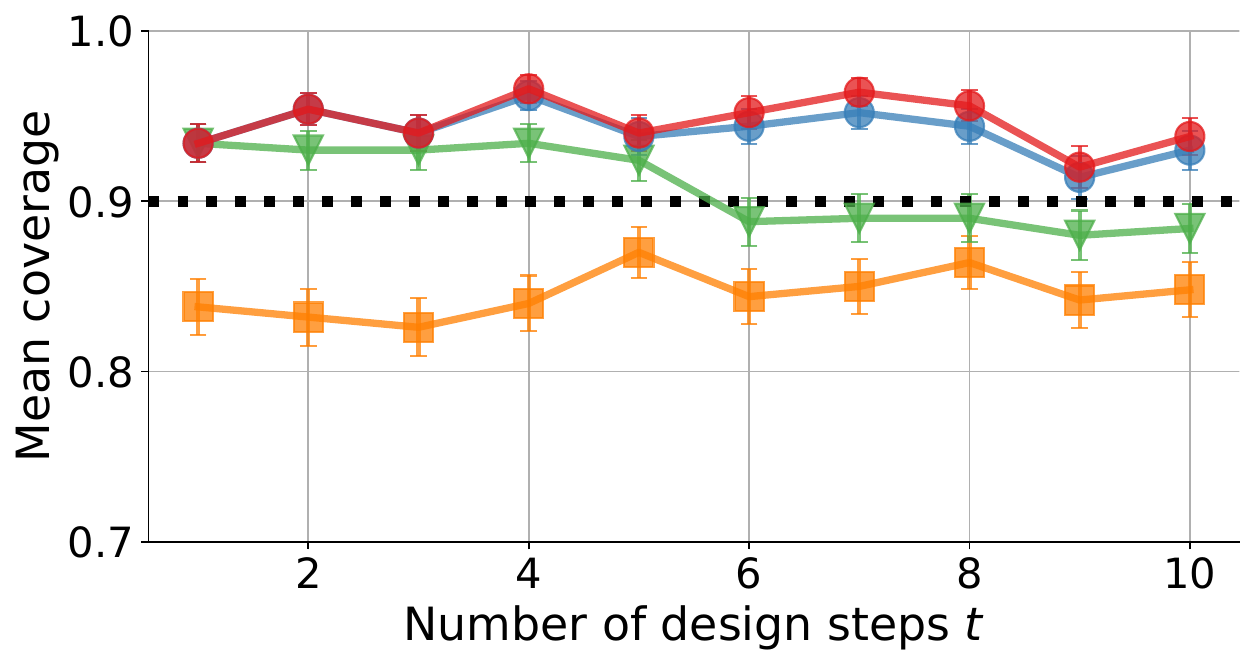}
    \end{subfigure}
    \hfill
    \begin{subfigure}{0.3\textwidth}
        \includegraphics[width=\textwidth]{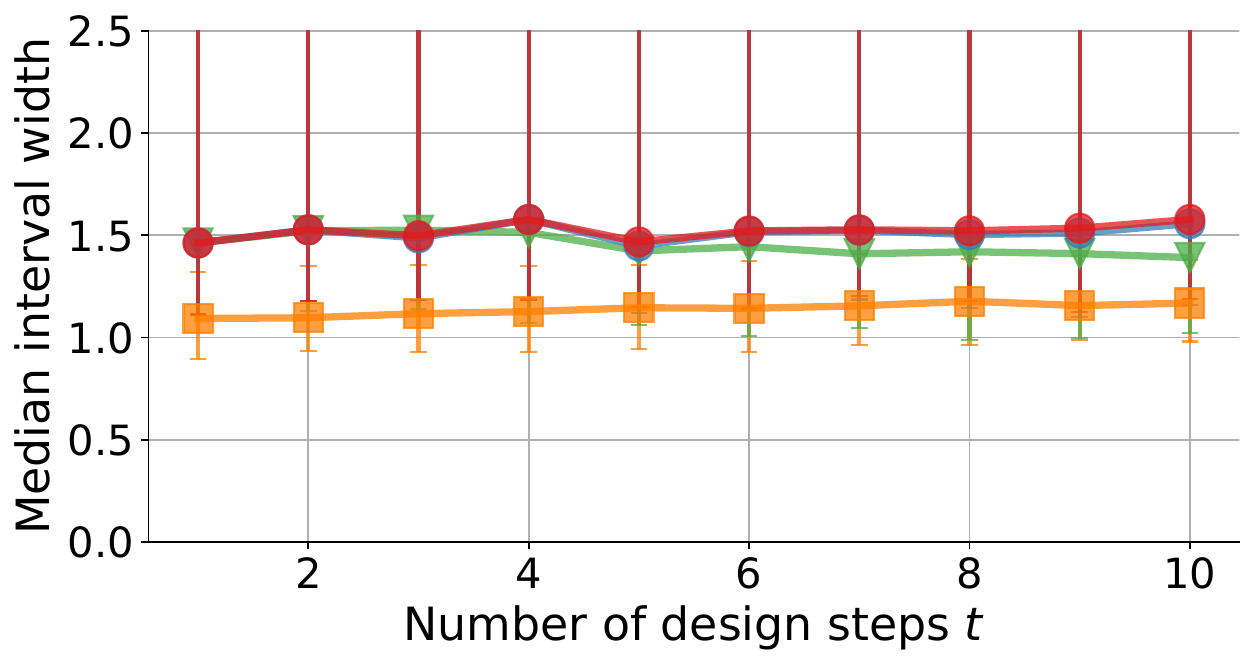}
    \end{subfigure}
    \hfill
    \begin{subfigure}{0.3\textwidth}
        \includegraphics[width=\textwidth]{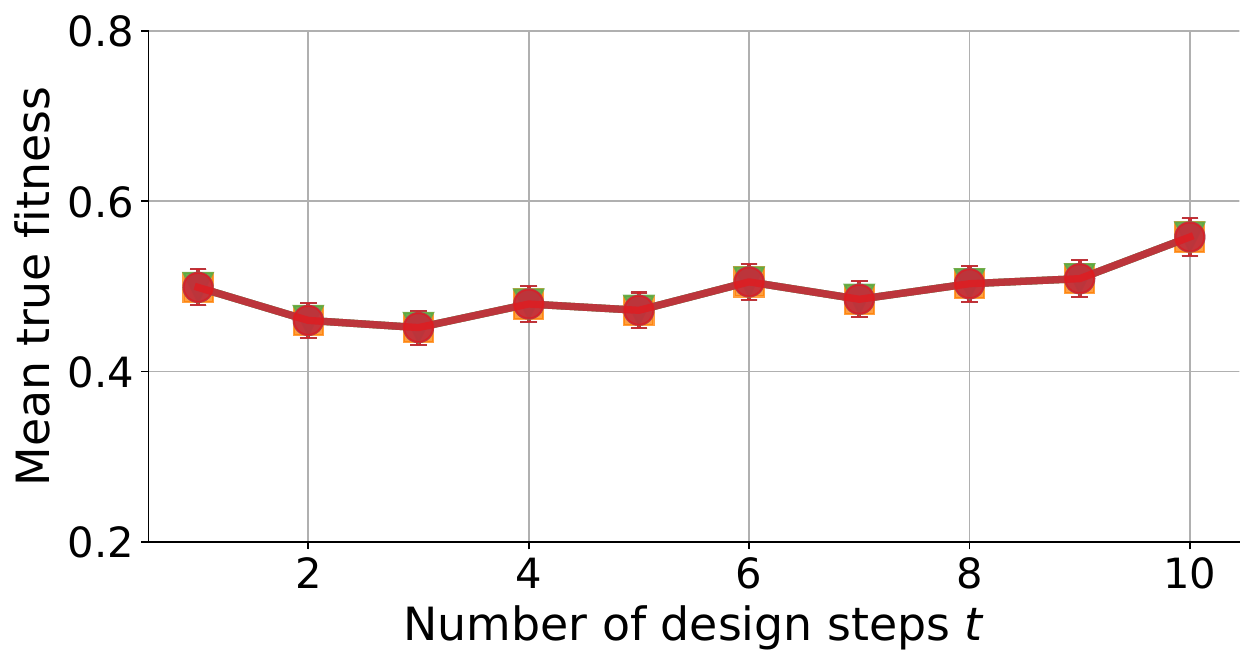}
    \end{subfigure}
    \hfill
    \\
    \begin{center}
    \begin{subfigure}{0.5\textwidth}\includegraphics[width=\textwidth]{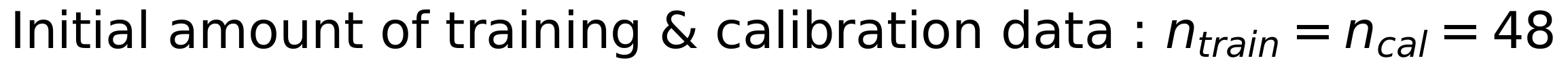}
    \end{subfigure}
    \end{center}
    \hfill
    \\
    \hfill
    \begin{subfigure}{0.3\textwidth}
        \includegraphics[width=\textwidth]{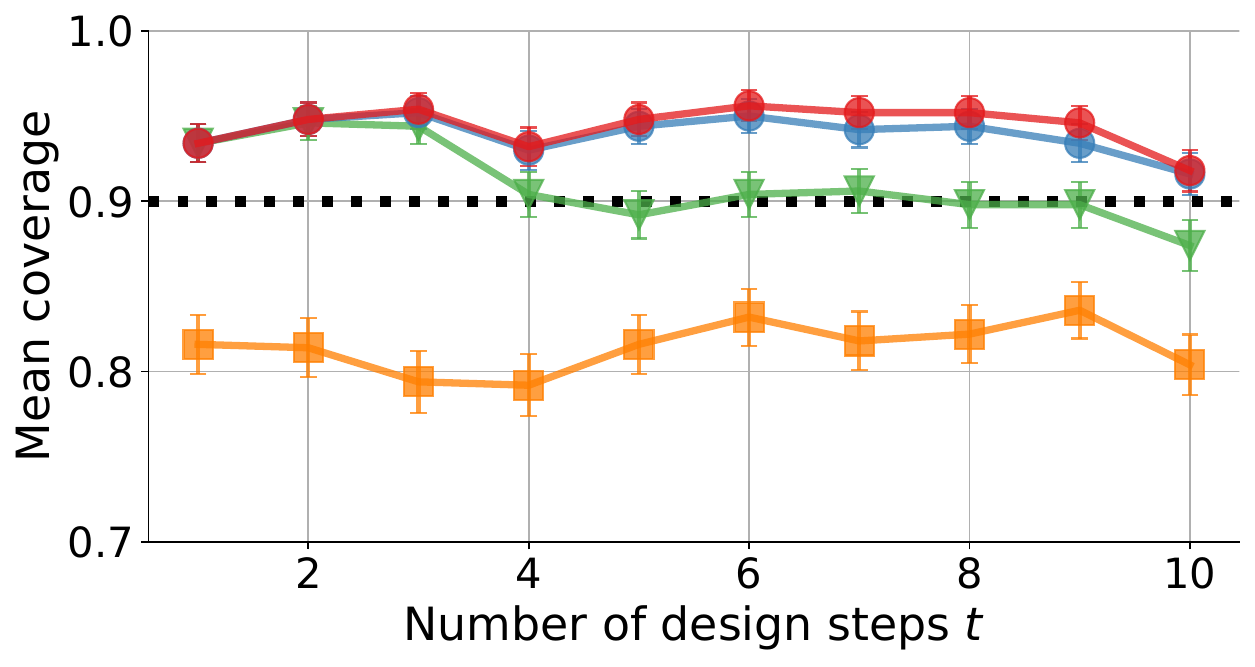}
    \end{subfigure}
    \hfill
    \begin{subfigure}{0.3\textwidth}
        \includegraphics[width=\textwidth]{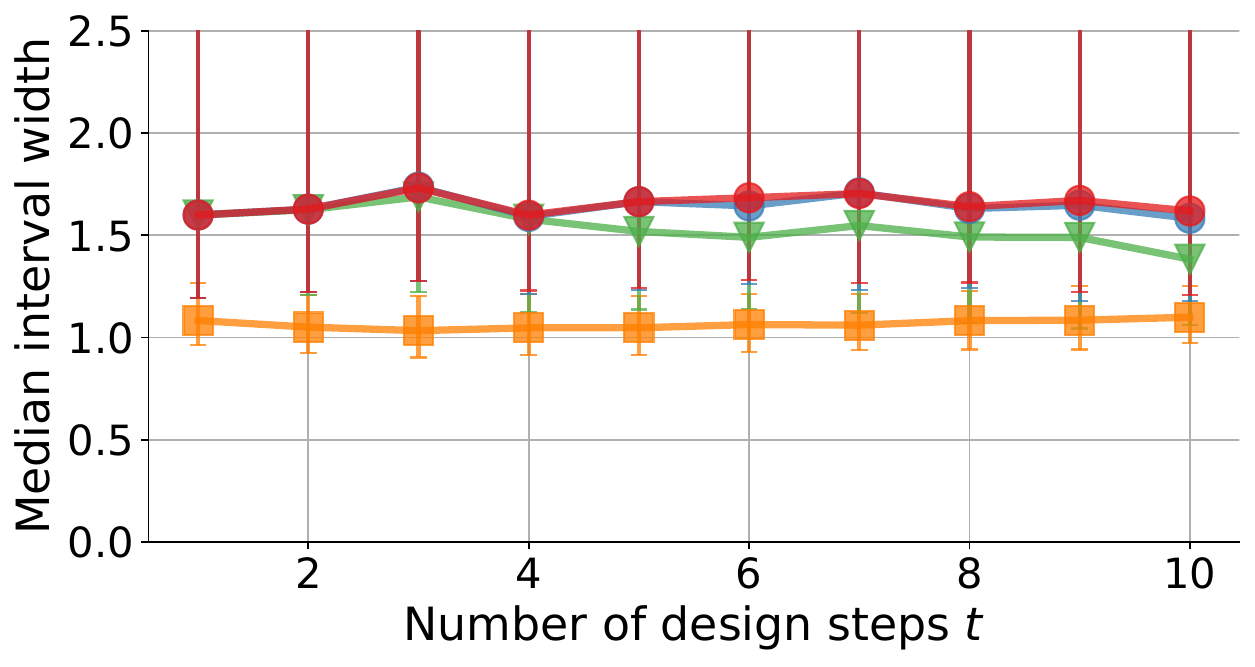}
    \end{subfigure}
    \hfill
    \begin{subfigure}{0.3\textwidth}
        \includegraphics[width=\textwidth]{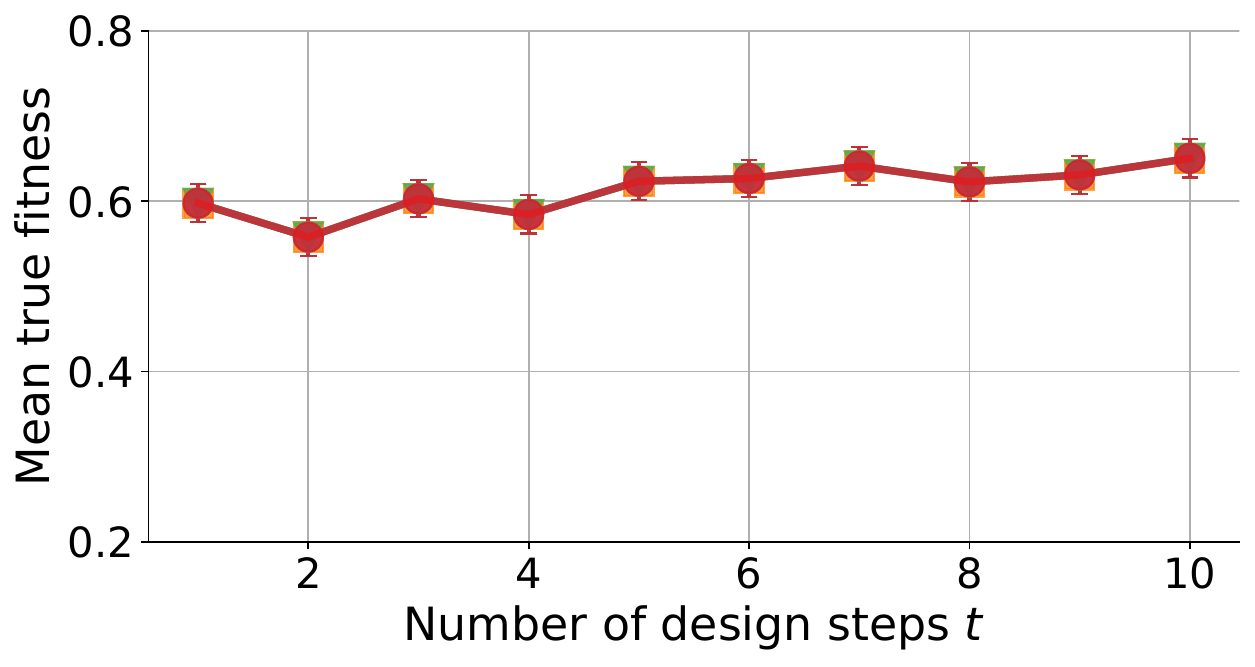}
    \end{subfigure}
    \hfill
    \\
        \begin{center}
    \begin{subfigure}{0.5\textwidth}\includegraphics[width=\textwidth]{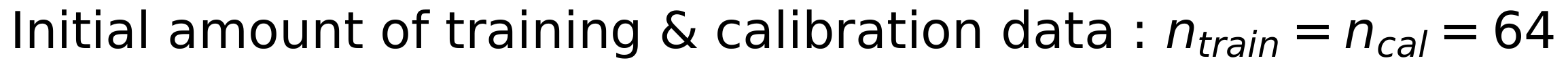}
    \end{subfigure}
    \end{center}
    \hfill
    \\
    \hfill
    \begin{subfigure}{0.3\textwidth}
        \includegraphics[width=\textwidth]{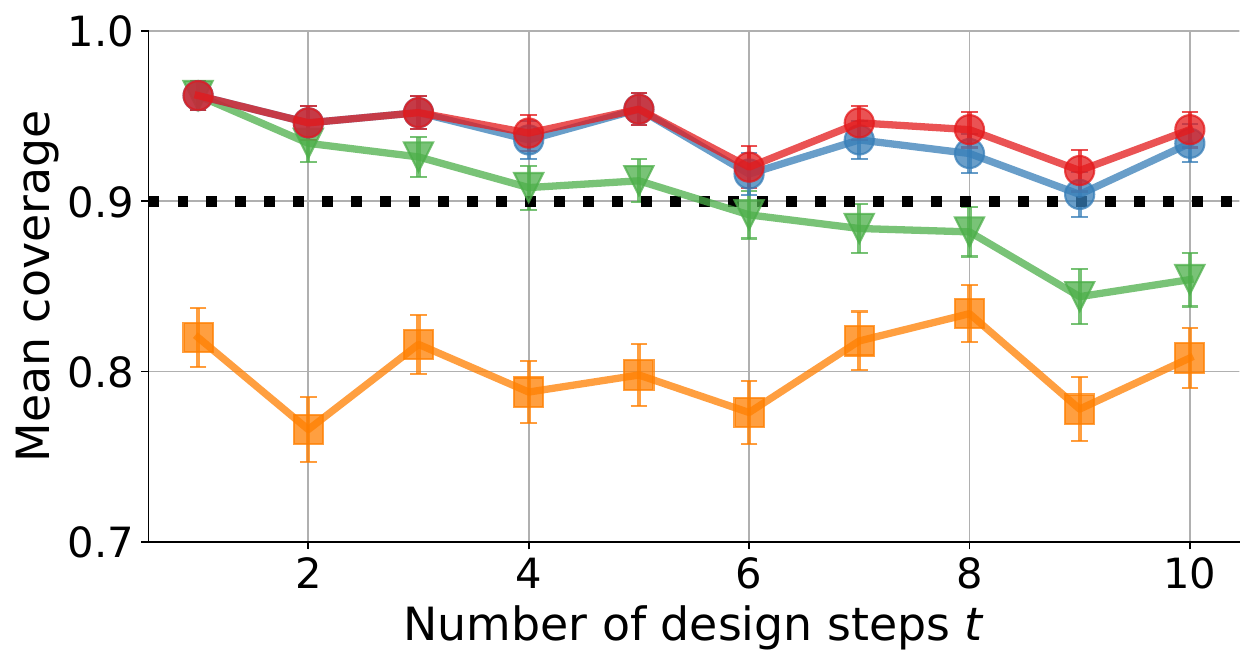}
    \end{subfigure}
    \hfill
    \begin{subfigure}{0.3\textwidth}
        \includegraphics[width=\textwidth]{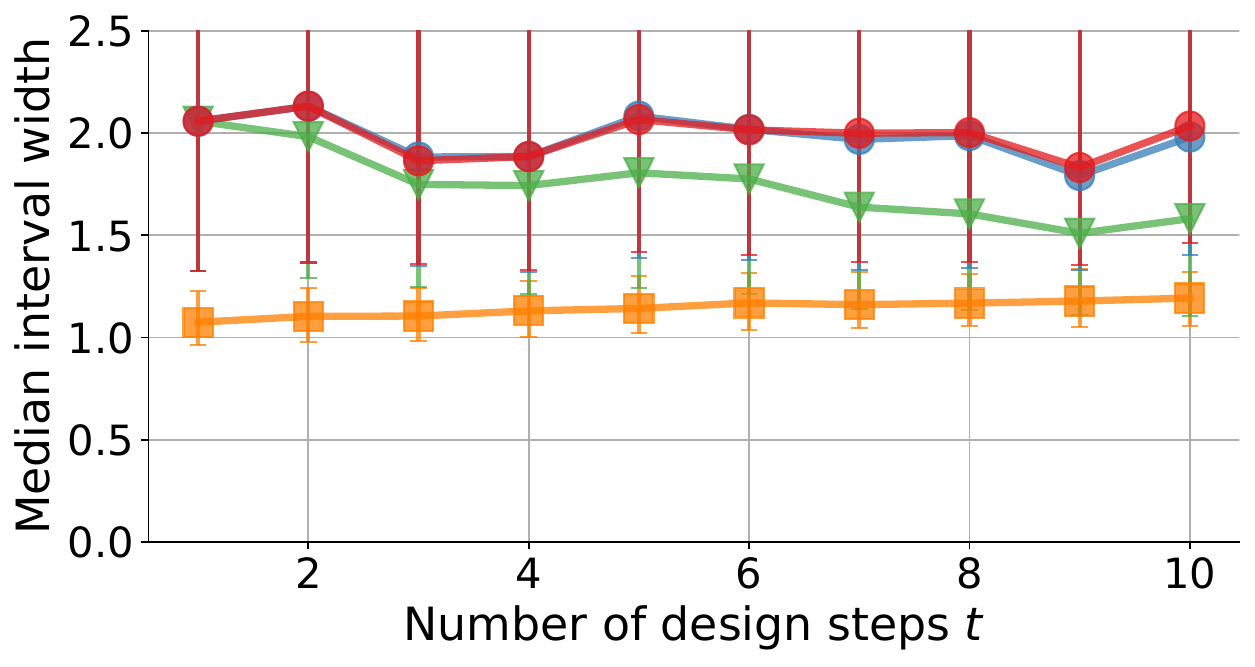}
    \end{subfigure}
    \hfill
    \begin{subfigure}{0.3\textwidth}
        \includegraphics[width=\textwidth]{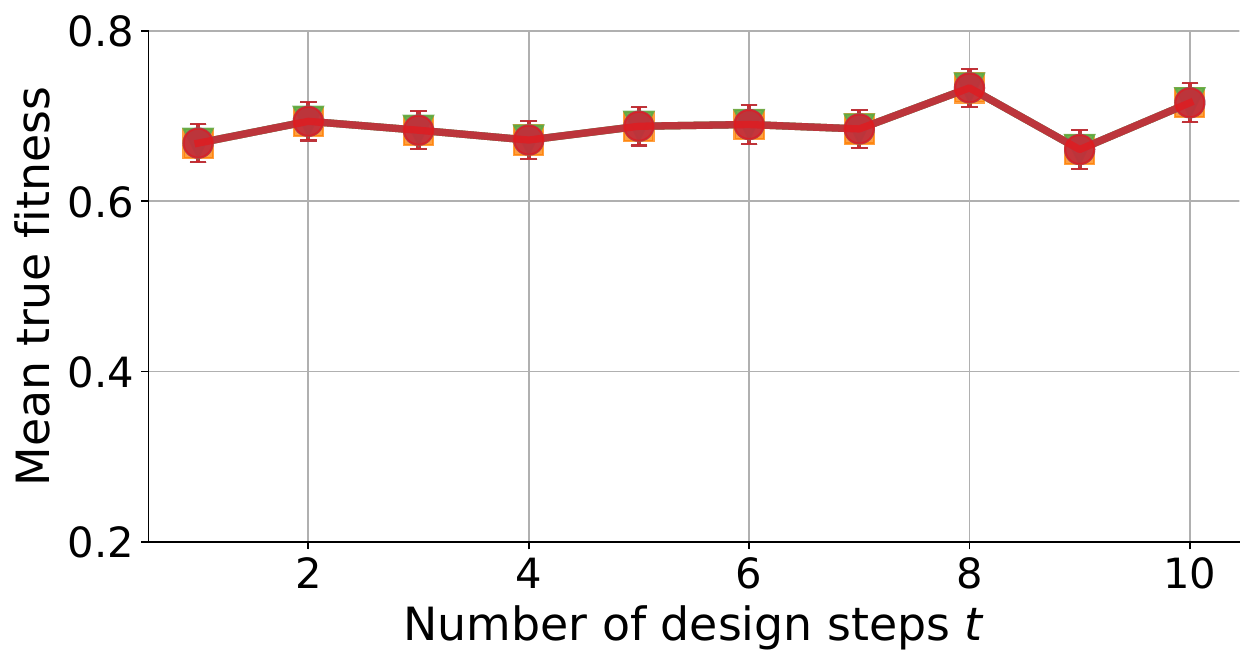}
    \end{subfigure}
    \hfill
    \\
\caption{Ablation studies for the initial amount of training and calibration data ($n_{\text{train}}=n_{\text{cal}}$,) focused on Split CP experiments on the red protein design dataset, where each row corresponds to a different initial sample size $n_{\text{train}}=n_{\text{cal}} \in \{16, 32, 48, 64\}$.
    The ML predictor is the  \texttt{MLPRegressor} from \texttt{scikit-learn} (with LBFGS solver, logistic activation, and default parameters otherwise). All values are computed over 500 random seeds. Other than $n_{\text{train}}$ and $n_{\text{cal}}$, full hyperparameters are as in Appendix \ref{subsec:black_box_opt_exp_details}, Table \ref{tab:fig2}.
}
\label{fig:n_init}
\end{figure*}

\newpage

\subsection{ACI Step Size Ablation Study}
\label{app:subsec:aci_level}

Here we report sensitivity analyses on an ACI baseline's ``step size'' hyperparameter that controls how strongly the method responds to recent over- or under-coverage \citep{gibbs2021adaptive} (by adjusting its empirical quantile level with the aim of having its long-run empirical coverage be close to the target coverage rate). Larger values of the ACI step size correspond to making larger adjustments to recent mistakes, and thus make the method more adaptive. These results are for the airfoil dataset, with hyperparameters as in Appendix \ref{subsec:active_learning_exp_details}, Table \ref{tab:fig3_hyperparams}, aside from the ACI adaptiveness and the reported number of active learning iterations (here $T=25$). In these experiments, increasing the adaptiveness (step size) of ACI tends to \textit{hurt} its coverage performance, possibly due to changes to the CP calibration set causing its retroactive updates to be ``out-of-date.''

\begin{figure*}[h]
    \centering
    \begin{subfigure}{0.85\textwidth}\includegraphics[width=\textwidth]{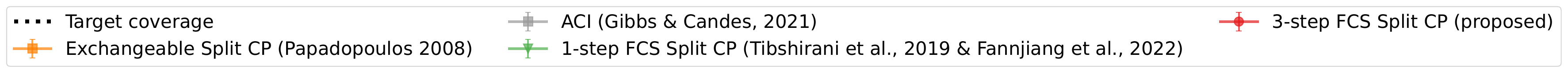}
    \end{subfigure}
    \\
    \begin{subfigure}{0.2\textwidth}\includegraphics[width=\textwidth]{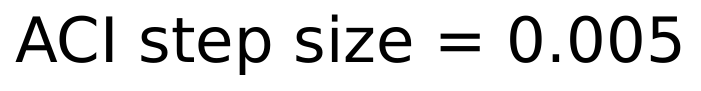}
    \end{subfigure}
    \\
    \hfill
    \begin{subfigure}{0.2\textwidth}
        \includegraphics[width=\textwidth]{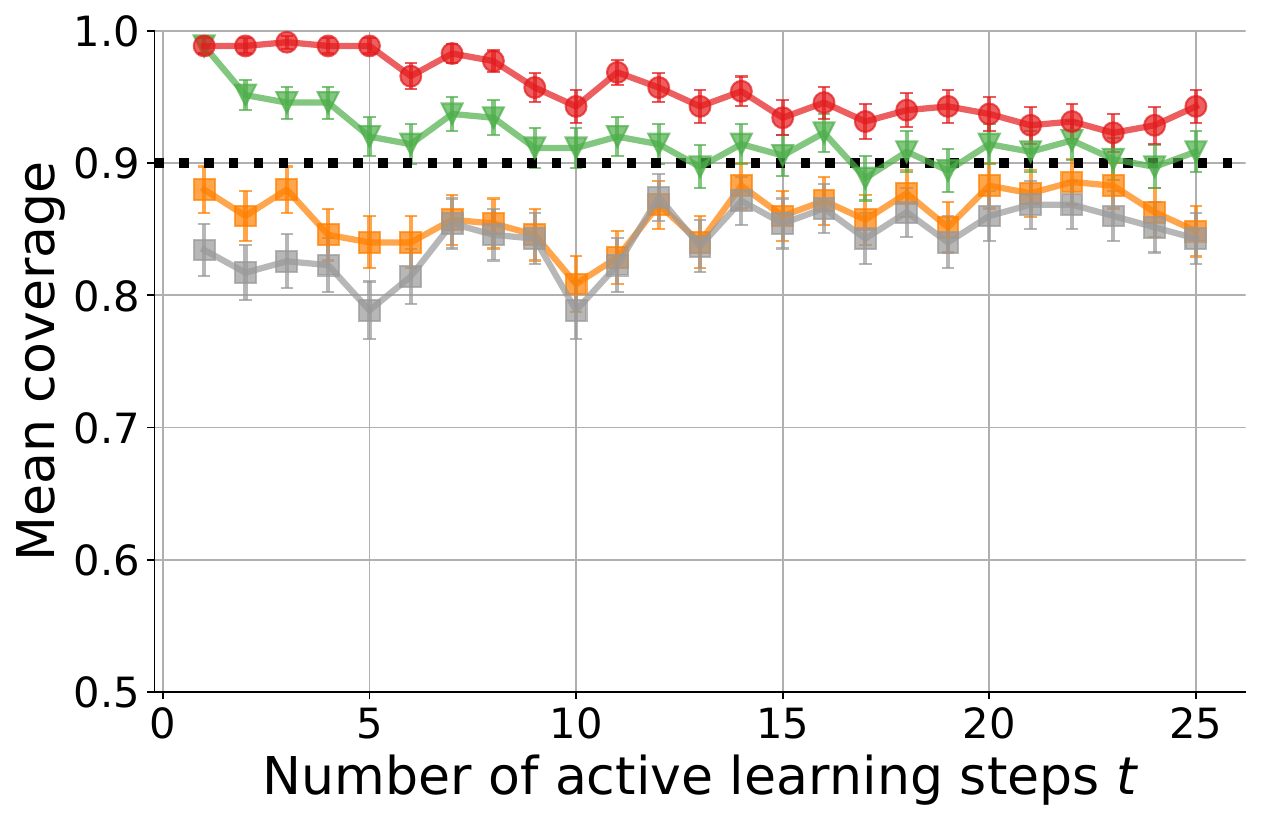}
    \end{subfigure}
    \hfill
    \begin{subfigure}{0.2\textwidth}
        \includegraphics[width=\textwidth]{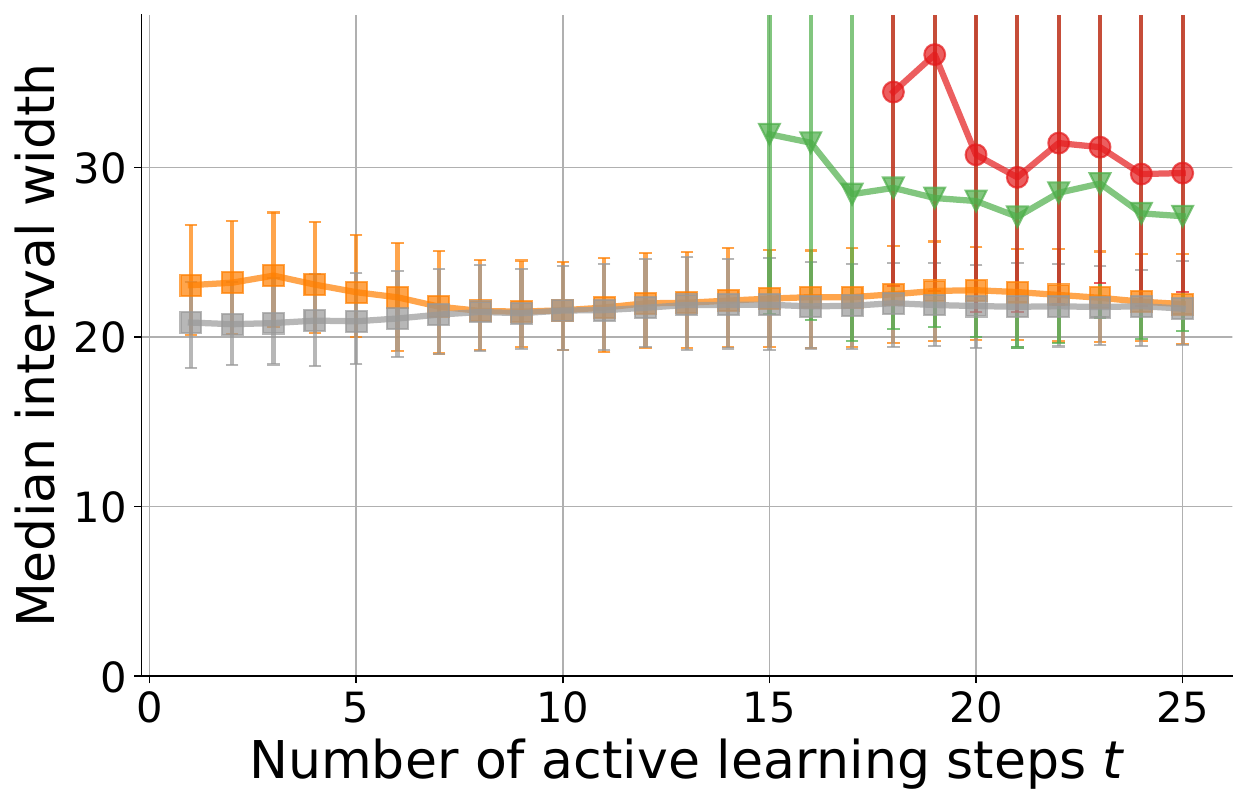}
    \end{subfigure}
    \hfill
    \begin{subfigure}{0.2\textwidth}
        \includegraphics[width=\textwidth]{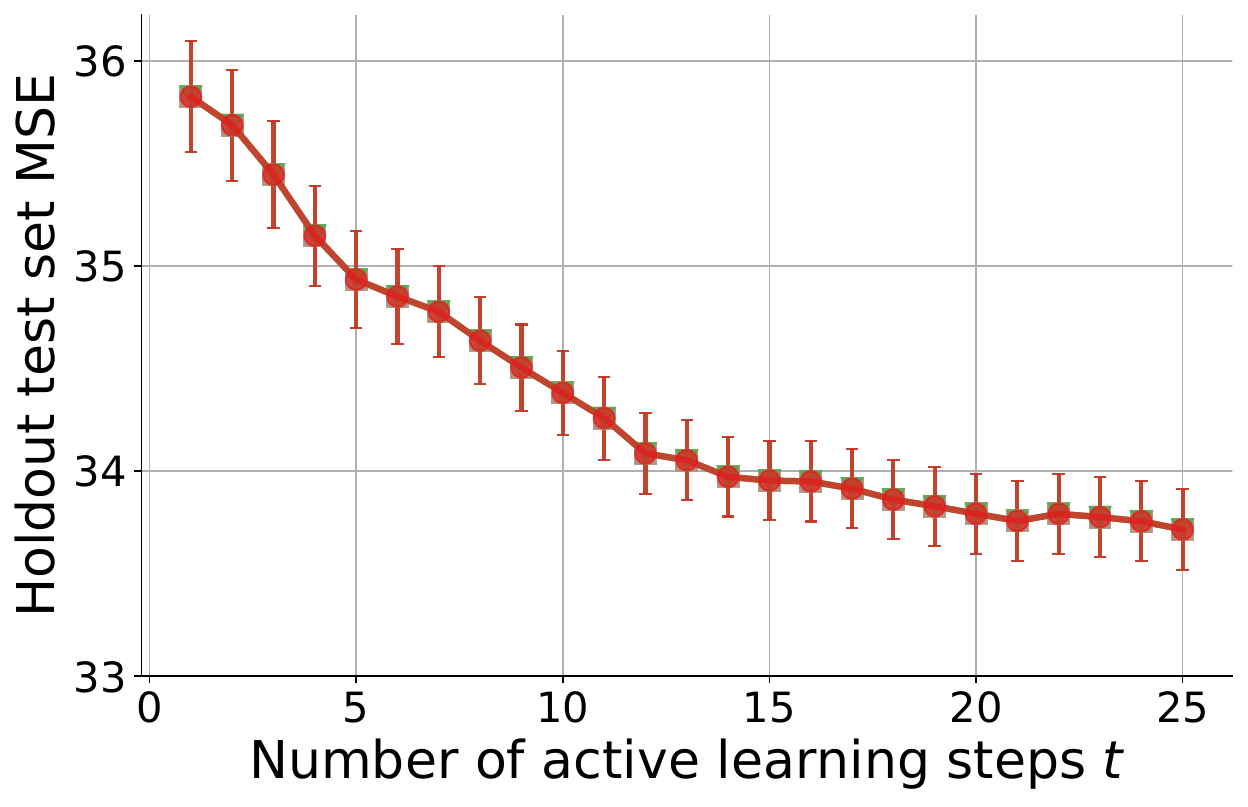}
    \end{subfigure}
    \hfill
    \begin{subfigure}{0.2\textwidth}
        \hfill
        \includegraphics[width=\textwidth]{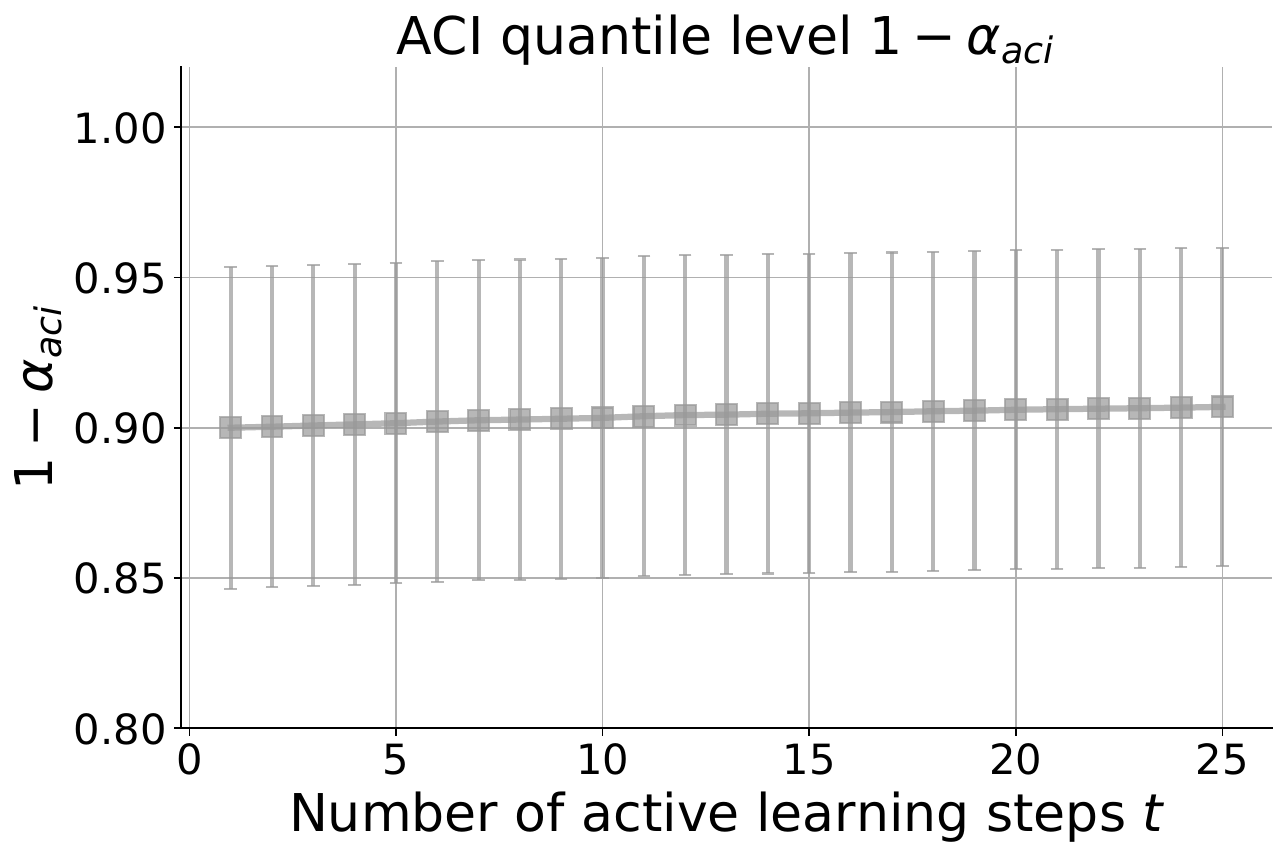}
    \end{subfigure}
    \hfill
    \begin{subfigure}{0.2\textwidth}\includegraphics[width=\textwidth]{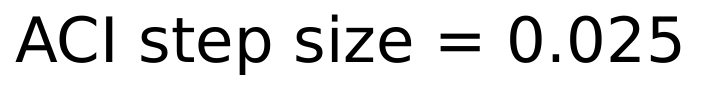}
    \end{subfigure}
    \\
    \hfill
    \begin{subfigure}{0.2\textwidth}
        \includegraphics[width=\textwidth]{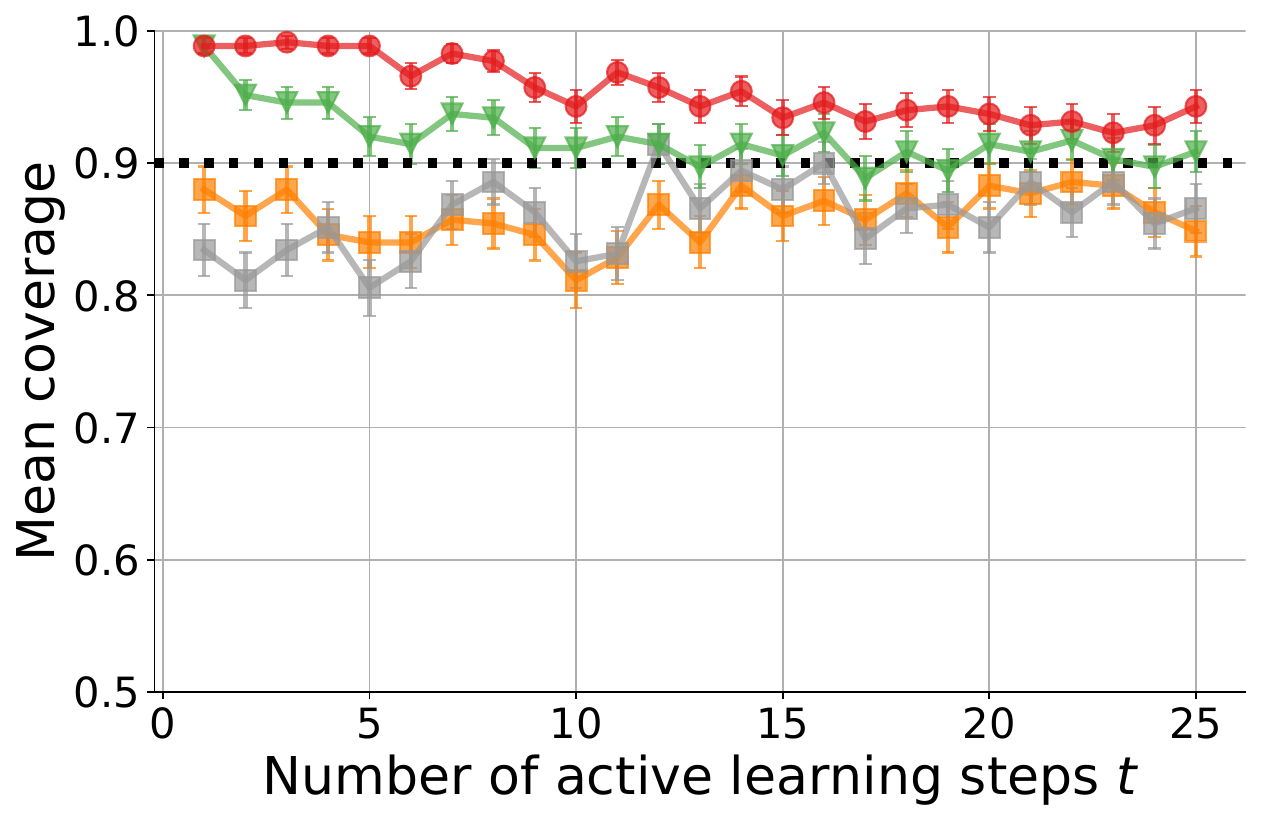}
    \end{subfigure}
    \hfill
    \begin{subfigure}{0.2\textwidth}
        \includegraphics[width=\textwidth]{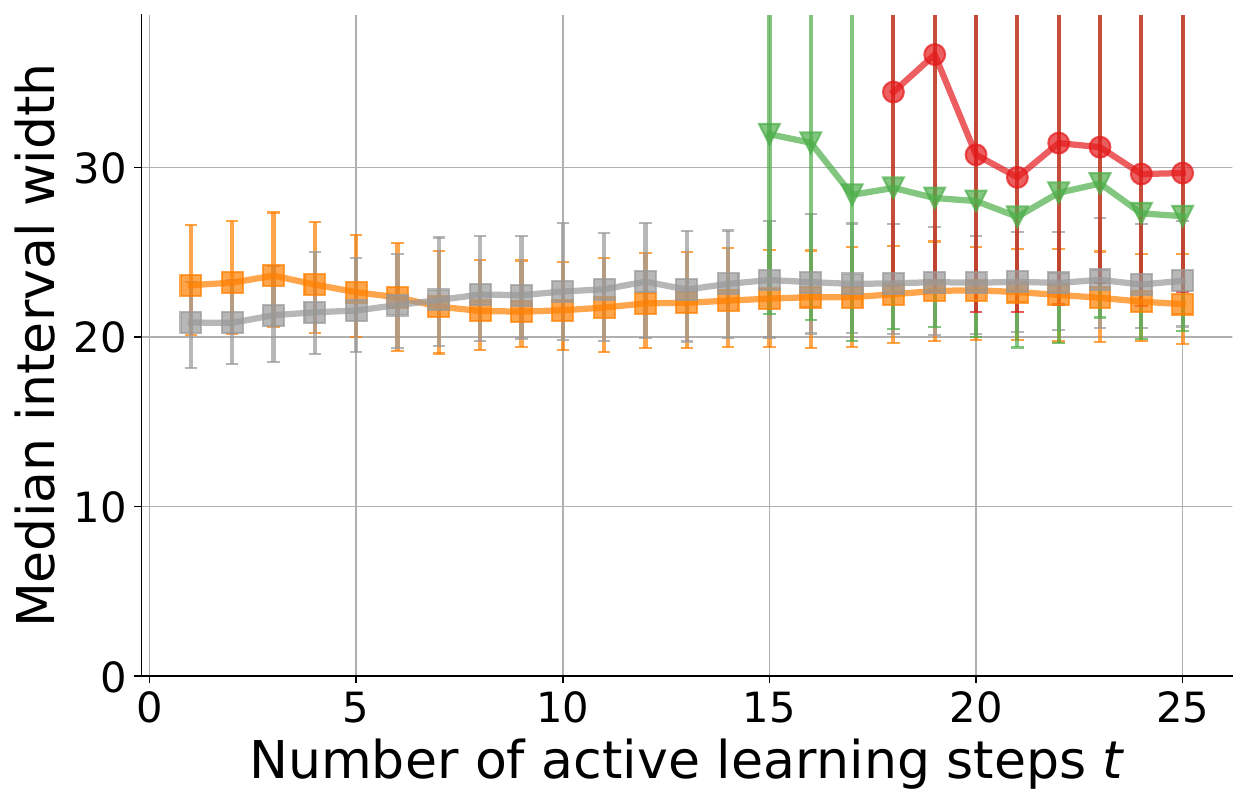}
    \end{subfigure}
    \hfill
    \begin{subfigure}{0.2\textwidth}
        \includegraphics[width=\textwidth]{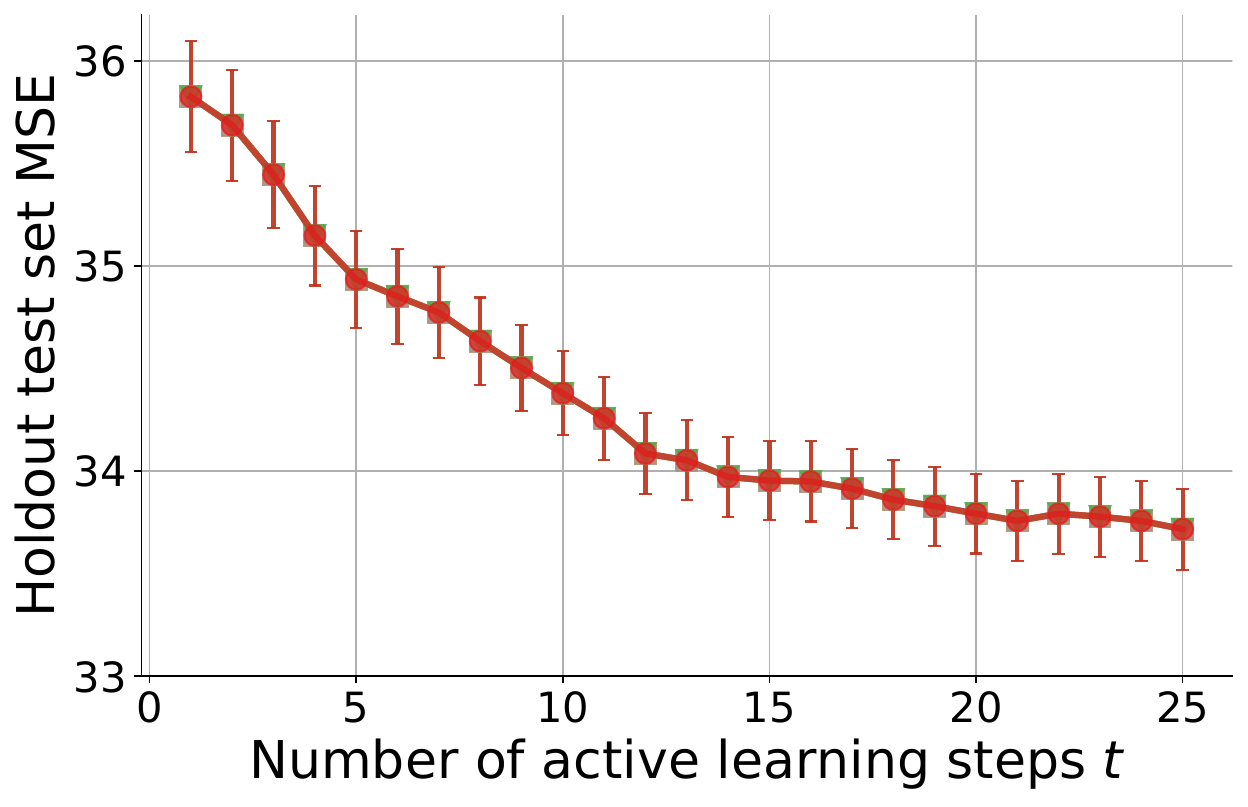}
    \end{subfigure}
    \hfill
    \begin{subfigure}{0.2\textwidth}
        \hfill
        \includegraphics[width=\textwidth]{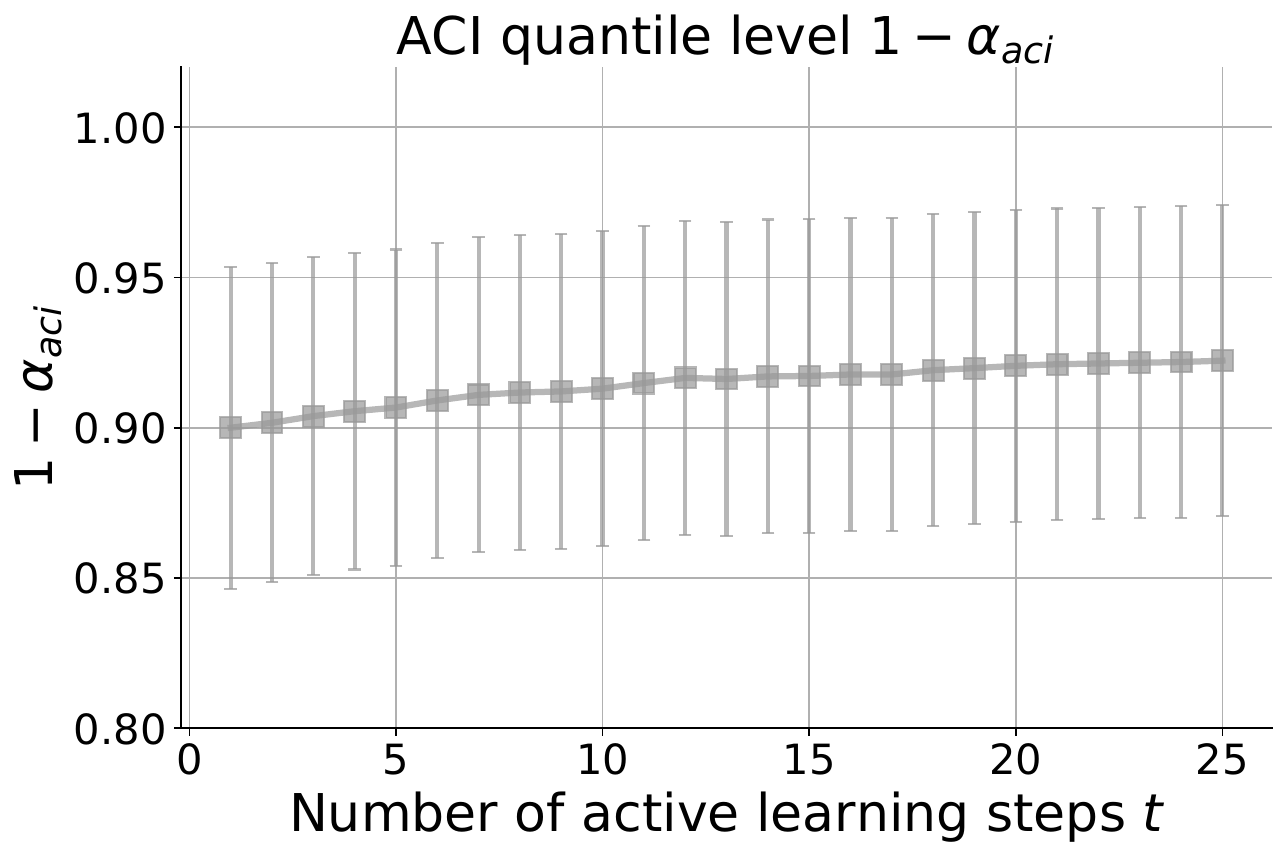}
    \end{subfigure}
    \\
    \begin{subfigure}{0.2\textwidth}\includegraphics[width=\textwidth]{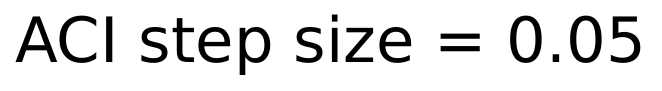}
    \end{subfigure}
    \\
    \hfill
    \begin{subfigure}{0.2\textwidth}
        \includegraphics[width=\textwidth]{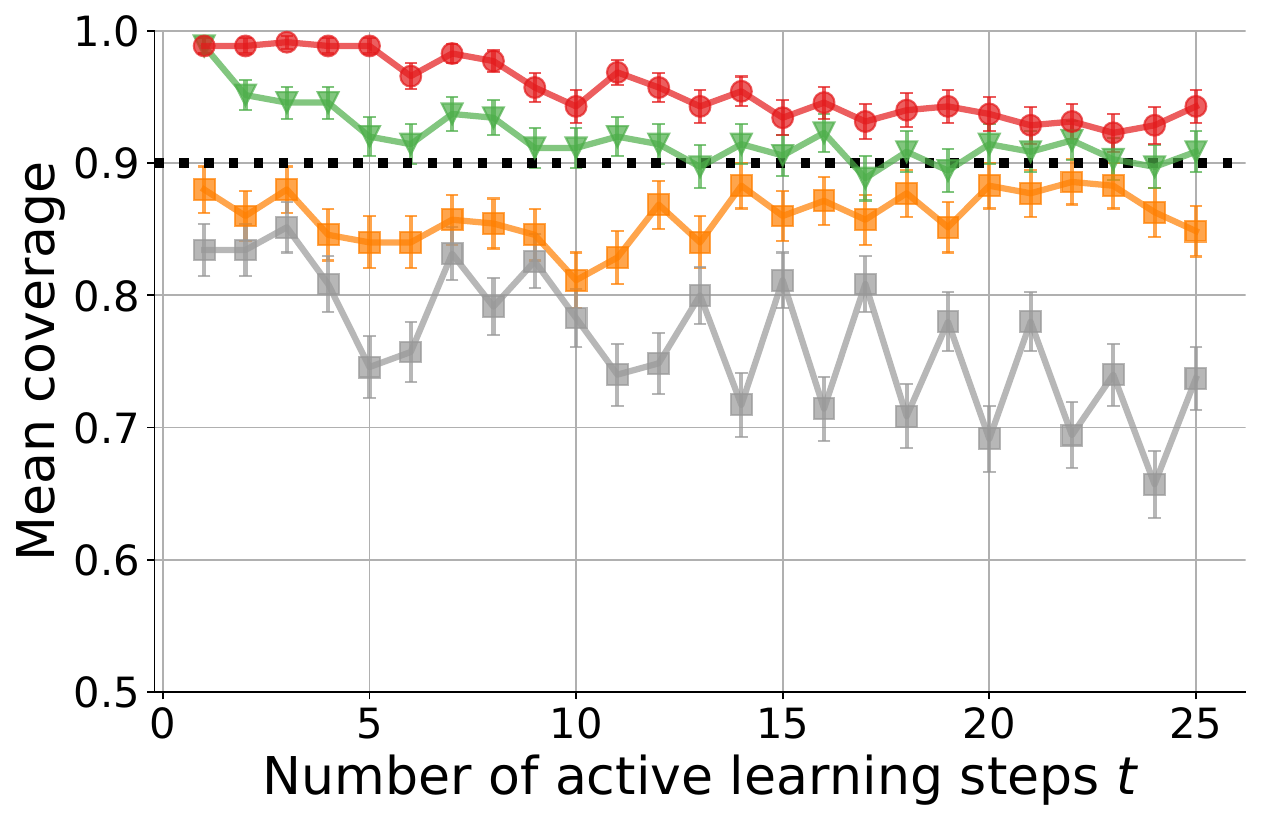}
    \end{subfigure}
    \hfill
    \begin{subfigure}{0.2\textwidth}
        \includegraphics[width=\textwidth]{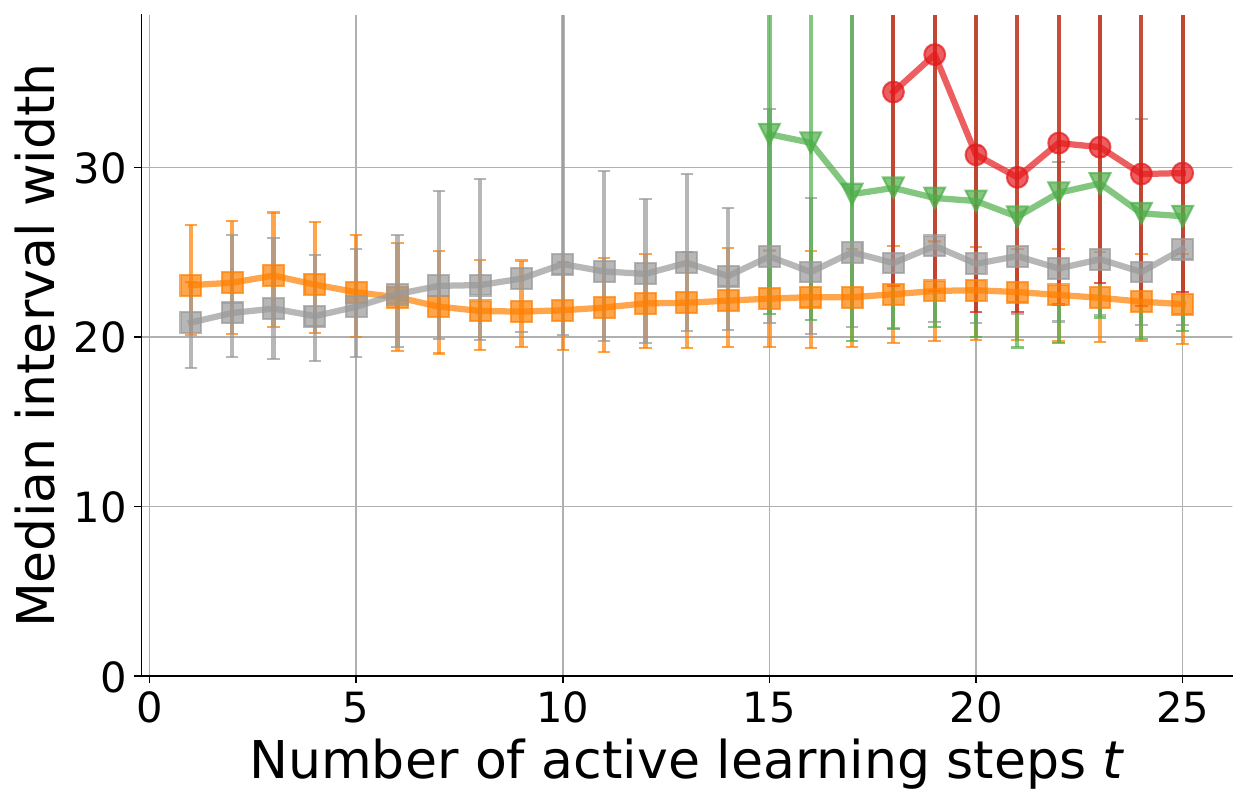}
    \end{subfigure}
    \hfill
    \begin{subfigure}{0.2\textwidth}
        \includegraphics[width=\textwidth]{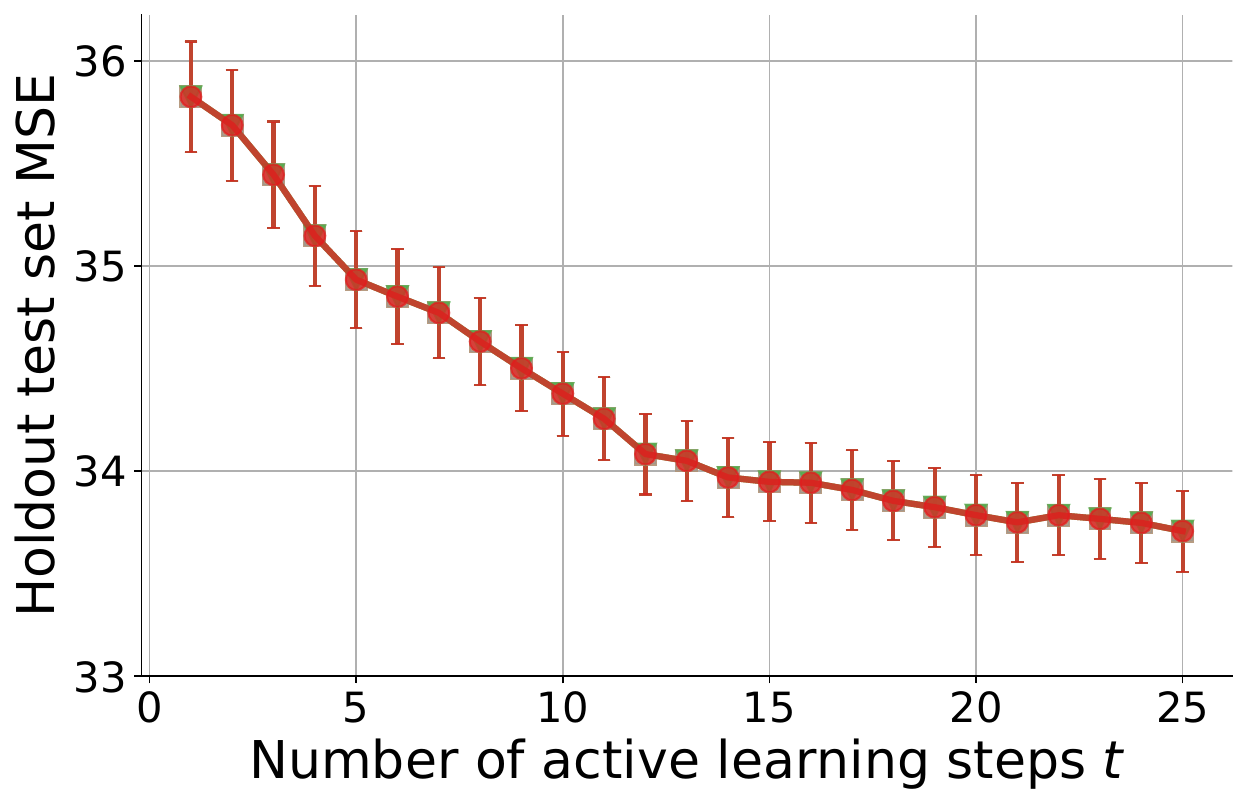}
    \end{subfigure}
    \hfill
    \begin{subfigure}{0.2\textwidth}
        \hfill
        \includegraphics[width=\textwidth]{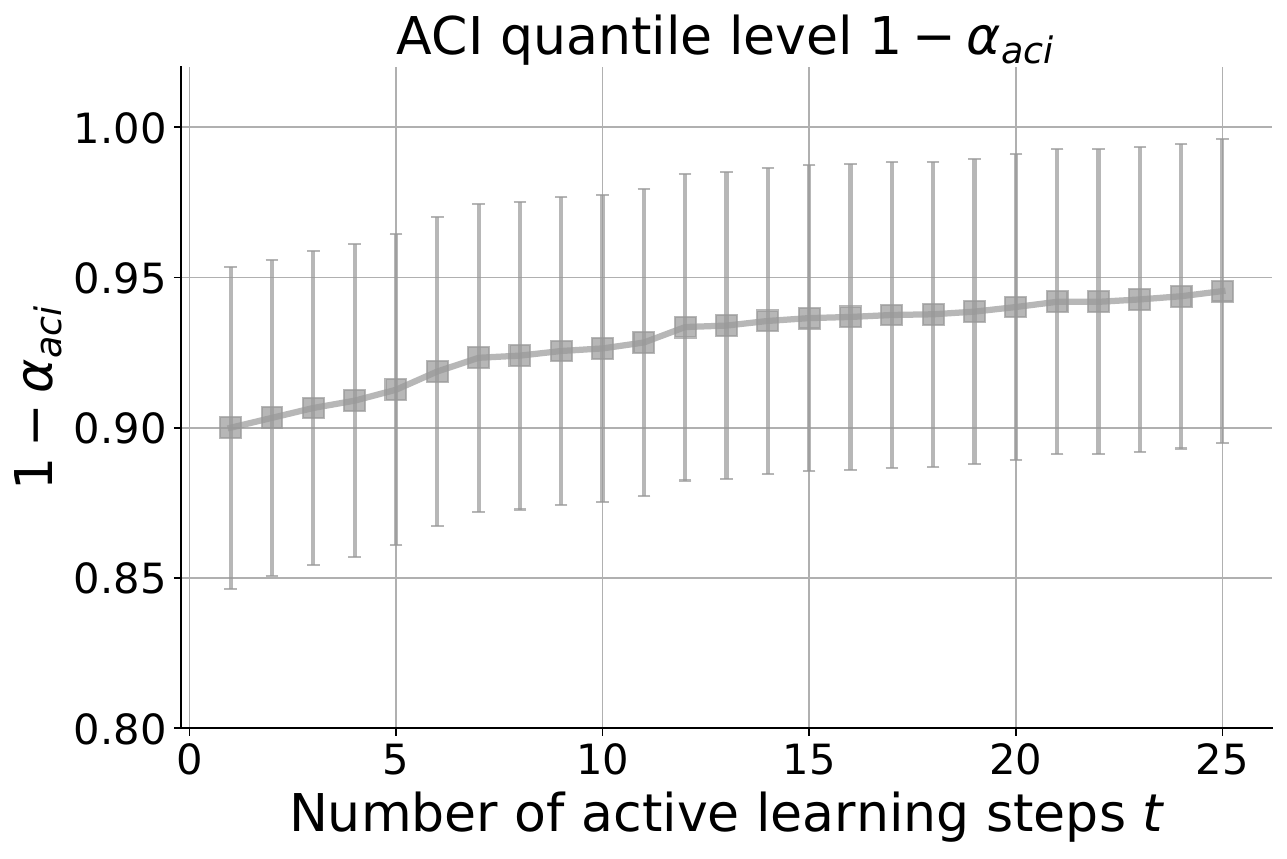}
    \end{subfigure}
    \hfill
    \\
\caption{Ablation studies for ACI baseline's step size over values in $\{0.005, 0.25, 0.05\}$. All these experiments are conducted on the airfoil dataset with the \texttt{GaussianProcessRegressor} from \texttt{scikit-learn}, initially trained on 64 datapoints and with 16 points in the calibration set, and at each active learning step the queried point is added to either training or calibration with equal probability. All values are computed over 350 random seeds. 
    }
\label{appfig:AblationActiveLearningExpts_airfoil}
\end{figure*}

\subsection{Recursion Depth Wall-Clock Runtimes}
\label{app:subsec:runtimes}

\begin{table}[!htb]
\caption{Wall-clock runtimes for computing estimated MFCS weights (Eq. \ref{eq:def_3rd_order_weights}) with different estimation or recursion depths $d$. In these experiments, initially $n_{\text{cal}}=16$ points are in the calibration set and every queried point is added to the calibration set deterministically.  Mean and standard error values are computed over 100 repeated experiments. Increasing the depth by one increases the wall-clock runtime by over an order of magnitude (for each evaluated recursion depth $d \in \{1,...,6\}$), which could quickly become intractable for larger $d$.}
\label{tab:runtimes}
\vskip 0.15in
\begin{center}
\begin{tabular}{cc}
\toprule
MFCS recursion depth $d$ & Wall-clock runtime in seconds ($\pm$ std. err.)  \\
\midrule
1  & 1.297 $(\pm 0.063) \ \mathrm{e}{-6}$ \\
2 & 9.595 $(\pm 0.042) \ \mathrm{e}{-5}$ \\
3 & 1.607 $(\pm 0.008) \ \mathrm{e}{-3}$ \\
4 & 3.194 $(\pm 0.036) \ \mathrm{e}{-2}$ \\
5 & 6.628 $(\pm 0.030) \ \mathrm{e}{-1}$ \\
6 & 1.444 $(\pm 0.001) \ \mathrm{e}{+1}$ \\
\bottomrule
\end{tabular}
\end{center}
\vskip -0.1in
\end{table}


\end{document}